\documentclass[fleqn,10pt]{wlscirep}

\usepackage[utf8]{inputenc}
\usepackage[T1]{fontenc}

\usepackage{amsmath, amssymb}
\usepackage{siunitx}      

\usepackage{graphicx}     
\usepackage{svg}          
\usepackage{float}        
\usepackage{caption}
\usepackage{subfig}       

\usepackage[table]{xcolor} 
\usepackage{array}
\usepackage{multirow}
\usepackage{adjustbox}
\usepackage{longtable}
\usepackage{booktabs}

\newcolumntype{L}[1]{>{\raggedright\arraybackslash}p{#1}}
\newcommand{\taskrule}{\cmidrule[\lightrulewidth](lr){2-5}}

\usepackage{fullpage}     
\usepackage{paralist}     
\usepackage{enumitem}     
\usepackage{pifont}

\usepackage{listings}     
\usepackage{algorithm}
\usepackage{algorithmicx}
\usepackage{algcompatible}
\usepackage{algpseudocode}

\usepackage{verbatim}     
\usepackage{marvosym}
\usepackage{etoc}
\usepackage[normalem]{ulem}
\usepackage{nameref}

\usepackage[numbers,sort]{natbib}
\usepackage{xr}           
\usepackage{xurl}         
\usepackage{hyperref}
\hypersetup{breaklinks=true}

\usepackage[nameinlink,noabbrev]{cleveref} 

\crefname{figure}{Fig.}{Figs.}         
\crefname{table}{Table}{Tables}
\crefname{section}{Section}{Sections}
\crefname{appendix}{Supplementary}{Supplementary}

\title{Learning Sparse Latent Predictive Foundation Model for Multimodal Neuroimaging}

\author[1\Letter]{Haoxu Huang}
\author[1]{Long Chen}
\author[2,9]{Jingyun Chen}  
\author[1]{Jinu Hyun}
\author[2]{James Ryan Loftus}
\author[3]{Kara Melmed}
\author[4,5]{Daniel Orringer}
\author[3]{Jennifer Frontera}
\author[2,10]{Seena Dehkharghani}
\author[3,6,7]{Arjun Masurkar}
\author[1,2,8\Letter]{Narges Razavian}
\affil[1]{New York University, Center for Data Science, New York, NY, 10001, USA}
\affil[2]{NYU Grossman School of Medicine, Department of Radiology, New York, NY, 10016, USA}
\affil[3]{NYU Grossman School of Medicine, Department of Neurology, New York, NY, 10016, USA}
\affil[4]{NYU Grossman School of Medicine, Department of Neurosurgery, New York, NY, 10016, USA}
\affil[5]{NYU Grossman School of Medicine, Department of Pathology, New York, NY, 10016, USA}
\affil[6]{NYU Grossman School of Medicine, Department of Neuroscience, New York, NY, 10016, USA}
\affil[7]{NYU Grossman School of Medicine, Neuroscience Institute, New York, NY, 10016, USA}
\affil[8]{NYU Grossman School of Medicine, Department of Population Health, New York, NY, 10016, USA}
\affil[9]{State University of New York at Binghamton, School of Computing, Binghamton, NY 13902, USA}
\affil[10]{Stanford University, Department of Radiology, Stanford, CA, 94305, USA}

\affil[\Letter]{\textbf{Corresponding author:}Narges.Razavian@nyulangone.org, hh2740@nyu.edu}

\begin{abstract}
Magnetic resonance imaging (MRI) of the brain routinely combines complementary contrasts—including T1-weighted (T1w), T2-weighted (T2w) and FLAIR sequences—to capture anatomy and pathology. Although foundation models trained on large neuroimaging datasets have shown promise, it remains unclear whether they effectively integrate these complementary contrasts across diverse clinical settings. Here we present \textit{Neuro-JEPA}, a multimodal foundation model that learns unified representations across structural brain MRIs. Neuro-JEPA combines latent predictive learning with a sparse Mixture-of-Experts architecture to encode multiple MRI contrasts within a single model. The model is pretrained on 1.55 million scans from 428,647 clinical imaging studies spanning T1w, T2w and FLAIR sequences. The model performance is evaluated on 47 downstream tasks across three independent health systems and 12 public datasets, encompassing single-contrast, multimodal and cross-domain applications. Across these evaluations, Neuro-JEPA achieved the strongest overall performance and generalized robustly across heterogeneous cohorts, outperforming existing neuroimaging foundation models and strong conventional baselines in the majority of comparisons despite using less than one-third of the pretraining data of leading models. Performance gains were particularly pronounced in data-scarce settings, reaching up to 15\% over prior foundation models for clinically important pathologies including gliosis and multiple sclerosis. Neuro-JEPA showed clinically relevant capabilities across domains, including prediction of amyloid status and progression from mild cognitive impairment to Alzheimer’s disease dementia, stroke identification and classification, and prediction of post-stroke functional outcome and length of stay. Together, these results show that data scale, curation and principled model design can act synergistically to produce broadly transferable neuroimaging representations, and establish Neuro-JEPA as a scalable framework toward general-purpose multimodal foundation models for neurological imaging.

\end{abstract}
\begin{document}

\flushbottom
\maketitle
%
%
\thispagestyle{empty}


\section*{Introduction}
Neuroimaging is central to the care of the patients with neurological conditions, with a multitude of clinical guidelines framed directly around the imaging manifestations of diseases. Magnetic Resonance Imaging (MRI) remains the favored approach for in-vivo tissue characterization, owing to the richness of biological contrasts afforded by carefully tuned acquisition parameters, yielding variable contrast weightings designed to accentuate desired structural, functional, or pathologic features. MRI has thus emerged as an indispensable tool for diagnosis, prognosis, and treatment monitoring for a vast array of neurological conditions, ranging from neurodegenerative diseases \cite{Qiu2022} and brain tumors \cite{Castellano2016-sl} to cerebrovascular events \cite{Warach1992-ky} and traumatic injuries \cite{Le2009-ky}. With nearly 40 million neuroimaging examinations performed annually in the United States alone \cite{patel2025mri}, the staggering volume of data requiring interpretation by highly specialized medical imaging specialists has spurred demand for scalable computational tools capable of imaging interpretation and decision-support.

Due to the multicontrast nature of brain MRI, comprising potentially dozens of pulse sequences each defined by carefully tuned radiofrequency and magnetic field gradient pulses, even the most routine clinical brain MRI can be viewed as inherently multimodal. These distinct MRI acquisition sequences yield complementary biophysical signals that together inform the interpretation of the exam, each operating along a unique axis sensitized. While considerable variability may exist between brain MRI studies, several pulse sequences stand as essentially standard core contrasts; specifically,T1-weighted (T1w) sequence offers exquisite anatomical contrast; T2-weighted (T2w) sequence are highly sensitized to fluid content; and FLAIR contrasts modifies base T2-weighted contrast through suppression of bulk/free aqueous pools such as the cerebrospinal fluid, in order to increase conspicuity for tissue edema and other pathologies. In clinical practice, expert diagnostic reasoning is informed by the combination of findings across such pulse sequences, which together inform the presence or absence of diseases.

There nevertheless remains a clear gap between this cognitive-clinical exercise and the underlying logic driving contemporary clinical AI paradigms. Most methods operate on a single sequence and cannot exploit cross-contrast synergies, and the multimodal methods that do exist are typically evaluated with all sequences present as input. This leaves two central questions unanswered: does jointly encoding multiple contrasts actually outperform the best single-sequence model, and does any advantage survive under varied disease presentations and differences across sites and scanners? Without controlled unimodal-versus-multimodal comparisons under these conditions, gains attributed to multimodality cannot be distinguished from the benefit of simply having more data at test time. Compounding this, the field lacks principled study of how design choices shape what these models learn, relying instead on direct application of existing algorithms or heuristic modifications. Closing these gaps is essential to move from heuristic design toward a clear understanding of the factors that drive representation learning in human neuroimaging. Under systematic studies on what drives the improved learning of neuroimaging foundation model, our model achieves a large margin of improvement over NeuroVFM~\cite{kondepudi2025healthlearningachievesgeneralist}, despite the latter being pretrained on $\sim$5 million scans (more than three times our pretraining data), underscoring that proper pretraining algorithm, not data scale alone, is decisive for building effective neuroimaging foundation models.


\begin{figure}[htbp]
    \centering
    \includegraphics[width=0.92\textwidth]{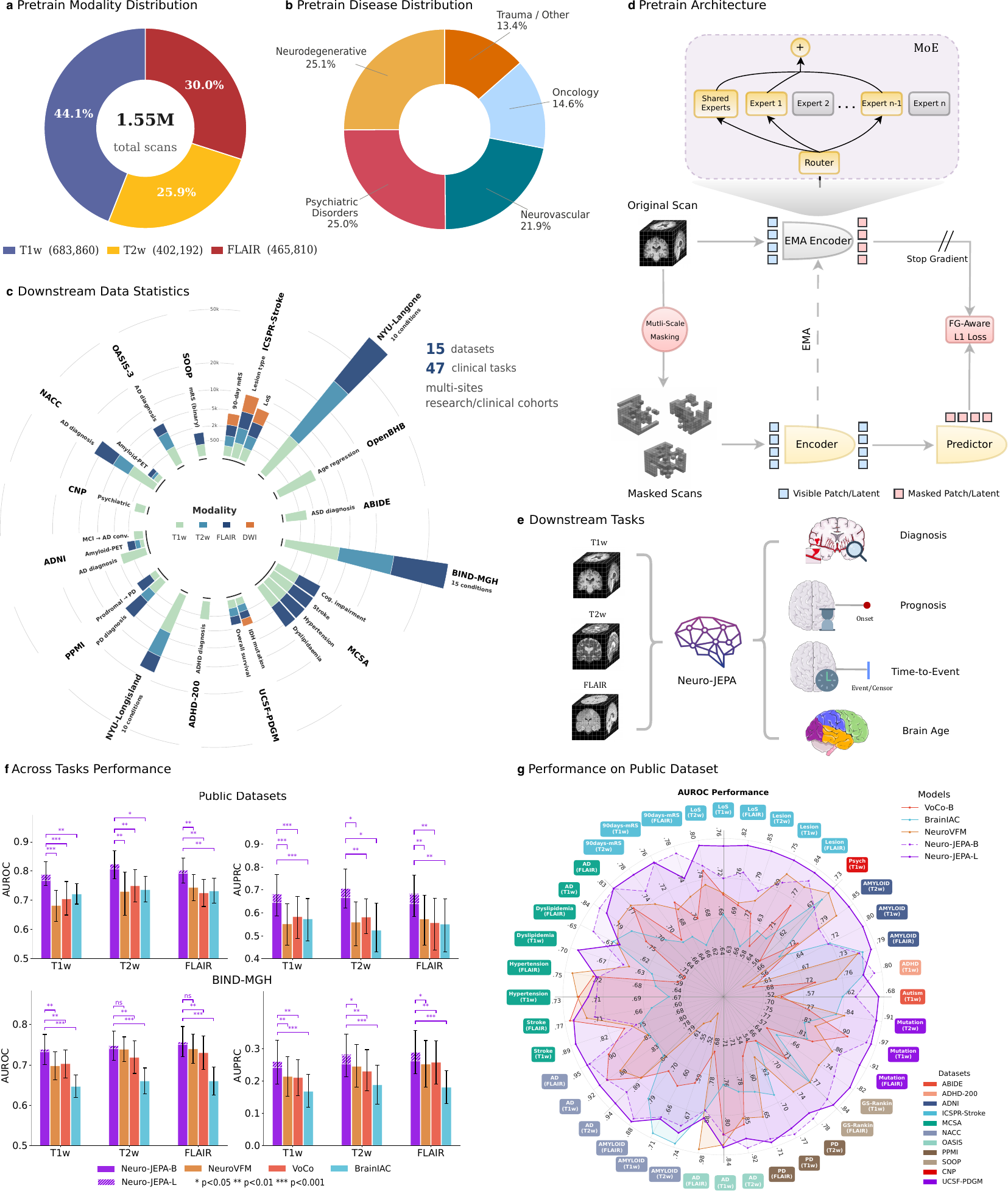}
    \caption{\textbf{Overview of the study} - the full pipeline on pre-training Neuro-JEPA with data distribution and performance evaluations. \textbf{a,} MRI modalities distribution on T1w, T2w and FLAIR. \textbf{b,} Disease distribution on pre-training data from five main categories. \textbf{c,} Number of patients for each task and modality on evaluated downstream datasets. \textbf{d,} Neuroimaging specialized pre-training architecture built upon JEPA with improvement on masking strategies, foreground-aware loss and backbone sparsification with MoE. \textbf{e,} Downstream evaluation with FM-NeuroSp on Diagnosis, Prognosis, Time-to-Event and Age Prediction by taking input from three modalities (T1w, T2w, FLAIR). \textbf{f,} Models average performance across tasks for public datasets and BIND-MGH with statistical significance on each modalities (reported on both AUROC and AUPRC). \textbf{g,} Per task AUROC performance on different tasks for different foundation models with public datasets. Full name and definition for each task abbreviation is present in Section \hyperref[sec:pub_dataset]{"Public Research Datasets"}}
    \label{fig:overview}
\end{figure}

Here we introduce Neuro-JEPA, a sparse multimodal neuroimaging foundation model integrating a Vision Transformer (ViT) \cite{dosovitskiy2021an}, joint-embedding predictive learning (JEPA) \cite{Assran23,assran2025vjepa2,lecun2022path}, and a Mixture-of-Experts (MoE) architecture \cite{JMLR:v23:21-0998,NEURIPS2021_48237d9f,jordan93,mustafa2022multimodal,shazeer2017}. After pretraining on 1,551,862 MRI sequences (T1w, T2w, FLAIR) from 428,647 studies and 282,693 patients, Neuro-JEPA consistently outperformed frontier neuroimaging foundation models — BrainIAC \cite{Tak2026}, VoCo \cite{voco,wu26}, and NeuroVFM \cite{kondepudi2025healthlearningachievesgeneralist} — across three health systems and twelve public cohorts. Rather than reporting only the all-sequences-present case for each study, we quantify multimodality performance through controlled unimodal, multimodal, and cross-domain comparisons, showing when and by how much joint encoding different modalities helps.

Importantly, Neuro-JEPA extends neuroimaging foundation models to clinically consequential disease settings and endpoints that have received comparatively limited attention in prior work. From routine structural sequences alone, Neuro-JEPA demonstrated strong performance across acute stroke and neurodegenerative disease tasks, including stroke identification and classification, prediction of 90-day modified Rankin Scale (mRS) outcome and hospital length of stay, identification of small or subtle lesions, estimation of cerebral amyloid status from standard T1w, T2w, and FLAIR imaging using amyloid PET as the reference standard, and prediction of progression from prodromal to clinically manifest neurodegenerative disease. These tasks address major unmet clinical needs. Rapid recognition of stroke is essential because ischemia, hemorrhage, and mimics require fundamentally different and often time-sensitive management, while early outcome prediction can inform treatment decisions, recovery expectations, and discharge planning. In Alzheimer’s disease, determination of cerebral amyloid burden is increasingly central to diagnosis and eligibility for disease-modifying therapies, yet current approaches rely on PET, cerebrospinal fluid assays, or emerging blood-based biomarkers that may be costly, invasive, or incompletely accessible. More broadly, predicting progression from prodromal to symptomatic stages in disorders such as Alzheimer’s and Parkinson’s disease could enable earlier risk stratification, anticipatory guidance, and enrichment of clinical trials targeting disease before substantial irreversible neurodegeneration occurs. Reliable inference of these molecular, diagnostic, and prognostic endpoints from sequences already acquired in routine clinical examinations could therefore extend clinically actionable information to settings where specialized testing is unavailable, delayed, or costly. Together, these capabilities highlight the promise of predictive latent-space foundation models not only for extracting clinically meaningful representations of current disease state, but also for learning trajectories of disease evolution that may ultimately support future-state prediction, longitudinal forecasting, and clinically informed planning.



\section*{Results}
\label{sec:results}
Although structural MRI sequences (T1w, T2w, FLAIR) share a common physical origin, their distinct statistical profiles, tissue contrasts, and clinical utilities present learning multi-sequence brain MRI signals as a multi-modal representation learning problem. Under this framing, we developed and validated Neuro-JEPA, a Mixture of Experts (MoE) transformer-based foundation model optimized for neuroimaging, accompanied by detailed analyses of multi-modal learning. \Cref{fig:overview} provides an overview of the study and its high-level results. Neuro-JEPA was pretrained on data drawn from the imaging informatics archive of a large health system, spanning diverse disease profiles, demographics, and imaging devices (N=282,693 patients; 1,551,862 multi-modal MRI sequences across T1w, T2w, and FLAIR) (\Cref{fig:overview}.a and \Cref{fig:overview}.b), with the characteristics of the training data reported in Supplementary \Cref{tab:pretrain-demo}. The model was trained to handle all three modalities within one architecture. To evaluate its capabilities, we assessed both how well the model encodes individual modality representations and how effectively it captures complementary information across modalities under multi-modal learning. These evaluations span heterogeneous neurological phenotypes and clinical endpoints, including neurodegenerative disease, cerebrovascular disease, inflammatory and structural abnormalities, and both diagnostic and prognostic tasks, enabling us to assess whether a single pretrained representation transfers across clinically distinct settings.

\begin{figure}[htbp]
    \centering
    \includegraphics[width=\textwidth]{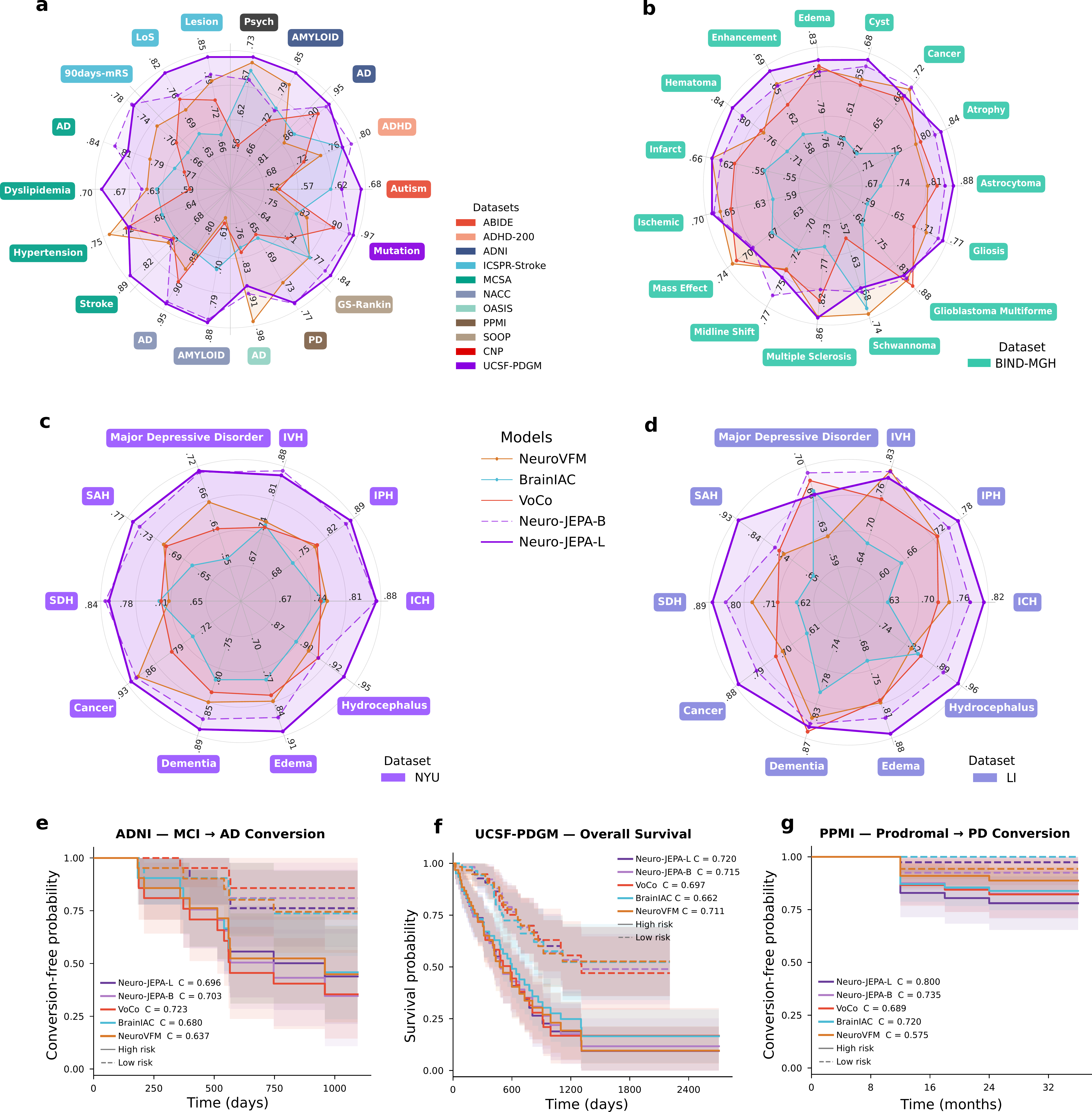}
    \caption{\textbf{Best Achievable Unimodal Performance} - AUROC for diagnosis/prognosis and C-index for time-to-event tasks are reported with performance from best modality across four foundation models (NeuroVFM, BrainIAC, VoCo and Neuro-JEPA) evaluated under full finetuning except NeuroVFM. The result shows that our model achieves overall best performance across the tasks. \textbf{a,} per task performance on best performance modality per task on public datasets. \textbf{b,} per task performance on best performance modality per task on BIND-MGH. \textbf{c,} per task performance on best performance modality per task on NYU Langone. \textbf{d,} per task performance on best performance modality per task on NYU Long Island. \textbf{e,} time-to-event on Mild Cognitive Impairment (MCI) to Alzheimer Dementia (AD) conversion with ADNI dataset. \textbf{f,} time-to-event on overall survival (until death) on Gliobastoma with UCSF-PDGM dataset. \textbf{g,} time-to-event on Prodromal to Parkinson (PD) conversion with PPMI dataset.}
    \label{fig:fig2-best-unimodal}
\end{figure}

\begin{figure}[p]
    \centering
    \includegraphics[width=0.80\textwidth]{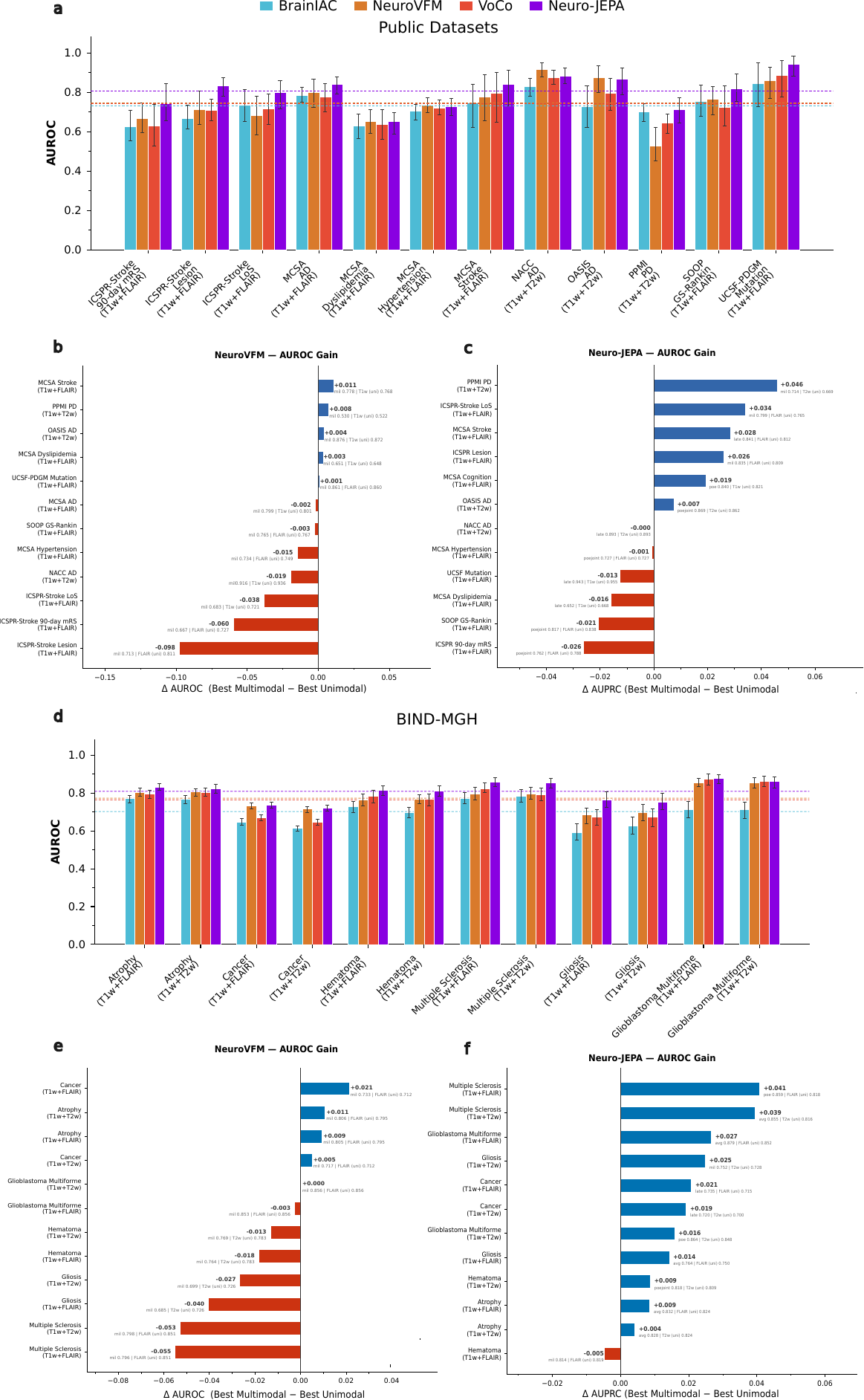}
    \caption{\textbf{Multimodal Performance and Gain over Unimodal Baselines.}}
    \label{fig:fig3-multimodal}
\end{figure}

We performed our main evaluations on both the ViT-Base (Neuro-JEPA-B) and ViT-Large (Neuro-JEPA-L) models with MoE (\Cref{fig:overview}f,g and \Cref{fig:fig2-best-unimodal}, with detailed configurations in Section \hyperref[sec:model_arch]{"Model Architecture"}). The remaining ablation experiments (\Cref{fig:fig3-multimodal,fig:fig4-ablation,fig:fig5-few-shot-analysis}) were conducted using the Neuro-JEPA-B (86 million activated parameters out of 122 million total). This ensures that the number of activated parameters is comparable to the baseline foundation models used in our evaluations for a fair comparison, including BrainIAC (88 million parameters), VoCo-Base (73 million parameters), and NeuroVFM (86 million parameters). Scaling the parameter count via the ViT-Large model with MoE (303 million activated parameters out of 429 million total) presented an average improvement of 1.5--1.9\% on AUROC and 1.9--4.1\% on AUPRC, as demonstrated in Section \hyperref[sec:scalability]{"Scalability Under Model Size"}, representing the effectiveness of our method under model size scaling. All performance comparisons mentioned in the following sections are calculated based on Neuro-JEPA-B model unless otherwise specified.

Across these evaluations, Neuro-JEPA achieved strong performance, with average improvements of 4.4--6.4\% in AUROC and 6.4--9.4\% in AUPRC over state-of-the-art alternatives (\Cref{fig:overview}.f,g; Section \hyperref[sec:unimodal]{"Uni-Modal Learning Capability"}; \Cref{fig:fig2-best-unimodal}). Importantly, these gains extended across clinically distinct disease domains and endpoint types rather than being concentrated within a single disease category. In neurodegenerative disease, Neuro-JEPA performed strongly in predicting amyloid status and progression from mild cognitive impairment to Alzheimer's disease dementia, and demonstrated competitive performance for Parkinson's disease. In cerebrovascular disease, the model supported both stroke identification and classification and prognostic endpoints, including 90-day modified Rankin Scale and length of stay. Performance gains were also pronounced for disorders including gliosis and multiple sclerosis, demonstrating transfer across heterogeneous neurological phenotypes.

To quantify multimodal gains, we measured the added-value in downstream task performances, comparing the best multi-modal trained model vs. the best uni-modal trained alternative for each task. Neuro-JEPA exhibits consistent higher improvements in uni- vs multi-modal setting, and strong overall multimodal performance compared to existing foundation models with 5.8--7.6\% improvement on AUROC and 6.2--8.5\% on AUPRC with evaluated public datasets (Section \hyperref[sec:multimodal]{"Multi-Modalities Learning Capability"}, \Cref{fig:fig3-multimodal}). 

We further investigated the design of self-supervised pretraining within the latent predictive modeling paradigm. We demonstrate that direct deployment of existing configurations, such as V-JEPA 2 \cite{assran2025vjepa2}, yields suboptimal performance on neuroimaging. Through systematic analysis and extensive experimentation, we identified three critical adaptations for robust and improved pretraining: (1) multiscale masking, which optimally calibrates the difficulty of the anatomical prediction task; (2) MoE-driven sparsity, which scales model capacity while disentangling heterogeneous neuroimaging anatomies and (3) background signal suppression, which forces the predictive loss to prioritize brain tissue. Comprehensive evaluations of these design choices are detailed in the Results (Section \hyperref[sec:opt_pretrain]{"Key Architectural Components for Effective Pre-training"}) and Methods (Section \hyperref[sec:model_arch]{"Model Architecture"}).

Beyond full-data evaluations, we assessed adaptation under limited supervision \cite{Tak2026,Vorontsov2024,chen2024uni}, a property of particular importance in neuroimaging, where expert annotations and disease-specific cohorts are frequently scarce. Neuro-JEPA consistently outperformed all comparison models across few-shot settings, with the largest gains observed for selected clinical phenotypes: AUROC improvements of up to 15\% over prior best neuroimaging foundation models for gliosis and multiple sclerosis, and of more than 5\% for several clinically important conditions, including Alzheimer's disease, IDH mutation status, and cancer (Section \hyperref[sec:label_eff]{``Label Efficiency Under Few Shots''}; \Cref{fig:fig5-few-shot-analysis}). These results indicate that the learned representations remain transferable even when task-specific supervision is limited.

We next asked whether the benefits of Neuro-JEPA extended beyond comparisons among pretrained foundation models. Each foundation model was therefore compared with a neuroimaging-specific CNN trained from scratch for the corresponding downstream task~\cite{Liu2022}. Across public datasets, Neuro-JEPA was the only evaluated foundation model to outperform the supervised CNN baseline on average, whereas other foundation models showed inconsistent advantages over task-specific supervised learning (Section~\hyperref[sec:cnn_baseline]{"Comparison with a supervised CNN baseline"}). This result indicates the importance of holistic neuroimaging foundation model evaluation beyond pretrained counterparts.

Finally, we assessed scaling with respect to pretraining data size using 10\%, 30\% and 100\% of the training corpus. Under controlled model architectures, downstream performance improved monotonically with increasing pretraining data (Section \hyperref[sec:scalability]{"Scalability Under Pretraining Data Size"}). Together with the architectural ablations and model-size scaling experiments, these findings show that Neuro-JEPA benefits from both increased data and model capacity.

\subsection*{Uni-Modal Learning Capability}
\label{sec:unimodal}
To assess the performance of individual modality encoding, we benchmarked on diagnosis and prognosis tasks with Neuro-JEPA across 105 dataset--task--modality combinations spanning 3 institutions and several publicly available research-grade datasets. Our study datasets include NYU Langone (NYU) and NYU Long Island (LI) with 10 EHR-driven detection tasks for Hematoma subtypes, Cancer, Edema, Dementia, Major Depressive Disorder, and Hydrocephalus; Massachusetts General Hospital (MGH) across 15 radiology-driven outcomes: Edema, Cyst, Enhancement, Hematoma, Infarct, Atrophy, Ischemic, Mass Effect, Multiple Sclerosis, Midline Shift, Glioblastoma Multiforme, Gliosis, Astrocytoma, and Schwannoma. In addition, we evaluated model quality on 12 research (public) datasets derived from large international studies of Dementia and Aging (ADNI, NACC, OASIS, and MCSA), Parkinson's Disease (PPMI), Stroke (SOOP and ICSPR-Stroke), Glioblastoma (USCF-PDGM), Attention-Deficit/Hyperactivity Disorder (ADHD-200), Autism Spectrum Disorder (ASD), Neuropsychiatric Phenomics (CNP) and Age prediction (OpenBHB). For each of these tasks and datasets where T1w, T2w or FLAIR modalities were present, we assessed performance of our model with that modality. Our MoE model is a single unified architecture capable of representing any (T1w, T2w, FLAIR) modalities, and we evaluated it via fine-tuning for each representative task/data/modality. Neuro-JEPA consistently outperformed three state-of-the-art neuroimaging foundation models (BrainIAC, VoCo, and NeuroVFM) across all settings (\cref{fig:overview,fig:fig2-best-unimodal}; Supplementary \cref{fig:sup-leftover-ap,fig:unimodal-all-auc}). The detailed averaged unimodal model performance across the tasks and datasets with corresponding 95\% confidence intervals is reported in Supplementary \cref{tab:unimodal_avg}.

On 41 dataset--task--modality combinations from public research-grade datasets, Neuro-JEPA improved mean AUROC by 4.4--6.4\% and mean AUPRC by 6.4--9.4\% over all baselines (all $p < 0.001$), outperforming other models by the majority of individual comparisons (AUROC: wins 32--35 out of 41 combinations; AUPRC: 31--37 out of 41 combinations). Advantages were substantially larger on internal institutional cohorts. On NYU Langone (30 combinations), Neuro-JEPA improved AUROC by 5.3--9.5\% and AUPRC by 11.4--16.1\%, with improvement on nearly all tasks (AUROC: wins 29--30 out of 30 combinations; AUPRC: wins 29--30 out of 30 combinations; all $p < 0.001$). Equivalent improvements were observed on NYU Long Island (30 combinations; AUROC: $+6.7$ to $+13.2$\% and wins 28--30 out of 30 combinations; AUPRC: $+8.8$ to $+17.1$\% and wins 27--30 out of 30 combinations; all $p < 0.001$). On BIND-MGH (45 combinations), Neuro-JEPA retained consistent advantages (AUROC: $+1.6$ to $+8.5$\%, wins 34--44 out of 45 combinations; AUPRC: $+1.6$ to $+7.4$\%, wins 29--45 out of 45 combinations), though improvements over NeuroVFM were more modest (AUROC $p<0.001$; AUPRC $p = 0.011$).

Across 6 time-to-event dataset--task--modality combinations, Neuro-JEPA achieved improved concordance indices (C-index) relative to BrainIAC ($+4.5$\%; 95\% CI: 2.7--6.3; $p < 0.001$; wins 6 out of 6 combinations), NeuroVFM ($+6.6$\%; 95\% CI: 2.7--12.2; $p <0.001$; wins 6 out of 6 combinations), and VoCo ($+3.1$\%; 95\% CI: 0.6--5.4; $p = 0.015$; wins 5 out of 6 combinations; Supplementary \cref{fig:sup-kaplan_meier}).

On the OpenBHB brain-age prediction benchmark ($n = 757$ for test set), Neuro-JEPA attained $R^{2} = 0.894$, $\mathrm{MAE} = 2.78$ years, and $\mathrm{RMSE} = 4.15$ years (Supplementary \cref{fig:sup-openbhb}). These metrics surpassed BrainIAC ($\Delta\mathrm{MAE} = -2.64$ years \footnote{$\Delta$ represents absolute difference}, 95\% CI: 2.37--2.93; $\Delta R^{2} = +0.372$, 95\% CI: 0.326--0.413; both $p<0.001$), NeuroVFM ($\Delta\mathrm{MAE} = -1.58$ years, 95\% CI: 1.35--1.75; $\Delta R^{2} = +0.221$, 95\% CI: 0.182--0.249; all $p<0.001$), and VoCo ($\Delta\mathrm{MAE} = -3.44$ years, 95\% CI: 2.90--3.92; $\Delta R^{2} = +0.783$, 95\% CI: 0.752--0.795; both $p<0.001$).

Collectively, these results demonstrate that Neuro-JEPA learns robust, transferable unimodal representations that generalize across diverse neuroimaging tasks, imaging modalities, and clinical populations.


\begin{figure}[t]
    \caption*{\textbf{\Cref{fig:fig3-multimodal} (continued).}
    We report AUROC for paired multimodal inputs and the corresponding multimodal gain, defined as the difference between the best multimodal fusion result and the best unimodal result. For fair comparison, unimodal baselines in this analysis were re-trained on the subset of cases with complete modalities availability on multimodal counterparts and Neuro-JEPA-B is used in the evaluation. Additional AUPRC results and comparisons with other evaluated models are provided in Supplementary \Cref{fig:sup-mm-auprc,fig:sup-mm-brainiac-voco,fig:sup-mm-bind-gain-neuro-jepa,fig:sup-mm-bind-gain-neurovfm,fig:sup-mm-bind-gain-brainiac,fig:sup-mm-bind-gain-voco}.
    \textbf{a,d,} AUROC of multimodal fusion across selected public datasets and BIND-MGH for four foundation models. For each model, performance is reported using the best fusion strategy among five evaluated methods. Dotted lines indicate mean performance across tasks, showing that Neuro-JEPA outperforms competing foundation models by a substantial margin.
    \textbf{b,c,e,f,} AUROC gain from multimodal fusion for NeuroVFM, the best previous foundation model, and Neuro-JEPA on public datasets and BIND-MGH. Neuro-JEPA yields overall higher and more consistent gains.}
\end{figure}

\subsection*{Multi-Modal Learning Capability}
\label{sec:multimodal}
Multimodal integration is known to yield nuanced benefits: prior work has identified cases in which multimodal fusion provides marginal gain over unimodal baselines~\cite{asadi2026mirage, Wang20,Remi19} and in which performance is highly sensitive to the selected fusion strategy~\cite{Maciej23,liang2024foundationsmultisensoryartificialintelligence}, reflecting differences in how models organize modality-specific representations in latent space. Despite these well-characterized challenges, existing multimodal foundation models have not been systematically evaluated for their multimodal learning capacity. We therefore conducted a comprehensive benchmark spanning five fusion strategies and diverse modality combinations across the evaluated models (\cref{fig:fig3-multimodal}; Supplementary \cref{fig:mm_methods_perf_fm,fig:mm_methods_perf_fm_cont,fig:mm_methods_perf_fm_neurovfm,fig:sup-mm-auprc,fig:sup-mm-brainiac-voco}), with mean multimodal performance averaged across tasks and datasets with corresponding 95\% confidence intervals reported in Supplementary \cref{tab:multimodal_avg}.

We assess multimodal capability using two complementary criteria. First, a robust multimodal model should achieve consistently strong performance across tasks under best performing multimodal fusion method. Second, effective multimodal learning should yield positive cross-modal transfer, whereby joint modeling improves performance relative to unimodal baselines (defined as the difference between best achievable multimodal and unimodal performance). For multimodal combination selection, we designate T1w as the structural anchor modality and assess multimodal learning by incorporating complementary contrasts from T2w and FLAIR sequences. This design yields two targeted experimental settings (T1w+T2w and T1w+FLAIR), enabling systematic evaluation of how distinct and clinically relevant signal complements contribute to representation learning. Across both criteria, Neuro-JEPA demonstrates improved performance compared with existing approaches (as demonstrated in \cref{fig:fig3-multimodal} and Supplementary \cref{fig:mm_methods_perf_fm,fig:mm_methods_perf_fm_cont,fig:mm_methods_perf_fm_neurovfm,fig:sup-mm-auprc,fig:sup-mm-brainiac-voco}), indicating that it captures complementary information across modalities in a stable and generalizable manner. Notably, Neuro-JEPA demonstrates consistent positive transfer across nearly all evaluated tasks in the clinical cohorts from the BIND-MGH dataset, highlighting its strong multimodal capability when deployed in real-world clinical settings.

On public-dataset multimodal tasks (12~combinations), Neuro-JEPA outperformed all baselines. Relative to BrainIAC, Neuro-JEPA improved AUROC by 7.6\% (95\% CI: 4.7--10.2; $p < 0.001$;  wins 12 out of 12~combinations) and AUPRC by 8.5\% (95\% CI: 4.9--12.0; $p = 0.0015$; wins 11 out of 12~combinations). Relative to NeuroVFM, improvements were 5.8\% on AUROC (95\% CI: 2.9--9.4; $p = 0.014$; wins 9 out of 12~combinations) and 6.2\% on AUPRC (95\% CI: 0.7--10.3; $p = 0.005$; wins 9 out of 12~combinations). Relative to VoCo, Neuro-JEPA improved AUROC by 6.2\% (95\% CI: 3.8--8.0; $p < 0.001$; wins 12 out of 12~combinations) and AUPRC by 7.5\% (95\% CI: 3.3--10.6; $p < 0.001$; wins 9 out of 12~combinations).

Consistent advantages were observed on BIND-MGH multimodal tasks (30~combinations). Relative to BrainIAC, Neuro-JEPA improved AUROC by 7.8\% (95\% CI: 6.5--9.4; $p < 0.001$; wins 28 out of 30~combinations) and AUPRC by 9.2\% (95\% CI: 7.1--11.5; $p < 0.001$; wins 30 out of 30~combinations). Relative to NeuroVFM, improvements were 2.1\% on AUROC (95\% CI: 1.2--2.8; $p < 0.001$; wins 25 out of 30~combinations) and 3.9\% on AUPRC (95\% CI: 2.0--6.6; $p < 0.001$; wins 24 out of 30~combinations). Relative to VoCo, Neuro-JEPA improved AUROC by 3.4\% (95\% CI: 2.7--4.3; $p < 0.001$; 28/30~combinations) and AUPRC by 5.3\% (95\% CI: 3.5--7.8; $p < 0.001$; wins 29 out of 30~combinations).

Collectively, the results confirmed that Neuro-JEPA excels in both multimodal evaluation criteria across all tested configurations. It achieves superior task performance under optimal fusion and consistently exhibits better positive multi-modal transfer relative to unimodal baselines in comparison to other evaluated models. 

\begin{figure}[htbp]
    \centering
    \includegraphics[width=0.9\textwidth]{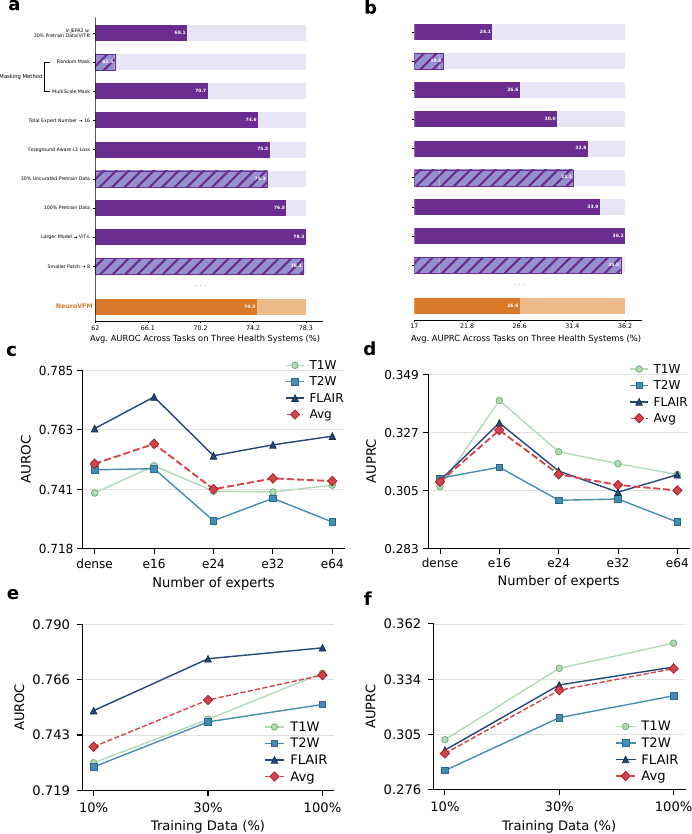}
    \caption{\textbf{Ablation of Model Design and Scaling} - AUROC and AUPRC are averaged across all tasks and modalities from three health systems with attentive probing: NYU Langone, NYU Long Island, and MGH. \textbf{a,b,} Stepwise ablation of model design \cite{convnext}, in which each modification is introduced sequentially from the original V-JEPA2 implementation to the final largest scale Neuro-JEPA model, showing each design choice brings meaningful contribution to overall performance. Hatched bar means the design choice is not applied. \textbf{c,d,} Ablation of the total number of experts, performed with all other design components applied and 30\% of pretrain data, showing that increasing the number of experts to 16 improves both AUROC and AUPRC relative to the dense model, whereas further expert increases yield diminishing returns. \textbf{e,f,} Ablation of pretraining data scale, showing that performance continues to improve as the proportion of pretraining data increases up to the full dataset.}
    \label{fig:fig4-ablation}
\end{figure}

\subsection*{Components for Effective Pre-training}
\label{sec:opt_pretrain}
We identified three key components improving the generalizability of JEPA pre-training in neuroimaging. First, masking strategy determines the quality of the predictive learning signal. Second, sparse computation via Mixture-of-Experts (MoE) architectures enhances the model's capacity to disentangle heterogeneous anatomical tokens by routing information through different pathways. Third, suppressing learning signals from background regions improves robustness, as 40--60\% of voxels in skull-stripped neuroimages are non-informative. Full implementation details are provided in \hyperref[sec:model_arch]{Model Architecture} and Supplementary~\cref{apd:alg_details}.

To quantify the contribution of each component, we performed controlled ablation studies using the V-JEPA~2 pre-training framework as baseline, incrementally introducing: (1)~multi-scale masking, (2)~multi-scale masking with MoE, and (3)~multi-scale masking with MoE and a foreground-aware $L_1$ loss to downweight background voxels (down-weighting ratio $\beta=0.1$). Each modification yielded consistent performance gains on attentive probing averaged across evaluated datasets on NYU Langone, NYU Long Island and Massachusetts General Hospital (\cref{fig:fig4-ablation}). Multi-scale masking alone improved mean AUROC by 1.5\% and AUPRC by 2.5\% ; adding MoE contributed a further 3.9\% (AUROC) and 3.4\% (AUPRC); incorporating the foreground-aware $L_1$ loss added 0.9\% (AUROC) and 2.8\% (AUPRC). More details on per dataset and modality performance on 30\% pretrain data are present in Supplementary \Cref{fig:sup-experts-abla}.



\subsection*{Label efficiency Under Few Shots}
\label{sec:label_eff}
\begin{figure}[htbp]
    \centering
    \includegraphics[width=\textwidth]{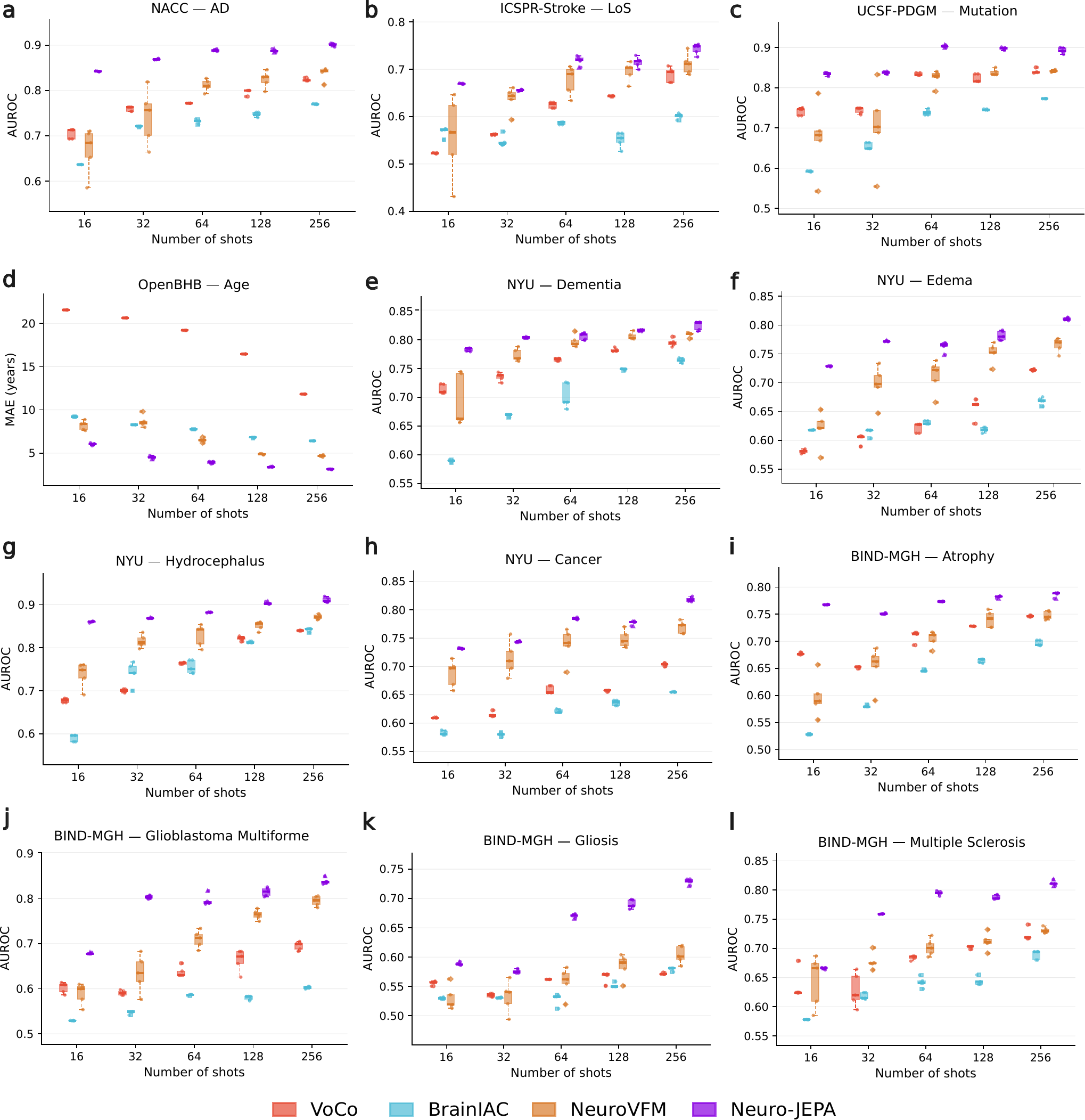}
    \caption{\textbf{Few-shot Analysis} - we examine the evaluated models label efficiency when only $k=\{16,32,64,128,256\}$ positive samples are provided on training. Neuro-JEPA-B is used in the evaluation. The performance in reported in AUROC for classification and MAE for regression. The result demonstrates that our model performs better than evaluated models by a large margin under majority of environment with limited labeled data. \textbf{a-d,} Few-shot performance on selected tasks from public datasets. All result is reported as averaged performance across all available modalities for each task \textbf{e-h,} Few-shot performance on selected tasks from NYU-Langone dataset. \textbf{i-l,} Few-shot performance on selected tasks from BIND-MGH dataset.}
    \label{fig:fig5-few-shot-analysis}
\end{figure}

We evaluate model label efficiency under few-shot conditions in \Cref{fig:fig5-few-shot-analysis}. The result is evaluated with $k=\{16,32,64,128,256\}$ and with fine-tuning on attentive layers, where $k$ is the number of positive samples used for classification and number of samples used for each quartile of the original age distribution for age prediction. The result is presented on four selected tasks on public datasets, NYU-Langone and BIND-MGH with AUROC average across all modalities. More diverse tasks evaluation on AUROC, AUPRC and per modality performance can be found in Supplementary \Cref{fig:sup-few-shot-avg-auroc,fig:sup-few-shot-avg-auprc,fig:sup-few-shot-permod-t1w-auroc,fig:sup-few-shot-permod-t2w-auroc,fig:sup-few-shot-permod-flair-auroc,fig:sup-few-shot-permod-t1w-auprc,fig:sup-few-shot-permod-t2w-auprc,fig:sup-few-shot-permod-flair-auprc}. The results demonstrate that Neuro-JEPA achieves superior label efficiency, consistent across all evaluated tasks and particularly for tasks derived from clinical datasets such as NYU-Langone and BIND-MGH. Across different values of k, several tasks showed improvements exceeding 5\% in both AUROC and AUPRC, including gliosis, multiple sclerosis, and hematomas. The label-efficiency gains were especially pronounced in tasks for which performance approached saturation under full-data training. For example, on NACC AD, Neuro-JEPA achieved approximately 15\% higher AUROC and AUPRC across different k values, despite all compared models attaining close performance when trained on full samples.

\subsection*{Cross-cohort and out-of-distribution modality generalization}
We assessed whether Neuro-JEPA generalizes across independent clinical cohorts and neuroimaging modalities. For cross-cohort evaluation, models were fine-tuned on a source cohort and evaluated without further adaptation on an external target cohort with matched label definitions. This setting approximates direct cross-institutional deployment, in which a model optimized at one institution is applied to data from another. As shown in Supplementary \Cref{fig:sup-ood_transfer}, Neuro-JEPA maintained stable performance across cohort transfers, including NACC-to-ADNI for Alzheimer’s disease and amyloid status prediction, and MGH-to-NYU for hematoma prediction, with no substantial degradation in majority of external test performance.

We further evaluated whether Neuro-JEPA can generalize to neuroimaging modalities not observed during pretraining. Specifically, we fine-tuned and evaluated the model on diffusion-weighted MRI (DWI), which was absent from the pretraining data. Experiments were conducted on 90-day modified Rankin Scale (mRS) prediction, lesion type classification, and length-of-stay prediction in ICSPR-Stroke, as well as IDH mutation prediction in UCSF-PDGM. Despite this modality shift, Neuro-JEPA remained the best-performing model among all evaluated baselines, achieving average improvements of 1.7\% in AUROC and 1.8\% in AUPRC (Supplementary \Cref{fig:sup-ood_modalities,tab:dwi_comparison}). These findings indicate that Neuro-JEPA learns representations that transfer robustly across both clinical institutions and previously unseen neuroimaging modalities.

\subsection*{Scalability Under Pretraining}
\label{sec:scalability}



Scalability is the central principle in the development of foundation models. We evaluate the scalability of our framework on both data and model size. Increasing the pretraining dataset size leads to consistent improvement in downstream performance, with gains of 1.3\% in AUROC and 1.1\% in AUPRC across evaluation benchmarks shown in \Cref{fig:fig4-ablation}, demonstrating effective data scaling. More details on performance change under increased data size can be found in Supplementary \Cref{fig:sup-percentabla_perdataset}. Same as data size scaling, we present study on model size scaling in Supplementary \Cref{fig:sup-modelsize_abla_clinical} and \Cref{fig:sup-modelsize_abla_public}, where increasing model size from ViT-B-MoE (121 million total parameters) to ViT-L-MoE (429 million total parameters) effectively improved model performance on both health-system and public datasets with averaged $1.5\%$ improvement on AUROC and $2.3\%$ improvement on AUPRC for clinical
datasets and averaged $1.9\%$ improvement on AUROC and $4.1\%$ improvement on AUPRC for public datasets.

\subsection*{Comparison with a Simple CNN Baseline}
\label{sec:cnn_baseline}
To determine whether foundation-model pretraining provides practical benefit beyond simple supervised learning from scratch, we compared each pretrained foundation model with a neuroimaging-specific CNN trained from scratch for the downstream task~\cite{Liu2022}. Across 41 task--modality combinations from 12 public datasets, existing foundation models did not consistently outperform this simple CNN baseline, whereas Neuro-JEPA achieved consistent gains, improving average AUROC by 3.7\% and AUPRC by 4.5\% (Supplementary \Cref{tab:simple_cnn_compare}; Supplementary \Cref{fig:simple_cnn_compare}). Neuro-JEPA was also the only foundation model to improve over the CNN baseline for age prediction on quasi-raw scans, increasing $R^2$ by $+2.8$ and reducing MAE and RMSE by $-0.37$ and $-0.50$, respectively. These results emphasize that claims of foundation-model superiority should be supported by comprehensive evaluations against simple, task-specific conventional baselines, rather than comparisons limited to other pretrained models.

\subsection*{MoE Routing Analysis and Visualization}
\label{sec:moe_routing}
\begin{figure}[htbp]
    \centering
    \includegraphics[width=\textwidth]{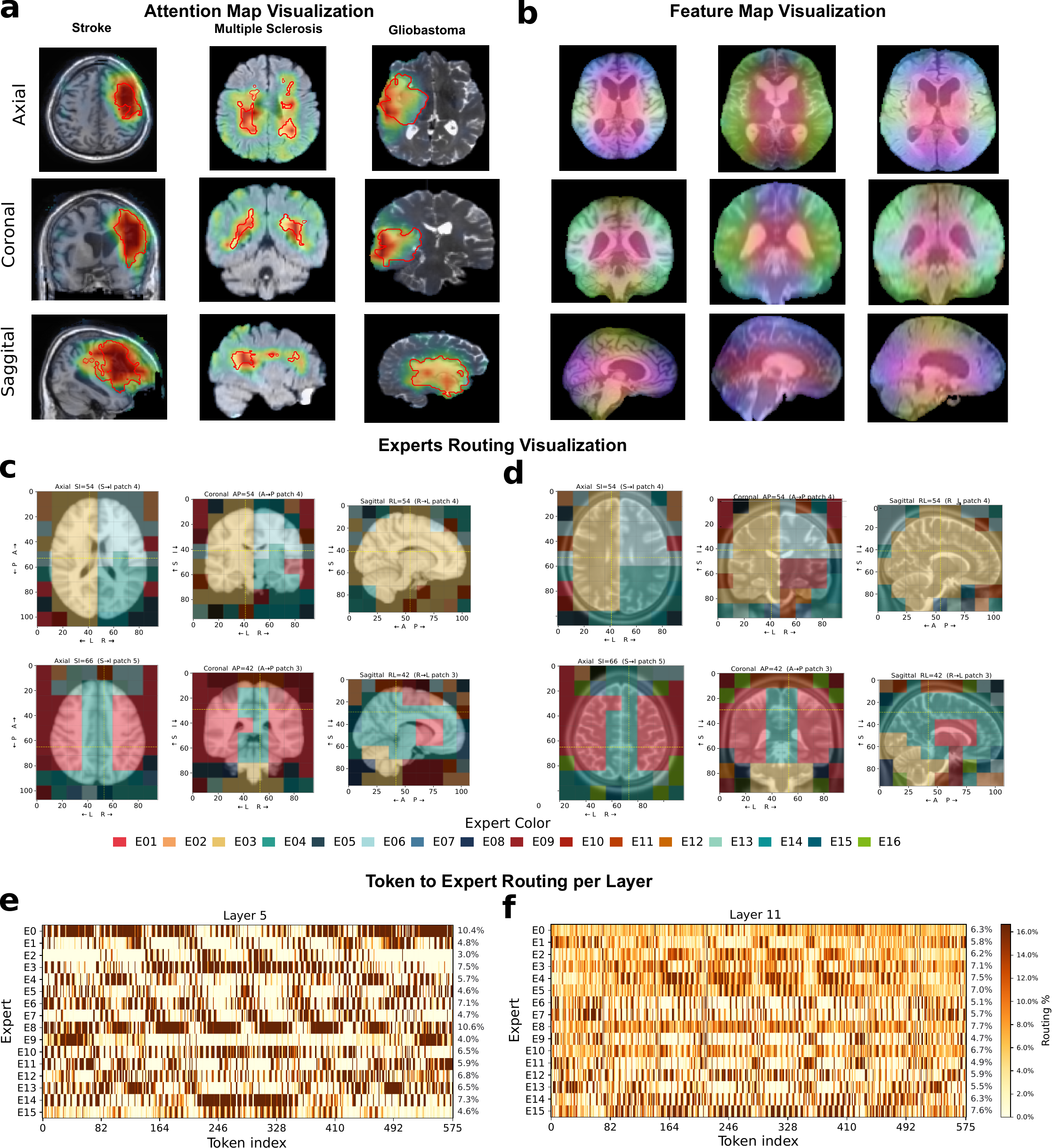}
    \caption{\textbf{MoE Routing Analysis and Visualization} - we explore visualization with attention map and feature map from Principal Components Analysis (PCA) to identify if Neuro-JEPA learns meaningful brain structure from pre-training. Then, we explore the MoE routing behavior by examining if different experts are routing to different regions. \textbf{a,} Attention map visualization on lesions of Stroke, Multiple Sclerosis and Gliobastoma, where the red circles indicate ground truth lesion masks \textbf{b,} Feature map on selected of pre-training data \textbf{c,d,} Routing regions on different experts \textbf{e,f,} Token routing frequency on different experts from different token indices.}
    \label{fig:fig6-moe-analysis}
\end{figure}

To evaluate the model's capacity for zero-shot pathological localization, we extracted spatial attention maps by querying patches within a selected lesion-containing region (\Cref{fig:fig6-moe-analysis}a). The final maps were computed by weighted averaging (higher weights if the selected patch contain more lesion voxels) the attention scores across these patches. The attention map evaluation is applied to scans exhibiting stroke, multiple sclerosis, and glioblastoma. The samples are selected from datasets with ground truth lesion masks on ATLAS \cite{Liew2022}, MSLesSeg \cite{Guarnera2025} and UCSF-PDGM \cite{calabrese2022ucsf_pdgm}. The results demonstrate that strong neuroimaging foundation model intrinsically learned the representation to localizes these diverse pathologies without requiring any disease-specific fine-tuning. Additional visualization result for attention map is present in Supplementary \Cref{fig:sup-att_vis-stroke,fig:sup-att_vis-ms,fig:sup-att_vis-glio}.

We analyzed representation feature maps in \Cref{fig:fig6-moe-analysis}b by smoothing stacked first three principal components from principal component analysis (PCA). The visualization reveals that representations inherently preserve distinct brain anatomical structures. Additional visualization result for feature map before smoothing is present in Supplementary \Cref{fig:sup-feature-map}. 

Different tokens routing visualization (Figures \ref{fig:fig6-moe-analysis}c-d) and heatmaps of token-level routing distributions (Figures \ref{fig:fig6-moe-analysis}e-f); token vs. expert index) confirm that individual experts preferentially route specific token subsets. Projecting averaged routing distributions onto T1w and T2w templates for representative NYU-Langone samples (Figures \ref{fig:fig6-moe-analysis}i–j) demonstrates clear spatial segregation, confirming that MoE mechanisms effectively disentangle regional processing by partitioning experts across anatomical structures (see Supplementary Figures \ref{fig:sup-moe-vis-t1w}–\ref{fig:sup-moe-vis-t2w-ps8} for extended examples).

Extended analyses further demonstrated foreground–background and cross-modal routing behavior. Most experts displayed pronounced routing imbalances between foreground and background tokens, highlighting strong functional separation (Supplementary Figures \ref{sup-moe-routing-fg-bg-1}–\ref{sup-moe-routing-fg-bg-2}). Conversely, within foreground regions across T1w, T2w, and FLAIR sequences, experts exhibited largely uniform routing (Supplementary Figures \ref{sup-moe-modalities-nyu}–\ref{sup-moe-modalities-nyu-l}). This contrasts with previous reports of rigid modality specialization for different experts \cite{tong2026beyond,mustafa2022multimodal}, suggesting instead that the learned experts from Neuro-JEPA leverage high cross-modal anatomical redundancy to disentangle information processing on different shared structural regions.

\section*{Discussion}

Modern neurological diagnosis depends on multimodal neuroimaging, which captures complementary biological information needed for clinical decision-making. Although recent multimodal neuroimaging foundation models have shown promise for integrating these signals \cite{kondepudi2025healthlearningachievesgeneralist,Lyu2026}, the effects of pretraining design and the robustness of multimodal representations remain insufficiently understood. Here, we introduce \textit{Neuro-JEPA}, a vision foundation model pretrained on approximately 1.55 million carefully curated brain MRI scans. Across clinical cohorts from three major health systems and research cohorts spanning 12 public datasets, Neuro-JEPA showed strong and consistent performance across unimodal and multimodal tasks, with robust generalization across heterogeneous imaging settings.

Building on the joint-embedding predictive architecture (JEPA), our results show that data scale, curation and model design can act synergistically to improve multimodal representation learning. Joint multimodal pretraining improved not only multimodal fusion but also the quality of unimodal representations, suggesting that learning across complementary imaging sequences yields features that transfer broadly across downstream tasks. Scaling and training-trajectory analyses further showed that Mixture-of-Experts architectures can remain stable during large-scale pretraining. Together with advances in multimodal learning, general-purpose architectures and scalable foundation-model training \cite{tong2026beyond,team2023gemini,dai2025qoqmed,zhou2025transfusion,bai2025qwen3vltechnicalreport,Shukor_2025_ICCV}, these findings support continued efforts to combine larger and more diverse datasets with principled architectural and algorithmic development.

The generalist nature of Neuro-JEPA is particularly relevant to clinical neuroimaging, where radiologists must simultaneously evaluate numerous pathologies, anatomical variants and technical artifacts. Building separate specialist models for each potential finding is difficult to scale, requires substantial computational and deployment infrastructure, and is poorly suited to the long-tailed distribution of neurological abnormalities encountered in practice. By contrast, a single pretrained model that transfers effectively across tasks could provide a common foundation for multiple diagnostic and prognostic applications. This is especially important in neuroimaging, where examination volumes continue to increase and interpretation often requires specialized expertise, including for time-sensitive and high-acuity conditions.

Neuro-JEPA also showed complementary strengths across major neurological disease domains. In neurodegenerative disease, it captured clinically meaningful imaging phenotypes associated with amyloid status, progression from mild cognitive impairment to Alzheimer’s disease dementia and Parkinson’s disease, supporting the potential use of routine MRI for opportunistic detection and longitudinal risk stratification. In stroke, Neuro-JEPA performed strongly in disease identification and classification and also predicted 90-day functional outcome and length of stay, extending its utility from diagnostic recognition to prognosis and care planning. Collectively, these results suggest that broadly transferable neuroimaging representations can support clinically distinct tasks across disease categories without requiring a separately pretrained model for each application.

Several limitations define important directions for future work. Neuro-JEPA was pretrained on three widely used MRI sequences—T1-weighted, T2-weighted and FLAIR—to enable controlled evaluation of multimodal learning. Clinical neuroimaging, however, also relies on diffusion-weighted imaging, susceptibility-weighted imaging, contrast-enhanced MRI, functional MRI and other modalities such as CT, each of which provides complementary pathophysiological information. Incorporating these additional inputs may further expand the clinical scope of general-purpose neuroimaging models. Our experiments also used down-sampled scans and fixed patch sizes to enable systematic ablations under computational constraints. Future work should evaluate higher spatial resolution and alternative patching strategies, which may be particularly important for small or fine-grained abnormalities \cite{oquab2024dinov,assran2025vjepa2}. 


Finally, the differences observed between internal and external evaluations highlight the importance of training on broader multi-institutional data. Neuro-JEPA outperformed NeuroVFM on both cohorts, but the margin was substantially larger on the internal NYU cohort than on the external MGH cohort, underscoring the limitations of single-system pretraining and the need for models exposed to greater institutional, demographic and acquisition heterogeneity.

In summary, Neuro-JEPA provides a scalable framework for multimodal brain MRI representation learning and a systematic analysis of how data scale, architecture and pretraining objectives influence downstream performance. Beyond the specific model introduced here, our findings support a broader shift from narrowly specialized imaging algorithms toward unified neuroimaging foundation models that can transfer across modalities, diseases and clinical tasks. Continued expansion to more diverse imaging sequences, institutions and patient populations will be essential to determine whether such models can ultimately support robust, generalizable and clinically useful AI systems for neurological imaging.

\section*{Methods}
\subsection*{Datasets}
\subsubsection*{Data Curation}
Registered MRI scans are used for all cohorts, where all brain MRI scans used for pretraining and downstream evaluation were affine registered to the MNI152 standard space and resampled to an isotropic resolution of $1.0 \times 1.0 \times 1.0~\mathrm{mm}^3$ using FSL \cite{JENKINSON2012782}. T1w and T2w sequences were registered to their corresponding MNI152 templates, whereas FLAIR sequence were registered to the MNI152 T2w template because of their closer contrast characteristics. We further applied bias-field correction and skull stripping with SynthStrip \cite{hoopes2022synthstrip}, which also facilitated robust anonymization by removing non-brain tissue. Similar registration strategies have been used in recent brain MRI modeling studies \cite{Xue2024,Tak2026}.

Clinical MRI cohorts acquired directly from hospital environments often contain artifacts that degrade image quality, including severe noise, signal dropout, and partial or near-complete loss of brain tissue, as illustrated in Supplementary \Cref{fig:noise_samples}. These artifacts are primarily attributable to patient-related factors, such as motion during acquisition. Although large-scale foundation models are often expected to acquire robustness from heterogeneous data, consistent with observations in other domains on the role of data quality in foundation model training \cite{NEURIPS2022_7b75da9b,refineweb,xu2024demystifying}, our empirical analysis indicated that severely degraded scans can compromise pretraining efficacy. Specifically, we conducted an ablation study whereby  $25{,}000$ previously excluded low-quality scans were reintroduced into a $30\%$ subset of the pretraining data, corresponding to approximately $5\%$ of the full pretraining corpus, reducing model performance as shown in \Cref{fig:fig4-ablation}a,b.

To mitigate the effects of severe image degradation, we implemented a quantitative quality-control pipeline after image registration. For each registered scan, we computed Mutual Information (MI), Peak Signal-to-Noise Ratio (PSNR), and Pearson correlation with the corresponding MNI152 reference template. Scans were excluded if they satisfied any of the following modality-specific criteria: for T1w, $\mathrm{MI}<0.30$, $\mathrm{PSNR}<10$, or Pearson correlation $<0.85$; for T2w, $\mathrm{MI}<0.30$, $\mathrm{PSNR}<11$, or Pearson correlation $<0.65$; and for FLAIR, $\mathrm{MI}<0.25$, $\mathrm{PSNR}<11$, or Pearson correlation $<0.65$. These thresholds were selected by manually reviewing the lowest-quality scans within each modality and identifying values that removed images with severe tissue loss or prohibitive noise while retaining scans with only mild quality degradation. This curation procedure reduced the pretraining data from $1{,}639{,}685$ to $1{,}551{,}862$ scans, corresponding to the removal of approximately $5\%$ of the data. We further confirmed that this data-curation step improved model performance, as shown in \Cref{fig:fig4-ablation}a,b and Supplementary \Cref{fig:sup-noiseabla_perdataset}.

\subsubsection*{Data Split}
We split all datasets at the patient level such that the training, validation, and test sets contain non-overlapping individuals. Although some previous studies \cite{kondepudi2025healthlearningachievesgeneralist,Lyu2026} have adopted time-based splits, reserving future time points for evaluation, this design still permits the same patient to appear in both training and evaluation sets, thereby introducing substantial risk of data leakage and overly optimistic performance. We therefore adopted patient-level splitting to provide a more rigorous assessment of representation generalizability. 

This choice is motivated by the following considerations. Medical imaging models are well-known to exploit spurious correlations or direct data leakage at the patient level \cite{Apicella2025,pmlr-v219-compton23a,WEN2020101694,DeGrave2021} and high-dimensional 3D scans often contain highly individual-specific anatomical patterns that can function as biometric-like signatures. For example, when the same patient appears in both training and evaluation sets, it is known that models can rely on patient identity instead of disease-relevant imaging features \cite{Samala21,ChaibubNeto2019} to perform diagnostic tasks. This issue can easily inflate model performance even without robust representation learning on disease related features. Additionally, patient-level splitting ensures a consistent evaluation framework across all datasets, where the temporal information for the public datasets is often intentionally anonymized by date shifting such as BIND-MGH and several clinical trial cohorts in our evaluation benchmark.

\subsubsection*{Pretraining Dataset}
The pretraining dataset is collected from NYU Langone Picture Archiving and Communication System (PACS) system. After data curation, the pretrain dataset consists of 1,551,862 scans across 282,693 patients and 428,647 studies with all scans performed between 2009 and 2025. We partitioned the full dataset by the patient IDs into training and held-out downstream evaluation sets to avoid the leakage of scans from the sample patients. This yielded 88,314 (35,183 unique patients) samples for held-out downstream evaluation, although the model was only pretrained on the 1,551,862 scans from the training set.

\subsubsection*{Downstream Datasets}
For unimodal experiments, we leveraged the full set of available samples for each dataset and task. For multimodal evaluation, we restricted analyses to the subset of samples with all modalities (T1, T2, FLAIR) availability. This design ensures balanced contributions from each modality during both training and evaluation, enabling a controlled and unbiased assessment of the difference between unimodal and multimodal performance. By standardizing modality availability, we minimize confounding effects that could otherwise arise from unequal sample distributions across modalities and bias comparative analyses. 

Our downstream benchmark spans a diverse set of clinical and research cohorts, including two internal hospital systems from distinct sites (NYU Langone and NYU Long Island), one geographically distinct external institution (Massachusetts General Hospital), and 12 high-quality public research datasets. Together, this collection represents, to our knowledge, one of the most comprehensive and heterogeneous benchmarks in the neuroimaging foundation model literature for evaluation of model generalizability across clinical sites, research cohorts, and imaging contexts. Detailed descriptions of dataset composition and downstream tasks are provided below. Multimodal evaluation was performed for datasets with sufficient overlap in samples across modalities, enabling controlled assessment of multimodal representation learning. Full dataset statistics are provided in Supplementary \Cref{tab:benchmark_stats}.

\paragraph{Health System Datasets}
\paragraph{NYU Langone Hopsitals Manhattan and Brooklyn --- 10 tasks} We used data from the NYU Langone main campus (Manhattan and Brooklyn) for internal in-domain (ID) evaluation (abbreviated as NYU Langone). Disease labels were derived from electronic health records (EHR) within a 3-month window centered on the imaging date, based on ICD-10 diagnostic codes and medication records (Supplementary \Cref{tab:disease_definition}). The dataset comprises 19,325 T1w scans from 10,004 patients, 29,237 T2w scans from 19,132 patients, and 23,302 FLAIR scans from 15,937 patients. Data are split at the patient level into training, validation, and test sets (60/20/20). We evaluated 10 clinical conditions. The conditions we examined include Cancer, Edema, Dementia, Major Depressive Disorder, Hydrocephalus (HCP), Intraparenchymal Hemorrhage (IPH), Intraventricular Hemorrhage (IVH), Subdural Hemorrhage (SDH), Subarachnoid Hemorrhage (SAH), Intracerebral Hemorrhage (ICH)

\paragraph{NYU Langone Hospital Long Island --- 10 tasks} We use data from NYU Langone Hospital Long Island as an internal out-of-domain (OOD) evaluation cohort (abbreviated as NYU Long Island), with labels defined identically to NYU Langone. The dataset includes 8,024 T1w scans from 3,928 patients, 5,050 T2w scans from 3,206 patients, and 3,376 FLAIR scans from 3,065 patients. Patient-level splits follow the same 60/20/20 protocol.

\paragraph{Massachusett General Hospital (MGH) --- 15 tasks} 
We evaluate external generalization using a curated subset of data from the Massachusetts General Hospital \cite{Maschke2025.10.01.25337054}. Disease labels are extracted from radiology reports using a large language model (Bio-Medical-Llama-3-8B \cite{ContactDoctor_Bio-Medical-Llama-3-8B}), achieving $>90\%$ accuracy as verified by expert manual annotation on a subset. To ensure fair comparisons, we construct balanced subsets with equal sample sizes across modalities for both unimodal and multimodal evaluations. The dataset includes 22,142 scans from 11,802 patients, with patient-level splits of 60/20/20. Fifteen tasks are selected based on label prevalence spanning tumor, vascular, inflammatory, and degenerative conditions. The tasks included Astrocytoma, Atrophy, Cyst, Edema, Enhancement, Hematoma, Infarct, Ischemic, Mass Effect, Midline Shift, Multiple Sclerosis, Schwannoma, White Matter Changes, Glioblastoma Multiforme, Gliosis.
\paragraph{Public Research Datasets}
\label{sec:pub_dataset}
\paragraph{ABIDE --- 1 task} The Autism Brain Imaging Data Exchange (ABIDE) \cite{DiMartino2014} aggregates multi-site neuroimaging data across 17 international cohorts. We evaluate binary classification of autism versus Healthy Controls using T1w MRI with 1,099 patients and equal number of samples as patients. Patient-level split is performed on 60/20/20.

\paragraph{ADHD-200 --- 1 task} The ADHD-200 \cite{Bellec2016-ml} dataset comprises structural MRI from eight international imaging  centers. We evaluated ADHD versus Healthy Control classification using T1w scans with 776 patients and equal number of samples as patients. Patient-level split is performed on 60/20/20.

\paragraph{ADNI --- 3 tasks} The Alzheimer’s Disease Neuroimaging Initiative (ADNI) \cite{Petersen2009-pj} is a multi-center longitudinal study across North America. We evaluated (1) Amyloid positivity (ADNI4 dataset), (2) Alzheimer's disease (AD) versus Healthy Control classification (ADNI1 standardized subset is used) and (3) Mild Cognitive Impairment (MCI) to AD conversion as a time-to-event task. For amyloid detection with amyloid positivity determined from PET imaging, The dataset include 167 patients for T1w, 205 patients for T2w and 317 patients for FLAIR. Equal numbers of samples as subjects are presented for amyloid detection. For AD classification, 1632 T1w scans from 455 patients is used. For MCI to AD conversion, 209 T1w scans from 209 patients is used. Data are split at the patient level with 60/20/20.

\paragraph{ICSPR-Stroke --- 3 tasks} The Annotated Clinical MRIs of Patients with Acute Stroke dataset (ICSPR-Stroke) \cite{Liu2023} is a high-density, longitudinal repository of 2,888 multimodal MRIs collected at a National Stroke Center. We define three tasks from this dataset. (1) 90-day functional outcome with modified Rankin Scale (mRS) as binary classification (0-2 mRS as 0 and 3-6 mRS as 1);(2) lesion type classification (ischemic, hemorrhagic or absent); and (3) length of stay at abinary classification threshold of 8 days. Because every task has missing labels and modalities, each presents different numbers of patients and samples as detailed in Supplementary \Cref{tab:benchmark_stats}. Only T1w and FLAIR are used for multimodal evaluation. The data are split by 60/20/20 at the patient level.

\paragraph{MCSA --- 4 tasks} The Mayo Clinic Study of Aging (MCSA) \cite{Roberts2008-ml} is an ongoing, longitudinal population study of residents in Olmsted County, Minnesota, dedicated to mapping the trajectories of cognitive impairment and successful aging. We framed a total of 4 tasks based on the availability of diagnosis labels from this dataset. This includes tasks are diseases vs. Healthy Control with diseases to be AD, stroke, hypertension and dyslipidemia. The dataset includes 2,873 scans from 1,715 patients in total with T1w and FLAIR subsets for unimodal and multimodal analyses. The data are split by 60/20/20 at the patient level.

\paragraph{NACC --- 2 tasks} The National Alzheimer's Coordinating Center (NACC) \cite{Beekly2007-iw} dataset is among the world's largest and most comprehensive longitudinal repositories for Alzheimer’s disease and related dementias (AD/ADRD). NACC aggregates and standardizes multimodal data from over 40 Alzheimer's Disease Research Centers (ADRCs) across the United States. We evaluate Alzheimer’s disease versus control and amyloid status prediction across modalities. For AD vs. Healthy Control under unimodal evaluation, it presents 4,994 total samples from 3,841 patients for T1w, 3,062 total samples from 2,538 patients for T2w and 3,755 samples from 3,024 patients for FLAIR. For Amyloid vs. Non-Amyloid PET positivity classification task, it presents 182 samples from 176 patients for T1w, 97 samples from 92 patients for T2w, 159 samples from 155 patients for FLAIR.  For multimodal evaluation, it presents 3,772 total samples from 3,132 patients for T1w and T2w. The data are split by 60/20/20 at the patient level.

\paragraph{OASIS3 --- 1 task} The Open Access Series of Imaging Studies (OASIS3) \cite{Marcus2010-qk} is a open-science repository providing three decades of longitudinal clinical, cognitive, and multimodal neuroimaging data. The data is collected from Washington University in St. Louis (Knight Alzheimer Disease Research Center). Our main focus was on classifying AD vs. Healthy Control for this dataset. For unimodal evaluation, it presents 1,924 total samples from 1,126 patients for T1w, 1,665 total samples from 1,004 patients for T2w and 1,028 samples from 727 patients for FLAIR. For multimodal evaluation, it presents 1,384 samples from 883 patients for T1w and T2w. The data are split by 60/20/20 at the patient level.

\paragraph{PPMI --- 2 task} The Parkinson's Progression Markers Initiative (PPMI) \cite{Parkinson_Progression_Marker_Initiative2011-fz} is a international observational study designed to identify and validate robust clinical, imaging, genetic, and biochemical biomarkers of Parkinson’s disease (PD) progression with data collected from nearly 50 international sites. Our focus was on (1) discriminating Parkinson's vs. Prodromal vs. Healthy Controls for this dataset (2) determining conversion from Prodromal to Parkinson as a time-to-event task. For Parkinson diagnosis, under unimodal evaluation, it presents 1,763 total samples from 1,000 patients for T1w, 1,296 samples from 346 patients for T2w and 2,365 samples from 1,577 patients for FLAIR. Under multimodal evaluation, it presents 1,339 samples from 302 patients for T1w and T2w. For Prodromal to Parkinson conversion, it presents 397 total samples from 397 patients for T1w, 882 samples from 882 patients for FLAIR. The data are split by 60/20/20 at the patient level.

\paragraph{SOOP --- 1 tasks} The Stroke Outcome Optimization Project (SOOP) \cite{Absher2024} dataset is a large-scale neuroimaging repository comprising acute clinical MRI scans and linked metadata collected in South Carolina. For this dataset, we evaluated functional outcome as modified Rankin Scale at patient discharge as binary classification (denoted as GS-Rankin for SOOP) with 0-2 as 0 and 3-6 as 1). It includes 647 patients with equal numbers of samples as patients. T1w and FLAIR are used in both unimodal and multimodal evaluation. The data are split by 60/20/20 at the patient level.

\paragraph{CNP --- 1 task} The Preprocessed Consortium for Neuropsychiatric Phenomics (CNP) \cite{Gorgolewski2017-fd} dataset is a rigorously curated neuroimaging resource designed to accelerate the study of brain-behavior relationships across the neuropsychiatric spectrum, encompassing healthy controls as well as individuals diagnosed with adult ADHD, bipolar disorder, and schizophrenia. We performed a three-way classification defined by healthy control, ADHD, and bipolar disorder or schizophrenia using T1w. It presents 265 patients with equal number of samples as patients. The data are split by 60/20/20 at the patient level.

\paragraph{UCSF-PDGM --- 2 task} The University of California San Francisco Preoperative Diffuse Glioma MRI (UCSF-PDGM) \cite{calabrese2022ucsf_pdgm} dataset is an open-access radiogenomic repository in neuro-oncology, focused on Gliobastoma. We evaluated IDH mutation status prediction and overall survival as a time-to-event task. The data includes 495 total samples from 295 patients, wherein all samples have equal number of T1w, T2w and FLAIR. The data are split by 60/20/20 at the patient level.

\paragraph{OpenBHB --- 1 task} OpenBHB is a large-scale, multi-site brain MRI dataset designed to support robust brain-age modeling and evaluation under acquisition-site variability. The dataset aggregates 5,330 three-dimensional T1w brain MRI scans from healthy controls across 10 publicly available cohorts, covering 71 acquisition sites and participants from European-American, European and Asian populations. Becuase ages were not made available for the original test set, we split the original training set by 80/20 producing 2581 patients for training and 646 patients for validation. The original validation set is used as the test set with 757 patients. In order to evaluate the model performance under minimal preprocessing, we used quasi-raw (preprocessed with bias-field correction, affine registration to MNI space) scans in all evaluations.

\subsection*{Model Architecture}
\label{sec:model_arch}
\subsubsection*{Backbone Model}
Motivated by previous studies, demonstrating strong model performance and scalability of 3D Vision Transformers (ViT) on volumetric data such as video \cite{assran2025vjepa2} or medical imaging \cite{zhu20253d}, we employed 3D ViT with Mixture of Experts (MoE) architecture as our backbone. Specifically, the 3D ViT backbone of Neuro-JEPA model includes 12 transformer layers, each based on 768 hidden dimensions, 3072 dimension feedforward layers, and 12 attention heads. To introduce sparsity, MoE routing is enabled on alternating layers (6 MoE layers out of the 12 total) \cite{mustafa2022multimodal}. The MoE configuration utilizes 2 shared experts and 16 total experts, with 6 experts activated per forward pass, softmax scoring gating and a routing scaling factor of 4.0 (detailed ablation studies on these MoE design choices are provided in Supplementary \Cref{apd:moe_routing_analysis}). For positional encoding, we applied 3D Rotary Position Embedding (RoPE) following \citep{assran2025vjepa2}. Input data from cropped standard templates are resized from $180\times216\times180$ to $100\times120\times100$ and then cropped to $96\times108\times96$. A patch size of $12\times12\times12$ is used, yielding 576 tokens per scan.

\subsubsection*{Joint Embedding Pretraining}
Our primary pretraining framework is built upon the Joint-Embedding Predictive Architecture (JEPA). We selected JEPA over alternative methods due to its rapid convergence (fewer than 200 epochs in our experiments) and high throughput, achieved by avoiding the computationally expensive 3D image interpolation required by frameworks like the DINO family \cite{caron2021emerging,oquab2024dinov, simeoni2025dinov3}. This efficiency enabled training on our complete dataset of approximately 1.5 million scans within one week under constrained computational resources. Additionally, we showed that this approach outperforms alternative reconstruction based methods (e.g. Masked Autoencoder) under equivalent settings as demonstrated in Supplementary \Cref{fig:mae_vs_jepa_abla}. Fundamentally, JEPA predicts the latent representations of masked regions based on visible context. The masked volume is processed by the online encoder, while the full volume target is generated by a momentum encoder updated via an Exponential Moving Average (EMA) of the online encoder's weights (see Figure \Cref{fig:overview} for an architectural overview). JEPA has recently demonstrated strong scalability and performance across diverse data in 3D domains, including video \cite{assran2025vjepa2}, physical planning \cite{terver2026drivessuccessphysicalplanning}, echocardiography \cite{munim2026echojepalatentpredictivefoundation}, and neuroimaging \cite{kondepudi2025healthlearningachievesgeneralist,NEURIPS2024_9c3828ad}.

Because our pretraining data underwent skull stripping and defacing, approximately $40-60\%$ of each resulting volume consists of background. To prevent the model from overfitting to low-information background signals while preserving essential spatial anchoring, we modified the standard $L_1$ loss in JEPA into a foreground-aware $L_1$ loss. Specifically, we down-weight the loss contributed by predicted background regions, which we define as voxels with intensities outside the 2nd--98th percentile range of the current scan. Applying an intensity-based cutoff incurs negligible computational cost in comparison to calculating the foreground with a segmentation algorithm. For a batch of $B$ samples (indexed by $b$) and $M$ masks, where each mask $m$ contains $K$ tokens (indexed by $k$) with predicted embedding $\hat{y}$ and target embedding $y$, the loss is formulated as:
$$L = \frac{1}{B \cdot M \cdot K} \sum_{b,m,k} w_{b,m,k} \cdot \left| \hat{y}_{b,m,k} - y_{b,m,k} \right|$$
where the normalized weight as
$$w_{b,m,k} = \frac{w'_{b,m,k}}{\frac{1}{K} \sum_{k=1}^K w'_{b,m,k}}$$
and the unnormalized weighting function as
$$w'_{b,m,k} = \beta + (1 - \beta)f_{b,m,k}$$
on the foreground map $f \in [0, 1]$ (i.e. when $f=1$ (Foreground), The weight becomes $1.0$. When $f=0$ (Background), the weight reduces to $\beta$). The normalization is important in this setup because it ensures that the average weight per mask is always 1.0, regardless of how much background is in the image. In our experiment, $\beta$ is set to be 0.1. Full details on pretraining hyperparameter setting is present in Supplementary \Cref{apd:pretrain_details}.

\subsubsection*{Mixture of Experts}
Introducing sparsity is crucial for efficient learning representations \cite{JMLR:v23:21-0998,NEURIPS2021_48237d9f,jordan93,mustafa2022multimodal,shazeer2017}. We systematically evaluated the role of sparsity in building a multimodal neuroimaging foundation model by integrating MoE. Our design incorporates 2 shared experts and routes tokens to 6 active experts out of 16 total, using a routing scale of 4.0 approximated via sampling \citep{kexuefm-10945}. This sparsity rate is intentionally lower than that of some recent Large Language Models (LLMs) based MoE models \cite{kimiteam2026kimik2openagentic,Guo2025,tong2026beyond}, as our ablation studies in \Cref{fig:fig4-ablation}c,d indicate diminishing performance returns with further sparsification. We apply MoE on every other layer for feedforward layers in attention blocks instead of in every layer, following \cite{mustafa2022multimodal} towards saving computational cost.

Mixture-of-experts (MoE) models are prone to expert collapse, in which routing concentrates on only a few experts \cite{shazeer2017}. This is conventionally mitigated by load balancing, a regularization applied to the routing distribution. Here we adopt auxiliary-loss-free bias-update for load balancing \cite{wang2024auxiliarylossfreeloadbalancingstrategy}, which achieves performance comparable to traditional auxiliary-loss methods \cite{JMLR:v23:21-0998}. We found, however, that the standard auxiliary-loss-free formulation produced unstable load balancing in our setting. To stabilize training, we introduce three modifications. First, token-averaged error correction divides the error correction by the average number of tokens per expert, so that experts further from a balanced distribution receive proportionally larger bias updates. Second, zero-mean projection is applied to the bias terms to prevent unbounded drift. Third, bias clipping constrains bias values that exceed a predefined threshold (set to 0.3 for softmax gating). Together, these modifications reliably prevent expert collapse during JEPA pretraining with MoE under our configuration. The complete modified bias-update algorithm is provided in Supplementary \Cref{apd:algo2}.

\subsubsection*{Multiscale Masking}
Different masking strategies can impose different generalization behaviors on JEPA pre-training as demonstrated from previous works \cite{NEURIPS2024_9c3828ad,nam2026causaljepalearningworldmodels}. We empirically observed that the original masking strategy utilized in V-JEPA 2 \cite{assran2025vjepa2} yields suboptimal performance across our benchmarks. 
The original masking strategy was designed for video, where strong temporal continuity and redundancy permit the removal of large spatiotemporal blocks. In volumetric neuroimaging, however, such aggressive masking along the depth axis may be detrimental, as it can obscure high-frequency spatial details required to encode small lesions and subtle anatomical variation. We therefore hypothesized that neuroimaging requires a masking strategy that better preserves fine-grained three-dimensional structure. As shown in \Cref{fig:fig4-ablation}a,b, random masking patterns that permit easy interpolation between masked targets and visible context substantially degrade downstream performance. Consequently, we propose a modified multiscale masking strategy optimized for masked latent representation prediction for 3D neuroimaging. To reduce reliance on such local interpolation while retaining sufficient depth context for fine-detail prediction, we randomly sample masking blocks across multiple spatial scales until the target masking ratio is reached. Specifically, each mask is iteratively sampled with a spatial scale drawn from one of three ranges, $[0.0, 0.2]$, $[0.2, 0.5]$, or $[0.5, 0.7]$, and a depth scale drawn from the range of $[0.0, 1.0]$, until the fixed masking ratio of $0.75$ is achieved. The complete multiscale sampling algorithm is provided in Supplementary \Cref{apd:multiscale_mask}.

\subsection*{Evaluation Setting}
\paragraph{Baseline Comparisons} We benchmark our proposed model against VoCo \cite{voco,wu26}, a self-supervised model pre-trained on computed tomography (CT) volumes with overall high performance across neuroimaging, and two domain-specific neuroimaging foundation models: BrainIAC \cite{Tak2026} and NeuroVFM \cite{kondepudi2025healthlearningachievesgeneralist}. VoCo is a self-supervised learning model designed to enhance 3D medical image analysis by leveraging the consistent geometric relationships of human anatomy. The model is pre-trained on a diverse collection of 160,167 unannotated 3D CT volumes collected from public datasets with SwinUNETR \cite{yucheng22} as its model backbone. Its training objective predicts crop positions as pseudo-labels under the assumption of anatomical consistency. The base model is used for VoCo across all evaluations to align parameter sizes. BrainIAC is a 3D brain MRI foundation model pre-trained on 48,965 diverse scans using the contrastive self-supervised framework SimCLR \cite{pmlr-v119-chen20j} with a Vision Transformer (ViT) backbone to learn universal anatomical representations. NeuroVFM utilizes a ViT backbone, train with JEPA objective on latent masked prediction on only foreground tokens with approximately 5.24 million scans on both MRI and CT volumes. Variable length flash attention \cite{dao2023flashattention2} is used for handling varying batch lengths per sample for foreground tokens. Although PRIMA \cite{Lyu2026} is also a relevant baseline, both PRIMA and NeuroVFM were pretrained on similar data from the University of Michigan Health System. We therefore selected a single representative model from this shared pretraining source to avoid redundant comparisons. NeuroVFM was chosen because its image-only latent predictive framework is more closely aligned with our method, enabling a more controlled and informative comparison.


\paragraph{Unimodal Encoding Evaluation} A robust multimodal foundation model must first demonstrate high-fidelity encoding of individual modalities. Consequently, we establish our baseline by evaluating unimodal encoding performance. We independently encode each modality (T1w, T2w, and FLAIR) and assess the performance of these representations across various downstream tasks with full fine-tuning on attentive layers. Specifically, we evaluated the model across three institutional datasets (NYU Langone, NYU Long Island, and MGH) alongside 12 public datasets. The experimental results are detailed in \cref{fig:overview,fig:fig2-best-unimodal}; Supplementary \cref{fig:sup-leftover-ap,fig:unimodal-all-auc,apd:avg_performance_unimodal}.

\paragraph{Multimodal Fusion Evaluation} Beyond robust unimodal encoding, an effective multimodal model must successfully leverage complementary information when distinct modalities are combined. We evaluated multimodal fusion by comparing the performance of combined modalities against their unimodal counterparts, ensuring all models are fine-tuned on an identical number of samples. This controlled setting prevents the introduction of confounding variables, offering a more rigorous comparison than evaluating against models trained with missing modalities. We selected 12 tasks from 7 public datasets based on sample size of available modalities and evaluated a total of 15 tasks using MGH as an external clinical cohort. 

As shown in previous studies \cite{liang2024foundationsmultisensoryartificialintelligence,liang2023quantifying,perez2019mfas}, achieving optimal multimodal learning requires tailoring fusion methods to specific models and task combinations to effectively capture cross-modal interactions. Accordingly, we applied five distinct fusion strategies and performed comprehensive hyperparameter sweeps for each task and model. Specifically, our evaluation incorporates late fusion with cross-attention, logit averaging, multiple instance learning-based fusion \cite{kondepudi2025healthlearningachievesgeneralist}, product-of-experts fusion, and inter-intra modality modeling \cite{madaan2024jointly} (detailed in Supplementary \Cref{apd:alg_details}). As T1w serves as the primary structural reference, while T2w and FLAIR sequences provide distinct complementary contrasts, we evaluate multimodal learning under two clinically grounded settings: (1) T1w+T2w, which integrates anatomical information with broad fluid-sensitive contrast, and (2) T1w+FLAIR, which combines anatomy with lesion-accentuated, cerebrospinal fluid-suppressed contrast. These configurations capture two clinically relevant forms of complementarity, enabling a robust assessment of the multimodal learning quality across different foundation models.

To quantify the benefits of multimodal fusion over unimodal approaches, we compared multimodal gain as defined by the difference between best achievable performance across all fusion methods and the optimal unimodal performance. Comprehensive results detailing individual fusion strategies and modality-specific performance are provided in Supplementary \Cref{fig:mm_methods_perf_fm,fig:mm_methods_perf_fm_cont,fig:mm_methods_perf_fm_neurovfm,fig:sup-mm-auprc,fig:sup-mm-brainiac-voco}.


\paragraph{Few-shots Evaluation} We evaluate model performance under limited data regimes with $K=16, 32, 64, 128, 256$, wherein the data is sampled such that positive and negative samples are equal to $K$. For age prediction, $K$ present number of samples bootstrapped from each quartile of the full age distribution in the training data. Because the performance for few-shots is highly dependent on the bootstrapped samples, we run each experiment five times on different bootstrapped samples and calculate the mean and $95\%$ confidence intervals. The detailed few-shot evaluation results across datasets are presented in \Cref{fig:fig5-few-shot-analysis} and Supplementary \Cref{fig:sup-few-shot-avg-auroc,fig:sup-few-shot-avg-auprc,fig:sup-few-shot-permod-t1w-auroc,fig:sup-few-shot-permod-t2w-auroc,fig:sup-few-shot-permod-flair-auroc,fig:sup-few-shot-permod-t1w-auprc,fig:sup-few-shot-permod-t2w-auprc,fig:sup-few-shot-permod-flair-auprc}.

\paragraph{Design Choices Evaluation} We investigate three main architectural adaptations to optimize the Joint-Embedding Predictive Architecture (JEPA) for neuroimaging tasks: integrating Mixture of Experts (MoE) across varying expert counts, adopting multiscale masking over standard strategies, and suppressing loss signals from background predictions. Additionally, we show that pretraining scales effectively with model size, whereas decreasing patch size yields no further improvement. The detailed design choice evaluation is presented in \Cref{fig:fig4-ablation}a-d and Supplementary \Cref{fig:sup-experts-abla,fig:sup-methods-abla}.

\paragraph{Scaling Analysis} We assessed the efficacy of our pre-training framework on data scaling regimes. The main performance improvements were  driven by data scaling and algorithmic improvement, although we did additionally explore parameter scaling. By analogy to the Chinchilla scaling laws \cite{hoffmann2022trainingcomputeoptimallargelanguage} and related literature \cite{pearce2024reconciling,kumar2025scaling}, we hypothesize that our current volume of pre-training data does not yet saturate the capacity of our model's parameter space. We empirically validate this conclusion by evaluating model performance when pre-training only on $10\%, 30\%, 100\%$ of the available data as shown in \Cref{fig:fig4-ablation}e,f and Supplementary \Cref{fig:sup-percentabla_perdataset}. 



\subsection*{Statistical analysis}
Model performance metrics are reported as the empirical mean alongside $95\%$ confidence intervals (CIs). CIs were estimated via non-parametric bootstrapping of the held-out validation set ($B = 1000$ resamples). To account for the sample-selection variance inherent to few-shot learning, training and evaluation procedures were repeated across five independent random samplings of the training distribution; aggregate means and corresponding CIs are reported. Statistical significance of performance differences between compared models was assessed using a two-sided paired permutation test ($N = 1000$ permutations).

\subsection*{Computing Hardware Software}
All experiments are performed in PyTorch (v2.10.0), MONAI (v1.5.1), Numpy (v1.26.4), Pandas (v2.2.3), NiBabel (v5.3.2), Omegaconf (v2.3.0), Hydra (v1.3.2). All plots and figures were created by Matplotlib (v3.10.1), Seaborn (v0.13.2) and Plotly (v6.6.0). We customized the implementation starting from original V-JEPA 2 original implementation \url{https://github.com/facebookresearch/vjepa2}. For evaluation environment on compared models (VoCo --- \url{https://github.com/Luffy03/VoCo}, BrainIAC --- \url{https://github.com/AIM-KannLab/BrainIAC}, NeuroVFM --- \url{https://github.com/MLNeurosurg/neurovfm}), we used the environment provided from their respective Github repositories. All downstream experiments were conducted on a single 80GB NVIDIA A100 GPU (graphics processing unit) or 48GB NVIDIA L40S GPU. All pre-training experiments on ViT Base were conducted on eight/twelve 48GB NVIDIA L40S GPUs in two/three nodes. All data storage and model development were performed on the NYU Langone UltraViolet High Performance Computing Core.

\subsection*{Data Availability}
The internal clinical data used in this study are unavailable due to privacy restrictions and institutional policy. The public datasets can be obtained by direct downloading or sending requests to the corresponding studies as follows.
\begin{sloppypar}
ABIDE (\url{https://fcon_1000.projects.nitrc.org/indi/abide/}),
ADHD-200 (\url{https://fcon_1000.projects.nitrc.org/indi/adhd200/}),
ADNI (\url{https://adni.loni.usc.edu/}),
ICSPR-Stroke (\url{https://www.icpsr.umich.edu/web/ICPSR/studies/38464}),
MCSA (\url{https://www.mayo.edu/research/centers-programs/alzheimers-disease-research-center/research-activities/mayo-clinic-study-aging/overview}),
NACC (\url{https://www.naccdata.org/}),
OASIS3 (\url{https://sites.wustl.edu/oasisbrains/home/oasis-3/}),
SOOP (\url{https://www.nature.com/articles/s41597-024-03667-5}),
CNP (\url{https://openneuro.org/datasets/ds000030/versions/1.0.0}),
UCSF-PDGM (\url{https://www.cancerimagingarchive.net/collection/ucsf-pdgm/}),
OpenBHB (\url{https://baobablab.github.io/bhb/dataset}), and
BIND-MGH (\url{https://bdsp.io/content/n1vba1x5qt62frfjem65/1.0/}).
\end{sloppypar} The data splits for all our evaluation on public datasets are available in our public code repository.

\section*{Data and Weight Availability}

All code required for model pre-training, downstream fine-tuning, and evaluation is publicly available at \url{https://github.com/NYUMedML/Neuro-JEPA}. Pretrained model weights are made publicly available upon reasonable request via Hugging Face at \url{https://huggingface.co/NYUMedML/Neuro-JEPA}. The model weights are distributed for non-commercial, non-derivatives, non-clinical, academic research purposes under the Creative Commons Attribution-NonCommercial-NoDerivatives 4.0 International License (CC BY-NC-ND 4.0).

\section*{Acknowledgements}
Our study was supported by various funding sources. H.H, L.C, A.M. and N.R. are supported by National Institute of Health National Institute on Aging study R01AG085617. A.M. and N.R are also supported by National Institute of Health National Institute on Aging award P30AG066512 and R01AG079175. We acknowledge resources and support from NYU Langone High Performance Computing team, and DataCore which undertook our PACS data request. The authors would like to thank constructive suggestions and feedback from Artie Shen, Carlos Fernandez-Granda, Divyam Madaan, Sumit Chopra and Un J. Kang.

\section*{Author contributions statement}
H.H. and N.R. conceived the study. H.H. developed the methodology, implemented the framework, conducted experiments, collected public datasets, preprocessed, analyzed and interpreted the data, and wrote the original manuscript. L.C. assisted with data preprocessing and organization. J.C. instructed data preprocessing pipeline. J.H. assisted on data visualization and model evaluation verification. J.L., J.M., D.O., J.F., S.D., A.M. instructed evaluation protocols and data processing. N.R. collected NYULH data, and supervised the project. All authors contributed to manuscript editing and discussion.


\bibliography{sample}

\newpage
\appendix

\renewcommand{\figurename}{Supplementary Figure}
\renewcommand{\tablename}{Supplementary Table}
\setcounter{figure}{0}
\setcounter{table}{0}

\newpage
\appendix

\section*{Appendix}

\tableofcontents


\section{Dataset Details}
\subsection{Pretrain Data Demographics}
\label{apd:pretrain_demo}
We present patient demographics information on age, ethnicity and gender for our pretrained data in Supplementary \Cref{tab:pretrain-demo} from available EHR data. Due to the asynchronous update of the institutional EHR and PACS systems, demographic metadata were successfully integrated for $155,064/282,693$ of the cohort (before $01/01/2023$). The remaining represents the most recent imaging acquisitions $(2023-2025)$. The entire cohort was utilized for pretraining, while the matched subset was used for demographic-stratified examination and analysis. In total, our pretrain data contains 282,693 Patients, 428,647 Studies and 1,551,862 Total Scans across multiple regions in New York Area as shown in Supplementary \Cref{fig:sup-patient-stats}.

\begin{table}[ht]
\centering
\small
\caption{Pre-training Data Demographics}
\label{tab:pretrain-demo}
\begin{tabular}{lrr}
\toprule
\textbf{Category} & \textbf{Count} & \textbf{\%} \\
\midrule

\multicolumn{3}{l}{\textit{Age}} \\
\quad Child (0--12)            &  1,891 &  1.2 \\
\quad Teenager (13--17)        &  2,694 &  1.7 \\
\quad Young Adult (18--24)     &  5,160 &  3.3 \\
\quad Adult (25--34)           & 15,085 &  9.6 \\
\quad Adult (35--44)           & 18,985 & 12.1 \\
\quad Middle-Aged (45--54)     & 20,436 & 13.0 \\
\quad Older Adult (55--64)     & 24,737 & 15.7 \\
\quad Senior (65+)             & 56,842 & 36.1 \\
\quad Unknown                  &  9,234 &  5.9 \\
\addlinespace

\multicolumn{3}{l}{\textit{Ethnicity}} \\
\quad White                              & 81,182 & 51.6 \\
\quad African / Black                    & 14,486 &  9.2 \\
\quad Asian / Pacific Islander           &  6,971 &  4.4 \\
\quad Excluded / Other / Unknown / Refused & 52,425 & 33.3 \\
\addlinespace

\multicolumn{3}{l}{\textit{Gender}} \\
\quad Female  & 91,812 & 59.5 \\
\quad Male    & 54,723 & 35.5 \\
\quad Unknown &  8,529 &  5.5 \\

\midrule
\textbf{Total (unique)} & \textbf{157,064} & \\
\bottomrule
\end{tabular}
\end{table}

\begin{figure}[htbp]
    \centering
    \includegraphics[width=0.6\textwidth]{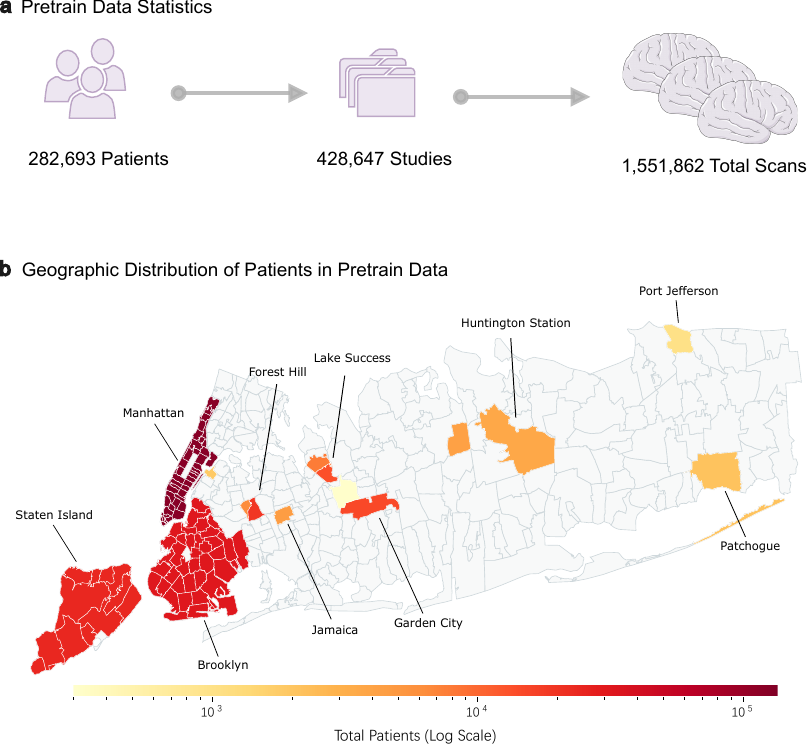}
    \caption{\textbf{Pretrain Data Patients Statistics} - \textbf{a,} Number of patients, studies and total scans used in pretrain data. \textbf{b,} Geographic distribution of patients in pretrain data}
    \label{fig:sup-patient-stats}
\end{figure}

\subsection{ICD10-Code Mapping}
\label{apd:icd-mapping}
We present ICD10 Code Mapping details in \Cref{tab:disease_definition}, where the mapped labels are used for NYU-Langone and Longisland downstream evaluation.

\begin{table}[!htpb]
    \centering
    \small
    \caption{The definition of diseases in EHR by diagnosis codes and medications.}
    \begin{tabular}{lr}
    \toprule
         Disease &  Definition in EHR \\
    \midrule
       IPH  &  I61.0, I61.1, I61.2, I61.3, I61.4, I61.8, I61.9 \\
       IVH  &  I61.5, P52.1, P52.2, P52.3  \\
       ICH  &  IPH + IVH + I61.6, I62.9, P10.9, P52.4, P52.9 \\
       SDH  &  S06.5, I62.0 \\
       SAH  &  I60.*, S06.6, P52.5, P10.3  \\
       Cancer  &  C71.*, C79.3, D33.0, D33.1, D33.2, D33.3, D33.7, D33.9  \\
       Hydrocephalus  &  G91.* \\
       Edema  &  G93.1, G93.5, G93.6, G93.82, S06.1 \\
       \multirow{2}{*}{Dementia}  &  G23.1, G30.*, G31.01, G31.09, G31.83, G31.85, G31.9, F01.*, F02.*, F03.*, G31.84, G31.1, \\ 
       & \textbf{Medication:} DONEPEZIL, RIVASTIGMINE, GALANTAMINE, MEMANTINE, TACRINE \\ 
       Major Depression Disorder  &  F32.*, F33.* \\
    \bottomrule
    \end{tabular}
    \label{tab:disease_definition}
\end{table}

\subsection{Sample Size and Label for Each Downstream Dataset}
\label{apd:num_samples}
We report the number of patients and samples for each downstream evaluation dataset in Supplementary \Cref{tab:benchmark_stats}. For public datasets, the unimodal and multimodal evaluations include different sample sizes because multimodal analyses were restricted to subjects with all required modalities available, enabling balanced and controlled comparisons across modalities. The exact dataset splits used for evaluation are provided in our public code repository.

\setlength\extrarowheight{0pt}
 
\small

\begin{longtable}{@{}p{1.8cm}p{3.3cm}p{2.5cm}rrp{3.1cm}@{}}
\caption{%
  \textbf{Benchmark dataset statistics.}
  \textit{Patients}: number of unique individuals.
  \textit{Samples}: total imaging sessions aggregated across training,
  validation, and test splits.
  For ICSPR, MCSA, and time-to-event tasks, counts are task-specific after
  excluding sessions with missing labels or missing event/censoring information.\textsuperscript{\,*}
  \textit{Positive labels}: for binary classification and time-to-event tasks, positive-class or event sample count and prevalence (\%); for 3-class tasks (CNP Psychiatric diagnosis, ICSPR Lesion type, MCSA Cognitive impairment, PPMI Parkinson diagnosis), class-0\,/\,class-1\,/\,class-2 sample counts (Healthy Control/Bipolar Disorder or Schizophrenia/ADHD for CNP Psychiatric diagnosis; Ischemic/Hemorrhagic/Absent for Lesion type; Cognitive Normal/Mild Cognitive Impairment/Alzheimer for MCSA Cognitive impairment; Healthy Control/Prodromal/Parkinson for PPMI diagnosis); for BIND-MGH multi-label findings see
  Table~\ref{tab:bindmgh}.%
}\label{tab:benchmark_stats}\\

\toprule
\textbf{Dataset} & \textbf{Task} & \textbf{Modality} &
\textbf{Patients} & \textbf{Samples} & \textbf{Positive labels} \\
\midrule
\endfirsthead

\endhead

\midrule
\multicolumn{6}{r@{}}{\footnotesize\textit{Continued on next page}}\\
\endfoot

\bottomrule
\multicolumn{6}{@{}p{15.0cm}@{}}{%
  \begin{minipage}{15.0cm}\vspace{3pt}\scriptsize
    \textsuperscript{\dag}\,BIND-MGH comprises 15 binary
    radiology-finding labels: Astrocytoma, Atrophy, Cyst, Edema,
    Enhancement, Hematoma, Infarct, Ischemic, Mass effect, Midline
    shift, Multiple sclerosis, Schwannoma, Cancer, Glioblastoma
    multiforme, and Gliosis. Per-finding prevalence reported in
    Table~\ref{tab:bindmgh}.\\[2pt]
    \textsuperscript{\ddag}\,NYU-Langone and NYU-Longisland each
    comprise 10 binary clinical-finding labels: Cancer,
    Hydrocephalus, Edema, Dementia, Intraparenchymal haemorrhage
    (IPH), Intraventricular haemorrhage (IVH), Subdural haematoma
    (SDH), Subarachnoid haemorrhage (SAH), Intracranial haemorrhage
    (ICH), and Major depressive disorder. There is also 1 combined finding Haemorrhage from all haemorrhage subtypes.
    Per-finding prevalence reported in
    Tables~\ref{tab:nyu_lg}--\ref{tab:nyu_li}.\\[2pt]
    \textsuperscript{*}\,ICSPR (90-day mRS, Lesion type,
    LoS\,$>$\,8~days) and MCSA (Cognitive impairment, Stroke,
    Hypertension, Dyslipidaemia): sessions with missing labels for a
    given task are excluded; patient and sample totals are
    task-specific.\\[2pt]
    \textbf{Abbreviations.}\;
    AD,~Alzheimer's disease;
    MCI,~mild cognitive impairment;
    ADHD,~attention-deficit/hyperactivity disorder;
    ASD,~autism spectrum disorder;
    DWI,~diffusion-weighted imaging;
    FLAIR,~fluid-attenuated inversion recovery;
    IDH,~isocitrate dehydrogenase;
    LOS,~length of hospital stay;
    mRS,~modified Rankin Scale;
    PD,~Parkinson's disease.
  \end{minipage}%
}\\
\endlastfoot

\multicolumn{6}{@{}l}{\textsc{Unimodal}}\\
\midrule

ABIDE & ASD diagnosis & T1w & 1,099 & 1,099 & 528 (48.1\%) \\
\midrule

ADHD-200 & ADHD diagnosis & T1w & 776 & 776 & 285 (36.7\%) \\
\midrule

\multirow[t]{5}{*}{ADNI}
  & AD diagnosis
    & T1w   &   455 & 1,632 & 784 (48.0\%) \\
& \multirow[t]{3}{3.3cm}{Amyloid-PET status}
    & FLAIR &   317 &   318 & 145 (45.6\%) \\
&   & T1w   &   167 &   167 &  78 (46.7\%) \\
&   & T2w   &   205 &   206 &  97 (47.1\%) \\
\taskrule
& MCI to AD conversion
    & T1w   &   209 &   209 &  94 (45.0\%) \\
\midrule

\multirow[t]{3}{*}{BIND-MGH}
  & \multirow[t]{3}{3.3cm}{Multi-label\textsuperscript{\dag}
      (15 conditions)}
    & FLAIR & 11,802 & 22,142 & See Table~\ref{tab:bindmgh} \\
&   & T1w   & 11,802 & 22,142 & See Table~\ref{tab:bindmgh} \\
&   & T2w   & 11,802 & 22,142 & See Table~\ref{tab:bindmgh} \\
\midrule

CNP & Psychiatric diagnosis & T1w & 265 & 265 & 125\,/\,99\,/\,41 \\
\midrule

\multirow[t]{12}{*}{ICSPR-Stroke}
  & \multirow[t]{4}{3.3cm}{90-day mRS\textsuperscript{*}}
    & DWI   & 1,139 & 1,139 & 439 (38.5\%) \\
&   & FLAIR & 1,099 & 1,099 & 421 (38.3\%) \\
&   & T1w   & 1,011 & 1,011 & 386 (38.2\%) \\
&   & T2w   & 1,019 & 1,019 & 376 (36.9\%) \\
\taskrule
& \multirow[t]{4}{3.3cm}{Lesion type\textsuperscript{*}}
    & DWI   & 2,643 & 2,643 & 1,878\,/\,295\,/\,470 \\
&   & FLAIR & 2,508 & 2,508 & 1,770\,/\,283\,/\,455 \\
&   & T1w   & 2,281 & 2,281 & 1,612\,/\,267\,/\,402 \\
&   & T2w   & 2,342 & 2,342 & 1,645\,/\,269\,/\,428 \\
\taskrule
& \multirow[t]{4}{3.3cm}{LOS\,$>$\,8~days\textsuperscript{*}}
    & DWI   & 1,892 & 1,892 & 398 (21.0\%) \\
&   & FLAIR & 1,823 & 1,823 & 372 (20.4\%) \\
&   & T1w   & 1,655 & 1,655 & 339 (20.5\%) \\
&   & T2w   & 1,675 & 1,675 & 327 (19.5\%) \\
\midrule

\multirow[t]{8}{*}{MCSA}
  & \multirow[t]{2}{3.3cm}{Cognitive impairment\textsuperscript{*}}
    & FLAIR & 1,713 & 2,866 & 2,486\,/\,337\,/\,43 \\
&   & T1w   & 1,713 & 2,866 & 2,486\,/\,337\,/\,43 \\
\taskrule
& \multirow[t]{2}{3.3cm}{Stroke\textsuperscript{*}}
    & FLAIR & 1,712 & 2,847 &    98 (3.4\%) \\
&   & T1w   & 1,712 & 2,847 &    98 (3.4\%) \\
\taskrule
& \multirow[t]{2}{3.3cm}{Hypertension\textsuperscript{*}}
    & FLAIR & 1,712 & 2,847 & 1,849 (64.9\%) \\
&   & T1w   & 1,712 & 2,847 & 1,849 (64.9\%) \\
\taskrule
& \multirow[t]{2}{3.3cm}{Dyslipidaemia\textsuperscript{*}}
    & FLAIR & 1,712 & 2,847 & 2,319 (81.5\%) \\
&   & T1w   & 1,712 & 2,847 & 2,319 (81.5\%) \\
\midrule

\multirow[t]{6}{*}{NACC}
  & \multirow[t]{3}{3.3cm}{AD diagnosis}
    & FLAIR & 3,024 & 3,755 &   710 (18.9\%) \\
&   & T1w   & 3,841 & 4,994 & 1,053 (21.1\%) \\
&   & T2w   & 2,538 & 3,062 &   535 (17.5\%) \\
\taskrule
& \multirow[t]{3}{3.3cm}{Amyloid-PET status}
    & FLAIR &   155 &   159 &  70 (44.0\%) \\
&   & T1w   &   176 &   182 &  83 (45.6\%) \\
&   & T2w   &    93 &    97 &  44 (45.4\%) \\
\midrule

\multirow[t]{3}{*}{OASIS-3}
  & \multirow[t]{3}{3.3cm}{AD diagnosis}
    & FLAIR &   727 & 1,028 & 107 (10.4\%) \\
&   & T1w   & 1,126 & 1,924 & 250 (13.0\%) \\
&   & T2w   & 1,004 & 1,665 & 210 (12.6\%) \\
\midrule

OpenBHB & Age regression & T1w & 3,984 & 3,984 & --- \\
\midrule

\multirow[t]{5}{*}{PPMI}
  & \multirow[t]{3}{3.3cm}{PD diagnosis}
    & FLAIR & 1,577 & 2,365 &   655\,/\,796\,/\,914 \\
&   & T1w   & 1,000 & 1,763 &   570\,/\,335\,/\,858 \\
&   & T2w   &   346 & 1,296 &   278\,/\,120\,/\,898 \\
\taskrule
& \multirow[t]{2}{3.3cm}{Prodromal to PD conversion}
    & FLAIR &   882 &   882 &    86 (9.8\%) \\
&   & T1w   &   397 &   397 &    30 (7.6\%) \\
\midrule

\multirow[t]{2}{*}{SOOP}
  & \multirow[t]{2}{3.3cm}{mRS (binary)}
    & FLAIR & 647 & 647 & 302 (46.7\%) \\
&   & T1w   & 647 & 647 & 302 (46.7\%) \\
\midrule

\multirow[t]{7}{*}{UCSF-PDGM}
  & \multirow[t]{4}{3.3cm}{IDH mutation}
    & DWI   & 495 & 495 & 103 (20.8\%) \\
&   & FLAIR & 495 & 495 & 103 (20.8\%) \\
&   & T1w   & 495 & 495 & 103 (20.8\%) \\
&   & T2w   & 495 & 495 & 103 (20.8\%) \\
\taskrule
& \multirow[t]{3}{3.3cm}{Overall survival}
    & FLAIR & 495 & 495 & 248 (50.1\%) \\
&   & T1w   & 495 & 495 & 248 (50.1\%) \\
&   & T2w   & 495 & 495 & 248 (50.1\%) \\

\midrule
\multicolumn{6}{@{}l}{\textsc{Multimodal}}\\
\midrule

BIND-MGH
  & Multi-label\textsuperscript{\dag} (15 conditions)
  & T1w\,+\,T2w/FLAIR
  & 11,802 & 22,142 & See Table~\ref{tab:bindmgh} \\
\midrule

\multirow[t]{3}{*}{ICSPR-Stroke}
  & 90-day mRS\textsuperscript{*}
  & \multirow[t]{3}{2.5cm}{T1w\,+\,FLAIR}
  &   993 &   993 & 377 (38.0\%) \\
& Lesion type\textsuperscript{*}       & & 2,212 & 2,212 & 1,561\,/\,259\,/\,392 \\
& LoS\,$>$\,8~days\textsuperscript{*}  & & 1,620 & 1,620 & 326 (20.1\%) \\
\midrule

\multirow[t]{4}{*}{MCSA}
  & Cognitive impairment\textsuperscript{*}
  & \multirow[t]{4}{2.5cm}{T1w\,+\,FLAIR}
  & 1,713 & 2,866 & 2,486\,/\,337\,/\,43 \\
& Stroke\textsuperscript{*}        & & 1,712 & 2,847 &    98 (3.4\%) \\
& Hypertension\textsuperscript{*}  & & 1,712 & 2,847 & 1,849 (64.9\%) \\
& Dyslipidaemia\textsuperscript{*} & & 1,712 & 2,847 & 2,319 (81.5\%) \\
\midrule

NACC      & AD diagnosis & T1w\,+\,T2w   & 3,132 & 3,772 & 525 (13.9\%) \\
\midrule

OASIS-3   & AD diagnosis & T1w\,+\,T2w   &   883 & 1,384 & 163 (11.8\%) \\
\midrule

PPMI      & PD diagnosis & T1w\,+\,T2w   &   302 & 1,339 & 286\,/\,121\,/\,932 \\
\midrule

SOOP      & mRS (binary) & T1w\,+\,FLAIR &   647 &   647 & 302 (46.7\%) \\
\midrule

UCSF-PDGM & IDH mutation & T1w\,+\,FLAIR &   495 &   495 & 103 (20.8\%) \\

\midrule
\multicolumn{6}{@{}l}{\textsc{Internal Cohorts}}\\
\midrule

\multirow[t]{3}{*}{NYU-Langone}
  & \multirow[t]{3}{3.3cm}{Multi-label\textsuperscript{\ddag}
      (11 conditions)}
    & T1w   &  10,004 & 19,325 & See Table~\ref{tab:nyu_lg} \\
&   & T2w   & 19,132& 29,237 & See Table~\ref{tab:nyu_lg} \\
&   & FLAIR & 15,937 & 23,302 & See Table~\ref{tab:nyu_lg} \\
\midrule

\multirow[t]{3}{*}{NYU-Long\-Island}
  & \multirow[t]{3}{3.3cm}{Multi-label\textsuperscript{\ddag}
      (11 conditions)}
    & T1w   & 3,306 & 8,024 & See Table~\ref{tab:nyu_li} \\
&   & T2w   & 2,828 & 5,050 & See Table~\ref{tab:nyu_li} \\
&   & FLAIR & 2,596 & 3,376 & See Table~\ref{tab:nyu_li} \\

\end{longtable}

\begin{table}[htbp]
\centering
\caption{%
  \textbf{BIND-MGH radiology finding prevalence.}
  Positive counts and prevalence are identical across T1w, T2w, and
  FLAIR unimodal splits, and across the multimodal split, because all
  modalities correspond to the same patient sessions with shared labels.
  Total: 22,142 sessions from 11,802 unique patients.
}\label{tab:bindmgh}
\begin{tabular}{@{}lrr@{}}
\toprule
\textbf{Finding} & \textbf{Positive ($n$)} & \textbf{Prevalence (\%)} \\
\midrule
Astrocytoma              & 1,152 &  5.2 \\
Atrophy                  & 2,375 & 10.7 \\
Cyst                     & 1,700 &  7.7 \\
Edema                    & 2,119 &  9.6 \\
Enhancement              & 5,169 & 23.3 \\
Hematoma                 & 1,733 &  7.8 \\
Infarct                  & 4,932 & 22.3 \\
Ischemic                 & 2,849 & 12.9 \\
Mass effect              &   743 &  3.4 \\
Midline shift            &   877 &  4.0 \\
Multiple sclerosis       & 1,382 &  6.2 \\
Schwannoma               &   282 &  1.3 \\
Cancer                   & 5,713 & 25.8 \\
Glioblastoma multiforme  &   545 &  2.5 \\
Gliosis                  &   722 &  3.3 \\
\bottomrule
\multicolumn{3}{@{}p{8.0cm}@{}}{%
  \scriptsize
  Prevalence $=$ positive sessions\,$/\,$total sessions.
  Findings listed in source-dataset column order.
}\\
\end{tabular}
\end{table}

\begin{table}[htbp]
\centering
\caption{%
  \textbf{NYU-Langone radiology and clinical finding prevalence by modality.}
  Each cell reports the number of positive sessions and prevalence
  (positive sessions\,/\,total sessions for that modality).
  T1w: 9,692 patients, 19,325 sessions.
  T2w: 17,485 patients, 29,237 sessions.
  FLAIR: 14,555 patients, 23,302 sessions.
}\label{tab:nyu_lg}
\begin{tabular}{@{}p{4.0cm}p{3.2cm}p{3.2cm}p{3.2cm}@{}}
\toprule
\textbf{Finding} &
\textbf{T1w} &
\textbf{T2w} &
\textbf{FLAIR} \\
\midrule
Cancer                    & 8,596 (44.5\%) &  9,360 (32.0\%) &  9,715 (41.7\%) \\
Hydrocephalus             & 1,305 (6.8\%)  &  1,753 (6.0\%)  &  1,817 (7.8\%)  \\
Edema                     & 2,707 (14.0\%) &  3,441 (11.8\%) &  3,099 (13.3\%) \\
Dementia                  & 2,193 (11.3\%) &  6,771 (23.2\%) &  3,694 (15.9\%) \\
IPH                       & 1,146 (5.9\%)  &  1,424 (4.9\%)  &  1,148 (4.9\%)  \\
IVH                       &   709 (3.7\%)  &    861 (2.9\%)  &    718 (3.1\%)  \\
SDH                       &   742 (3.8\%)  &  1,067 (3.6\%)  &    985 (4.2\%)  \\
SAH                       &   622 (3.2\%)  &    969 (3.3\%)  &    880 (3.8\%)  \\
ICH                       & 1,318 (6.8\%)  &  1,621 (5.5\%)  &  1,311 (5.6\%)  \\
Major depressive disorder & 2,597 (13.4\%) &  5,099 (17.4\%) &  3,641 (15.6\%) \\
Haemorrhage               & 2,006 (10.4\%) &  2,678 (9.2\%)  &  2,267 (9.7\%)  \\
\bottomrule
\multicolumn{4}{@{}p{13.6cm}@{}}{%
  \scriptsize
  Prevalence $=$ positive sessions\,$/\,$total sessions per modality.
  IPH, intraparenchymal haemorrhage; IVH, intraventricular haemorrhage;
  SDH, subdural haematoma; SAH, subarachnoid haemorrhage;
  ICH, intracranial haemorrhage.
}\\
\end{tabular}
\end{table}

\begin{table}[htbp]
\centering
\caption{%
  \textbf{NYU-Long Island radiology and clinical finding prevalence by modality.}
  Each cell reports the number of positive sessions and prevalence
  (positive sessions\,/\,total sessions for that modality).
  T1w: 3,306 patients, 8,024 sessions.
  T2w: 2,828 patients, 5,050 sessions.
  FLAIR: 2,596 patients, 3,376 sessions.
}\label{tab:nyu_li}
\begin{tabular}{@{}p{4.0cm}p{3.2cm}p{3.2cm}p{3.2cm}@{}}
\toprule
\textbf{Finding} &
\textbf{T1w} &
\textbf{T2w} &
\textbf{FLAIR} \\
\midrule
Cancer                    & 1,762 (22.0\%) &   587 (11.6\%) &   452 (13.4\%) \\
Hydrocephalus             &   345 (4.3\%)  &   199 (3.9\%)  &   104 (3.1\%)  \\
Edema                     & 1,377 (17.2\%) &   662 (13.1\%) &   416 (12.3\%) \\
Dementia                  & 1,314 (16.4\%) & 1,192 (23.6\%) &   601 (17.8\%) \\
IPH                       &   719 (9.0\%)  &   363 (7.2\%)  &   195 (5.8\%)  \\
IVH                       &   419 (5.2\%)  &   205 (4.1\%)  &   111 (3.3\%)  \\
SDH                       &   323 (4.0\%)  &   202 (4.0\%)  &   103 (3.1\%)  \\
SAH                       &   260 (3.2\%)  &   162 (3.2\%)  &    80 (2.4\%)  \\
ICH                       &   825 (10.3\%) &   413 (8.2\%)  &   226 (6.7\%)  \\
Major depressive disorder & 1,589 (19.8\%) & 1,035 (20.5\%) &   590 (17.5\%) \\
Haemorrhage               & 1,030 (12.8\%) &   583 (11.5\%) &   311 (9.2\%)  \\
\bottomrule
\multicolumn{4}{@{}p{13.6cm}@{}}{%
  \scriptsize
  Prevalence $=$ positive sessions\,$/\,$total sessions per modality.
  IPH, intraparenchymal haemorrhage; IVH, intraventricular haemorrhage;
  SDH, subdural haematoma; SAH, subarachnoid haemorrhage;
  ICH, intracranial haemorrhage.
}\\
\end{tabular}
\end{table}

\newpage

\section{Algorithm and Evaluation Details}
\label{apd:alg_details}
\subsection{Multiscale Masking}
\label{apd:multiscale_mask}
We empirically observed that masking strategies can largely influence JEPA pre-training generalization for neuroimaging. Supplementary \Cref{apd:algo1} presents one of masking implementations we attempted that presents stable training and improved performance over original V-JEPA 2 masking implementation under our experimental setups. While it does not represent the optimal masking strategy for neuroimaging, it indicates the importance of careful masking design for special data type such as neuroimaging, where model can easily learn shortcut in latent space such as interpolating nearby pixels (anatomy of the brain is highly structured). 

The algorithm operates on a 3D patch grid derived from a volume. Phase 1 places $K$ multiscale blocks with log-uniform aspect ratios. Phase 2 reaches the exact target context size by eroding/dilating along block boundaries, preserving spatial coherence. Fixed output lengths is applied for direct batch stacking without padding. Phase 3 shuffles indices to ensure spatial information is conveyed only through position embeddings, not through sequence order.

\begingroup
  \floatstyle{plain}\restylefloat{algorithm}
\begin{algorithm}
\caption{Multiscale Volumetric Block Masking.}
\label{apd:algo1}
\begin{adjustbox}{width=0.63\columnwidth, center=\columnwidth}
\begin{minipage}{\linewidth}
\hrule height 0.8pt\vspace{2pt}
\begin{algorithmic}[1]
\Require Patch-grid dimensions $(H,W,D)$; total mask ratio $\rho \in (0,1)$
\Require Number of blocks $K$; spatial-scale range $[s_{\min}, s_{\max}]$, depth-scale range $[\delta_{\min}, \delta_{\max}]$, aspect-ratio range $[\alpha_{\min}, \alpha_{\max}]$, $\mathcal{N}_6(v) = \{u \in [H]\times[W]\times[D] : \|u - v\|_1 = 1\}$, the
6-connected neighbourhood of patch $v$.
\Require (Optional) Foreground map $\mathcal{F} \in \{0,1\}^{H\times W\times D}$
\Ensure Shuffled context indices $\mathbf{I}_{\text{enc}}$ and target indices 
$\mathbf{I}_{\text{pred}}$ of fixed lengths $N_{\text{enc}}$ and $N_{\text{pred}}$
\Statex
\vspace{2pt}\hrule height 0.8pt
\State $N_{\text{pred}} \leftarrow \lfloor \rho\, HWD \rfloor$; \quad $N_{\text{enc}} \leftarrow HWD - N_{\text{pred}}$
\State $M \leftarrow \mathbf{1}^{H\times W\times D}$ \Comment{$M_v{=}1$: context (kept); $M_v{=}0$: target (masked)}
\Statex
\Statex \textit{Phase 1.\ Multiscale block placement}
\For{$k = 1,\ldots,K$}
    \State $s \sim \mathcal{U}(s_{\min}, s_{\max})$, \quad $\delta \sim \mathcal{U}(\delta_{\min}, \delta_{\max})$
    \State $A \leftarrow \max(1,\, \lfloor s\, HW \rfloor)$, \quad $d \leftarrow \max\!\bigl(1,\, \min(\lfloor \delta\, D \rfloor,\, D)\bigr)$
    \State $\log \alpha \sim \mathcal{U}(\log \alpha_{\min}, \log \alpha_{\max})$ \Comment{log-uniform aspect ratio}
    \State $h \leftarrow \mathrm{clip}\!\bigl(\lfloor\sqrt{A\,\alpha}\rceil,\, 1,\, H\bigr)$, \quad $w \leftarrow \mathrm{clip}\!\bigl(\lfloor\sqrt{A/\alpha}\rceil,\, 1,\, W\bigr)$
    \State Sample origin $(y, x, z)$ uniformly from valid placements
    \State $M[\,y{:}y{+}h,\; x{:}x{+}w,\; z{:}z{+}d\,] \leftarrow 0$
\EndFor
\Statex
\Statex \textit{Phase 2.\ Boundary-coherent count adjustment}
\State $n \leftarrow \textstyle\sum_v M_v$
\While{$n > N_{\text{enc}}$} \Comment{Erode: expand masked region inward}
    \State $\mathcal{B} \leftarrow \{\,v : M_v{=}1 \text{ and } \exists\, u \in \mathcal{N}_6(v),\; M_u{=}0\,\}$ \Comment{outer boundary}
    \If{$\mathcal{B} = \emptyset$} \textbf{break} \EndIf
    \If{$\mathcal{F}$ provided}
        \State $\mathcal{B}_0 \leftarrow \{b \in \mathcal{B} : \mathcal{F}_b{=}0\}$, \quad $\mathcal{B}_1 \leftarrow \{b \in \mathcal{B} : \mathcal{F}_b{=}1\}$
        \State $\mathcal{B} \leftarrow \mathrm{shuffle}(\mathcal{B}_0) \,\Vert\, \mathrm{shuffle}(\mathcal{B}_1)$ \Comment{background removed first}
    \Else
        \State $\mathcal{B} \leftarrow \mathrm{shuffle}(\mathcal{B})$
    \EndIf
    \State $\Delta \leftarrow \min(n - N_{\text{enc}},\; |\mathcal{B}|)$
    \State $M[\mathcal{B}[1{:}\Delta]] \leftarrow 0$, \quad $n \leftarrow n - \Delta$
\EndWhile
\While{$n < N_{\text{enc}}$} \Comment{Dilate: shrink masked region}
    \State $\mathcal{I} \leftarrow \{\,v : M_v{=}0 \text{ and } \exists\, u \in \mathcal{N}_6(v),\; M_u{=}1\,\}$ \Comment{inner boundary}
    \If{$\mathcal{I} = \emptyset$} \textbf{break} \EndIf
    \If{$\mathcal{F}$ provided}
        \State $\mathcal{I}_1 \leftarrow \{i \in \mathcal{I} : \mathcal{F}_i{=}1\}$, \quad $\mathcal{I}_0 \leftarrow \{i \in \mathcal{I} : \mathcal{F}_i{=}0\}$
        \State $\mathcal{I} \leftarrow \mathrm{shuffle}(\mathcal{I}_1) \,\Vert\, \mathrm{shuffle}(\mathcal{I}_0)$ \Comment{foreground restored first}
    \Else
        \State $\mathcal{I} \leftarrow \mathrm{shuffle}(\mathcal{I})$
    \EndIf
    \State $\Delta \leftarrow \min(N_{\text{enc}} - n,\; |\mathcal{I}|)$
    \State $M[\mathcal{I}[1{:}\Delta]] \leftarrow 1$, \quad $n \leftarrow n + \Delta$
\EndWhile
\Statex
\Statex \textit{Phase 2b.\ Random fallback (guarantees exact target counts)}
\If{$n > N_{\text{enc}}$}
    \State Randomly select $n - N_{\text{enc}}$ patches from $\{v : M_v{=}1\}$ and set to $0$
\ElsIf{$n < N_{\text{enc}}$}
    \State Randomly select $N_{\text{enc}} - n$ patches from $\{v : M_v{=}0\}$ and set to $1$
\EndIf
\Statex
\Statex \textit{Phase 3.\ Index extraction}
\State $\mathbf{I}_{\text{enc}} \leftarrow \mathrm{shuffle}\bigl(\{v : M_v{=}1\}\bigr)$
\State $\mathbf{I}_{\text{pred}} \leftarrow \mathrm{shuffle}\bigl(\{v : M_v{=}0\}\bigr)$
\State \Return $\mathbf{I}_{\text{enc}},\; \mathbf{I}_{\text{pred}}$
\vspace{2pt}\hrule height 0.8pt
\end{algorithmic}
\end{minipage}
\end{adjustbox}
\end{algorithm}
\endgroup

\newpage

\subsection{Multiscale Masking Visualization}
\label{apd:mask_vis}
\begin{figure}[htbp]
    \centering
    \includegraphics[width=0.85\textwidth]{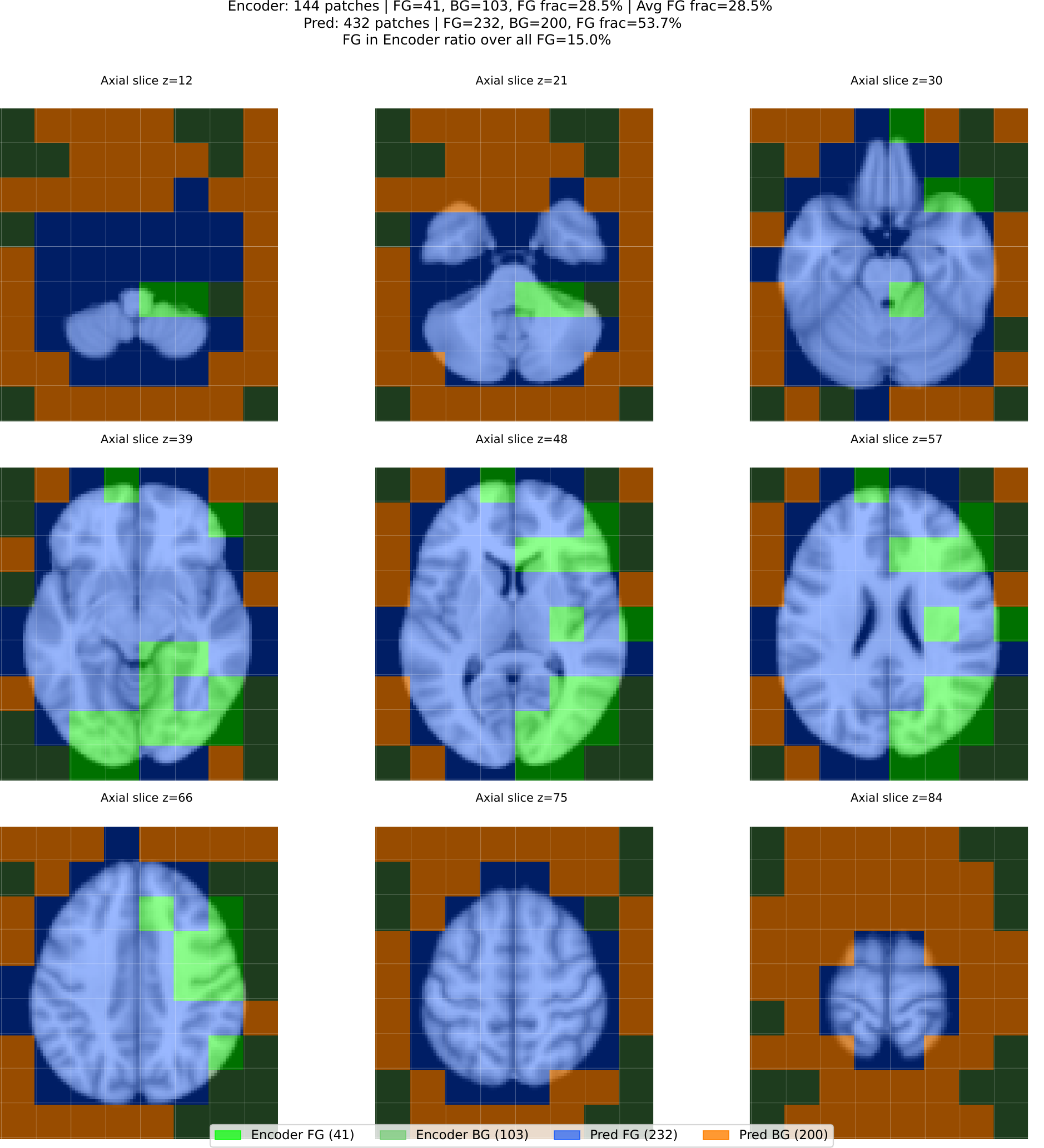}
    \caption{\textbf{Multiscale Masking Sample.} Foreground is denoted as FG, background is denoted as BG and the foreground ratio for encoder/predictor are separately shown.}
    \label{fig:mask_vis_1}
\end{figure}

\begin{figure}[htbp]
    \centering
    \includegraphics[width=0.85\textwidth]{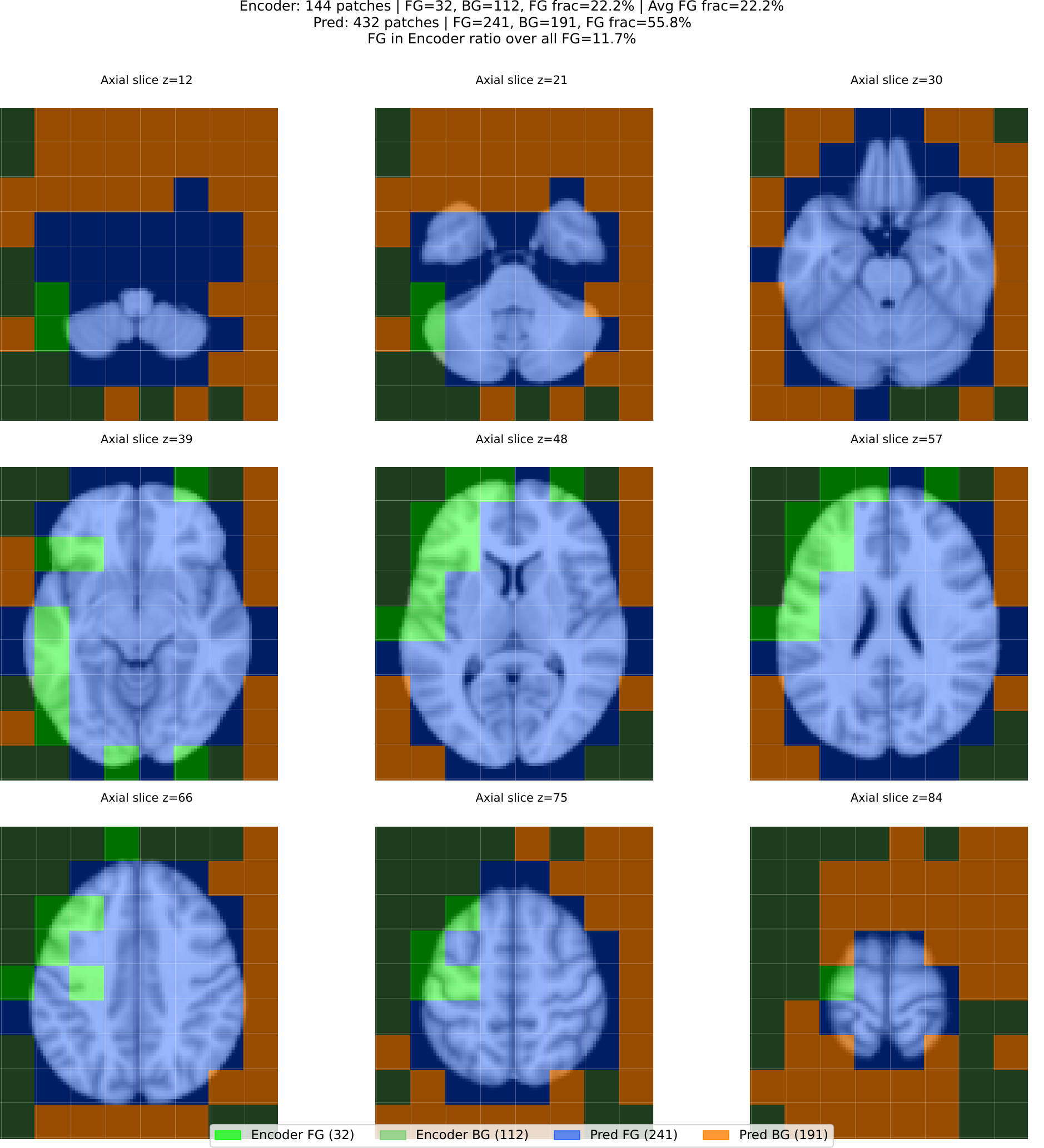}
    \caption{\textbf{Multiscale Masking Sample.} Foreground is denoted as FG, background is denoted as BG and the foreground ratio for encoder/predictor are separately shown.}
    \label{fig:mask_vis_2}
\end{figure}

\begin{figure}[htbp]
    \centering
    \includegraphics[width=0.85\textwidth]{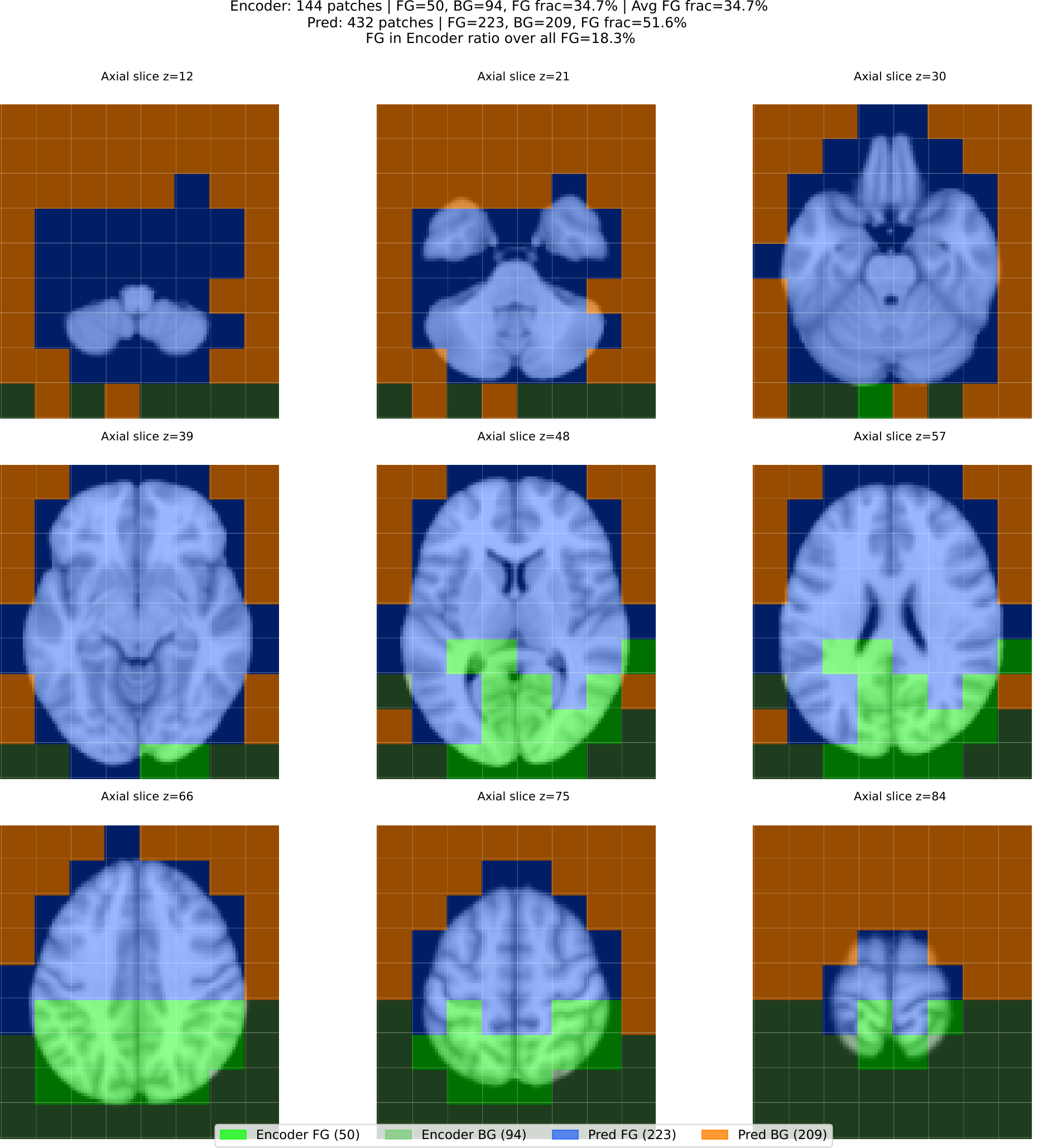}
    \caption{\textbf{Multiscale Masking Sample.} Foreground is denoted as FG, background is denoted as BG and the foreground ratio for encoder/predictor are separately shown.}
    \label{fig:mask_vis_3}
\end{figure}

Supplementary \Cref{fig:mask_vis_1,fig:mask_vis_2,fig:mask_vis_3} show sample visualization for multiscale masking, with foreground and background in encoder/predictor shown as different colors. The ratio of foreground in encoder and predictor is also calculated to monitor the given context.

\subsection{Feature Map Visualization}
\label{apd:feature_map}
\begin{figure}[htbp]
    \centering
    \includegraphics[width=0.85\textwidth]{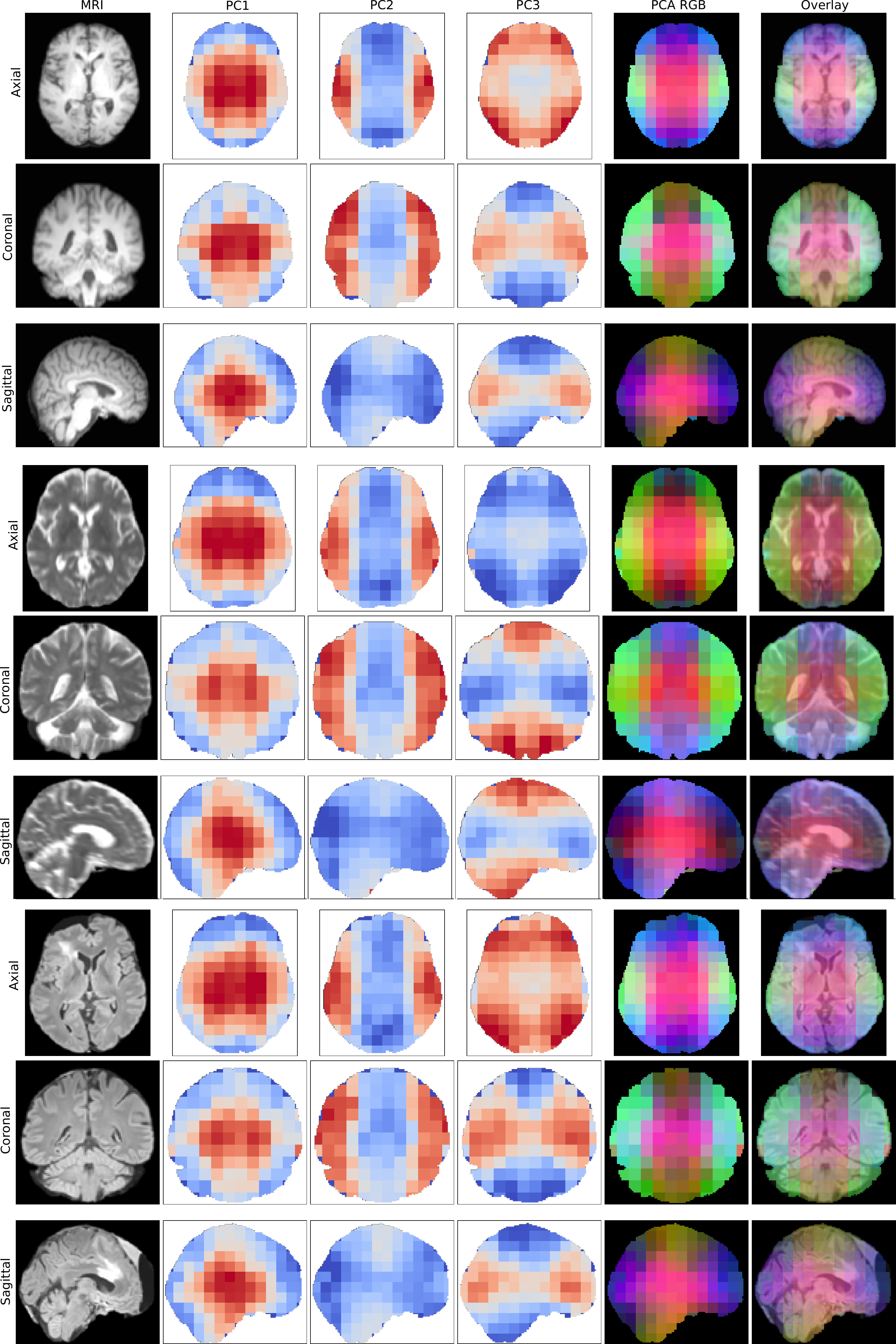}
    \caption{\textbf{Feature Map Visualization with PCA - ViT-L-MoE with Path Size 8}}
    \label{fig:sup-feature-map}
\end{figure}

Supplementary \Cref{fig:sup-feature-map} displays feature map visualizations obtained via Principal Component Analysis (PCA) on output token features, following the methodology of DINO \cite{oquab2024dinov,simeoni2025dinov3}. For each scan (T1w, T2w, and FLAIR), features were extracted using a ViT-Large backbone with Mixture-of-Experts (MoE) and a patch size of 8 to capture fine-grained spatial details. The resulting visualizations demonstrate that the learned representations sharply delineate distinct anatomical structures within the brain.

\newpage
\subsection{Modified Auxiliary Loss Free Load Balancing}
We observe original auxiliary loss free load balancing does not preserve ideal load balancing during pre-training. We therefore modified the algorithm with new bias update rule based on error correction, zero-mean projection and bias clipping to preserve more balanced bias update as demonstrated in Supplementary \Cref{apd:algo2}.
\begin{algorithm}
\caption{Modified MoE Load Balancing Bias Update}
\label{apd:algo2}
\begin{algorithmic}[1]

\Require Tokens routed to expert $i$: $C_i$, Number of experts: $N$,
         Learning rate: $\eta$, Smoothing factor: $\delta$,
         Clipping threshold: $b_{\max}$

\State $\bar{C} \gets \frac{1}{N} \sum_{i=1}^{N} C_i$
       \Comment{Calculate average tokens per expert}

\For{each expert $i \in \{1, \dots, N\}$}
    \State $\epsilon_i \gets \bar{C} - C_i$ \Comment{Error Calculation}
    \State \textit{Original Aux Loss Free Bias Update:}
           \sout{$\Delta b_i \gets \eta \cdot \text{sgn}(\epsilon_i)$}
    \State \textcolor{red}{\textit{New Bias Update Step:}}
           \textcolor{red}{$\Delta b_i \gets \eta \cdot \frac{\epsilon_i}{\bar{C} + \delta}$}
    \State $b_i \gets b_i + \Delta b_i$
\EndFor

\Statex \textit{// Zero-Mean Projection to prevent infinite drift}
\For{each expert $i \in \{1, \dots, N\}$}
    \State \textcolor{red}{$b_i \gets b_i - \frac{1}{N} \sum_{j=1}^{N} b_j$}
\EndFor

\Statex \textit{// Bias Clipping to avoid exploding bias terms}
\For{each expert $i \in \{1, \dots, N\}$}
    \State \textcolor{red}{$b_i \gets \max(-b_{\max}, \min(b_i, b_{\max}))$}
\EndFor

\end{algorithmic}
\end{algorithm}

\subsection{Multi-Modal Learning Methods}
There are five multimodal learning methods we attempted on investigating optimal achievable performance for each task. The formulation for each of them is detailed below:

\begin{description}
  \item[Logits-Averaging Ensemble] This is the simplest baseline where independent classification followed by logit-level averaging. Each modality $m$ has its own linear head $f_m:\mathbb{R}^D\rightarrow R^{C}$ for hidden dimension $D$ and number of classes $C$. For each modality $m$, given weights $W_m\in\mathbb{R}^{C\times D}$, latent representation $z^{(m)}\in\mathbb{R}^D$, the linear head map latent representation to logits by
  $$\ell^{(m)} = W_m\, \mathbf{z}^{(m)}$$
  The, the final prediction is the average of per-modality logits as
  $$\hat{y} = \frac{1}{m}\sum_{m=1}^{M} \ell^{(m)}$$
  This method serves as a strong baseline because it makes no assumptions about inter-modal interactions and ensemble averaging reduces variance in the logit space.

  \item[Gated Cross-Attention Late Fusion] Given two modalities producing patch-level feature sequences $\mathbf{Z}^{(1)} \in \mathbb{R}^{B \times N_1 \times D}$ and $\mathbf{Z}^{(2)} \in \mathbb{R}^{B \times N_2 \times D}$ from a shared backbone, each is first mapped through a modality-specific projection head with residual connection:
  $$\mathbf{H}^{(m)} = \text{Proj}_m(\mathbf{Z}^{(m)}) = \sigma\bigl(W_1^{(m)} \mathbf{Z}^{(m)}\bigr) W_2^{(m)} + W_1^{(m)} \mathbf{Z}^{(m)}, \quad m \in \{1,2\}$$ where $\sigma$ is GELU activation.
  Bidirectional cross-attention then enables each modality to attend to the other. For modality 1 attending to modality 2:
  $$\text{CrossAttn}(\mathbf{Q}, \mathbf{K}, \mathbf{V}) = \text{softmax}\!\left(\frac{\mathbf{Q}\mathbf{K}^\top}{\sqrt{d_h}}\right)\mathbf{V}$$
  where $\mathbf{Q} = \mathbf{H}^{(1)}W_Q$, $\mathbf{K} = \mathbf{H}^{(2)}W_K$, $\mathbf{V} = \mathbf{H}^{(2)}W_V$, and $d_h = D/H$ is the per-head dimension across $H$ heads. A symmetric operation computes modality 2's attention over modality 1. Residual connections and layer normalization (LN) are applied:
  $$\hat{\mathbf{H}}^{(1)} = \text{LN}\!\left(\mathbf{H}^{(1)} + \text{LN}\!\left(\text{CrossAttn}_{1\to2}(\mathbf{H}^{(1)}, \mathbf{H}^{(2)})\right)\right)$$
  $$\hat{\mathbf{H}}^{(2)} = \text{LN}\!\left(\mathbf{H}^{(2)} + \text{LN}\!\left(\text{CrossAttn}_{2\to1}(\mathbf{H}^{(2)}, \mathbf{H}^{(1)})\right)\right)$$
  
  After mean-pooling over the token dimension to obtain $\bar{\mathbf{h}}^{(m)} = \frac{1}{N_m}\sum_{i=1}^{N_m} \hat{\mathbf{H}}^{(m)}_i \in \mathbb{R}^D$, a learned gating performs soft feature selection by
  $$\mathbf{g} = \tanh\!\left(W_g^{(2)}\,\text{ReLU}\!\left(W_g^{(1)} [\bar{\mathbf{h}}^{(1)};\, \bar{\mathbf{h}}^{(2)}]\right)\right) \in \mathbb{R}^D$$
  $$\mathbf{f} = \mathbf{g} \odot \bar{\mathbf{h}}^{(1)} + (1 - \mathbf{g}) \odot \bar{\mathbf{h}}^{(2)}$$
  
  where $[\cdot\,;\,\cdot]$ denotes concatenation, $\odot$ is the Hadamard product, and $\mathbf{g}$ acts as a dimension-wise interpolation coefficient between the two modality representations. The final prediction is $\hat{y} = W_c \mathbf{f} + b_c$.
  
  This design allows fine-grained, element-wise arbitration of how much each modality contributes to each feature dimension.

  \item[Classify-Then-Aggregate Multiple Instance Learning \cite{kondepudi2025healthlearningachievesgeneralist}] This method treats the multimodal problem as a Multiple Instance Learning (MIL) task. Patch tokens from all modalities are concatenated into a single bag by
  $$\mathbf{X} = [\mathbf{Z}^{(1)}_1, \ldots, \mathbf{Z}^{(1)}_{N_1}, \mathbf{Z}^{(2)}_1, \ldots, \mathbf{Z}^{(2)}_{N_2}] \in \mathbb{R}^{N \times D}$$
  where $N = N_1 + N_2$. Each patch (instance) $\mathbf{x}_i$ undergoes two parallel computations
  
  1) \textit{Gated attention scoring.} An attention score per class $c$ is computed via a gated attention mechanism:
  $$a_{i,c} = \mathbf{w}_c^\top \left[\tanh(V \mathbf{x}_i) \odot \sigma(U \mathbf{x}_i)\right]$$
  where $V, U \in \mathbb{R}^{D' \times D}$ are shared attention and gating projections, $\sigma$ is the sigmoid function providing a soft gate, and $\mathbf{w}_c \in \mathbb{R}^{D'}$ is a class-specific attention vector. Per-class attention weights are obtained via segment softmax over each bag $\mathcal{B}_b$:
  $$\alpha_{i,c} = \frac{\exp(a_{i,c})}{\sum_{j \in \mathcal{B}_b} \exp(a_{j,c})}$$
  2) \textit{Instance-level classification.} A shared MLP produces patch-level logits:
  $$\hat{y}_{i,c} = \text{MLP}(\mathbf{x}_i)_c$$
  Bag-level aggregation. The final bag prediction for class $c$ is the attention-weighted sum of instance predictions.
  $$\hat{Y}_{b,c} = \gamma_c \left(\sum_{i \in \mathcal{B}_b} \alpha_{i,c} \cdot \hat{y}_{i,c}\right) + \beta_c$$
  where $\gamma_c$ and $\beta_c$ are learnable scale and bias parameters. This "classify-then-aggregate" formulation enables interpretable per-patch importance attribution across modalities.

  \item[Product-of-Experts Fusion] This method frames multimodal fusion as a Product of Experts (PoE) \cite{hinton02}, where each modality acts as an independent expert that contributes a probabilistic opinion over the label space. The joint posterior is obtained by multiplying per-modality softmax distributions and renormalizing --- equivalent to summation in log-probability space.
  
  Given $M$ modalities, a shared backbone $f_\theta$ extracts features $\mathbf{z}^{(m)} = f_\theta(\mathbf{x}^{(m)})$, and each modality $m$ has a dedicated classification head $g_m$ that produces logits $\boldsymbol{\ell}^{(m)} = g_m(\mathbf{z}^{(m)}) \in \mathbb{R}^C$. Each expert's belief is a categorical distribution:
  
  $$p_m(y \mid \mathbf{x}^{(m)}) = \text{softmax}(\boldsymbol{\ell}^{(m)})$$
  The PoE joint distribution is defined as the normalized product of all expert distributions:
  $$p_{\text{PoE}}(y \mid \mathbf{x}^{(1)}, \ldots, \mathbf{x}^{(M)}) = \frac{\prod_{m=1}^{M} p_m(y \mid \mathbf{x}^{(m)})}{\sum_{c=1}^{C} \prod_{m=1}^{M} p_m(y=c \mid \mathbf{x}^{(m)})}$$
  
  For numerical stability, all computations are carried out in log-space. Defining $\boldsymbol{\lambda}^{(m)} = \log\,\text{softmax}(\boldsymbol{\ell}^{(m)})$:
  
  $$\log p_{\text{PoE}}(y=c) = \sum_{m=1}^{M} \lambda_c^{(m)} - \log \sum_{c'=1}^{C} \exp\!\left(\sum_{m=1}^{M} \lambda_{c'}^{(m)}\right)$$
  
  The model is trained with negative log-likelihood loss (NLLLoss) on the fused log-probabilities:
  
  $$\mathcal{L} = -\log p_{\text{PoE}}(y = y^* \mid \mathbf{x}^{(1)}, \ldots, \mathbf{x}^{(M)})$$
  
  For ViT backbones, each per-modality head is Attentive Classifier (cross-attention pooler + linear layer); for CNN/SwinUNETR/BrainIAC backbones, each head is a simple linear layer applied to mean-pooled features.

  \item[Product-of-Experts with Joint Head \cite{madaan2024jointly}] This variant extends the standard PoE by introducing an additional joint expert $g_{\text{joint}}$ that operates on the concatenated features of all modalities, capturing cross-modal interactions that unimodal experts cannot model.
  
  For ViT backbones, the joint head receives the token-level concatenation along the sequence dimension $\mathbf{Z}_{\text{cat}} = [\mathbf{Z}^{(1)};\, \mathbf{Z}^{(2)};\, \ldots;\, \mathbf{Z}^{(M)}] \in \mathbb{R}^{B \times (N \cdot M) \times D}$, processed by an Attentive Classifier. For non-ViT backbones, pool-level features are concatenated along the feature dimension $\mathbf{z}_{\text{cat}} = [\mathbf{z}^{(1)};\, \ldots;\, \mathbf{z}^{(M)}] \in \mathbb{R}^{D \cdot M}$, processed by a linear head.
  
  The joint expert produces logits $\boldsymbol{\ell}^{(\text{joint})} = g_{\text{joint}}(\mathbf{Z}_{\text{cat}})$ and participates in the PoE as an $(M{+}1)$-th expert that is always considered valid:
  
  $$p_{\text{PoE+Joint}}(y=c) \propto \left(\prod_{m=1}^{M} p_m(y=c \mid \mathbf{x}^{(m)})\right) \cdot p_{\text{joint}}(y=c \mid \mathbf{x}^{(1)}, \ldots, \mathbf{x}^{(M)})$$
  
  In log-space, with same definition on $\lambda$ as Product-of-Experts,
  $$\log p_{\text{PoE+Joint}}(y=c) = \left(\sum_{m=1}^{M} \tilde{\lambda}_c^{(m)} + \lambda_c^{(\text{joint})}\right) - \log \sum_{c'=1}^{C} \exp\!\left(\sum_{m=1}^{M} \tilde{\lambda}_{c'}^{(m)} + \lambda_{c'}^{(\text{joint})}\right)$$
  
  The joint expert captures multimodal synergies that arise only when multiple modalities are observed together, while the unimodal experts ensure that informative predictions can still be made when synergies are missing. 
  
\end{description}

\subsection{Model Pretraining Details}
\label{apd:pretrain_details}
We present our pretraining hyperparameter configuration in Supplementary \Cref{tab:pretrain_base_config,tab:pretrain_large_config} for our base model. The learning rate linearly increase from $1.0\times 10^{-4}$ to $5.25\times 10^{-4}$ in first $40$ epochs, stay on $5.25\times 10^{-4}$ for $160$ (ViT-Base) or $200$ (ViT-Large) epochs and then cooldown with cosine decay to $1.0\times 10^{-6}$ with $40$ (ViT-Base) or $80$ (ViT-Large) epochs. Three different block size sampling is performed on multiscale masking with masking size demonstrated in Masking Spatial Scale and Masking Depth Scale. 



\subsection{Model Evaluation Details}
\label{apd:model_eval_details}
To ensure a fair comparison across models and downstream tasks, we performed model-specific hyperparameter sweeps and reported the test-set performance obtained from the configuration selected by validation-set performance. This procedure accounts for the fact that different foundation models may require different optimization settings to achieve their best downstream transfer performance. For both unimodal and multimodal evaluations, we swept learning rate, weight decay and batch size for each model. For BrainIAC, we evaluated learning rates of $\{3\times10^{-4}, 1\times10^{-4}, 1\times10^{-3}\}$, weight decay values of $\{1\times10^{-5}, 1\times10^{-2}, 1\times10^{-1}\}$, epochs number of $\{15, 30\}$ and batch sizes of $\{16,32,64\}$. For VoCo, we evaluated learning rates of $\{1.5\times10^{-4}, 1.5\times10^{-3}\}$, weight decay values of $\{1\times10^{-2}, 1\times10^{-1}\}$ and a batch size of $4$. For NeuroVFM, we evaluated learning rates of $\{5\times10^{-4}, 1\times10^{-4}, 1\times10^{-3}\}$, weight decay values of $\{5\times10^{-2}, 1\times10^{-2}, 1\times10^{-1}\}$, epochs number of $\{30, 50\}$ and a batch size of $8$. For Neuro-JEPA, we evaluated learning rates of $\{1.5\times10^{-5}, 1.5\times10^{-4}, 3\times10^{-5}\}$, weight decay values of $\{1\times10^{-5}, 1\times10^{-2}, 1\times10^{-1}\}$, epochs number of $\{15, 30\}$ and batch sizes of $\{16,32,64\}$. For CNN, we we evaluated learning rates of $\{3\times10^{-4}, 1\times10^{-4}, 1\times10^{-3}\}$, weight decay values of $\{1\times10^{-5}, 1\times10^{-2}, 1\times10^{-1}\}$, epochs number of $\{30,50\}$ and batch sizes of $\{16,32,64\}$. The search space for VoCo was necessarily smaller because of its substantially higher computational cost. For VoCo and NeuroVFM, the batch sizes were constrained by the maximum feasible per-GPU memory capacity. For NeuroVFM, we consistently used multiple instance learning (MIL) pooling across all evaluations with frozen backbone following the optimal setup from original manuscript and the suggestions by consulting the authors. While full fine-tuning is also experimented for NeuroVFM, we observe diminishing or not improved performance across the tasks.

\subsection{Evaluation Scope}
\label{apd:evaluation_scope}
The objective of Neuro-JEPA is to learn image-level representations for clinically relevant tasks. We therefore benchmarked the model on diagnosis, prognosis, time-to-event and age prediction, and deliberately restricted claims to this intended use. We did not include segmentation or open-ended vision–language evaluation because these task families require distinct supervision, inference interfaces, baselines and validity criteria. Medical image segmentation is a dense-prediction problem with strong task-specific standards; recent benchmarking \cite{Isensee2021,nnunet-revisit} shows that fair segmentation claims require carefully configured baselines, adequate dataset diversity and resource-matched comparisons, since many apparent architectural gains disappear under rigorous validation. Moreover, dedicated segmentation foundation models such as MedSAM \cite{Ma2024}, BiomedParse \cite{Zhao2025}, and VISTA3D \cite{he25} are trained and evaluated with mask-level objectives and segmentation-specific metrics, making segmentation a separate methodological question rather than an auxiliary image-level diagnostic task. We also did not report generic medical VQA metrics because recent studies \cite{asadi2026mirageillusionvisualunderstanding,madaan2026multimodal,Gu2026} show that high Vision-Language Model (VLM) accuracy can persist when images are absent, blank or mismatched, and therefore may not establish causal use of visual information without explicit grounding controls. Thus, adding superficial segmentation or VQA experiments would expand the apparent scope of the study without providing a valid test of the proposed methods. We therefore view segmentation transfer and grounded vision–language reasoning as important future directions requiring dedicated protocols and claims.

\begin{table}[H]
\centering
\caption{Neuro-JEPA-Base pretraining configuration.}
\label{tab:pretrain_base_config}
\begin{tabular}{ll}
\toprule
\textbf{Parameter} & \textbf{Value} \\
\midrule
Architecture           & ViT-Base-MoE (128M total params w. 86M activated params) \\
Number of total experts & $16$ \\
Number of shared experts & $2$ \\
Number of activated experts & $6$ \\
Auxiliary loss free bias update & $1e$-$4$ \\
MoE score function & softmax \\
Decoder depth          & $4$ \\
Exponential Moving Average Ratio & $(0.99925, 0.99925)$ \\
Per-GPU batch          & $48$ \\
Number of GPUs         & NVIDIA L40s $\times$ 12  \\
Gradient accumulation  & 1 $\rightarrow$ global batch $= 576$ \\
Iteration per epoch    & $\lceil 921,600/576 \rceil = 1600$ \\
Warmup epochs          & $\lceil 64,000/1600 \rceil = 40$ \\
Scheduling epochs      & $\lceil 256,000/1600 \rceil = 160$ \\
Cooldown epochs        & $\lceil 64,000/1600 \rceil = 40$ \\
Start learning rate & $1.0 \times 10^{-4}$ \\
Max learning rate & $5.25 \times 10^{-4}$ \\
Min learning rate & $1.0 \times 10^{-6}$ \\
Weight decay           & $0.04$ \\
Optimizer              & AdamW, $\beta = (0.9, 0.999)$ \\
Gradient Clip          & $3.0$ \\
Background Weight      & $0.1$ \\
Precision              & bfloat16 \\
Masking Aspect Ratio   & $(0.75, 1.5)$ \\
Masking Spatial Scale  & $(0.0, 0.2), (0.2, 0.5), (0.5, 0.7)$ \\
Masking Depth Scale    & $(0.0, 1.0), (0.0, 1.0), (0.0, 1.0)$ \\
\bottomrule
\end{tabular}
\end{table}

\begin{table}[H]
\centering
\caption{Neuro-JEPA-Large pretraining configuration.}
\label{tab:pretrain_large_config}
\begin{tabular}{ll}
\toprule
\textbf{Parameter} & \textbf{Value} \\
\midrule
Architecture           & ViT-Large-MoE (429M total params w. 303M activated params) \\
Number of total experts & $16$ \\
Number of shared experts & $2$ \\
Number of activated experts & $6$ \\
Auxiliary loss free bias update & $1e$-$4$ \\
MoE score function & softmax \\
Decoder depth          & $4$ \\
Exponential Moving Average Ratio & $(0.99925, 0.99925)$ \\
Per-GPU batch          & $30$ \\
Number of GPUs         & NVIDIA L40s $\times$ 12  \\
Gradient accumulation  & 1 $\rightarrow$ global batch $= 360$ \\
Iteration per epoch    & $\lceil 288,000/360 \rceil = 800$ \\
Warmup epochs          & $\lceil 32,000/800 \rceil = 40$ \\
Scheduling epochs      & $\lceil 160,000/800 \rceil = 200$ \\
Cooldown epochs        & $\lceil 64,000/800 \rceil = 80$ \\
Start learning rate & $1.0 \times 10^{-4}$ \\
Max learning rate & $5.25 \times 10^{-4}$ \\
Min learning rate & $1.0 \times 10^{-6}$ \\
Weight decay           & $0.04$ \\
Optimizer              & Muon, $\beta_{muon}=0.95$, $\beta_{adamw} = (0.9, 0.999)$ \\
Gradient Clip          & $0.0$ \\
Background Weight      & $0.1$ \\
Precision              & bfloat16 \\
Masking Aspect Ratio   & $(0.75, 1.5)$ \\
Masking Spatial Scale  & $(0.0, 0.2), (0.2, 0.5), (0.5, 0.7)$ \\
Masking Depth Scale    & $(0.0, 1.0), (0.0, 1.0), (0.0, 1.0)$ \\
\bottomrule
\end{tabular}
\end{table}

\newpage
\section{Unimodal Learning Experiments}
\label{apd:uni_sec}
\subsection{Average Performance Across Tasks}
\label{apd:avg_performance_unimodal}
Supplementary~\Cref{tab:unimodal_avg} reports mean unimodal performance averaged across all dataset--task--modality combinations within each cohort group: Public datasets (41 tasks), NYU Langone (30 tasks), NYU Long Island (30 tasks), BIND-MGH (45 tasks), time-to-event tasks (6 tasks), and brain-age prediction (OpenBHB; $n = 757$). Diagnosis and prognosis tasks are evaluated by AUROC and AUPRC; time-to-event tasks by C-index; and brain-age prediction by $R^{2}$, MAE, and RMSE. Across all cohorts and metrics, Neuro-JEPA consistently achieves the highest average performance relative to all baselines.

\begin{table*}[t]
\centering
\footnotesize
\setlength{\tabcolsep}{4pt}
\renewcommand{\arraystretch}{1.15}
\caption{%
\textbf{Average unimodal performance across cohorts and task types.}
Values are mean [95\% CI]. \textbf{Bold}: best per metric.
$\uparrow$~higher is better; $\downarrow$~lower is better. "combs" here refer to the number of combinations on dataset-task-modality. "n" for brain age represents number of samples in the test set. \underline{underlining} indicates the second-best model.
}
\label{tab:unimodal_avg}

\begin{tabular}{lcccccccc}
\toprule
& \multicolumn{2}{c}{\textbf{Public} (41 combs)}
& \multicolumn{2}{c}{\textbf{NYU} (30 combs)}
& \multicolumn{2}{c}{\textbf{LI} (30 combs)}
& \multicolumn{2}{c}{\textbf{MGH} (45 combs)} \\
\cmidrule(lr){2-3}\cmidrule(lr){4-5}\cmidrule(lr){6-7}\cmidrule(lr){8-9}
\textbf{Model}
& AUROC$\uparrow$ & AUPRC$\uparrow$
& AUROC$\uparrow$ & AUPRC$\uparrow$
& AUROC$\uparrow$ & AUPRC$\uparrow$
& AUROC$\uparrow$ & AUPRC$\uparrow$ \\
\midrule
VoCo
& 0.721 {\tiny[0.689, 0.754]} & 0.573 {\tiny[0.513, 0.635]}
& 0.762 {\tiny[0.726, 0.793]} & 0.338 {\tiny[0.265, 0.412]}
& 0.735 {\tiny[0.702, 0.764]} & \underline{0.295} {\tiny[0.240, 0.354]}
& 0.717 {\tiny[0.693, 0.741]} & 0.233 {\tiny[0.196, 0.269]} \\
BrainIAC
& 0.728 {\tiny[0.705, 0.751]} & 0.555 {\tiny[0.494, 0.618]}
& 0.730 {\tiny[0.697, 0.761]} & 0.296 {\tiny[0.226, 0.370]}
& 0.674 {\tiny[0.643, 0.704]} & 0.212 {\tiny[0.162, 0.266]}
& 0.655 {\tiny[0.636, 0.673]} & 0.179 {\tiny[0.148, 0.211]} \\
NeuroVFM
& \underline{0.741} {\tiny[0.706, 0.774]} & \underline{0.585} {\tiny[0.526, 0.648]}
& \underline{0.772} {\tiny[0.743, 0.801]} & \underline{0.343} {\tiny[0.271, 0.422]}
& \underline{0.739} {\tiny[0.713, 0.762]} & 0.217 {\tiny[0.175, 0.260]}
& \underline{0.725} {\tiny[0.705, 0.745]} & \underline{0.237} {\tiny[0.199, 0.277]} \\
\rowcolor{gray!12}
\textbf{Neuro-JEPA-B}
& 0.785 {\tiny[0.760, 0.811]} & 0.649 {\tiny[0.588, 0.705]}
& 0.825 {\tiny[0.796, 0.850]} & 0.457 {\tiny[0.382, 0.535]}
& 0.806 {\tiny[0.780, 0.828]} & 0.383 {\tiny[0.322, 0.442]}
& 0.741 {\tiny[0.720, 0.761]} & 0.253 {\tiny[0.216, 0.292]} \\
\rowcolor{gray!12}
\textbf{Neuro-JEPA-L}
& \textbf{0.801} {\tiny[0.775, 0.828]} & \textbf{0.687} {\tiny[0.631, 0.740]}
& \textbf{0.838} {\tiny[0.812, 0.862]} & \textbf{0.498} {\tiny[0.430, 0.567]}
& \textbf{0.810} {\tiny[0.779, 0.838]} & \textbf{0.407} {\tiny[0.344, 0.465]}
& \textbf{0.748} {\tiny[0.726, 0.770]} & \textbf{0.277} {\tiny[0.239, 0.317]} \\
\bottomrule
\end{tabular}

\smallskip

\begin{tabular}{lcccc}
\toprule
& \multicolumn{1}{c}{\textbf{Time-to-Event} (6 combs)}
& \multicolumn{3}{c}{\textbf{Brain Age} ($n = 757$)} \\
\cmidrule(lr){2-2}\cmidrule(lr){3-5}
\textbf{Model}
& C-index$\uparrow$
& $R^{2}\uparrow$ & MAE$\downarrow$ (yr) & RMSE$\downarrow$ (yr) \\
\midrule
VoCo
& \underline{0.663} {\tiny[0.616, 0.699]}
& 0.111 {\tiny[0.065, 0.165]} & 6.22 {\tiny[5.46, 6.94]} & 12.03 {\tiny[10.54, 13.36]} \\
BrainIAC
& 0.650 {\tiny[0.615, 0.686]}
& 0.522 {\tiny[0.447, 0.591]} & 5.42 {\tiny[4.93, 5.95]} & 8.82 {\tiny[7.77, 9.78]} \\
NeuroVFM
& 0.629 {\tiny[0.584, 0.673]}
& \underline{0.673} {\tiny[0.611, 0.735]} & \underline{4.36} {\tiny[3.91, 4.77]} & \underline{7.29} {\tiny[6.35, 8.12]} \\
\rowcolor{gray!12}
\textbf{Neuro-JEPA-B}
& 0.695 {\tiny[0.664, 0.718]}
& \textbf{0.894} {\tiny[0.860, 0.917]} & 2.78 {\tiny[2.56, 3.02]} & \textbf{4.15} {\tiny[3.69, 4.70]} \\
\rowcolor{gray!12}
\textbf{Neuro-JEPA-L}
& \textbf{0.705} {\tiny[0.6501, 0.754]}
& 0.893 {\tiny[0.862, 0.915]} & \textbf{2.74} {\tiny[2.50, 2.95]} & 4.18 {\tiny[3.73, 4.61]} \\
\bottomrule
\end{tabular}

\end{table*}

\subsection{AUPRC for Public Datasets and Best Achievable Unimodal Performance Across Datasets}
\begin{figure}[htbp]
    \centering
    \includegraphics[width=0.85\textwidth]{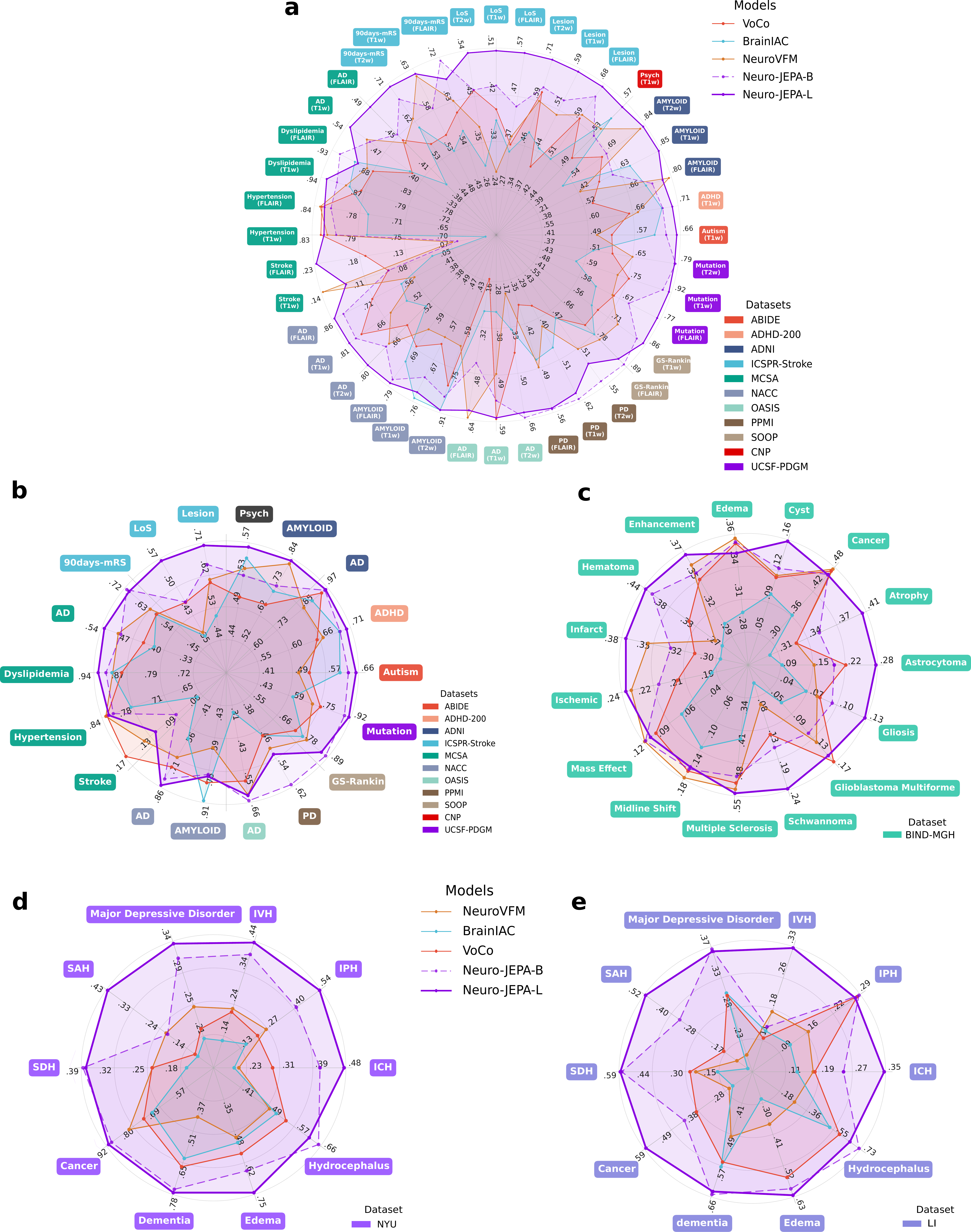}
    \caption{\textbf{Per Task AUPRC Across Public Dataset Tasks and AUPRC for Best Achievable Unimodal Performance.} AUPRC for public datasets on each unimodal performance and best achiable AUPRC performance across datasets and tasks. The result shows that our model demonstrates consistent performance improvement in comparison to other foundation models on AUPRC. \textbf{a,} AUPRC performance on different tasks with different modalities on public datasets. \textbf{b,} AUPRC performance for tasks with best achievable modalities on public datasets. \textbf{c,} AUPRC performance for tasks with best achievable modalities on BIND-MGH dataset.  \textbf{d,} AUPRC performance for tasks with best achievable modalities on NYU Langone dataset.  \textbf{e,} AUPRC performance for tasks with best achievable modalities on NYU Longisland dataset.}
    \label{fig:sup-leftover-ap}
\end{figure}

Supplementary \Cref{fig:sup-leftover-ap} presents AUPRC for (a) model performance across different modalities on public datasets for different evaluated models (b-c) all evaluated models on best achievable unimodal performance across all evaluated tasks and datasets (Public datasets, NYU-Langone, NYU-Longisland and BIND-MGH). 

\subsection{AUROC and AUPRC for Health System Datasets with Per Modality Performance}


Supplementary \Cref{fig:unimodal-all-auc} presents detailed unimodal performance on all modalities (T1w, T2w, FLAIR) across clinical cohorts datasets (NYU-Langone, NYU-Longisland, BIND-MGH). The result is present with both AUROC and AUPRC. The result shows that Neuro-JEPA consistently perform among the best across tasks and modalities.

\begin{figure}[htbp]
    \centering
    \includegraphics[width=0.85\textwidth]{figures/Supplementary-Unimodal-All/unimodal_all_comparison_combined.pdf}
    \caption{\textbf{Unimodal Per Task AUROC and AUPRC Across Three Health System Datasets.} AUROC and AUPRC with per modality performance for each task on NYU Langone, NYU Longisland and BIND-MGH datasets across all evaluated foundation models. All tasks are evaluated by full fine-tuning. Our model show improved performance across majority of tasks on different modalities \textbf{a,} AUROC for NYU Langone dataset. \textbf{b,} AUROC for NYU Longisland dataset. \textbf{c,} AUROC for BIND-MGH dataset. \textbf{d,} AUPRC for NYU Langone dataset. \textbf{e,} AUPRC for NYU Longisland dataset. \textbf{f,} AUPRC for BIND-MGH dataset.}
    \label{fig:unimodal-all-auc}
\end{figure}

\subsection{Kaplan Meier Curve and C-index for Time-to-Event Tasks}
\begin{figure}[htbp]
    \centering
    \includegraphics[width=1.0\textwidth]{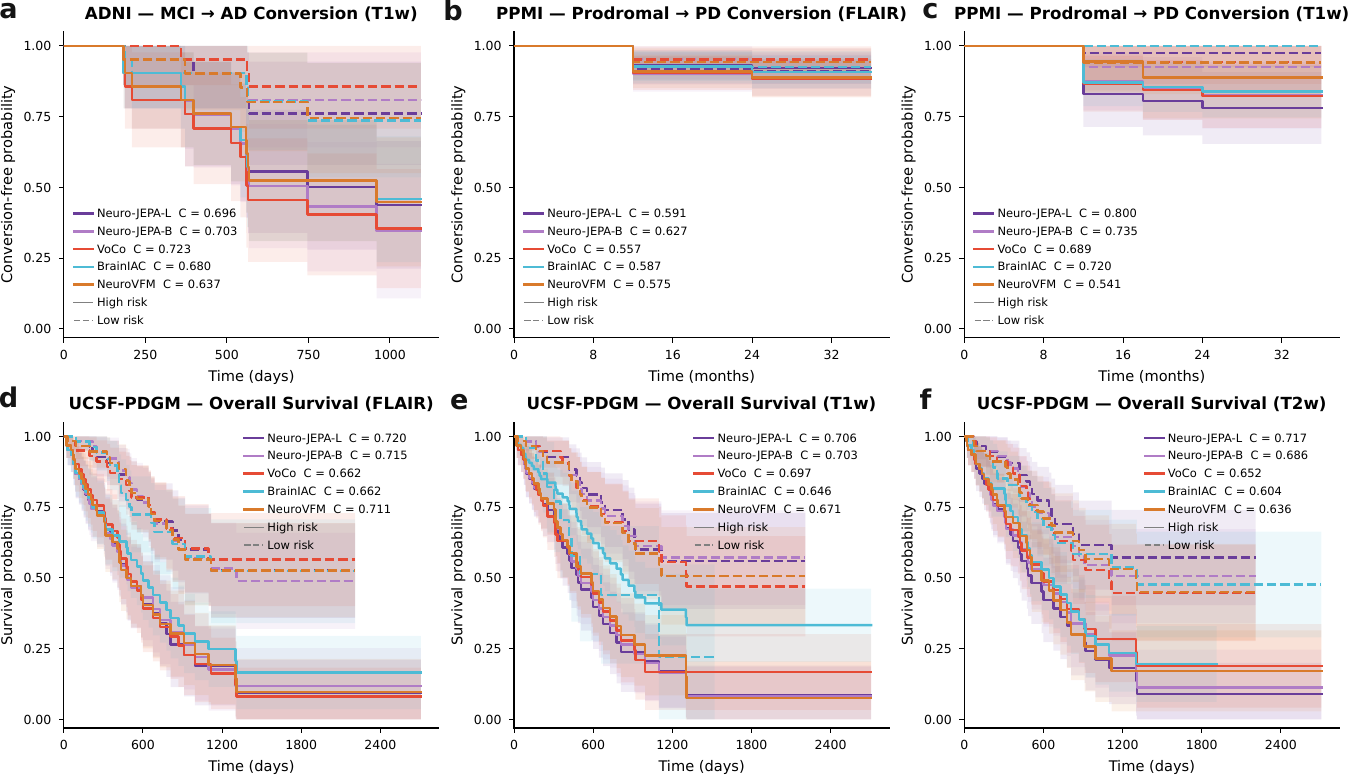}
    \caption{\textbf{Kaplan Meier Curve for Time-to-Event on All Modalities.} result is reported with Concordance Index (C-index) Prodromal to PD conversion for PPMI dataset and Overall Survival for UCSF-PDGM dataset. \textbf{a,} MCI to AD conversion within 3 years for ADNI dataset with T1w. \textbf{b,c,} Prodromal to PD conversion within 3 years for PPMI dataset with T1w and FLAIR. \textbf{e-g,} Overall Survival for UCSF-PDGM dataset with T1w, T2w and FLAIR.}
    \label{fig:sup-kaplan_meier}
\end{figure}

Supplementary \Cref{fig:sup-kaplan_meier} presents Kaplan Meier Curve for evaluated time-to-event tasks on all modalities. 

\subsection{Age Prediction Performance as Regression}
\begin{figure}[htbp]
    \centering
    \includegraphics[width=0.65\textwidth]{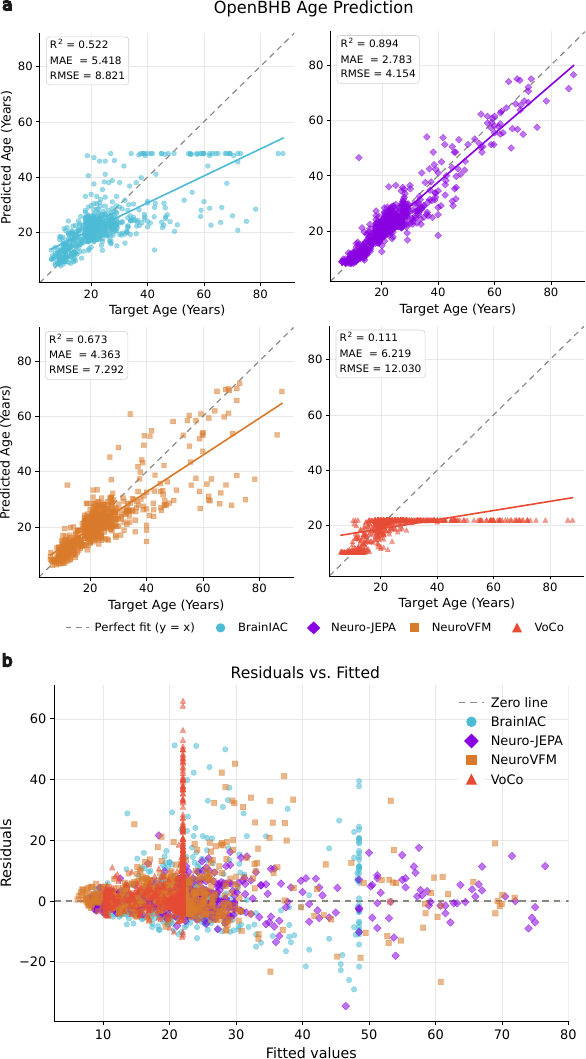}
    \caption{\textbf{Age Prediction Comparison on OpenBHB Dataset.} Age prediction as regression on OpenBHB dataset on Quasi-Raw T1w scans with performance reported on $R^2$, Mean Absolute Error (MAE) and Rooted Mean Squared Error (RMSE). The result shows that our model outperform existing foundation models especially with stronger fitting on patients with older age. \textbf{a,} regression goodness of fit for each individual foundation models. \textbf{b,} residuals vs. fitted values for all foundation models in one plot.}
    \label{fig:sup-openbhb}
\end{figure}

\begin{figure}[htbp]
    \centering
    \includegraphics[width=0.6\textwidth]{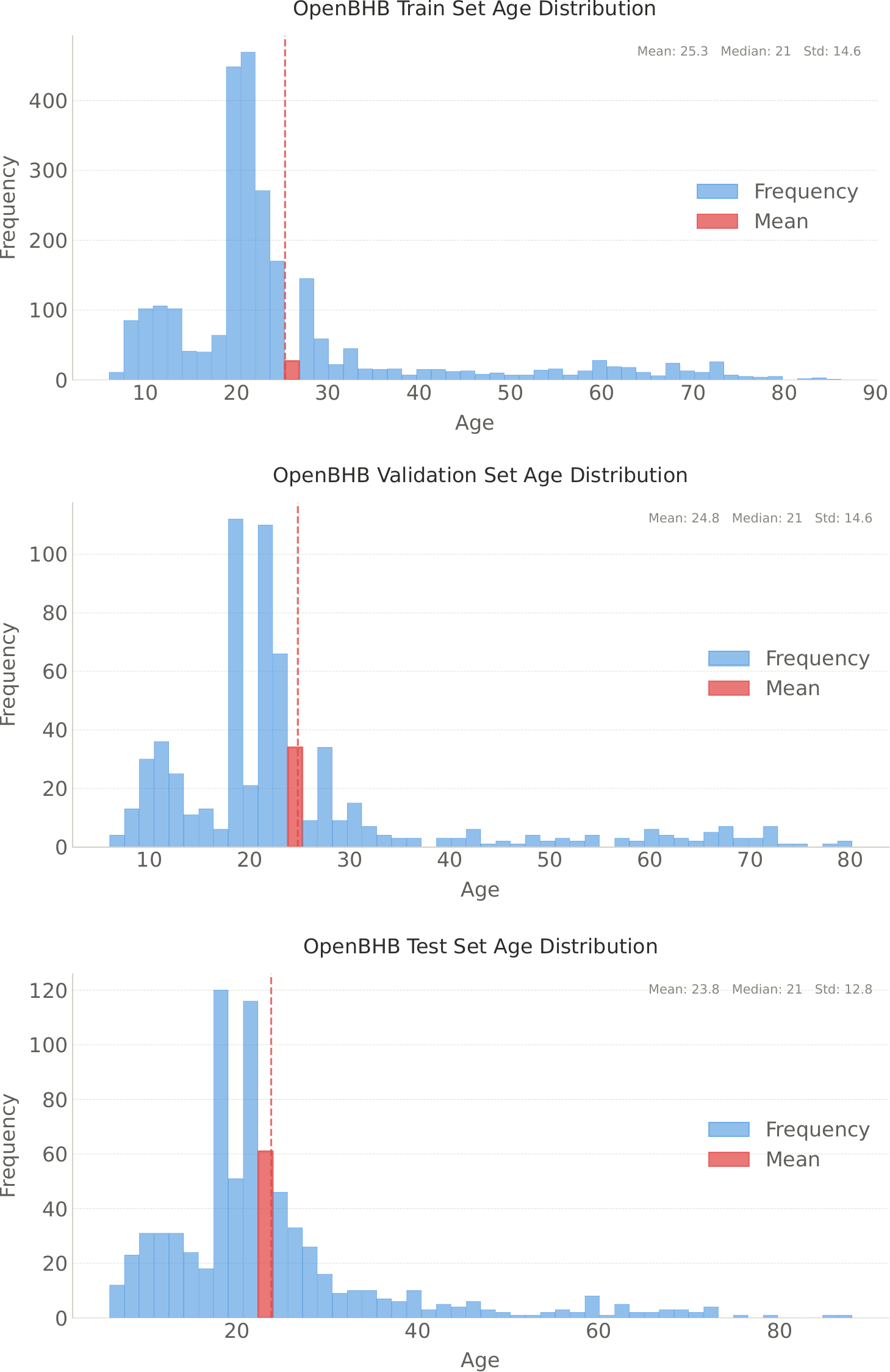}
    \caption{\textbf{Age Distribution on OpenBHB Dataset.} We show age distribution on train, validation and test set on OpenBHB dataset. As it demonstrates, the dataset presents heavy long-tailed distribution on elder age, where the evaluated models not training on large scale clinical dataset fail to generalize.}
    \label{fig:sup-openbhb-dist}
\end{figure}

Supplementary \Cref{fig:sup-openbhb} reports age regression on the healthy OpenBHB cohort for all evaluated models, using quasi-raw scans. Neuro-JEPA generalizes best across $R^2$, MAE, and RMSE. Notably, models pre-trained without clinical neuroimaging data (BrainIAC, VoCo) fail to generalize beyond a certain age, corresponding to the long tail of the OpenBHB age distribution (Supplementary \Cref{fig:sup-openbhb-dist}), whereas Neuro-JEPA and NeuroVFM do not—underscoring the value of large-scale clinical pre-training. (NeuroVFM originally reports ~2.8 years MAE, which we could not reproduce under our protocol of quasi-raw scans and predefined splits; since its preprocessing and splits are unspecified, we report our reproduced numbers to keep the evaluation setting consistent across all models.)

\newpage
\section{Multi-Modal Learning Experiments}
\label{apd:mm_sec}

\subsection{Average Performance Across Tasks}
\label{apd:avg_performance_multimodal}
Supplementary~\Cref{tab:multimodal_avg} reports mean performance for dataset-task combinations (combs) in which two modalities were jointly available, specifically T1w$+$T2w and T1w$+$FLAIR. The result is averaged across all dataset--task combinations within each cohort (Public datasets, 12~tasks; BIND-MGH, 30~tasks). Performance is evaluated by AUROC and AUPRC. Neuro-JEPA achieves the highest average AUROC and AUPRC in both cohorts, demonstrating that its representational advantage extends to multimodal settings.

\begin{table}[t]
\centering
\footnotesize
\setlength{\tabcolsep}{4pt}
\renewcommand{\arraystretch}{1.15}
\caption{%
\textbf{Average multimodal performance across cohorts.}
Values are mean [95\% CI]. \textbf{Bold}: best per metric.
$\uparrow$~higher is better. The number of tasks here refer to the number of combination on dataset-task-modality. \underline{underlining} indicates the second-best model.
}
\label{tab:multimodal_avg}
\begin{tabular}{lcccc}
\toprule
& \multicolumn{2}{c}{\textbf{Public} (12 combs)}
& \multicolumn{2}{c}{\textbf{MGH} (30 combs)} \\
\cmidrule(lr){2-3}\cmidrule(lr){4-5}
\textbf{Model}
& AUROC$\uparrow$ & AUPRC$\uparrow$
& AUROC$\uparrow$ & AUPRC$\uparrow$ \\
\midrule
VoCo
& 0.743 {\tiny[0.698, 0.789]} & 0.562 {\tiny[0.443, 0.673]}
& 0.729 {\tiny[0.701, 0.757]} & 0.241 {\tiny[0.194, 0.288]} \\
BrainIAC
& 0.730 {\tiny[0.693, 0.766]} & 0.552 {\tiny[0.428, 0.673]}
& 0.684 {\tiny[0.662, 0.707]} & 0.203 {\tiny[0.160, 0.245]} \\
NeuroVFM
& \underline{0.748} {\tiny[0.684, 0.804]} & \underline{0.574} {\tiny[0.449, 0.685]}
& \underline{0.742} {\tiny[0.721, 0.765]} & \underline{0.255} {\tiny[0.205, 0.305]} \\
\rowcolor{gray!12}
\textbf{Neuro-JEPA-B}
& \textbf{0.805} {\tiny[0.759, 0.849]} & \textbf{0.637} {\tiny[0.505, 0.749]}
& \textbf{0.763} {\tiny[0.739, 0.789]} & \textbf{0.295} {\tiny[0.248, 0.343]} \\
\bottomrule
\end{tabular}
\end{table}

\subsection{Multi-Modal Learning Result Across Fusion Methods - Public Datasets}
\begin{figure}[htbp]
    \centering
    \includegraphics[height=0.40\textwidth]{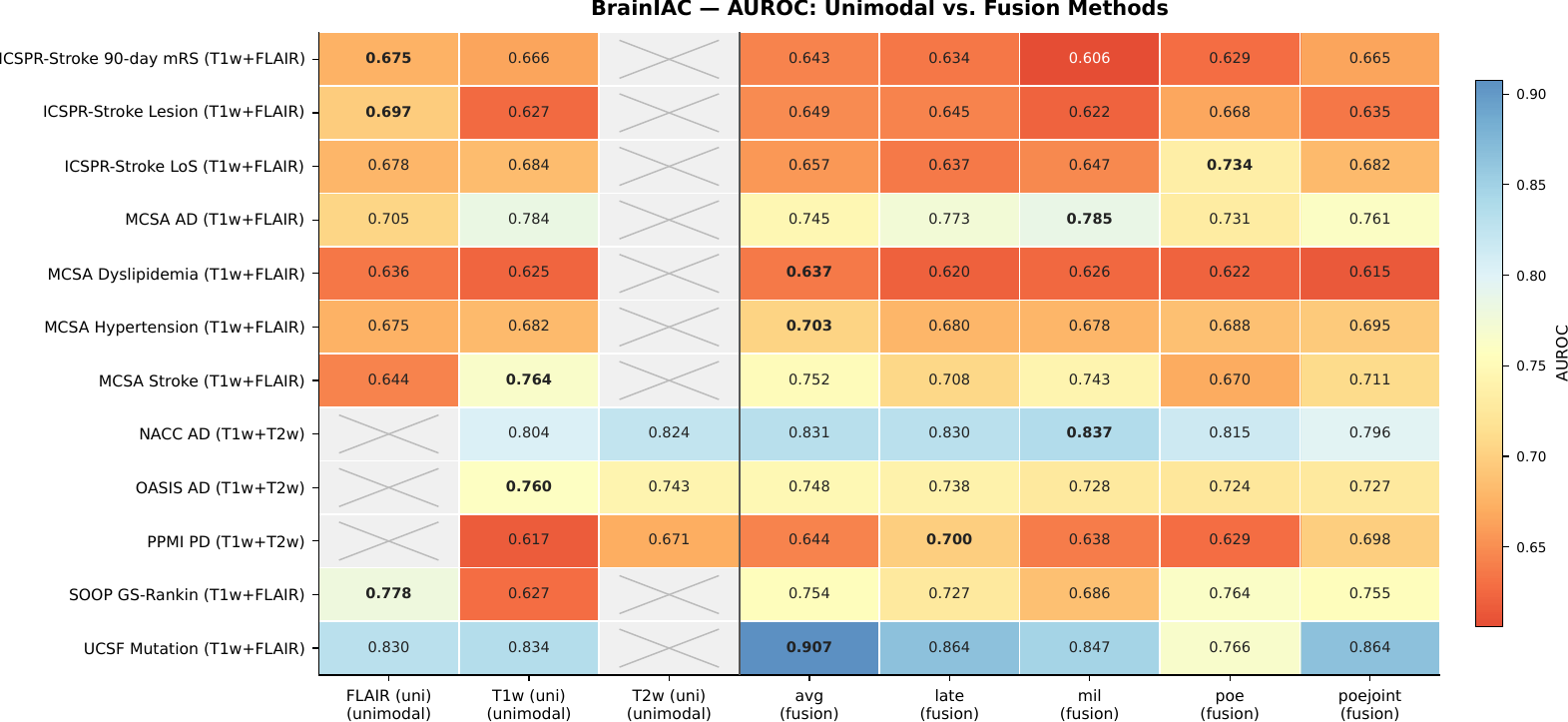} \\
    \vspace{0.03\textwidth}
    \includegraphics[height=0.40\textwidth]{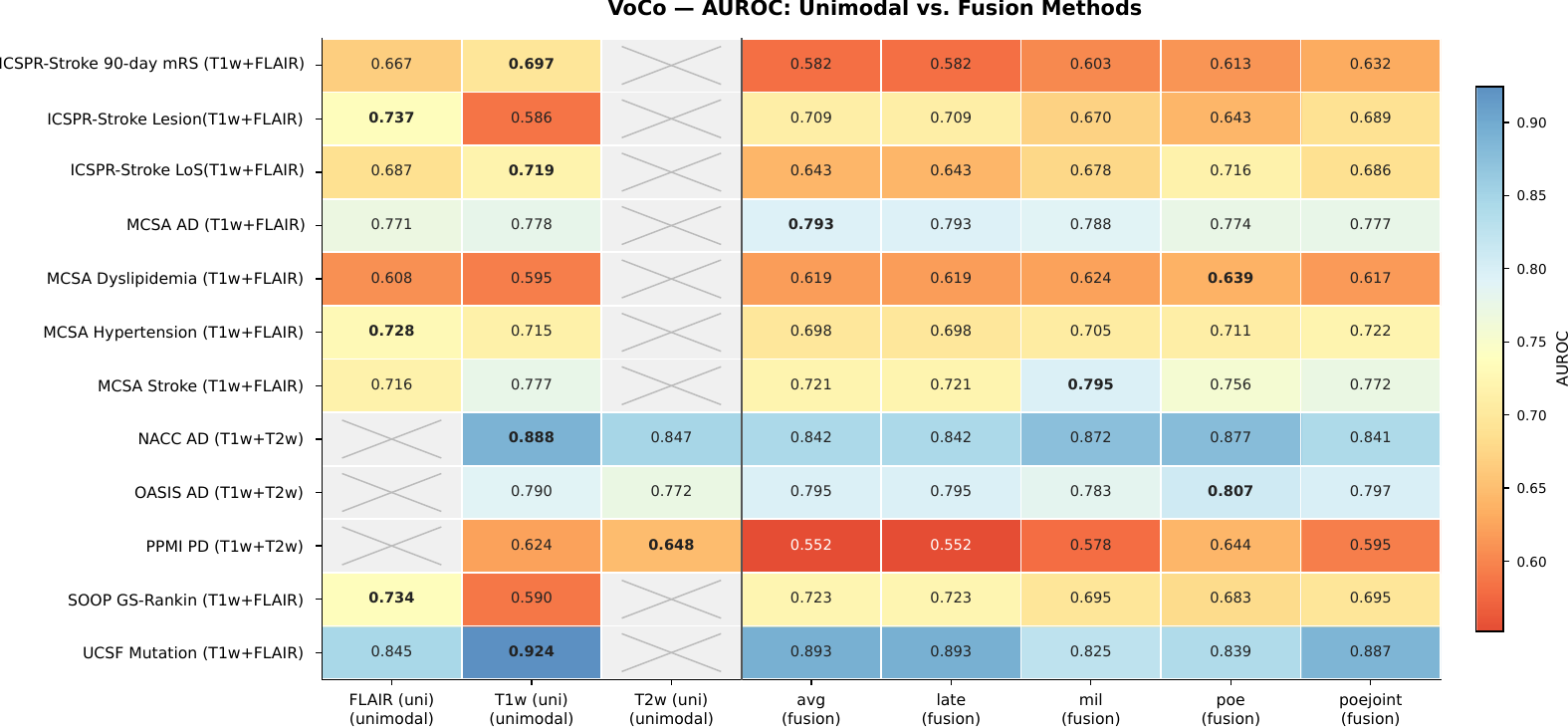} \\
    \vspace{0.03\textwidth}
    \includegraphics[height=0.40\textwidth]{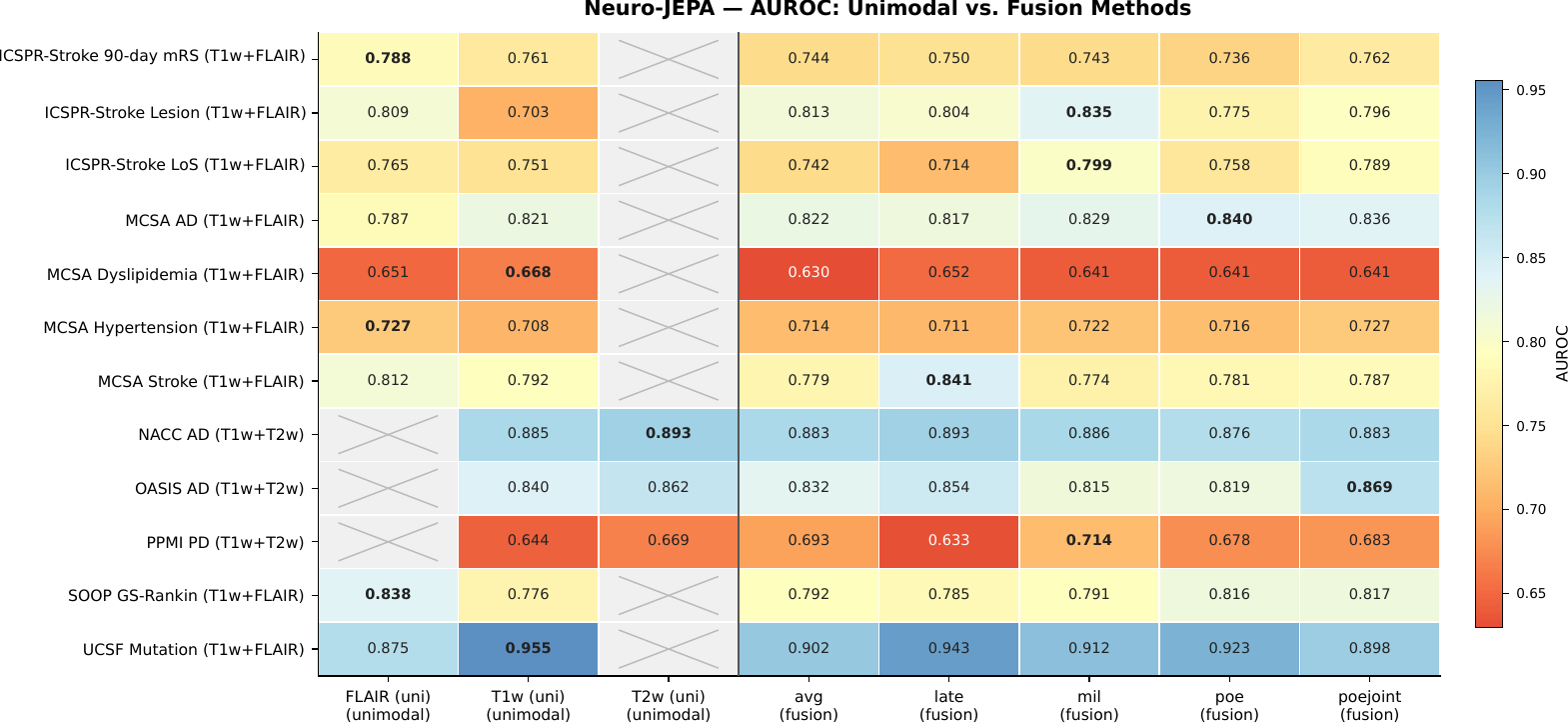}
    \caption{\textbf{Multimodal Learning Performance Across Fusion Methods for BrainIAC, VoCo and Neuro-JEPA on Public Datasets - AUROC.}}
    \label{fig:mm_methods_perf_fm}
\end{figure}

\begin{figure}[htbp]
    \centering
    \includegraphics[height=0.40\textwidth]{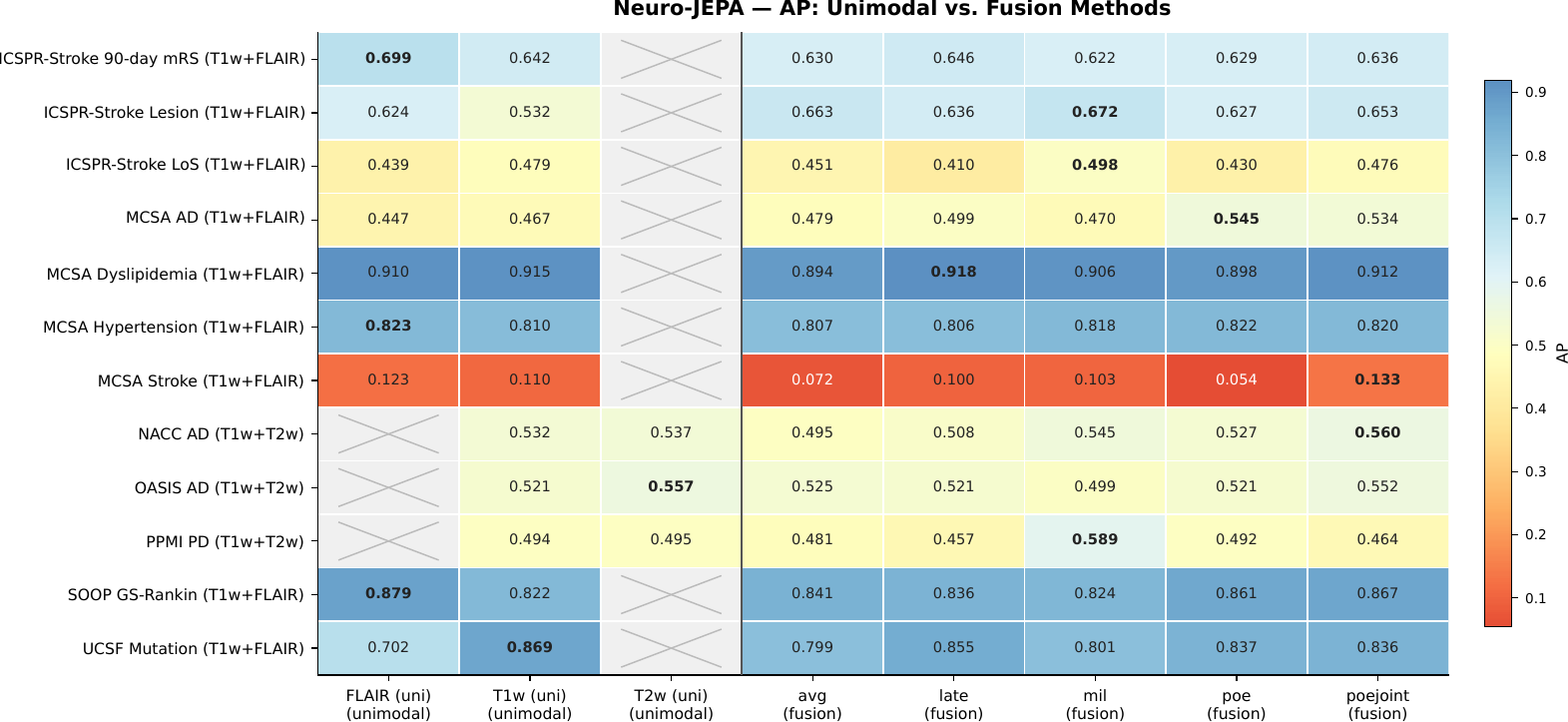} \\
    \vspace{0.03\textwidth}
    \includegraphics[height=0.40\textwidth]{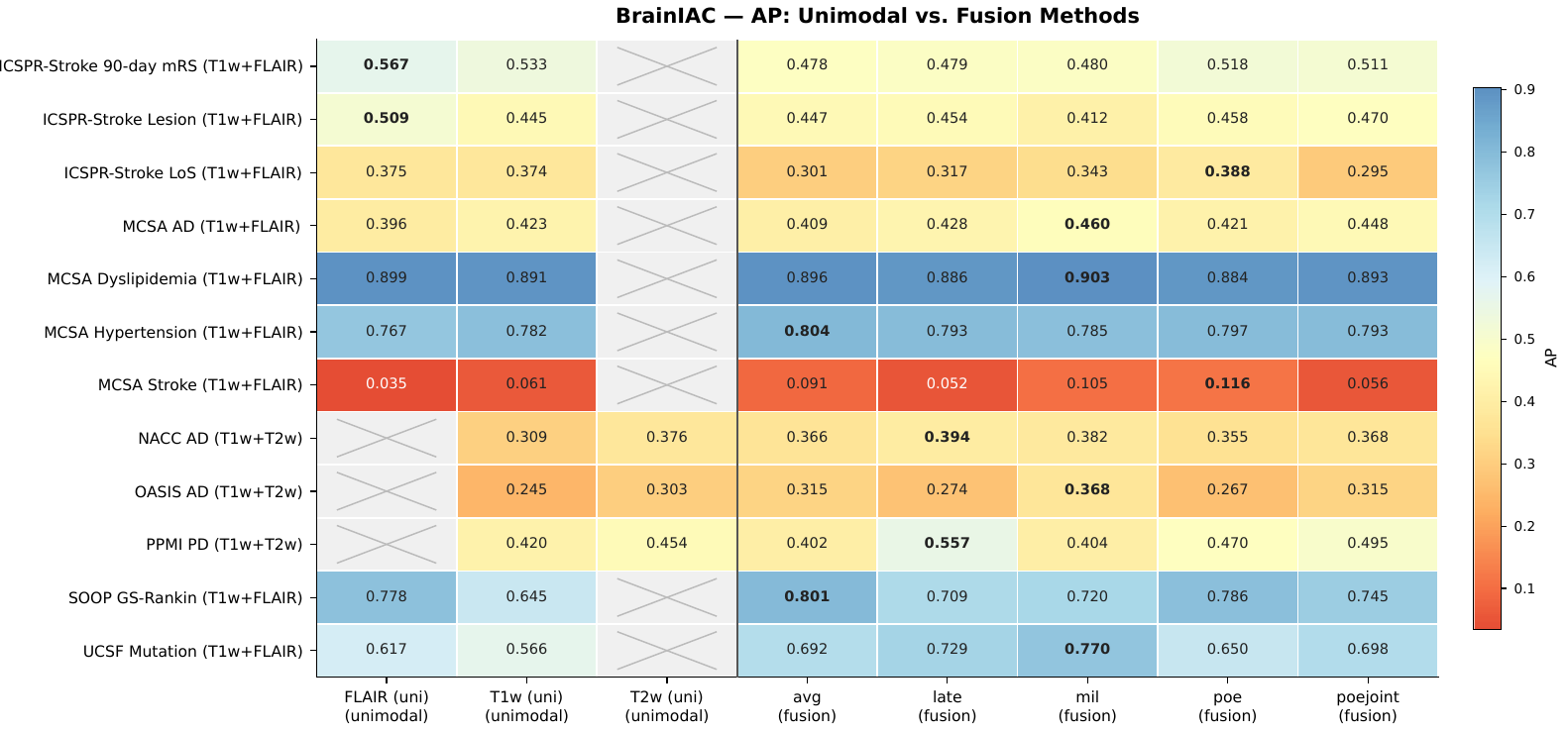} \\
    \vspace{0.03\textwidth}
    \includegraphics[height=0.40\textwidth]{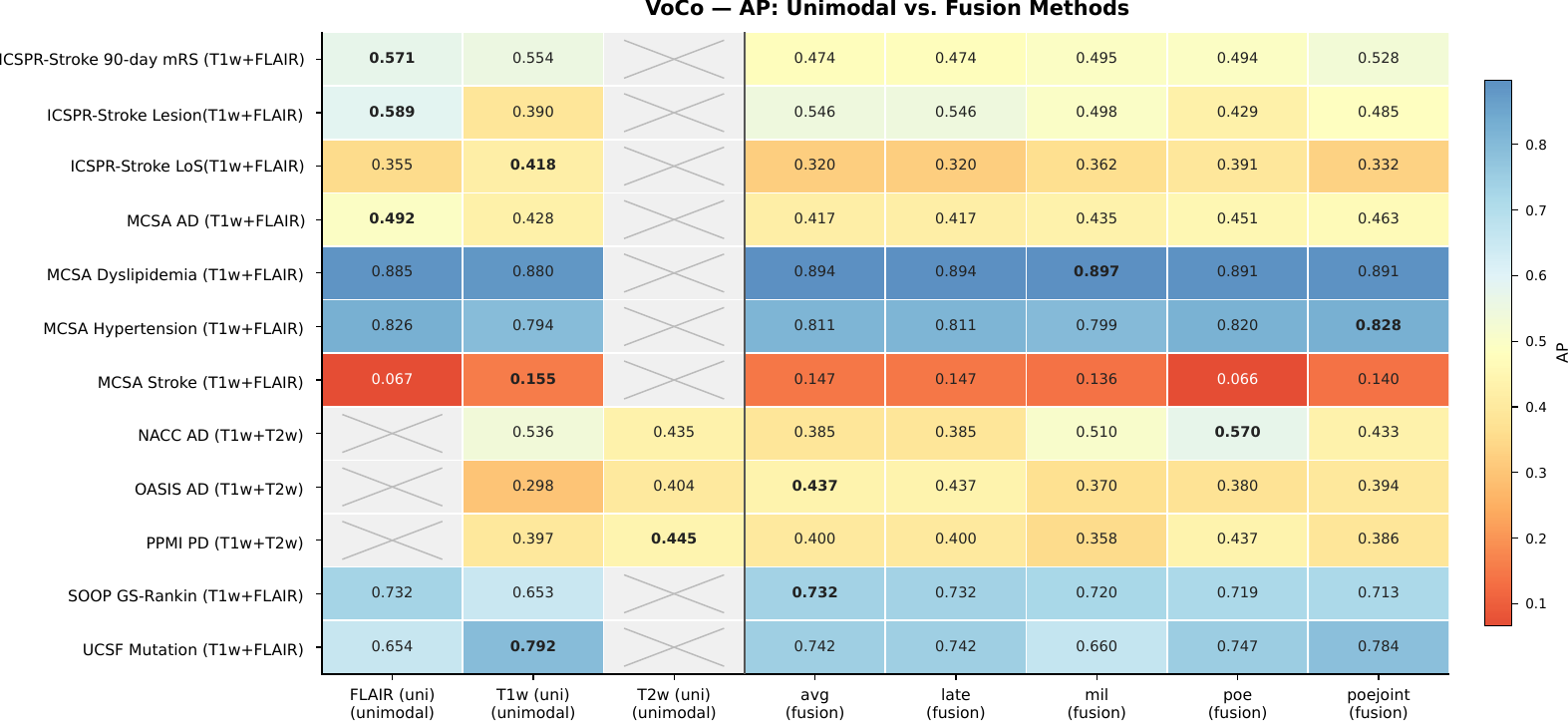} \\
    \caption{\textbf{Multimodal Learning Performance Across Fusion Methods for BrainIAC, VoCo and Neuro-JEPA on Public Datasets - AP.}}
    \label{fig:mm_methods_perf_fm_cont}
\end{figure}

\begin{figure}[htbp]
    \centering
    \includegraphics[height=1.0\textwidth]{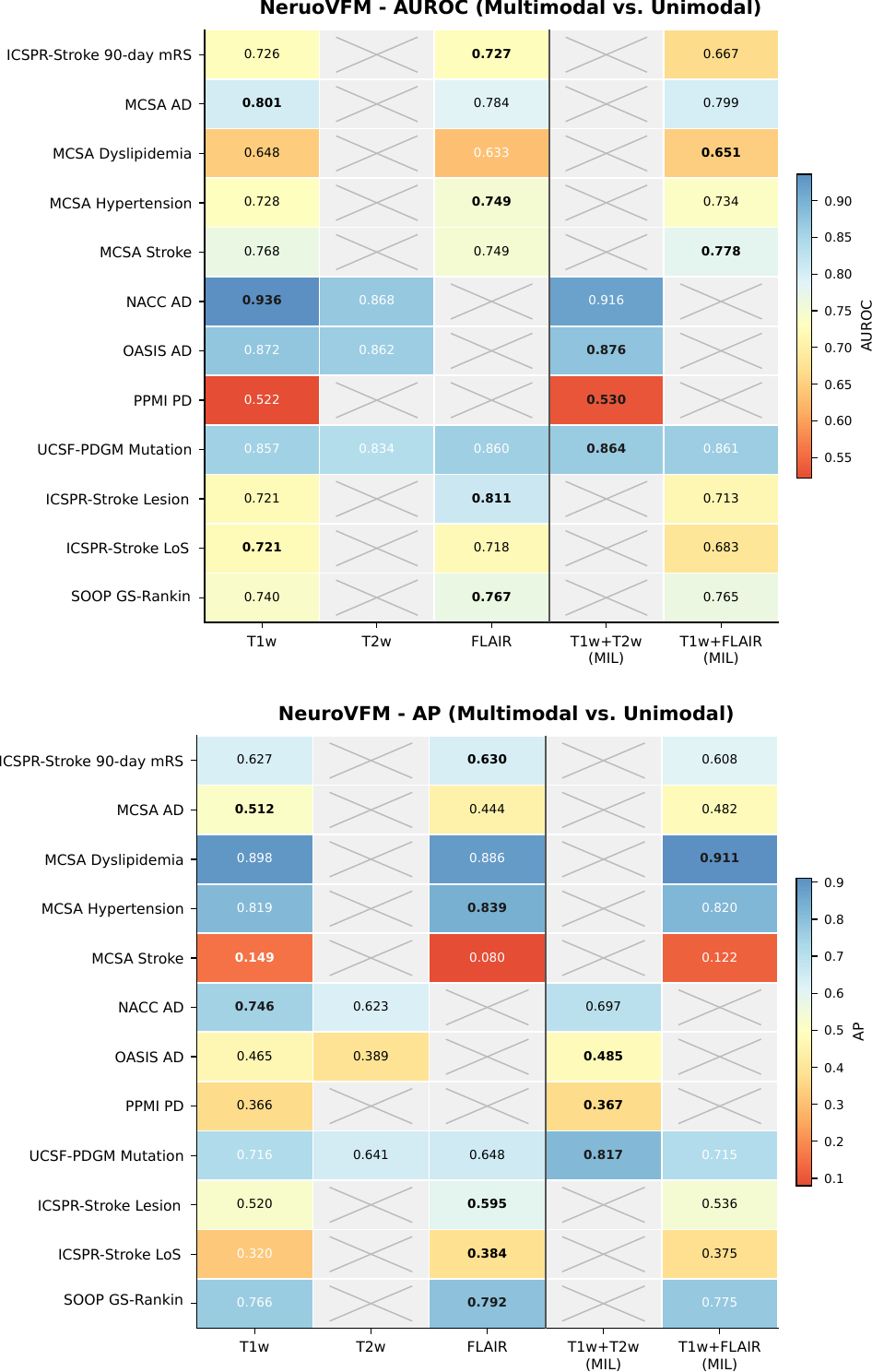} \\
    \caption{\textbf{Multimodal Learning Performance Across Fusion Methods for NeuroVFM on Public Datasets - AUROC and AP.} AUROC and AUPRC for unimodal and multimodal performance for NeuroVFM. The best suggested method (MIL) from original paper for multimodal fusion is applied in this evaluation.}
    \label{fig:mm_methods_perf_fm_neurovfm}
\end{figure}

Supplementary \Cref{fig:mm_methods_perf_fm,fig:mm_methods_perf_fm_cont,fig:mm_methods_perf_fm_neurovfm} present multi-modal learning performance on different models across different fusion methods on different public datasets tasks and multi-modal combinations (T1w+T2w or T1w+FLAIR) reported in AUROC and AUPRC. Each row in the plot shows present modalities for multimodal fusion and the corresponding performance on different fusion methods. The method with best performance is selected for the main evaluation. The result shows that Neuro-JEPA present higher multi-modal gain over other models in majority of the tasks.

\subsection{Multi-Modal Gain Result - Public Datasets}
\begin{figure}[htbp]
    \centering
    \includegraphics[width=1.0\textwidth]{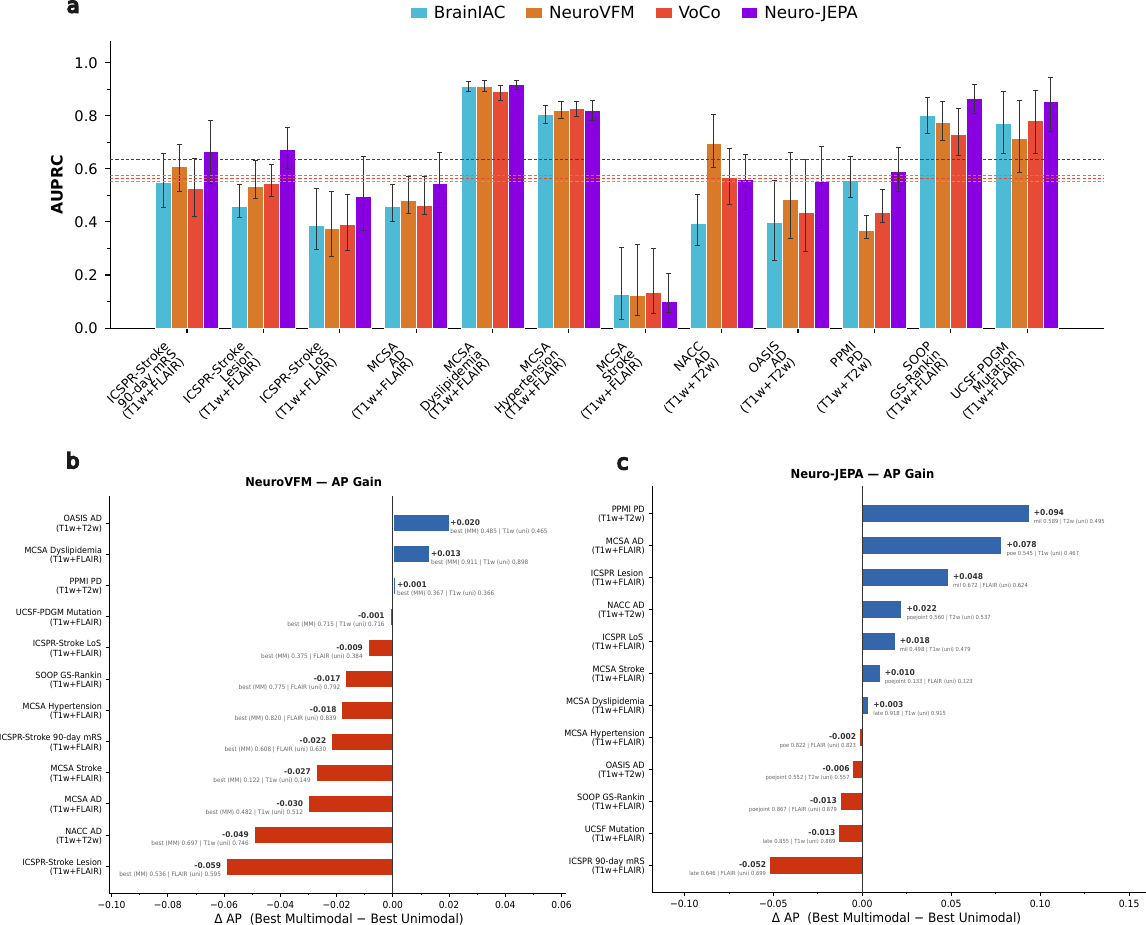}
    \caption{\textbf{Multimodal Performance and Gain Over Unimodal on AUPRC.} We report AUPRC on multimodal performance when two different modalities are combined and multimodal performance gain over unimodal defined as the difference between best multimodal combination and best unimodal performance. \textbf{a,} Multimodal performance on AUPRC across selected tasks on public datasets for all four compared foundation models. The result is reported by best performance multimodal fusion method among five different methods. Dotted horizontal line present average performance across tasks, where the result shows our model outperforms other foundation models with a large margin. \textbf{b,c,} Multimodal gain on AUPRC for best previous foundation model (NeuroVFM) and ours. The result shows that our model present better performance gain under multimodal fusion over previous model.}
    \label{fig:sup-mm-auprc}
\end{figure}

Supplementary \Cref{fig:sup-mm-auprc} present (a) performance on best multimodal method across tasks for each model reported in AUPRC. (b,c) multi-modal learning gain over uni-modal on public datasets for NeuroVFM and Neuro-JEPA reported reported in AUPRC.

\begin{figure}[htbp]
    \centering
    \includegraphics[width=1.0\textwidth]{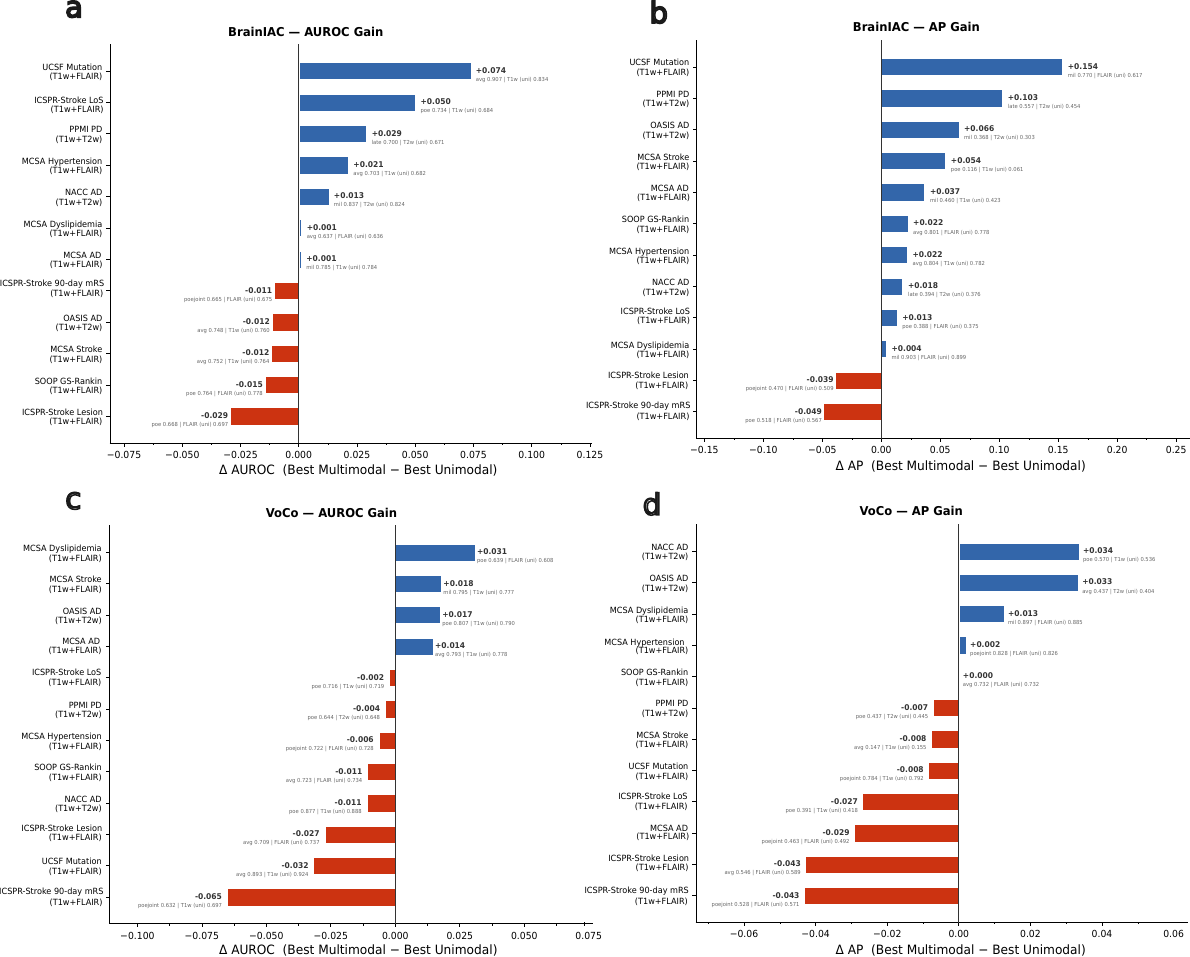}
    \caption{\textbf{Mutlimodal Gain Over Unimodal for BrainIAC and VoCo.} The difference between best multimodal fusion vs. best unimodal performance on AUROC and AUPRC. \textbf{a,b,} AUROC and AUPRC multimodal performance gain on the difference for BrainIAC. \textbf{c,d,} AUROC and AUPRC multimodal performance gain on the difference for VoCo.}
    \label{fig:sup-mm-brainiac-voco}
\end{figure}

Supplementary \Cref{fig:sup-mm-brainiac-voco} represent multi-modal learning gain over uni-modal on public datasets for BrainIAC and VoCo reported reported in both AUROC and AUPRC.

\subsection{Multi-Modal Learning Result Across Fusion Methods - BIND-MGH}
\begin{figure}[htbp]
    \centering
    \includegraphics[width=1.0\textwidth]{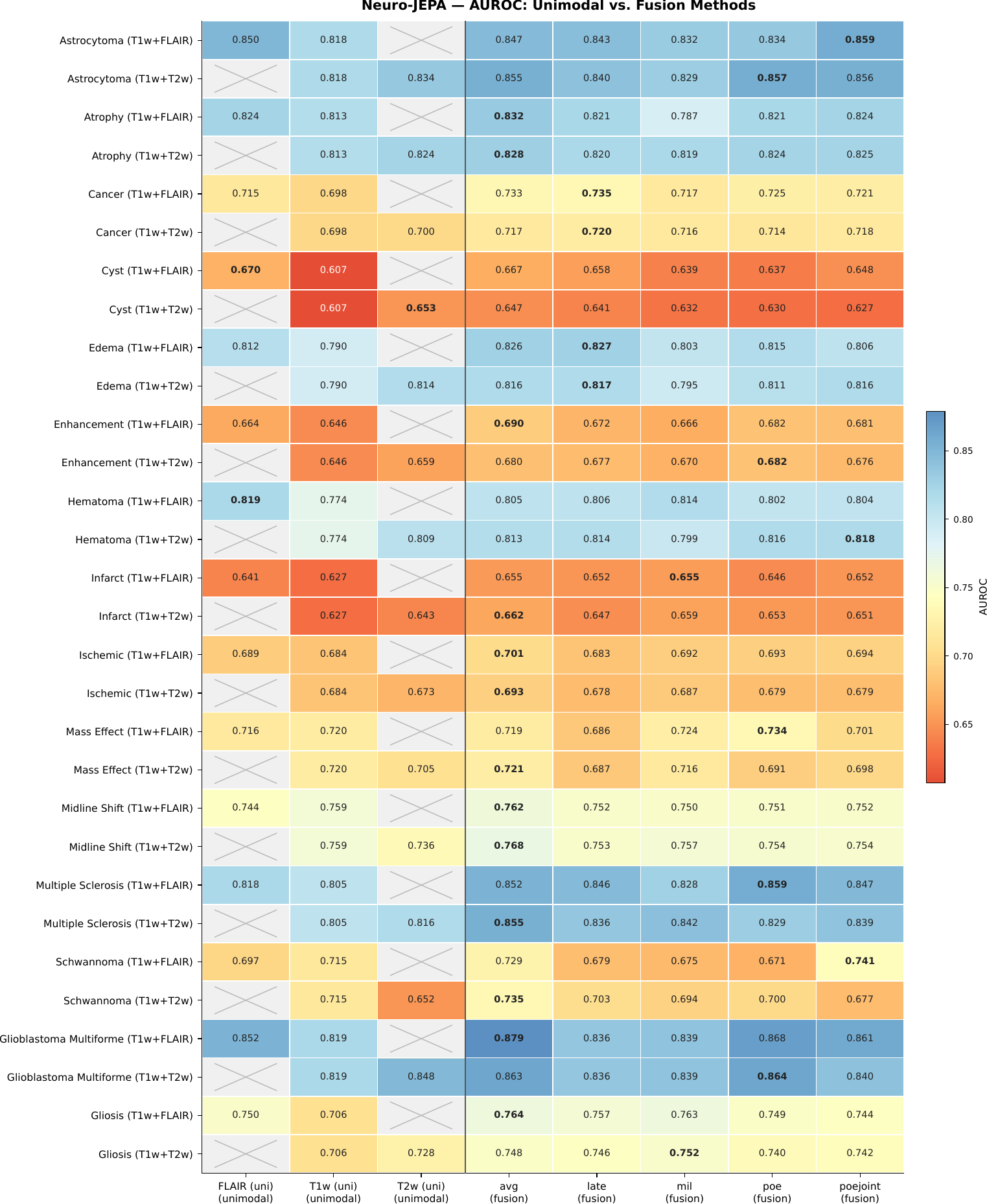}
    \caption{\textbf{Multimodal Learning Performance Across Fusion Methods for Neuro-JEPA on BIND-MGH - AUROC.}}
    \label{fig:sup-mm-bind-heatmap-neuro-jepa-auc}
\end{figure}

\begin{figure}[htbp]
    \centering
    \includegraphics[width=1.0\textwidth]{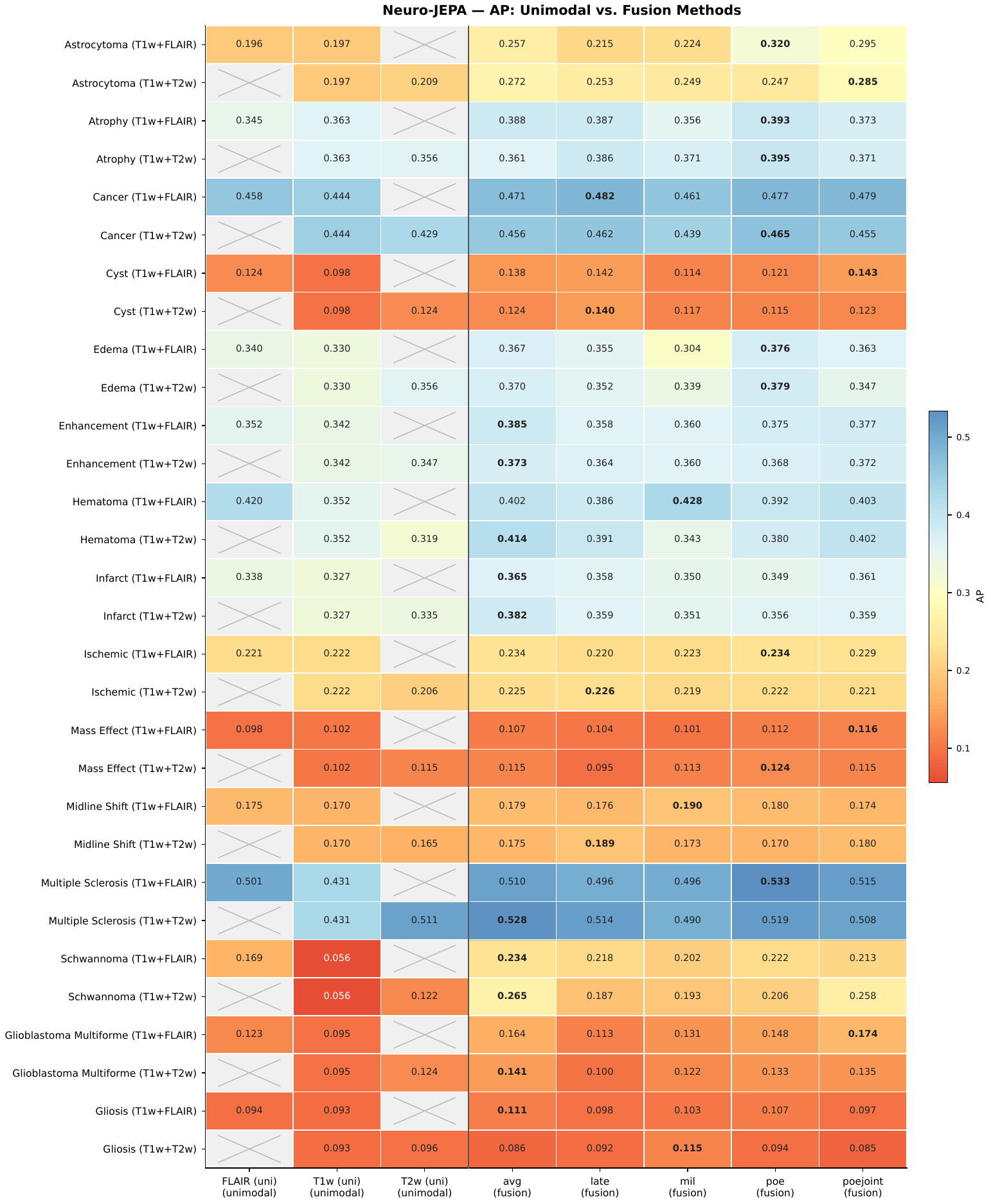}
    \caption{\textbf{Multimodal Learning Performance Across Fusion Methods for Neuro-JEPA on BIND-MGH - AP.}}
    \label{fig:sup-mm-bind-heatmap-neuro-jepa-ap}
\end{figure}

\begin{figure}[htbp]
    \centering
    \includegraphics[width=1.0\textwidth]{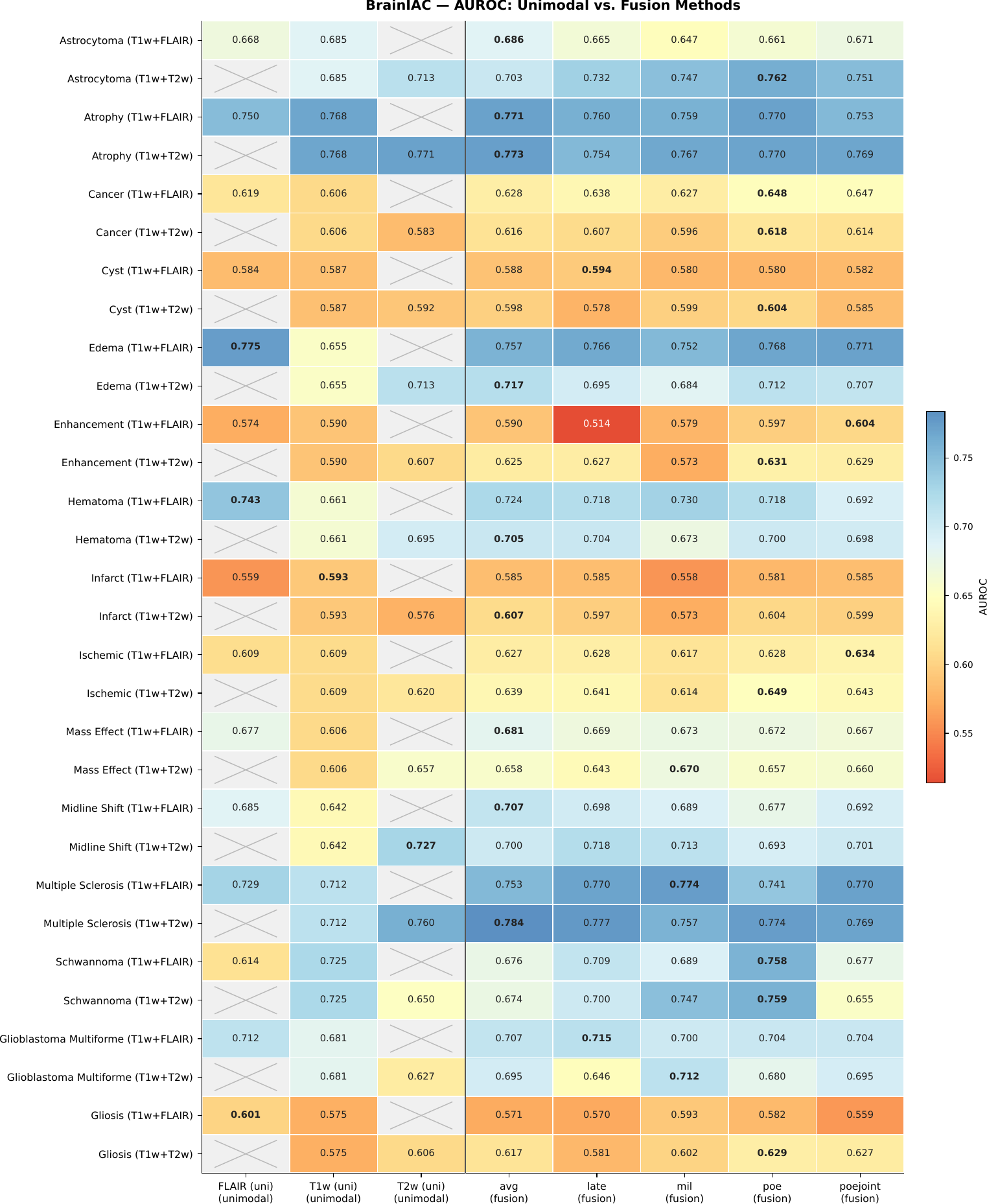}
    \caption{\textbf{Multimodal Learning Performance Across Fusion Methods for BrainIAC on BIND-MGH - AUROC.}}
    \label{fig:sup-mm-bind-heatmap-brainiac-auc}
\end{figure}

\begin{figure}[htbp]
    \centering
    \includegraphics[width=1.0\textwidth]{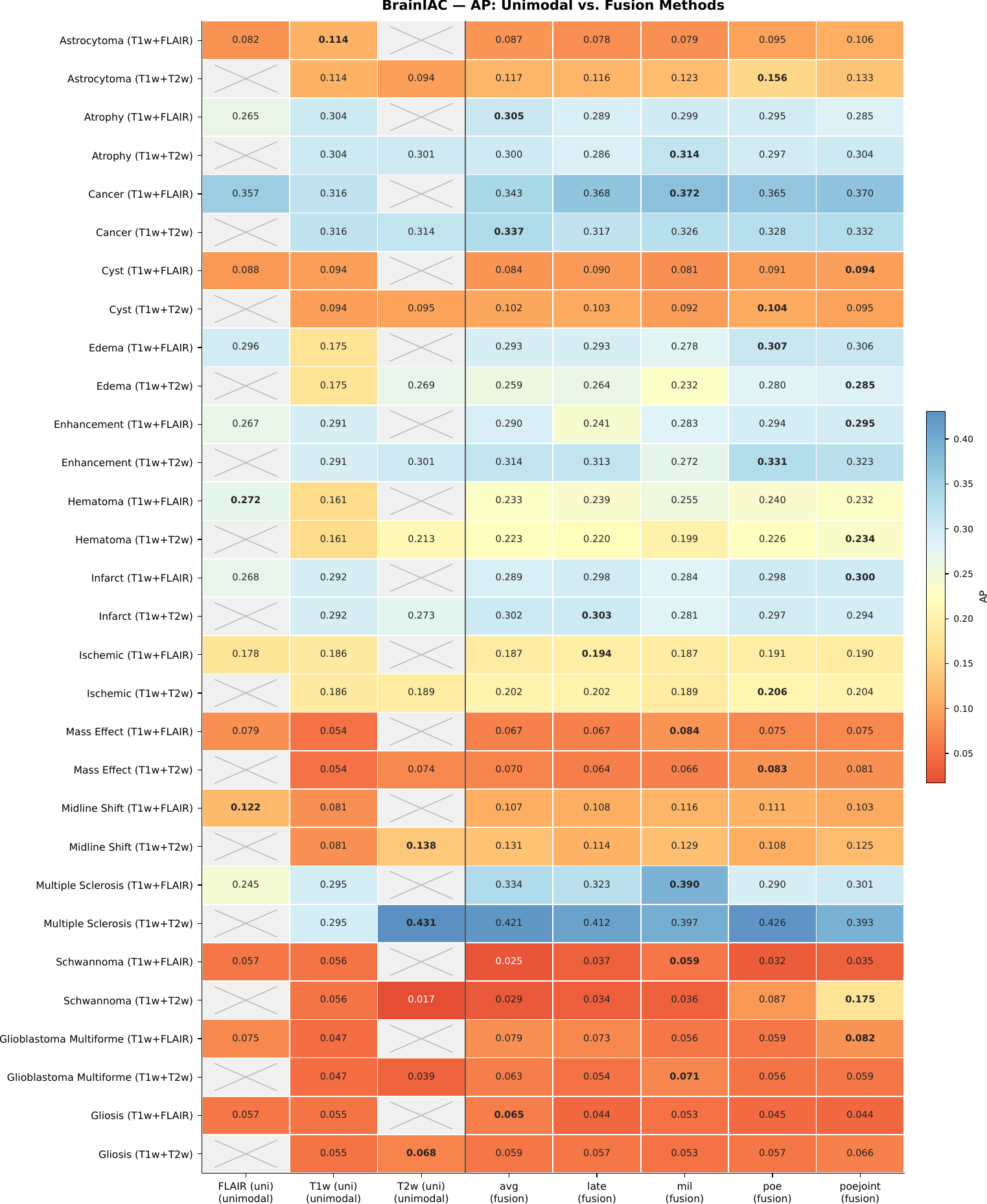}
    \caption{\textbf{Multimodal Learning Performance Across Fusion Methods for BrainIAC on BIND-MGH - AP.}}
    \label{fig:sup-mm-bind-heatmap-brainiac-ap}
\end{figure}

\begin{figure}[htbp]
    \centering
    \includegraphics[width=1.0\textwidth]{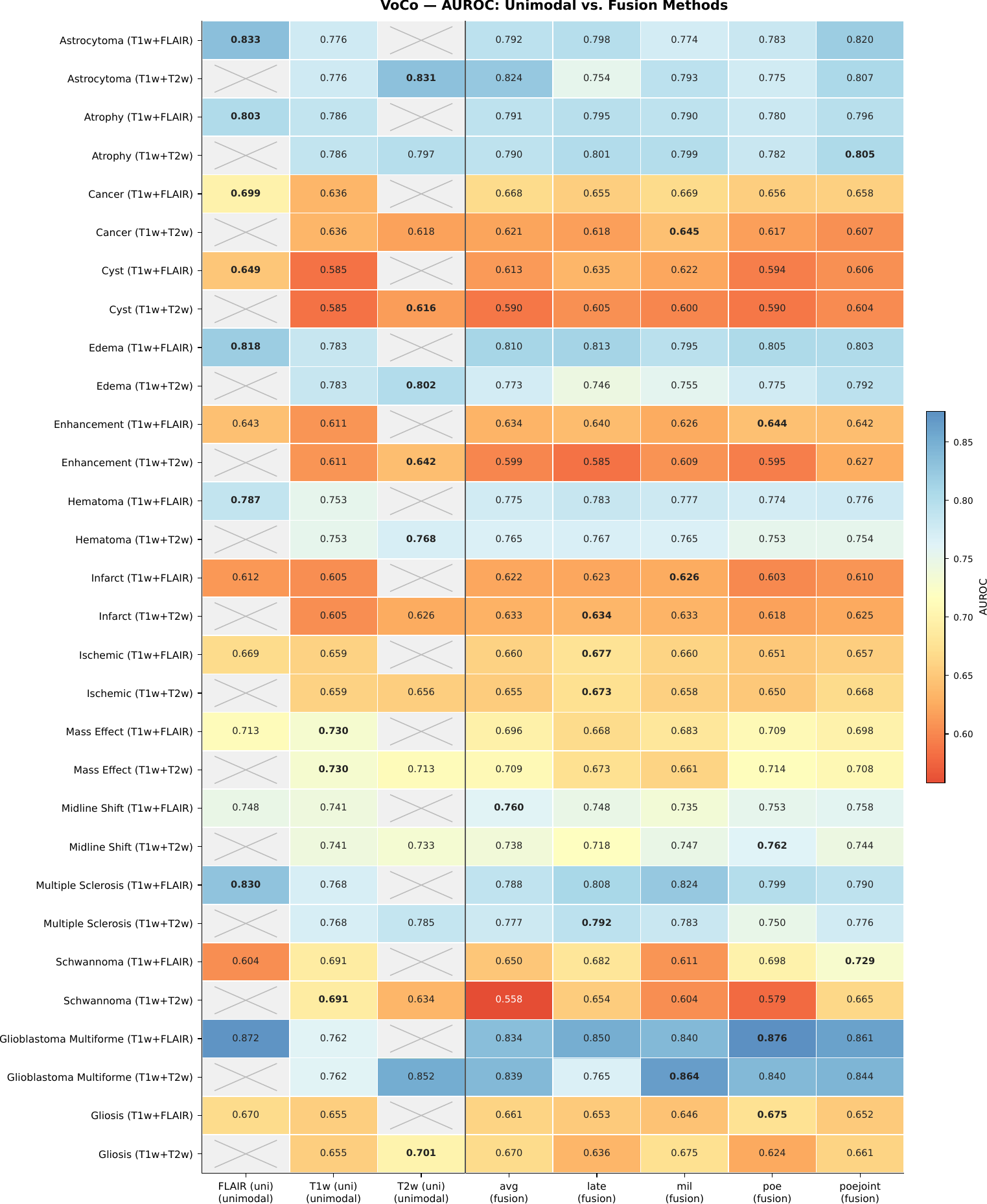}
    \caption{\textbf{Multimodal Learning Performance Across Fusion Methods for VoCo on BIND-MGH - AUROC.}}
    \label{fig:sup-mm-bind-heatmap-voco-auc}
\end{figure}

\begin{figure}[htbp]
    \centering
    \includegraphics[width=1.0\textwidth]{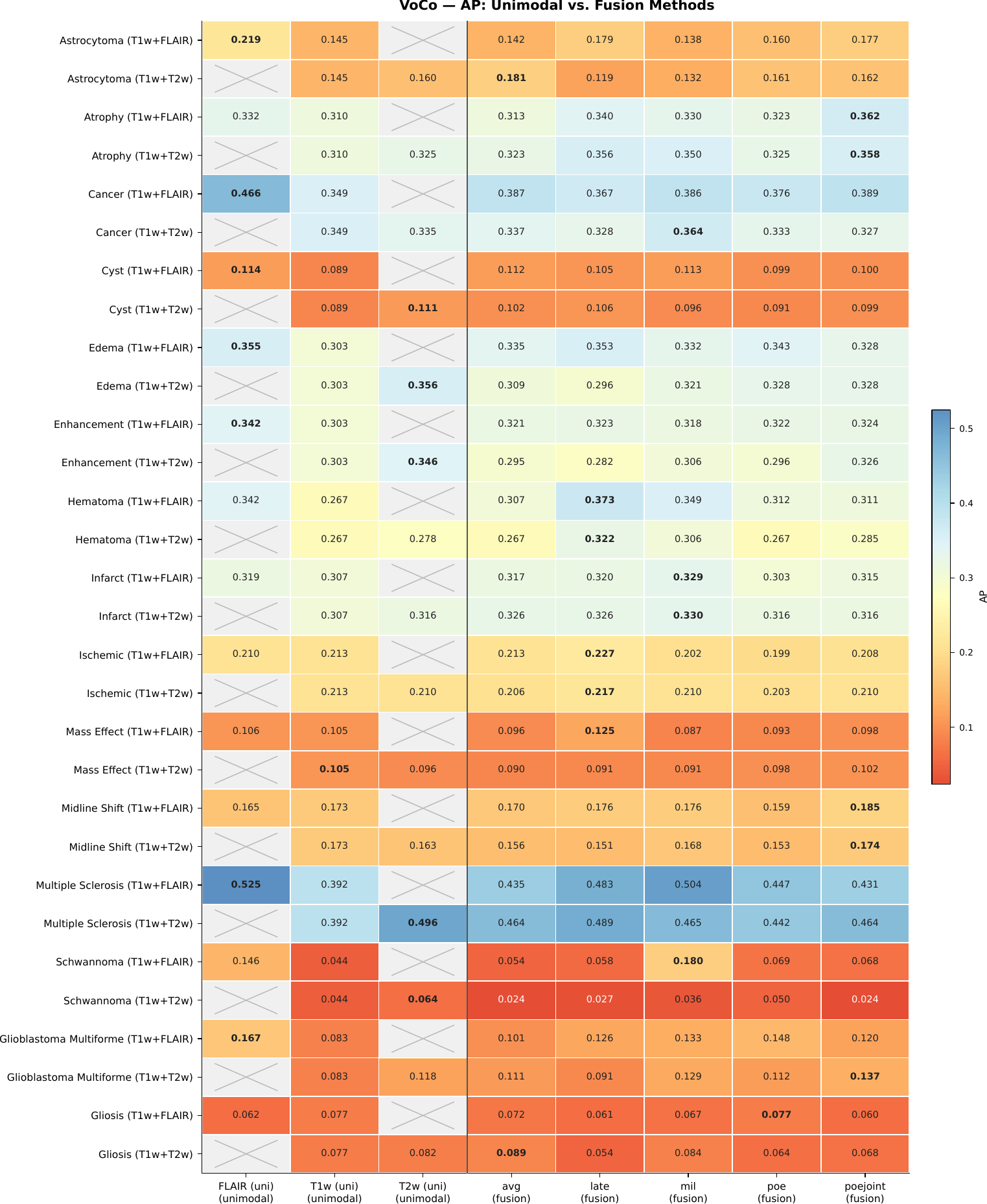}
    \caption{\textbf{Multimodal Learning Performance Across Fusion Methods for VoCo on BIND-MGH - AP.}}
    \label{fig:sup-mm-bind-heatmap-voco-ap}
\end{figure}

\begin{figure}[htbp]
    \centering
    \includegraphics[width=1.0\textwidth]{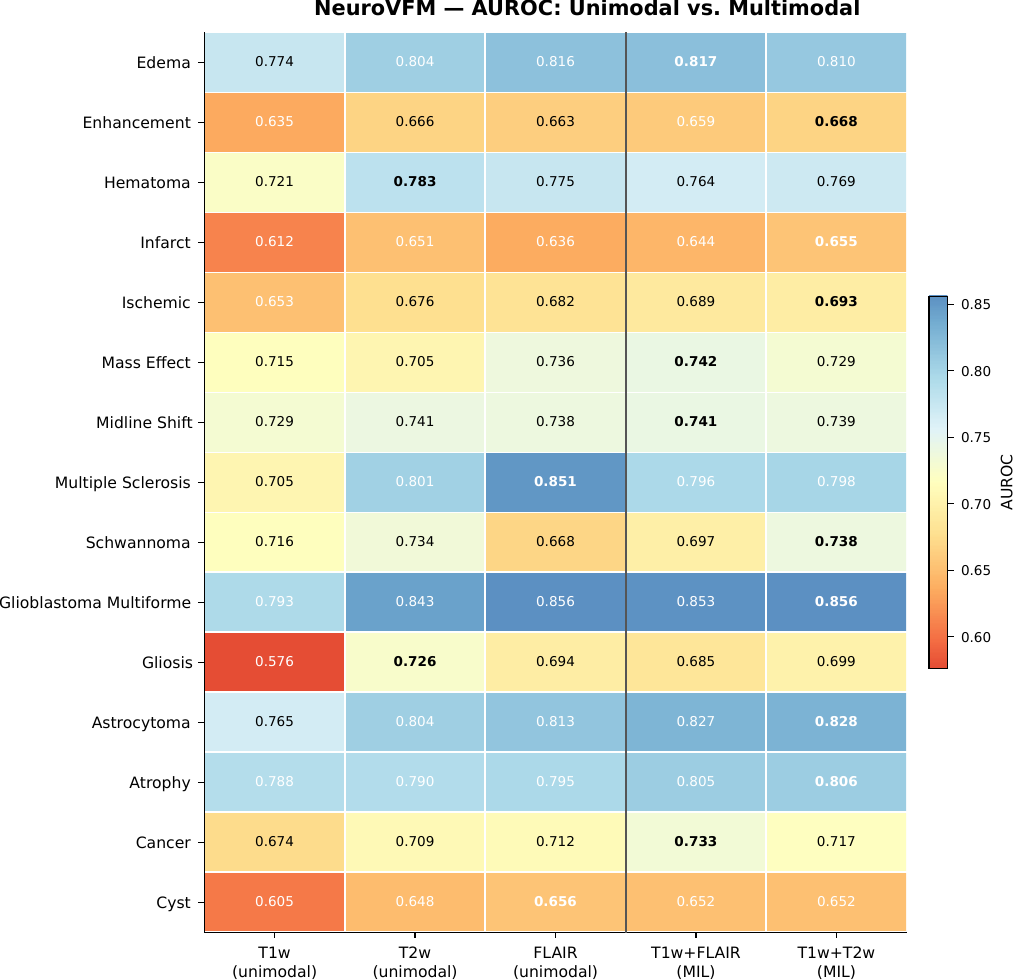}
    \caption{\textbf{Multimodal Learning Performance Across Fusion Methods for NeuroVFM on BIND-MGH - AUROC.}}
    \label{fig:sup-mm-bind-heatmap-neurovfm-auc}
\end{figure}

\begin{figure}[htbp]
    \centering
    \includegraphics[width=1.0\textwidth]{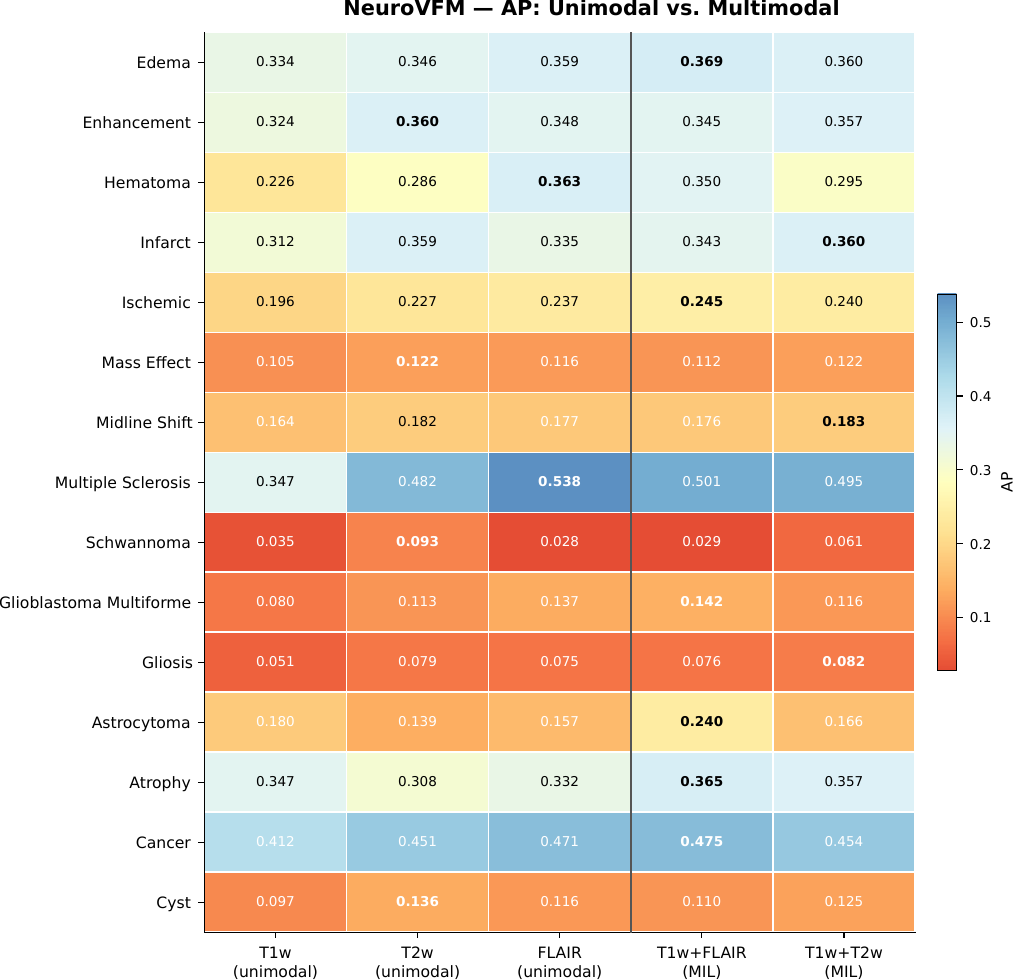}
    \caption{\textbf{Multimodal Learning Performance Across Fusion Methods for NeuroVFM on BIND-MGH - AP.}}
    \label{fig:sup-mm-bind-heatmap-neurovfm-ap}
\end{figure}

Supplementary \Cref{fig:sup-mm-bind-heatmap-neuro-jepa-auc,fig:sup-mm-bind-heatmap-neuro-jepa-ap,fig:sup-mm-bind-heatmap-brainiac-auc,fig:sup-mm-bind-heatmap-brainiac-ap,fig:sup-mm-bind-heatmap-voco-auc,fig:sup-mm-bind-heatmap-voco-ap,fig:sup-mm-bind-heatmap-neurovfm-auc,fig:sup-mm-bind-heatmap-neurovfm-ap} show multimodal learning performance (T1w+T2w or T1w+FLAIR) on BIND-MGH dataset across different fusion methods for different models reported in both AUROC and AUPRC. Each row in the plot shows present modalities for multimodal fusion and the corresponding performance on different fusion methods.

\subsection{Multi-Modal Gain Result - BIND-MGH}
\begin{figure}[htbp]
    \centering
    \includegraphics[width=1.0\textwidth]{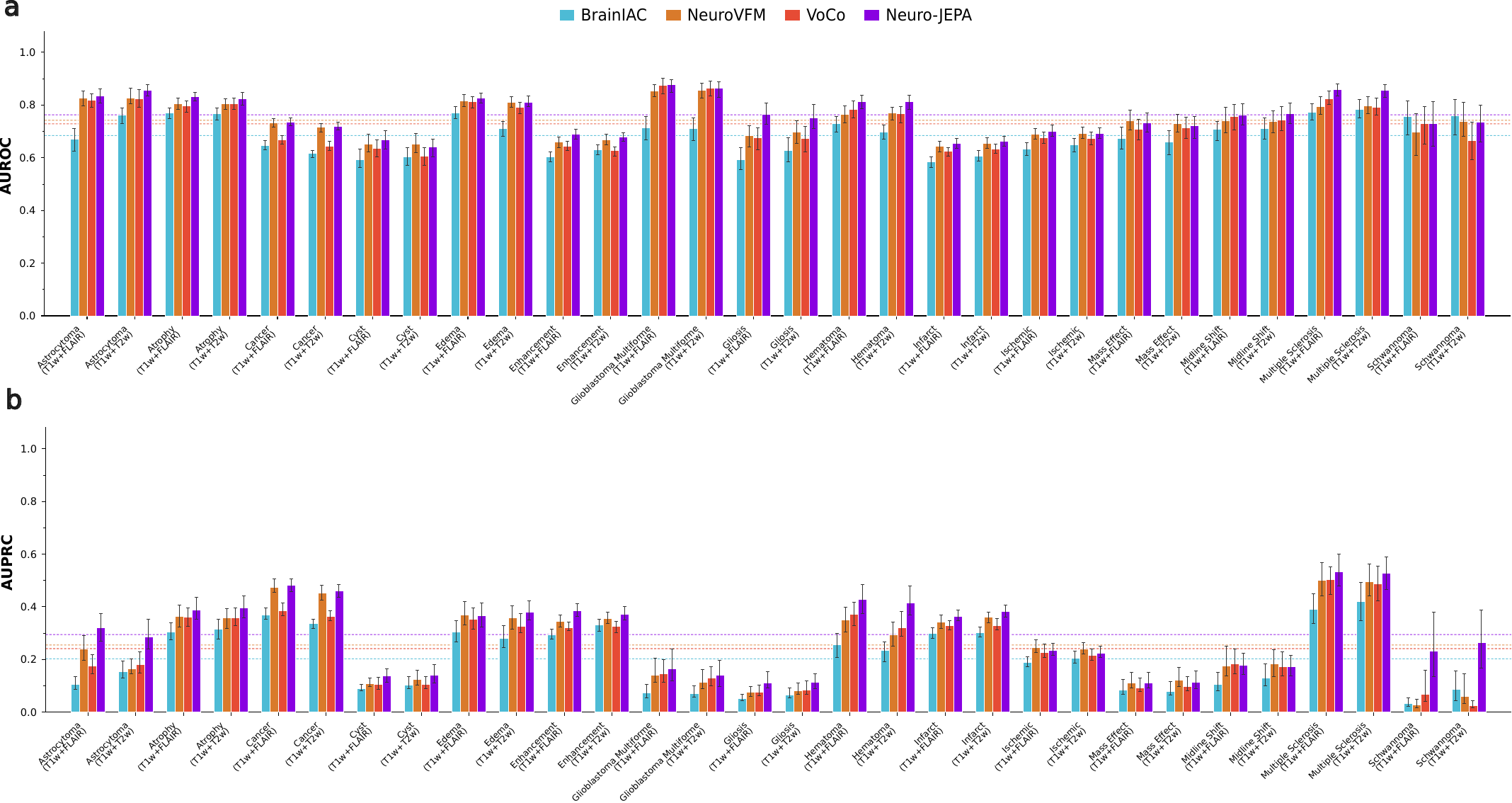}
    \caption{\textbf{Multimodal Performance on BIND-MGH.} We report AUROC and AUPRC for multimodal performance on all models when two different modalities are combined. The result is reported by best performance multimodal fusion method among five different methods. Dotted horizontal line present average performance across tasks, where the result shows our model outperforms other foundation models with a large margin. \textbf{a,} AUROC for different tasks across evaluated models \textbf{b,} AUPRC for different tasks across evaluated models.}
    \label{fig:sup-mm-bind-overall}
\end{figure}

Supplementary \Cref{fig:sup-mm-bind-overall} shows best multimodal learning performance across all 15 tasks on BIND-MGH datasets for all evaluated models. The result is reported in both AUROC and AUPRC. The result shows that Neuro-JEPA present larger multi-modal gain over other models in majority of the tasks.

\begin{figure}[htbp]
    \centering
    \includegraphics[width=1.0\textwidth]{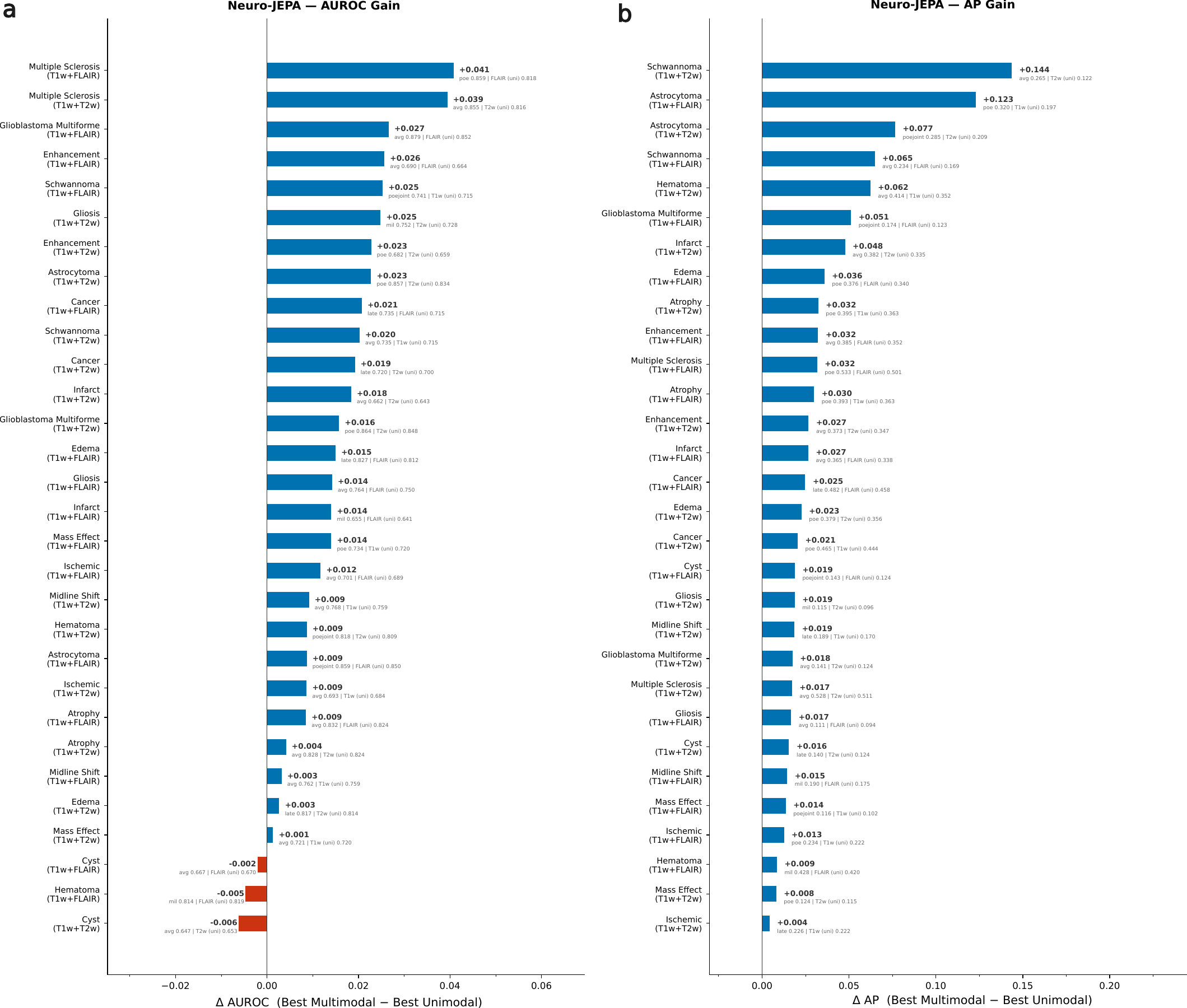}
    \caption{\textbf{Multimodal Gain Over Unimodal for Neuro-JEPA on BIND-MGH.} The difference between best multimodal fusion vs. best unimodal performance on Neuro-JEPA reported with AUROC and AUPRC. \textbf{a,} AUROC and AUPRC multimodal performance gain on the difference. \textbf{b,} AUPRC multimodal performance gain on the difference.}
    \label{fig:sup-mm-bind-gain-neuro-jepa}
\end{figure}

\begin{figure}[htbp]
    \centering
    \includegraphics[width=1.0\textwidth]{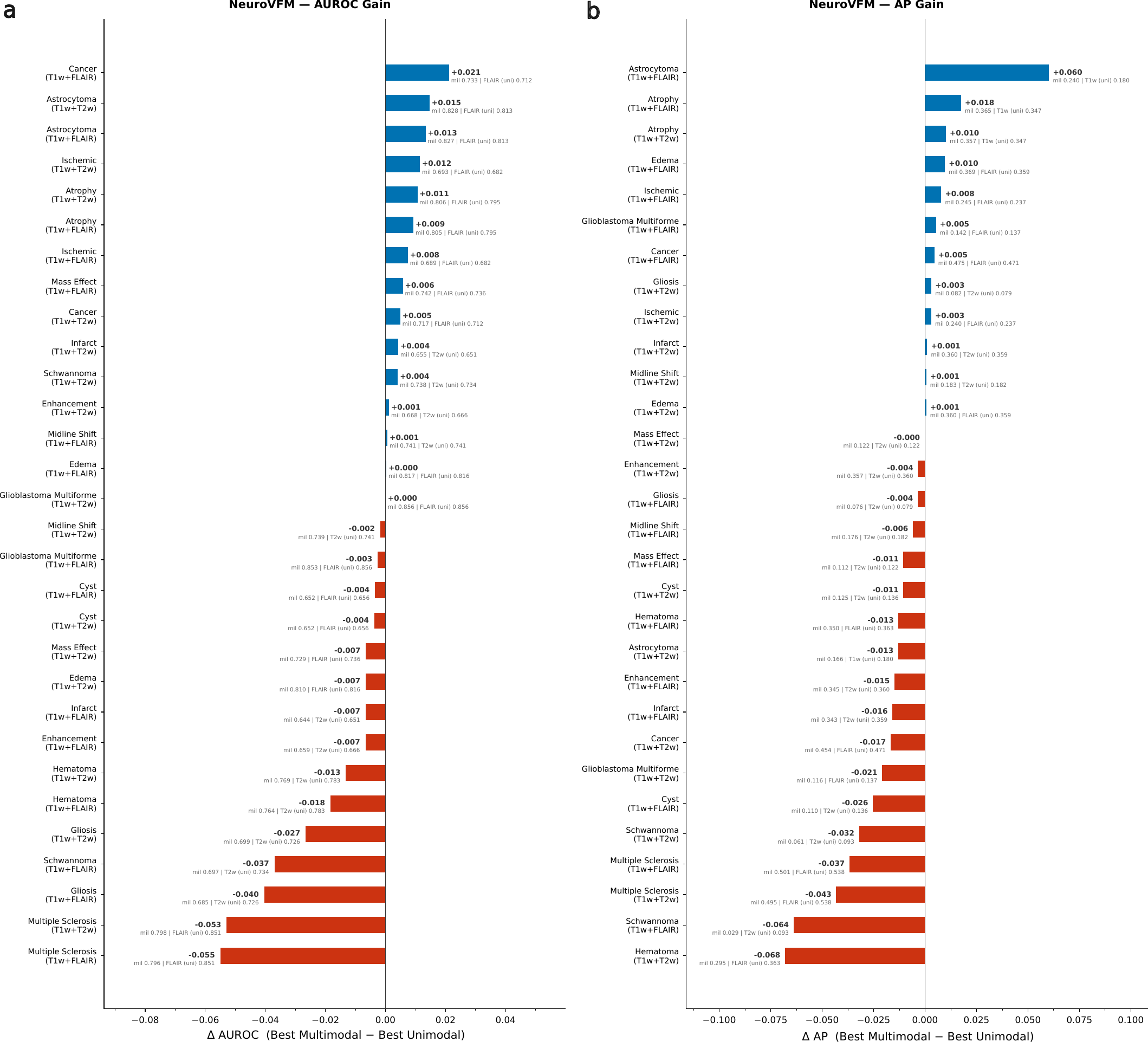}
    \caption{\textbf{Multimodal Gain Over Unimodal for NeuroVFM on BIND-MGH.} The difference between best multimodal fusion vs. best unimodal performance on NeuroVFM reported with AUROC and AUPRC. \textbf{a,} AUROC and AUPRC multimodal performance gain on the difference. \textbf{b,} AUPRC multimodal performance gain on the difference.}
    \label{fig:sup-mm-bind-gain-neurovfm}
\end{figure}

\begin{figure}[htbp]
    \centering
    \includegraphics[width=1.0\textwidth]{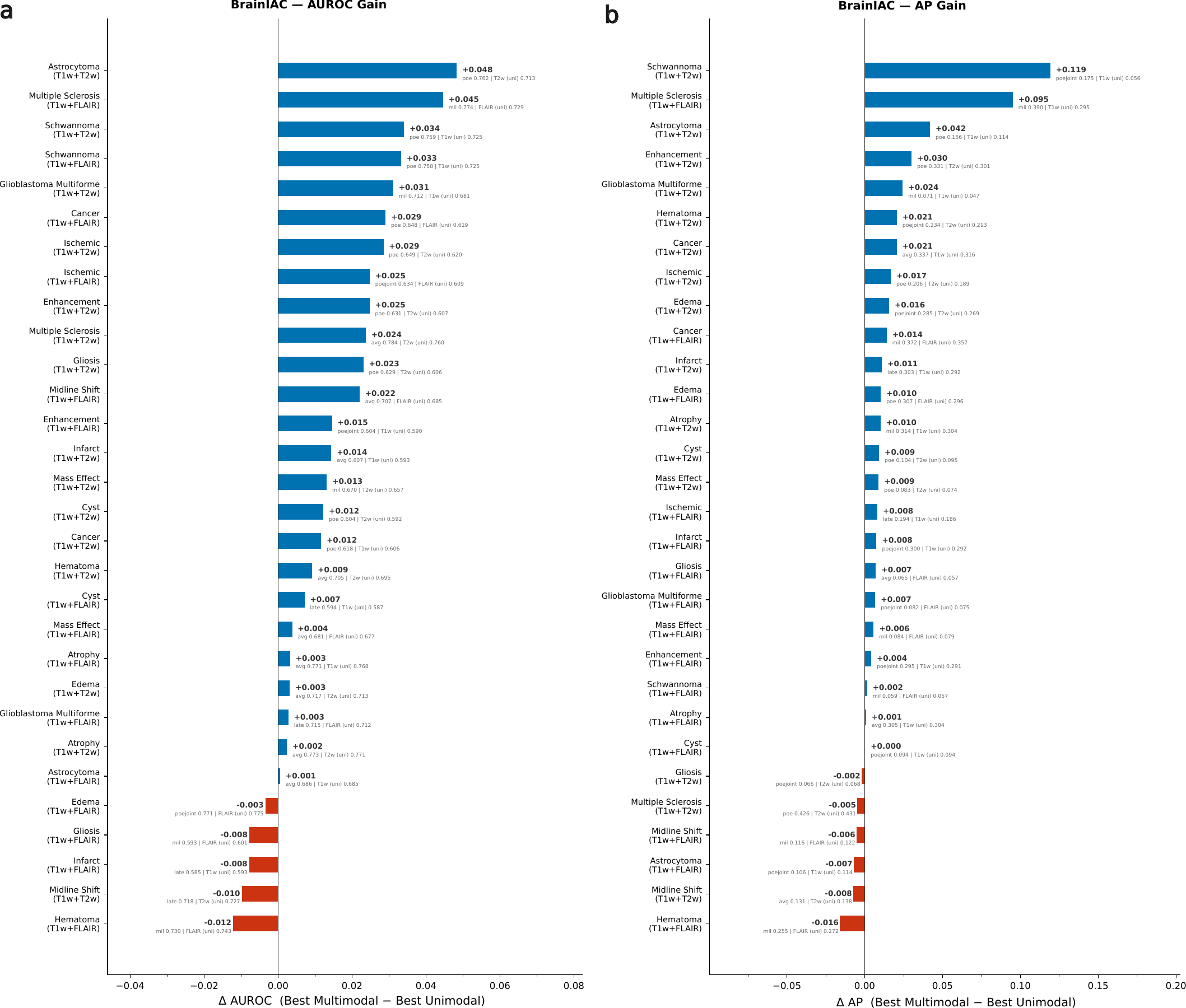}
    \caption{\textbf{Multimodal Gain Over Unimodal for BrainIAC on BIND-MGH.} The difference between best multimodal fusion vs. best unimodal performance on BrainIAC reported with AUROC and AUPRC. \textbf{a,} AUROC and AUPRC multimodal performance gain on the difference. \textbf{b,} AUPRC multimodal performance gain on the difference.}
    \label{fig:sup-mm-bind-gain-brainiac}
\end{figure}

\begin{figure}[htbp]
    \centering
    \includegraphics[width=1.0\textwidth]{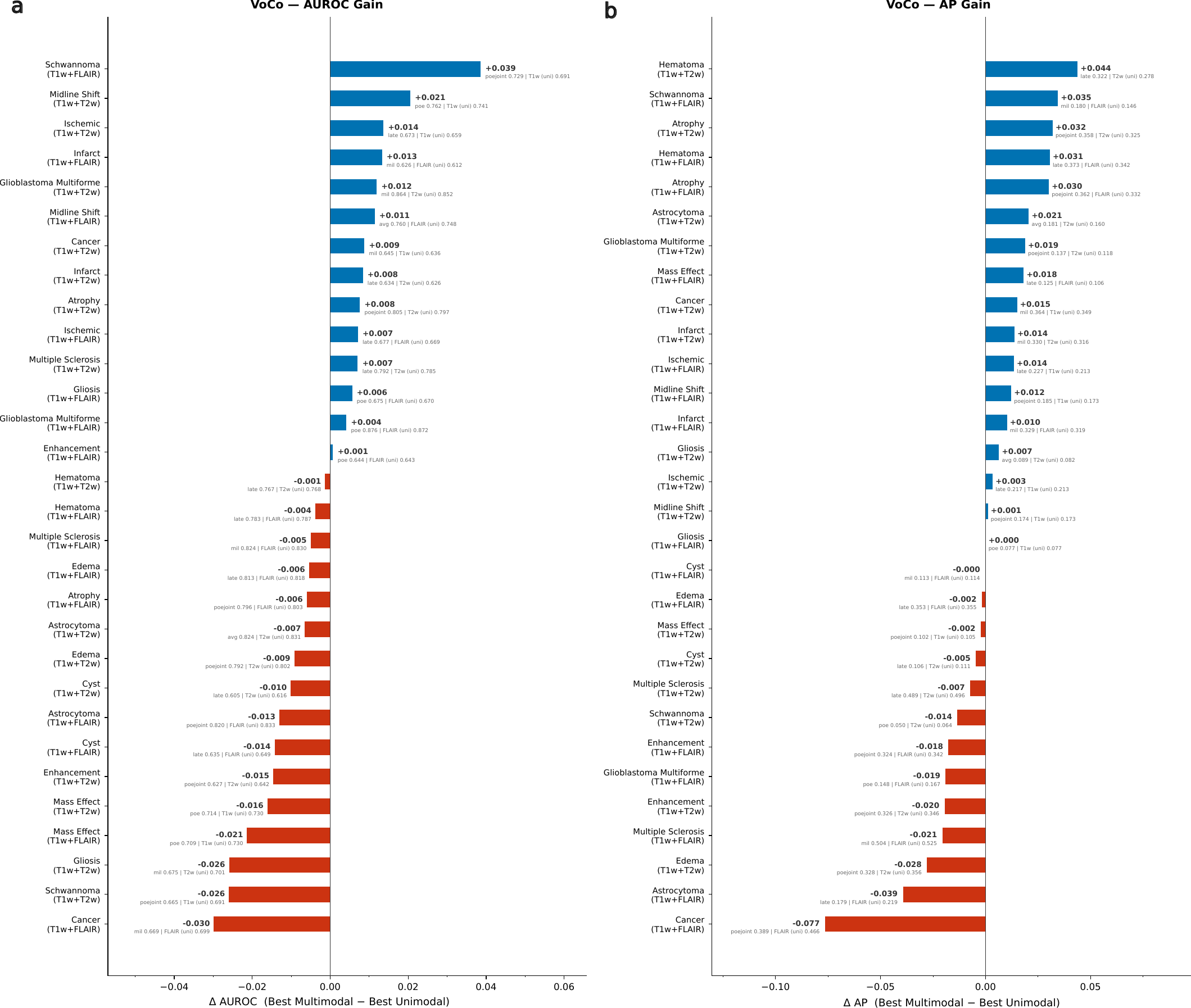}
    \caption{\textbf{Multimodal Gain Over Unimodal for VoCo on BIND-MGH.} The difference between best multimodal fusion vs. best unimodal performance on NeuroVFM reported with AUROC and AUPRC. \textbf{a,} AUROC and AUPRC multimodal performance gain on the difference. \textbf{b,} AUPRC multimodal performance gain on the difference.}
    \label{fig:sup-mm-bind-gain-voco}
\end{figure}

Supplementary \Cref{fig:sup-mm-bind-gain-neuro-jepa,fig:sup-mm-bind-gain-neurovfm,fig:sup-mm-bind-gain-brainiac,fig:sup-mm-bind-gain-voco} present multimodal gain over unimodal for all fifteen tasks on BIND-MGH dataset for all evaluated models. The result is reported in both AUROC and AUPRC. The result shows that Neuro-JEPA present most positive transfer while maintaining highest overall performance across the tasks.

\newpage
\section{Ablation Studies Experiments}
\label{apd:ablation_sec}
\subsection{Ablation Study on Number of Experts}
\begin{figure}[htbp]
    \centering
    \includegraphics[width=1.0\textwidth]{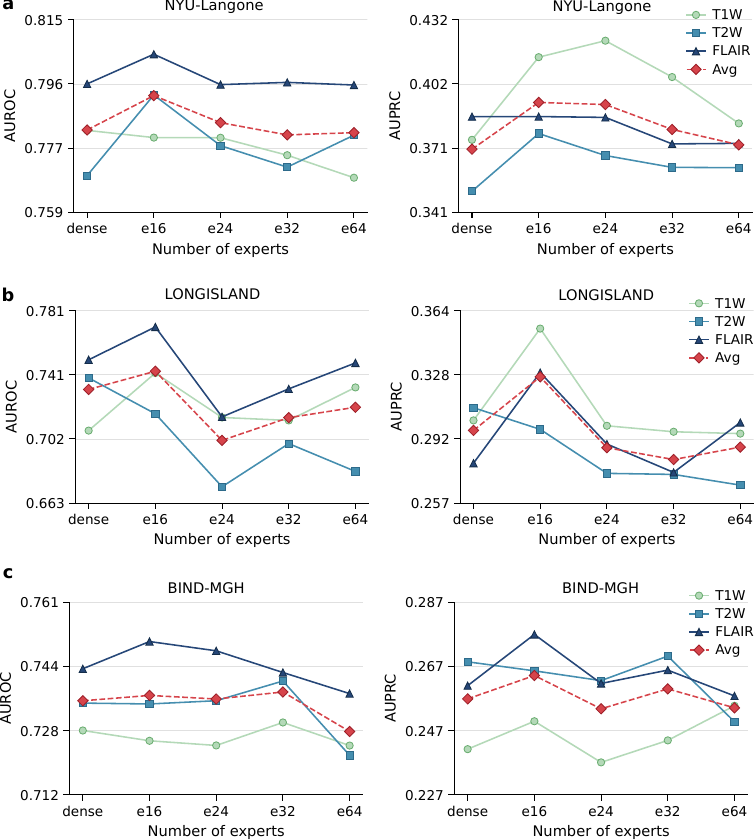}
    \caption{\textbf{Per Dataset Result for Number of Expert Ablation Study.} Per dataset result for number of experts with attentive probing across three evaluated datasets (NYU Langone, Longisland and BIND-MGH). Consistent with average performance show in main manuscript [ref], we observed improved performance on both AUROC and AUPRC when comparing model performance on dense model vs. model with same setting on 16 total experts. However, the performance improvement is diminishing when number of experts is increase from 24 to 64.}
    \label{fig:sup-experts-abla}
\end{figure}

Supplementary \Cref{fig:sup-experts-abla} presents detailed result for mixture of experts and dense model across evaluated clinical cohorts datasets with attentive probing. As the result shown, changing from dense model to 16 total experts give most performance improvement, while increasing number of total experts presents diminishing returns.

\subsection{Ablation Study on Methods}
\begin{figure}[htbp]
    \centering
    \includegraphics[width=1.0\textwidth]{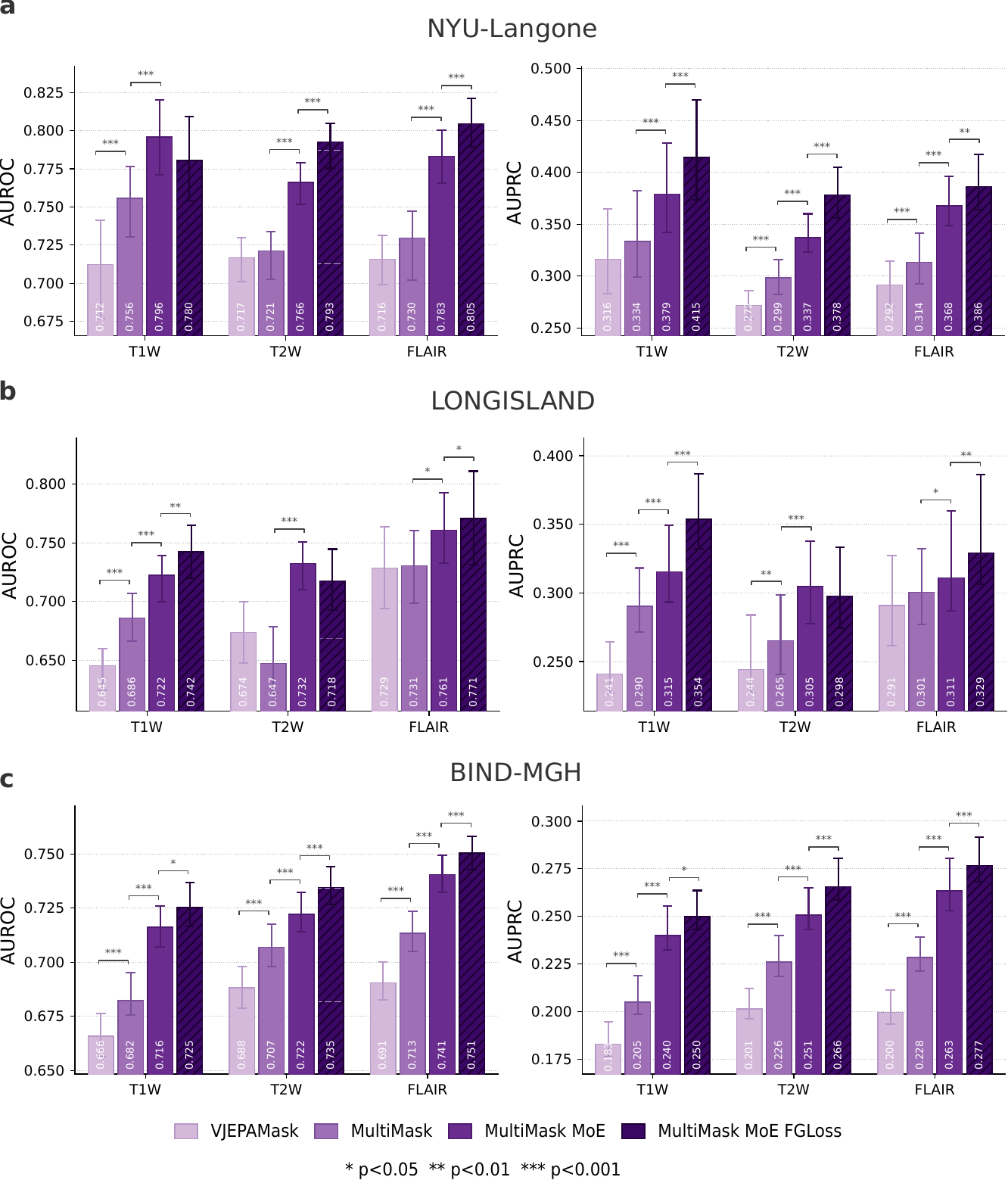}
    \caption{\textbf{Per Dataset Result for Design Choices Ablation Study.} Per dataset result for design choices with attentive probing across three evaluated datasets (NYU Langone, Longisland and BIND-MGH). The color indicates different combination of methods used during pretraining. Consistent with average performance show in main manuscript [ref], incrementally adding multiscale masking, mixture of experts and foreground aware L1 loss improve overall probing performance across the datasets.}
    \label{fig:sup-methods-abla}
\end{figure}

Supplementary \Cref{fig:sup-methods-abla} presents result for the impact of modifying the algorithm details starting from original V-JEPA 2 across the evaluated clinical cohorts datasets. The result is present as incrementally adding Multiscale Masking, Mixture of Experts and Foreground-aware L1 Loss. As the result shown, every modification presents meaningful improvement on the model overall performance. 

\subsection{Ablation Study on NeuroVFM}
\begin{figure}[htbp]
    \centering
    \includegraphics[width=0.9\textwidth]{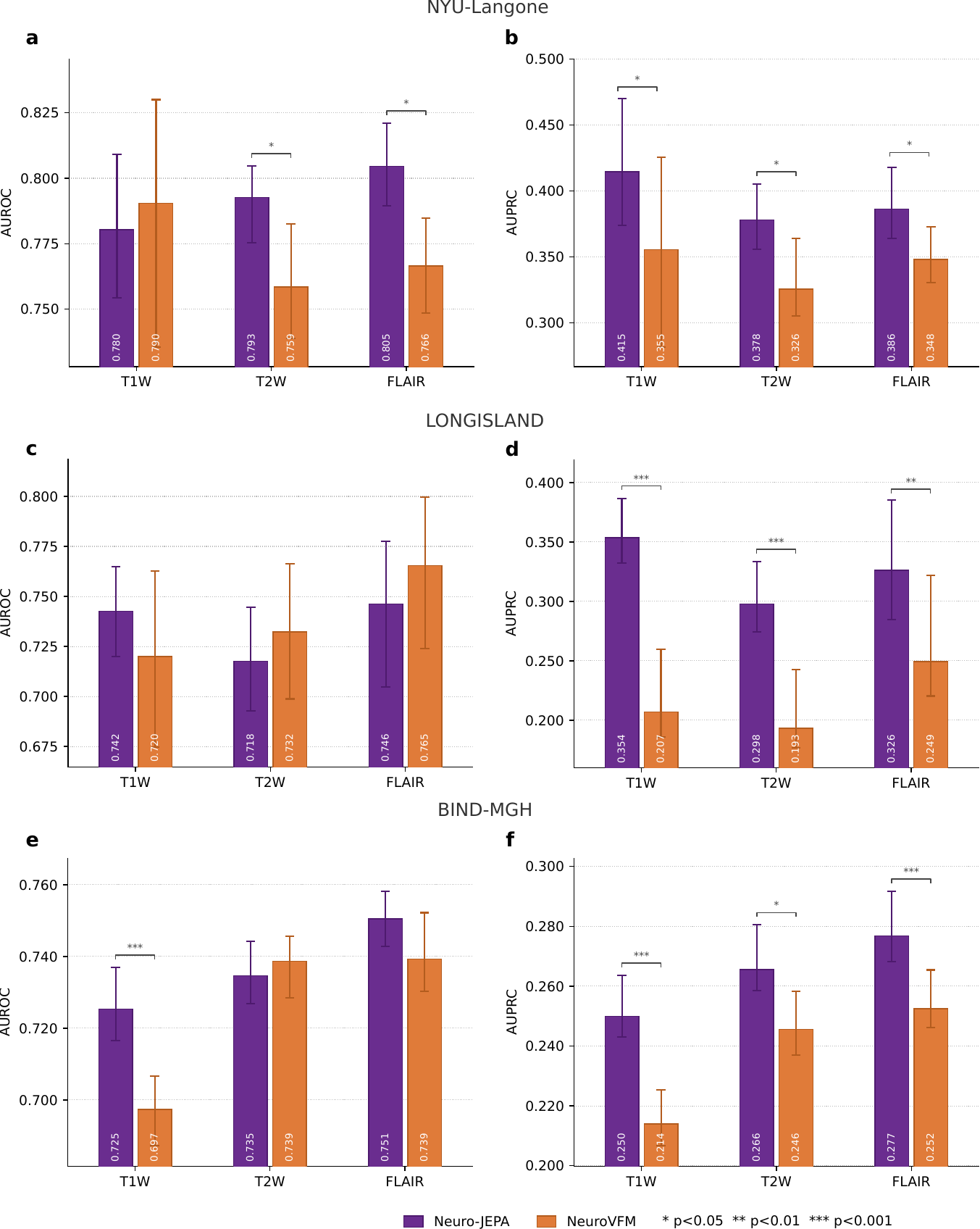}
    \caption{\textbf{Per Dataset Comparison with NeuroVFM.} Per dataset performance comparison with NeuroVFM. The performance is reported as AUROC and AUPRC average across all tasks on each dataset. The result shows that although both models have similar performance on AUROC, our model shows large performance improvement on AUPRC. This is critical on indicating the superior of our model as AUPRC better reflect the correctness on classification for the diagnosis where the positive cases are usually rare.}
    \label{fig:sup-neurovfm-abla}
\end{figure}

Supplementary \Cref{fig:sup-neurovfm-abla} presents per-dataset model performance comparison with clinical cohorts datasets for Neuro-JEPA and NeuroVFM with attentive probing at per modality level (T1w, T2w, FLAIR). As the result demonstrates, Neuro-JEPA show improved performance over NeuroVFM on a majority of cases, highlighting the important of algorithmic improvement over data scaling.

\subsection{Ablation Study on Different Percent of Pretrain Data}
\begin{figure}[htbp]
    \centering
    \includegraphics[width=1.0\textwidth]{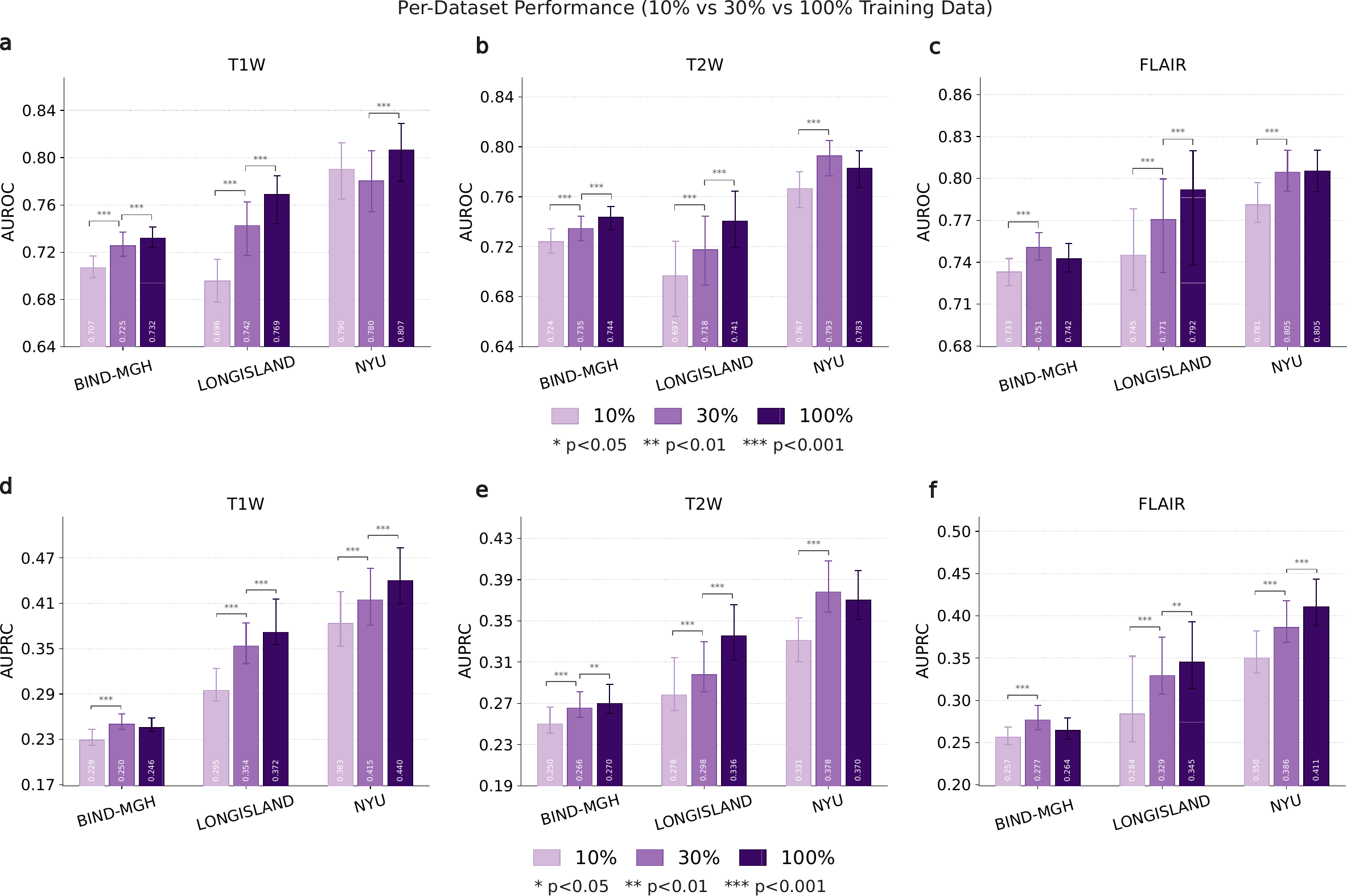}
    \caption{\textbf{Model Performance on Different Percentage of Pretrain Data.} AUROC and AUPRC averaged across tasks for each dataset and modality. \textbf{a-c,} AUROC for different datasets on different percentage for T1w, T2w and FLAIR. \textbf{e-f,} AUPRC for different datasets on different percentage for T1w, T2w and FLAIR.}
    \label{fig:sup-percentabla_perdataset}
\end{figure}

Supplementary \Cref{fig:sup-percentabla_perdataset} presents model per dataset and modality performance on model pre-trained on different percentage of pre-training data evaluated on clinical cohorts datasets with attentive probing. The result shows that our method is scalable with increasing data size across the tasks with diminishing return on data scaling similar to V-JEPA 2 \cite{assran2025vjepa2}.

\subsection{Ablation Study on Model Size}
\begin{figure}[htbp]
    \centering
    \includegraphics[width=0.9\textwidth]{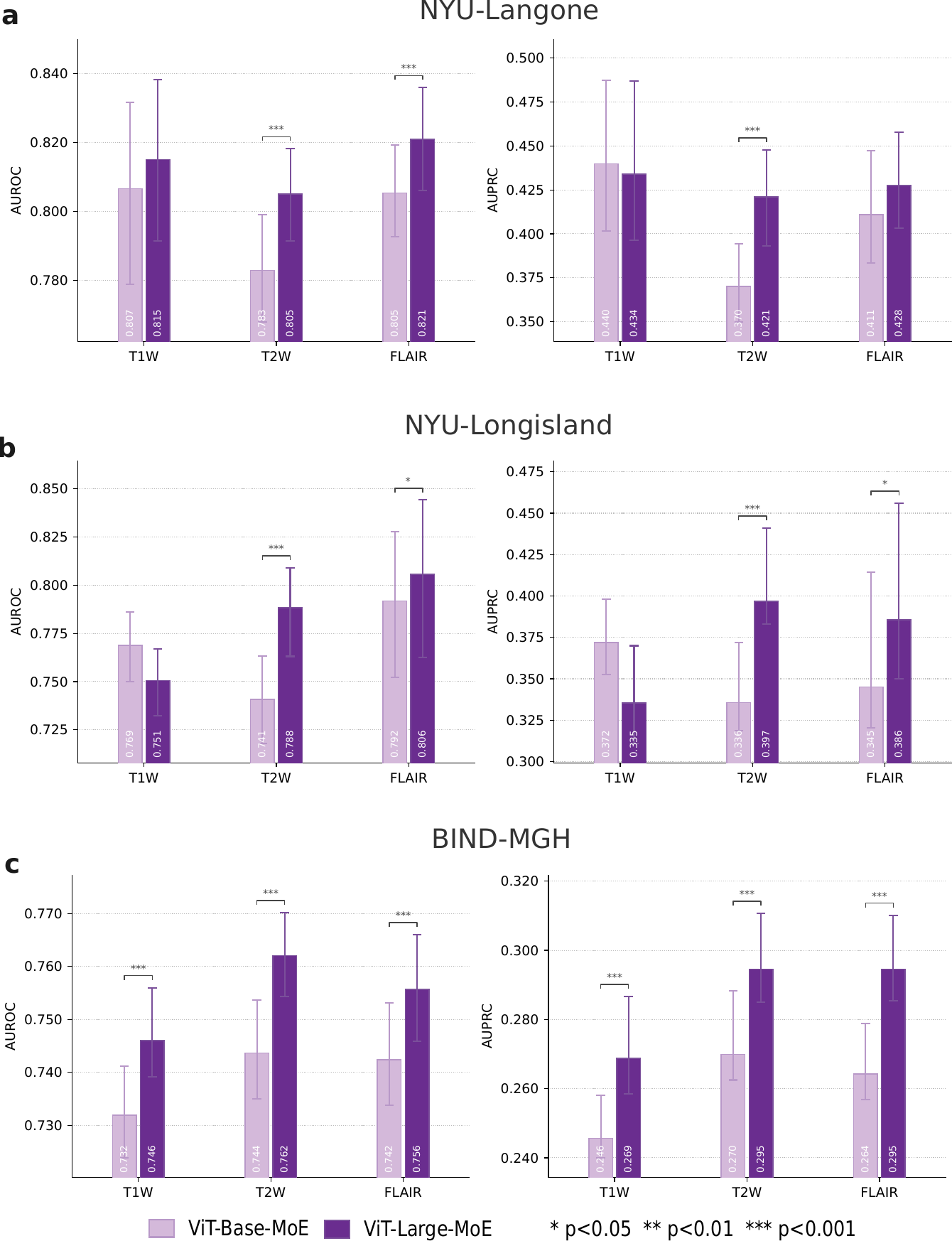}
    \caption{\textbf{Model Size Ablation - Health System Datasets.} This ablation experiment show per health-system dataset and modality performance (AUROC and AUPRC for Neuro-JEPA with Vit-Base and ViT-Large with \textbf{a,} NYU-Langone \textbf{b,} NYU-Longisland \textbf{c,} MGH}
    \label{fig:sup-modelsize_abla_clinical}
\end{figure}

\begin{figure}[htbp]
    \centering
    \includegraphics[width=1.0\textwidth]{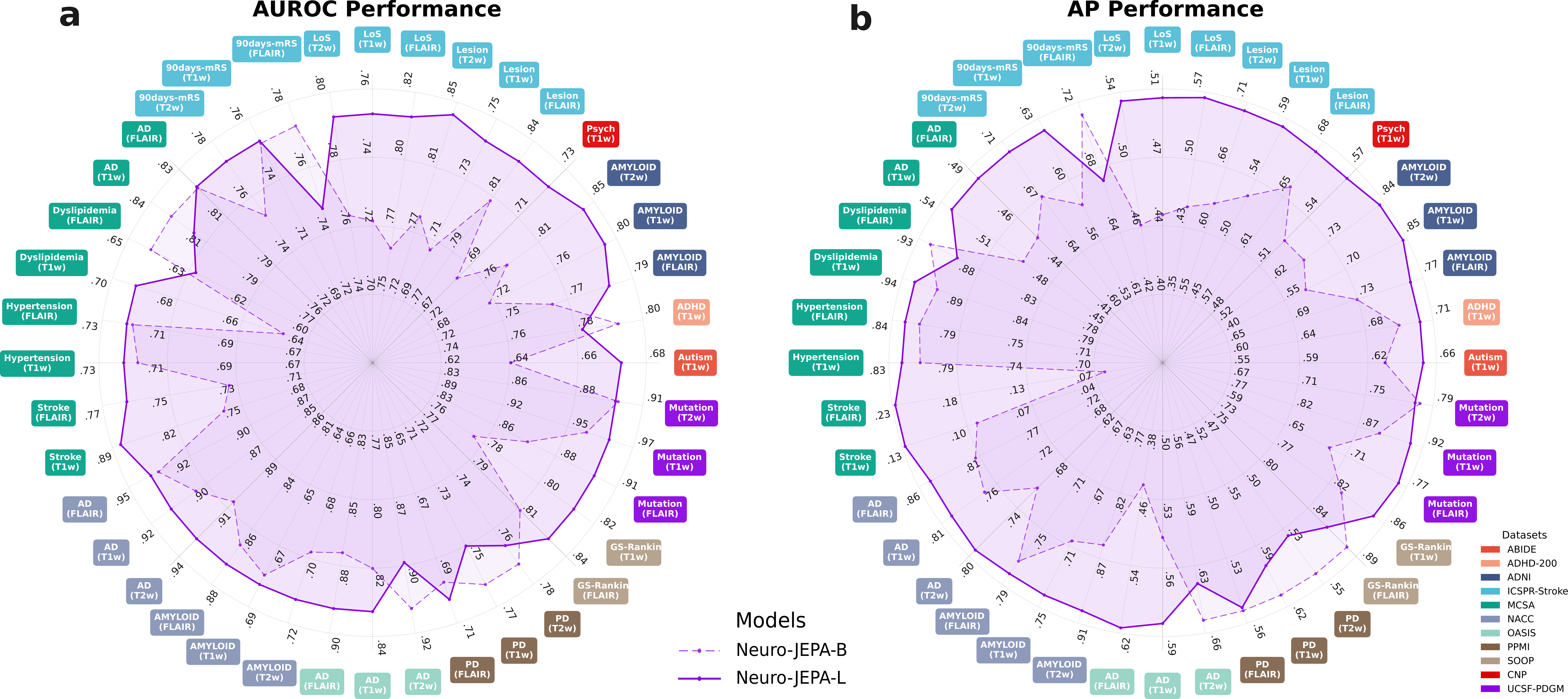}
    \caption{\textbf{Model Size Ablation - Public Datasets.} This ablation experiment show per public dataset and modality performance for Neuro-JEPA with ViT-Base and ViT-Large with \textbf{a,} AUROC \textbf{b,} AUPRC}
    \label{fig:sup-modelsize_abla_public}
\end{figure}

We present ablation study on model performance for different parameter size on ViT-B-MoE (121 million total parameters) and ViT-L-MoE (429 million total parameters) both pretrained with same Neuro-JEPA receipt on Supplementary \Cref{fig:sup-modelsize_abla_clinical,fig:sup-modelsize_abla_public}. The result presents that ViT-L-MoE outperform ViT-B-MoE in majorty of cases, highlighting the effectiveness of our pretraining receipt unnder model parameters size scaling.

\subsection{Ablation Study on Patch Size}
\begin{figure}[htbp]
    \centering
    \includegraphics[width=0.9\textwidth]{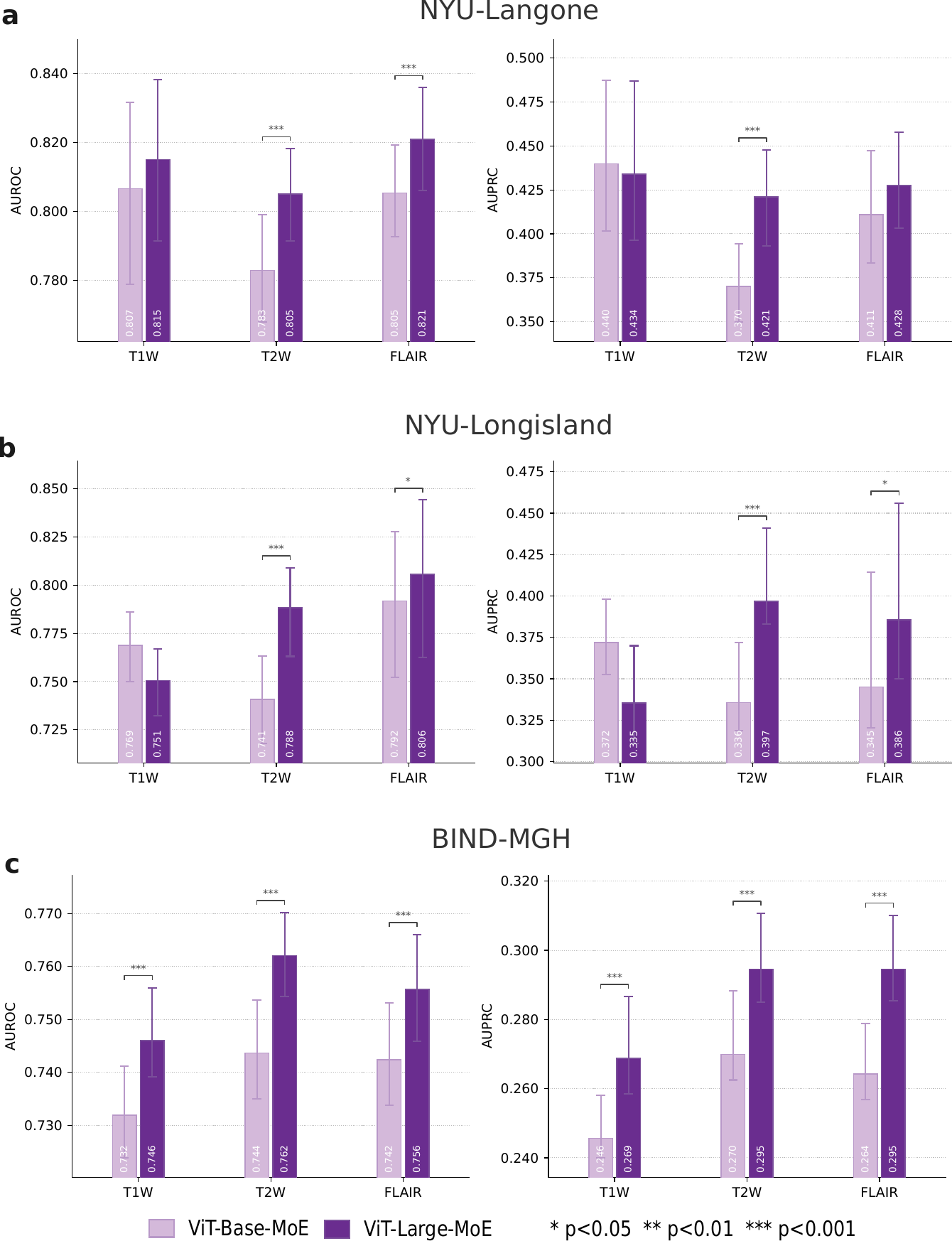}
    \caption{\textbf{Patch Size Ablation - Health System Datasets.} This ablation experiment show per health-system dataset and modality performance (AUROC and AUPRC for Neuro-JEPA with Vit-Base and ViT-Large with \textbf{a,} NYU-Langone \textbf{b,} NYU-Longisland \textbf{c,} MGH}
    \label{fig:sup-patchsize_abla_clinical}
\end{figure}

We present ablation study on model patch size for ViT-Large with 12 patches versus 8 patches in \Cref{fig:sup-patchsize_abla_clinical}. We observe the significant performance difference across tasks with the more fine-grained patch size of 8.

\subsection{Ablation Study on Pretraining with Uncurated vs. Curated Data}
\begin{figure}[htbp]
    \centering
    \includegraphics[width=1.0\textwidth]{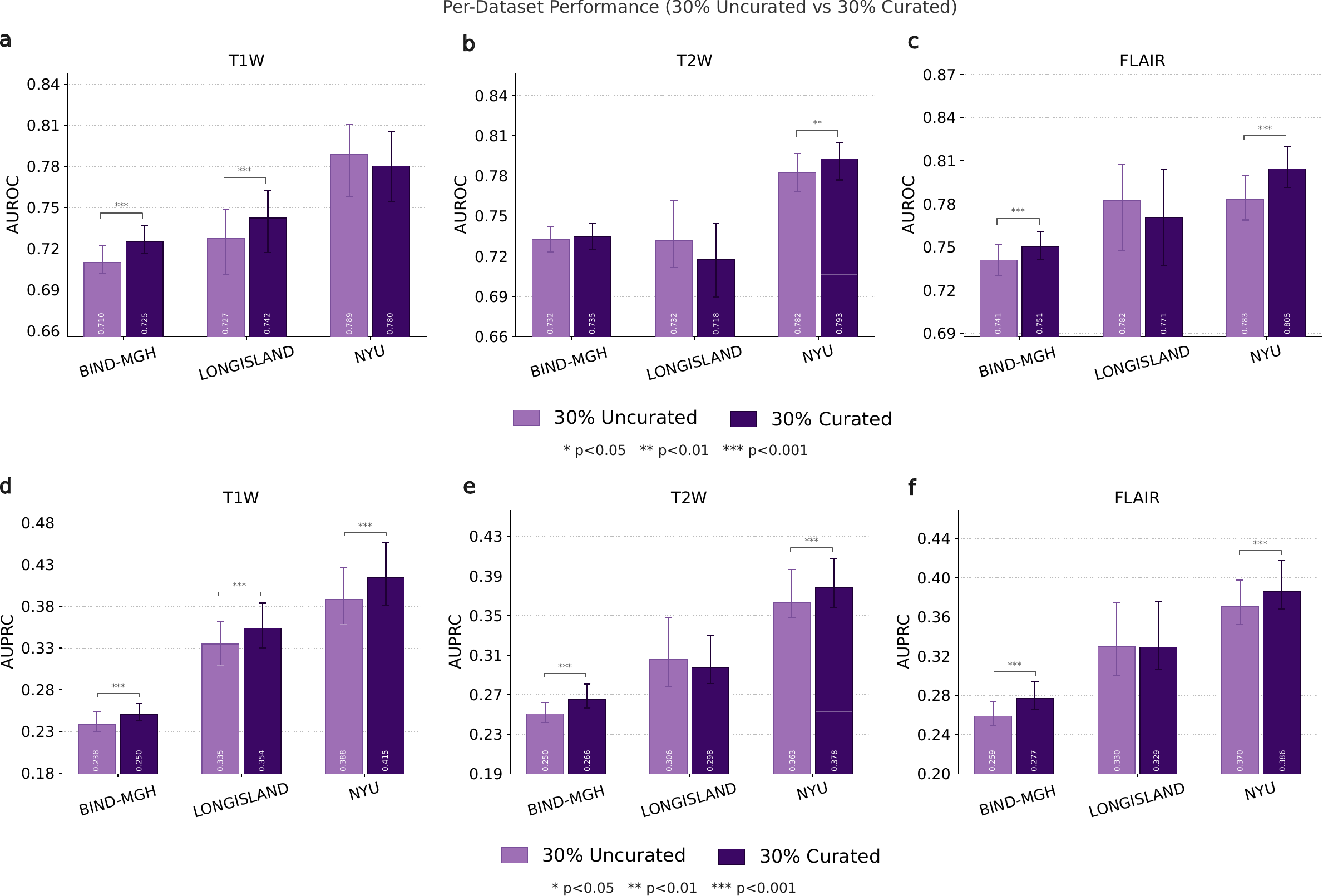}
    \caption{\textbf{Model Performance on Pretrain with Uncurated vs. Curated Data.} This ablation experiments are run on $30\%$ of pretrain data with Multi-Scale Masking, MoE and Foreground-Aware Masking all enabled. AUROC and AUPRC are averaged across tasks for each dataset and modality. \textbf{a-c,} AUROC on different datasets for model pretrained with uncurated vs.curated data for T1w, T2w and FLAIR. \textbf{e-f,} AUPRC on different datasets for model pretrained with uncurated vs.curated data for T1w, T2w and FLAIR.}
    \label{fig:sup-noiseabla_perdataset}
\end{figure}

Supplementary \Cref{fig:sup-noiseabla_perdataset} presents the performance difference on pretraining model with curated vs. uncurated data with $30\%$ pretraining data. Uncurated data present $5\%$ more noisy samples that are filtered out from curated data. The result demonstrates that including noisy samples for neuroimaging pre-training can present negative impact on overall model performance. 

\begin{figure}[htbp]
    \centering
    \includegraphics[width=0.18\textwidth]{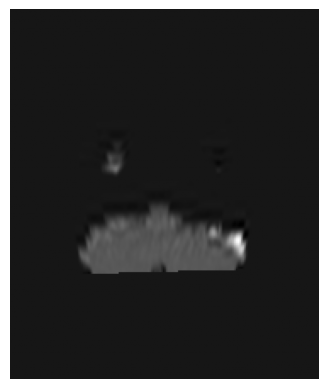}
    \includegraphics[width=0.18\textwidth]{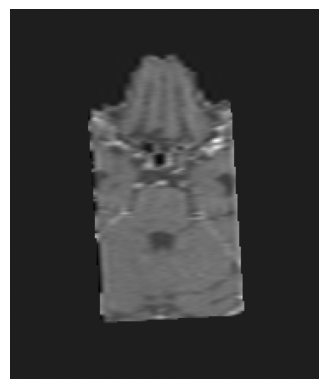}
    \includegraphics[width=0.18\textwidth]{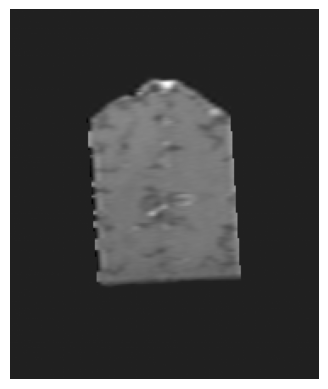}
    \includegraphics[width=0.18\textwidth]{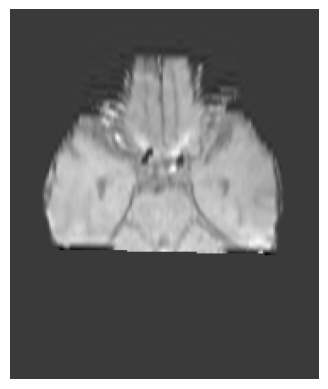}
    \includegraphics[width=0.18\textwidth]{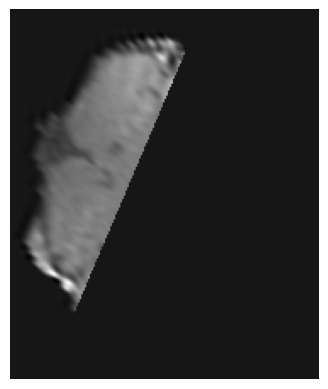}\\[6pt]
    \includegraphics[width=0.18\textwidth]{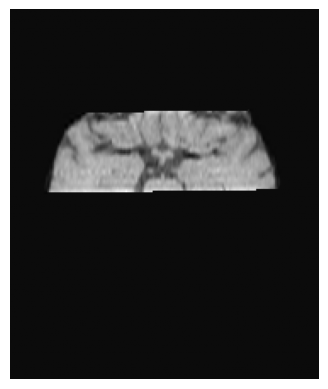}
    \includegraphics[width=0.18\textwidth]{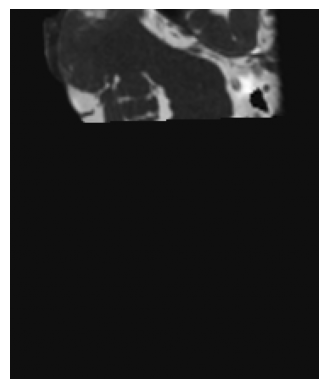}
    \includegraphics[width=0.18\textwidth]{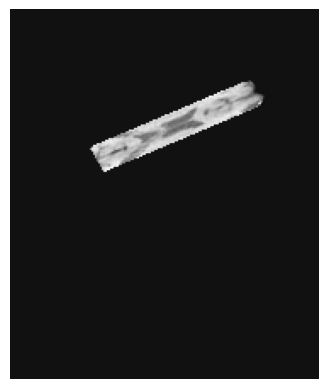}
    \includegraphics[width=0.18\textwidth]{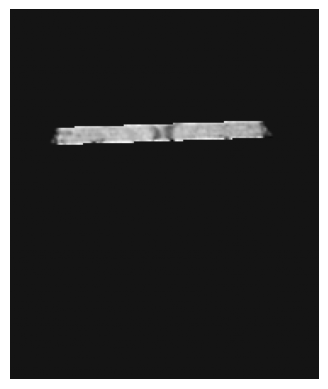}
    \includegraphics[width=0.18\textwidth]{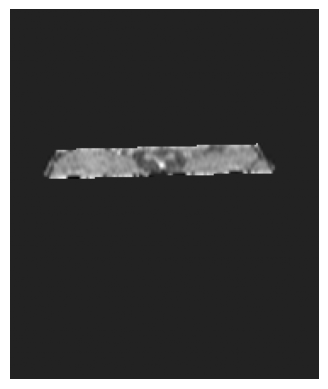}\\[6pt]
    \includegraphics[width=0.18\textwidth]{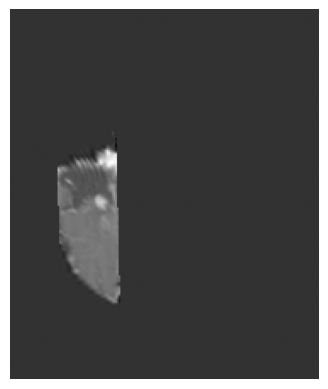}
    \includegraphics[width=0.18\textwidth]{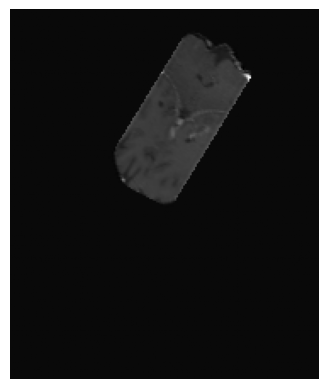}
    \includegraphics[width=0.18\textwidth]{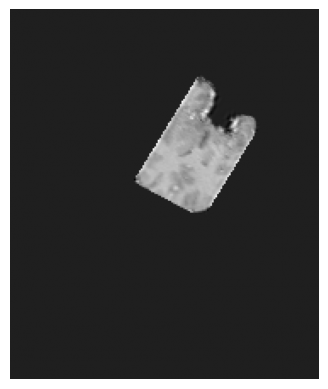}
    \includegraphics[width=0.18\textwidth]{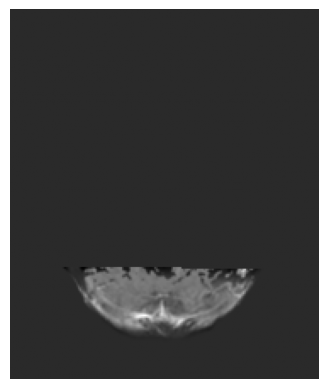}
    \includegraphics[width=0.18\textwidth]{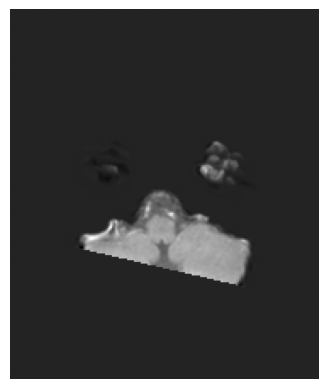}
    \caption{\textbf{Samples of Filtered Out Noisy Scans.} We show representative slices from scans excluded during scan-level quality control. Many of these scans contained limited usable anatomical information after registration, primarily due to restricted or incorrect fields of view, severe motion artifacts, acquisition or reconstruction failures, and other technical issues. Empirically, we found that pretraining with such scans can lead to unstable training dynamics and degraded downstream model performance. These observations underscore the importance of systematic data curation when developing neuroimaging foundation models from large-scale clinical cohorts collected directly from routine hospital practice.}
    \label{fig:noise_samples}
\end{figure}

Supplementary \Cref{fig:noise_samples} shows examples on some filtered out samples. Most samples present misaligned or missing anatomical brain structure that can potentially introduce pure noisy signals on model pre-training and make JEPA training trajectory to be unstable due to lack of visible context information on predicting the masked regions.

\subsection{Ablation Study on MAE vs. JEPA Pre-training Performance}
\begin{figure}[htbp]
    \centering
    \includegraphics[width=1.0\textwidth]{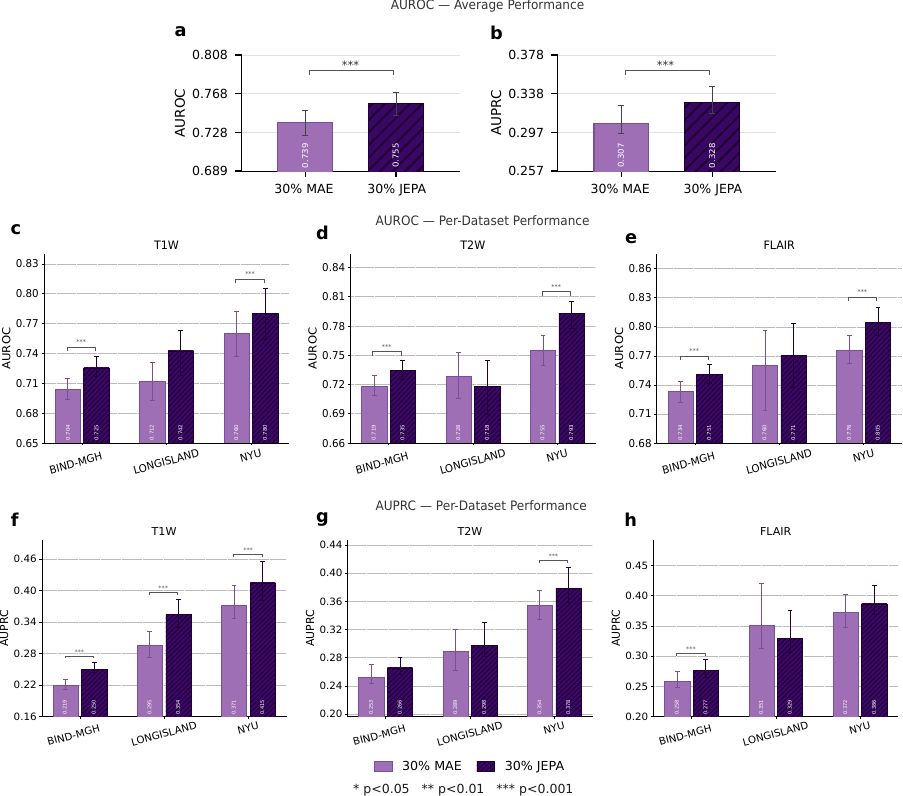}
    \caption{\textbf{MAE vs. JEPA Performance Across Datasets.} We compare the performance trained on $30\%$ of pretrain data, where we show JEPA on our full configurations with Multi-Scale Masking, MoE and Foreground Aware L1 Loss consistently outperform MAE. \textbf{a,b,} AUROC and AUPRC performance comparison averaged across all datasets and modalities \textbf{c-e,} AUROC performance comparison on each dataset and modality. \textbf{f-g,} AUPRC performance comparison on each dataset and modality.}
    \label{fig:mae_vs_jepa_abla}
\end{figure}

Supplementary \Cref{fig:mae_vs_jepa_abla} presents comparison on MAE vs. JEPA pre-training on $30\%$ data with same architecture. The result is evaluated on clinical cohorts datasets with attentive probing. The result demonstrates that our improved JEPA pre-training can effectively improve model overall performance over MAE across the datasets and modalities.

\newpage
\section{Generalization Under Cohort and Modality Shifts}
\subsection{Cross-Cohort Transfer Across Matched Clinical Endpoints}
\begin{figure}[htbp]
    \centering
    \includegraphics[width=0.85\textwidth]{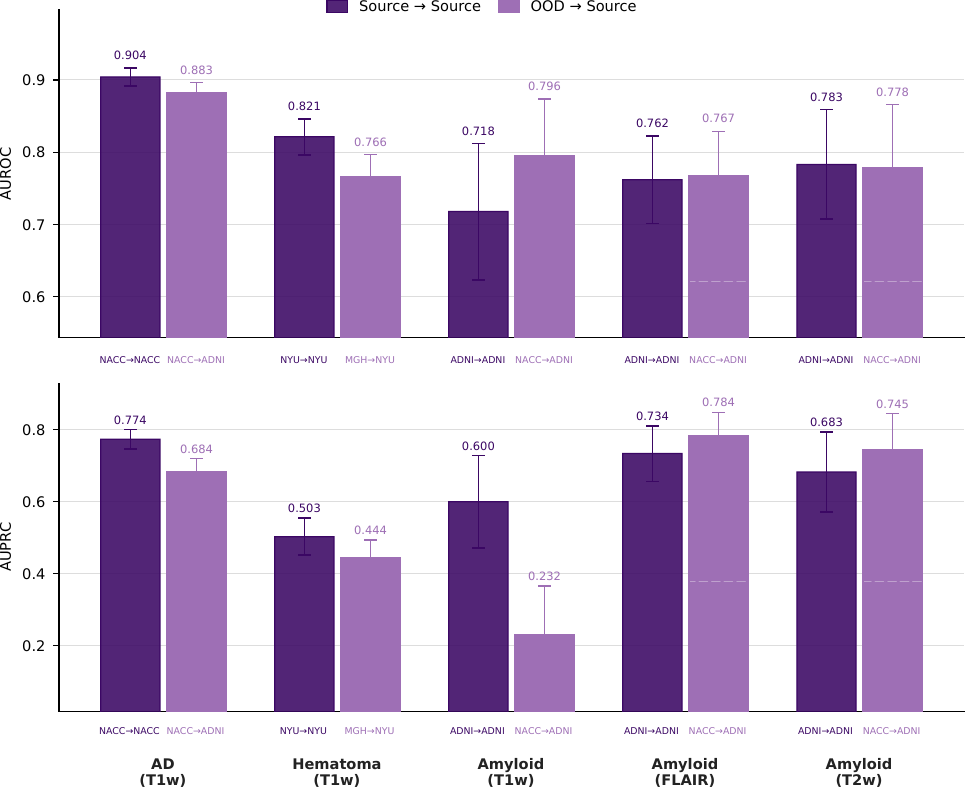}
    \caption{\textbf{Cross-cohort out-of-domain transfer performance.}
    AUROC and AUPRC are reported for models fine-tuned on one cohort and evaluated on an external cohort with matched task definitions. The evaluated transfer settings include NACC to ADNI for Alzheimer's disease and amyloid prediction, and MGH to NYU for hematoma prediction. Transfer performance is compared with in-domain performance, where models are trained and evaluated within the same cohort. Neuro-JEPA preserves strong predictive performance under external cohort shift in a majority of tasks, supporting its robustness in out-of-domain transfer settings.}
    \label{fig:sup-ood_transfer}
\end{figure}

Supplementary \Cref{fig:sup-ood_transfer} evaluates the out-of-domain transfer performance of Neuro-JEPA across independent clinical cohorts. In this setting, the model is fine-tuned on a source cohort and evaluated directly on an external target cohort with the same label definition, providing a clinically relevant assessment of robustness under cohort and institutional distribution shifts. Across the evaluated tasks, Neuro-JEPA maintains strong transfer performance relative to in-domain evaluation, demonstrating effective generalization beyond the source-domain data.

\subsection{Cross-Modality Generalization to DWI}
\begin{figure}[htbp]
    \centering
    \includegraphics[width=1.0\textwidth]{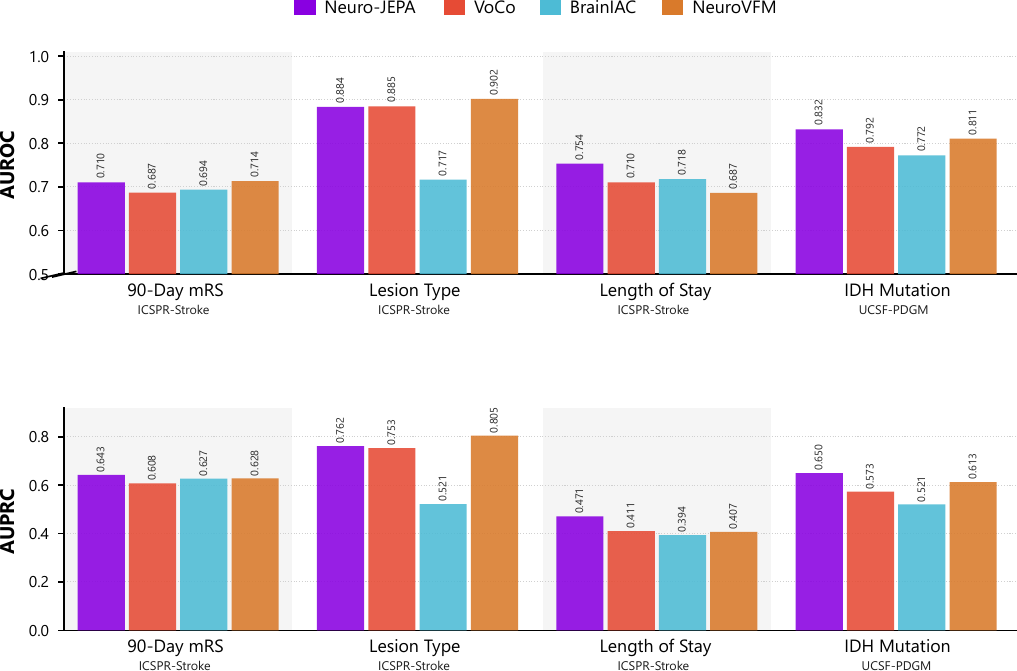}
    \caption{\textbf{Performance comparison on DWI downstream tasks.}
    Metrics are macro-averaged AUROC and AUPRC. For the three-class lesion type task, both metrics are computed using a one-versus-rest macro average.
    \textbf{Bold} indicates the best-performing model for each metric, and \underline{underlining} indicates the second-best model.}
    \label{fig:sup-ood_modalities}
\end{figure}

Supplementary \Cref{fig:sup-ood_modalities,tab:dwi_comparison} evaluates the ability of Neuro-JEPA to generalize to diffusion-weighted MRI (DWI), a modality that was not included during pretraining. Models are fine-tuned on DWI scans and evaluated on downstream tasks from the ICSPR-Stroke and UCSF-PDGM datasets. Despite the absence of DWI during pretraining, Neuro-JEPA achieves competitive performance across tasks and obtains the best macro-averaged AUROC and AUPRC among the evaluated models. These results indicate that Neuro-JEPA learns representations that transfer beyond the imaging modalities observed during pretraining.

\begin{table}[htbp]
  \centering
  \caption{%
    \textbf{Performance comparison on DWI downstream tasks.}
    Metrics are macro-averaged AUROC and AUPRC. For the three-class lesion type task, both metrics are computed using a one-versus-rest macro average.
    \textbf{Bold} indicates the best-performing model for each metric, and \underline{underlining} indicates the second-best model.
  }
  \label{tab:dwi_comparison}
  \begin{tabular}{lrrrrrrrr}
    \toprule
     & \multicolumn{2}{c}{\textbf{Neuro-JEPA-B}} & \multicolumn{2}{c}{\textbf{VoCo}} & \multicolumn{2}{c}{\textbf{BrainIAC}} & \multicolumn{2}{c}{\textbf{NeuroVFM}} \\
    \cmidrule(lr){2-3} \cmidrule(lr){4-5} \cmidrule(lr){6-7} \cmidrule(lr){8-9}
    \textbf{Task} & \textbf{AUROC} & \textbf{AUPRC} & \textbf{AUROC} & \textbf{AUPRC} & \textbf{AUROC} & \textbf{AUPRC} & \textbf{AUROC} & \textbf{AUPRC} \\
    \midrule
    \multicolumn{9}{l}{\textit{\small ICSPR-Stroke}} \\[-4pt]
    \quad 90-Day mRS & \underline{0.710} & \textbf{0.643} & 0.687 & 0.608 & 0.694 & 0.627 & \textbf{0.714} & \underline{0.628} \\
    \quad Lesion Type & 0.884 & \underline{0.762} & \underline{0.885} & 0.753 & 0.717 & 0.521 & \textbf{0.902} & \textbf{0.805} \\
    \quad Length of Stay & \textbf{0.754} & \textbf{0.471} & 0.710 & \underline{0.411} & \underline{0.718} & 0.394 & 0.687 & 0.407 \\
    \addlinespace[2pt]
    \multicolumn{9}{l}{\textit{\small UCSF-PDGM}} \\[-4pt]
    \quad IDH Mutation & \textbf{0.832} & \textbf{0.650} & 0.792 & 0.573 & 0.772 & 0.521 & \underline{0.811} & \underline{0.613} \\
    \midrule
    \textbf{Average} & \textbf{0.795} & \textbf{0.631} & 0.769 & 0.586 & 0.725 & 0.516 & \underline{0.778} & \underline{0.613} \\
    \bottomrule
  \end{tabular}
\end{table}

\newpage
\section{Pre-Training Optimization Dynamics}
\label{apd:pretrain_opt}
\begin{figure}[htbp]
    \centering
    \includegraphics[width=0.9\textwidth]{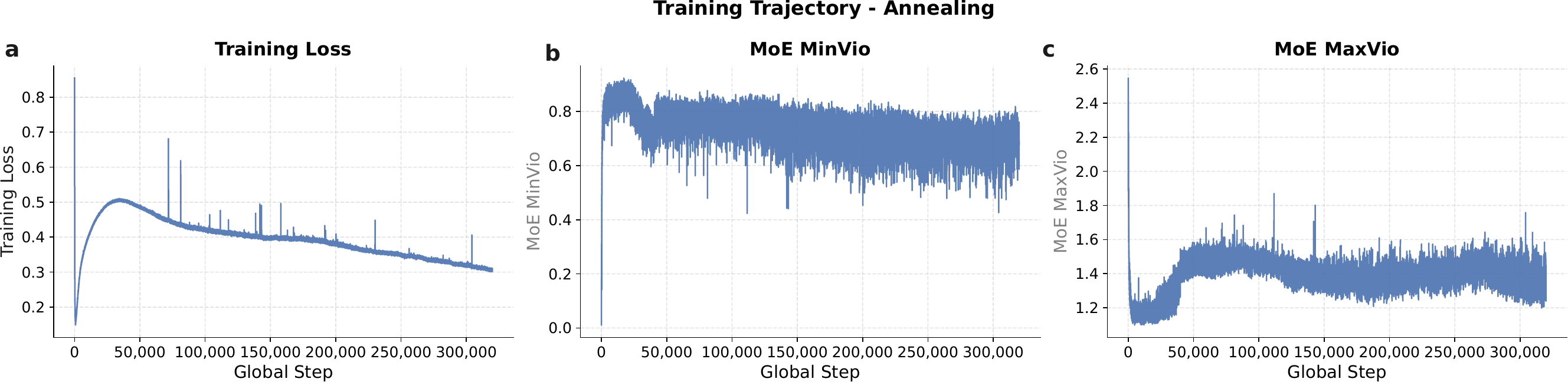}
    \caption{\textbf{Training Dynamics on Full-data Annealing Pretraining.}
    We show the optimization trajectory of Neuro-JEPA trained for 200 epochs annealing on the full pretraining dataset with our base model, using multiscale masking, Mixture-of-Experts (MoE) routing, and foreground-aware L1 latent predictive loss. \textbf{a,} Latent predictive L1 loss over training steps. \textbf{b,} Minimum MoE load-balancing violation, used to monitor under-utilization of experts. \textbf{c,} Maximum MoE load-balancing violation, used to monitor over-utilization of experts.}
    \label{fig:training_trajectory_annealing}
\end{figure}

\begin{figure}[htbp]
    \centering
    \includegraphics[width=0.9\textwidth]{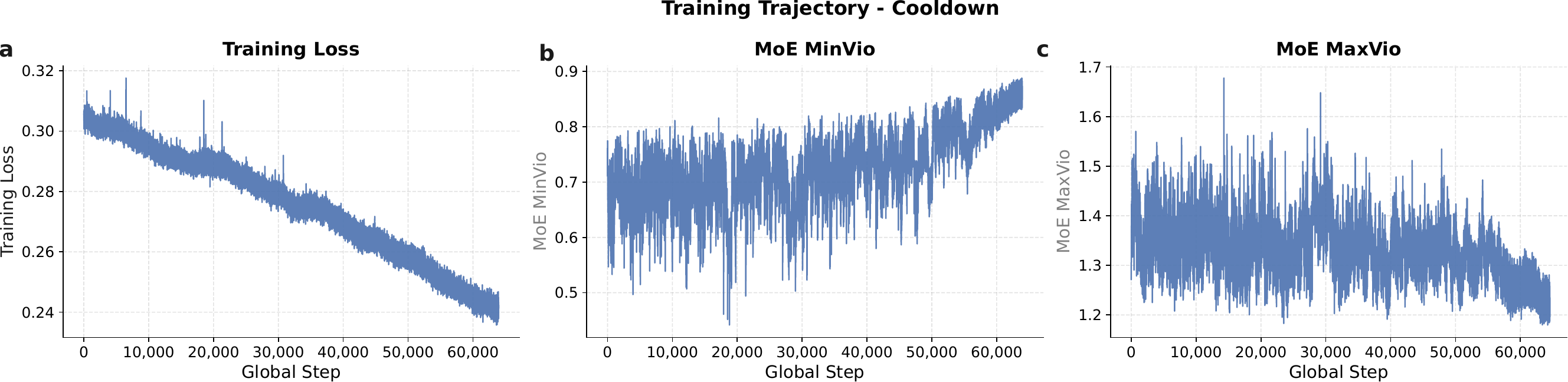}
    \caption{\textbf{Training Dynamics on Full-data Cooldown Pretraining.}
    We show the optimization trajectory of Neuro-JEPA trained for 40 epochs cooldown on the full pretraining dataset with our base model, using multiscale masking, Mixture-of-Experts (MoE) routing, and foreground-aware L1 latent predictive loss. \textbf{a,} Latent predictive L1 loss over training steps. \textbf{b,} Minimum MoE load-balancing violation, used to monitor under-utilization of experts. \textbf{c,} Maximum MoE load-balancing violation, used to monitor over-utilization of experts.}
    \label{fig:training_trajectory_cooldown}
\end{figure}

Supplementary \Cref{fig:training_trajectory_annealing,fig:training_trajectory_cooldown} presents training dynamics on training loss and Mixture of Experts load balancing (Minimum Violation and Maximum Violation) with $100\%$ pretraining data on our base model for 200 epochs annealing and 40 epochs cooldown. The result demonstrates that our model can be stably pre-trained under proper hyper-parameters setup. More details on pretraining dynamics can be found in the public wandb report \url{https://api.wandb.ai/links/notody/7t7d4dks}.

\newpage

\section{Additional Validation Analyses}
\subsection{Comparison with Simple CNN Baseline}
\label{apd:cnn_baseline}
\begin{figure}[htbp]
    \centering
    \includegraphics[width=0.95\textwidth]{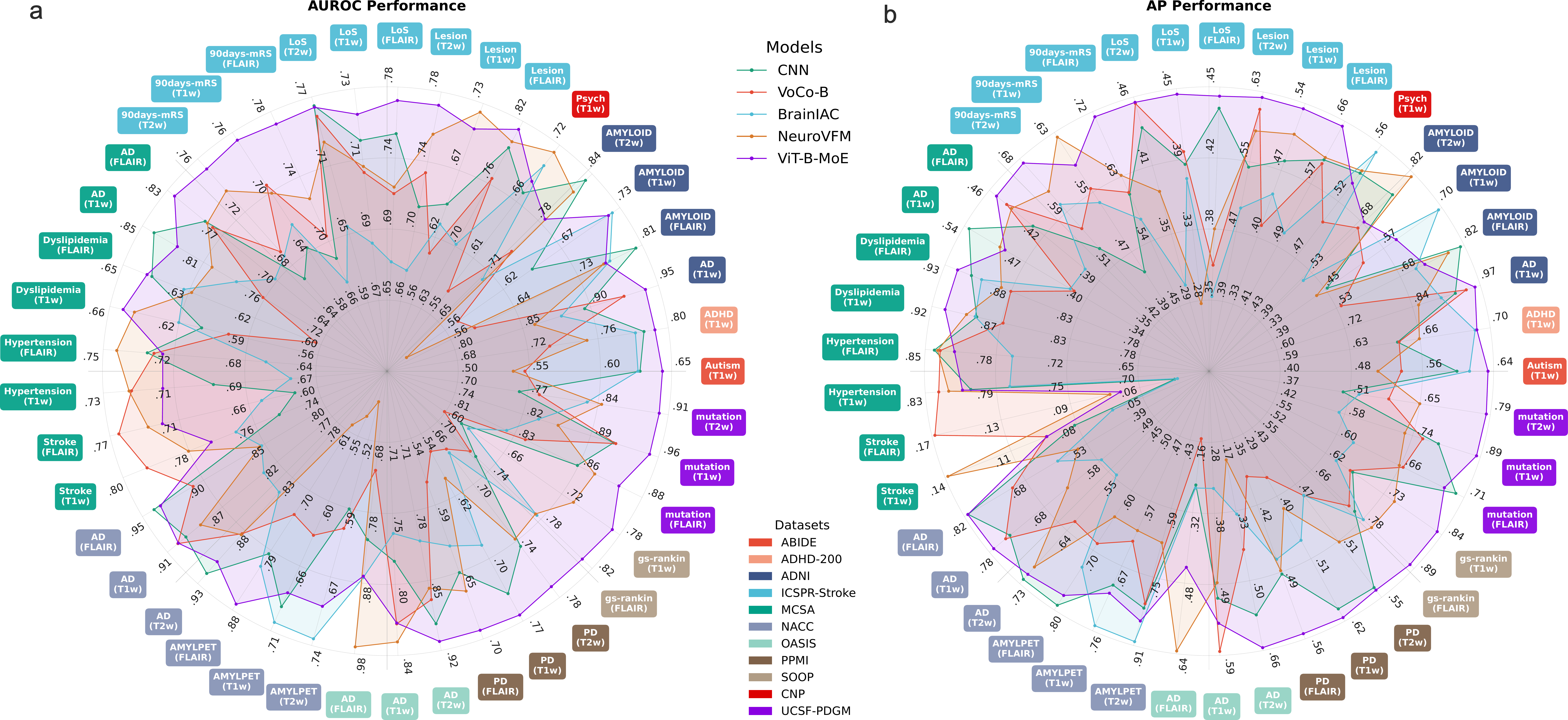}
    \caption{\textbf{All Evaluated Models vs. Simple CNN Baseline.} AUROC and AUPRC for 41 different combinations on public datasets. \textbf{a,} AUROC comparison across tasks and modalities. \textbf{b,} AUPRC comparison across tasks and modalities.}
    \label{fig:simple_cnn_compare}
\end{figure}

To assess whether foundation-model pretraining provides practical value beyond conventional simple models, we benchmarked a specific designed CNN baseline for neuroimaging~\cite{Liu2022} (a wide 4-layers CNN with 32 million parameters) trained directly for the downstream task from scratch. All evaluations follow hyperparameters sweep detailed in Supplementary \Cref{apd:model_eval_details}. This comparison provides an important calibration on if pre-existing foundation models bring any benefits beyond simple methods. The evaluation is performed on the 41 combinations from 12 public datasets and age prediction. Across these evaluations, existing foundation models did not consistently surpass the CNN baseline, whereas Neuro-JEPA achieved consistent gains over the CNN on both averaged AUROC (3.7\% improvement) and AUPRC (4.5\% improvement) (Supplementary \Cref{tab:simple_cnn_compare}; Supplementary \Cref{fig:simple_cnn_compare}). On age prediction, Neuro-JEPA presents the only foundation model performs better with Quasi-RAW scans (+2.8 on $R^2$, -0.37 on $MAE$ and -0.50 on $RMSE$). The results indicate that Neuro-JEPA represents first successful foundation model on surpassing simple CNN baseline with proper scaling and algorithmic design.

\begin{table*}[t]
\centering
\footnotesize
\setlength{\tabcolsep}{4pt}
\renewcommand{\arraystretch}{1.15}
\caption{%
\textbf{Average unimodal performance across cohorts and task types (Public datasets).}
Values are mean [95\% CI]. \textbf{Bold}: best per metric.
$\uparrow$~higher is better; $\downarrow$~lower is better. "combs" refer to the number of dataset-task-modality combinations.
"n" for brain age represents number of samples in the test set. \underline{underlining} indicates the second-best model.
}
\label{tab:simple_cnn_compare}

\begin{minipage}[t]{0.42\linewidth}
\centering
\textbf{(a) Classification / Diagnosis / Prognosis} (41 combs)

\smallskip
\begin{tabular}{lcc}
\toprule
\textbf{Model} & AUROC$\uparrow$ & AUPRC$\uparrow$ \\
\midrule
VoCo-B
& 0.721 {\tiny[0.689, 0.754]} & 0.573 {\tiny[0.513, 0.635]} \\
BrainIAC
& 0.728 {\tiny[0.705, 0.751]} & 0.555 {\tiny[0.494, 0.618]} \\
NeuroVFM
& 0.741 {\tiny[0.706, 0.774]} & 0.585 {\tiny[0.526, 0.648]} \\
CNN
& \underline{0.748} {\tiny[0.718, 0.779]} & \underline{0.604} {\tiny[0.539, 0.668]} \\
\rowcolor{gray!12}
\textbf{Neuro-JEPA}
& \textbf{0.785} {\tiny[0.760, 0.811]} & \textbf{0.649} {\tiny[0.588, 0.705]} \\
\bottomrule
\end{tabular}
\end{minipage}
\hfill
\begin{minipage}[t]{0.54\linewidth}
\centering
\textbf{(b) Brain Age Prediction} ($n = 757$)

\smallskip
\begin{tabular}{lccc}
\toprule
\textbf{Model} & $R^{2}\uparrow$ & MAE$\downarrow$ (yr) & RMSE$\downarrow$ (yr) \\
\midrule
VoCo-B
& 0.111 {\tiny[0.065, 0.165]} & 6.22 {\tiny[5.46, 6.94]} & 12.03 {\tiny[10.54, 13.36]} \\
BrainIAC
& 0.522 {\tiny[0.447, 0.591]} & 5.42 {\tiny[4.93, 5.95]} & 8.82 {\tiny[7.77, 9.78]} \\
NeuroVFM
& 0.673 {\tiny[0.611, 0.735]} & 4.36 {\tiny[3.91, 4.77]} & 7.29 {\tiny[6.35, 8.12]} \\
CNN
& \underline{0.858} {\tiny[0.827, 0.883]} & \underline{3.25} {\tiny[2.99, 3.50]} & \underline{4.81} {\tiny[4.39, 5.21]} \\
\rowcolor{gray!12}
\textbf{Neuro-JEPA}
& \textbf{0.894} {\tiny[0.860, 0.917]} & \textbf{2.78} {\tiny[2.56, 3.02]} & \textbf{4.15} {\tiny[3.69, 4.70]} \\
\bottomrule
\end{tabular}
\end{minipage}
\end{table*}

\newpage
\subsection{Few-shot Analysis}
\label{apd:few_shot}
\subsubsection{AUROC and AUPRC on Few-shot Performance Averaged Across Modalities}
We report few-shot model performance on AUROC, AUPRC and MAE for age prediction regression averaged across differnt modalities on more diverse tasks in Supplementary \Cref{fig:sup-few-shot-avg-auroc,fig:sup-few-shot-avg-auprc}. Consistently with the finding in the main article Figure 5. We found consistently improved few-shot performance for Neuro-JEPA across majority of tasks on both AUROC and AUPRC.

\subsubsection{Per-modality AUROC and AUPRC Few-short Performance}
We report few-shot model performance on AUROC, AUPRC and MAE for age prediction regression per modality (T1w, T2w, FLAIR) in Supplementary \Cref{fig:sup-few-shot-permod-t1w-auroc,fig:sup-few-shot-permod-t2w-auroc,fig:sup-few-shot-permod-flair-auroc,fig:sup-few-shot-permod-t1w-auprc,fig:sup-few-shot-permod-t2w-auprc,fig:sup-few-shot-permod-flair-auprc}. Per modality performance is consistent with the averaged performance where Neuro-JEPA improved in comparison to other evaluated models.

\begin{figure}[htbp]
    \centering
    \includegraphics[width=0.8\textwidth]{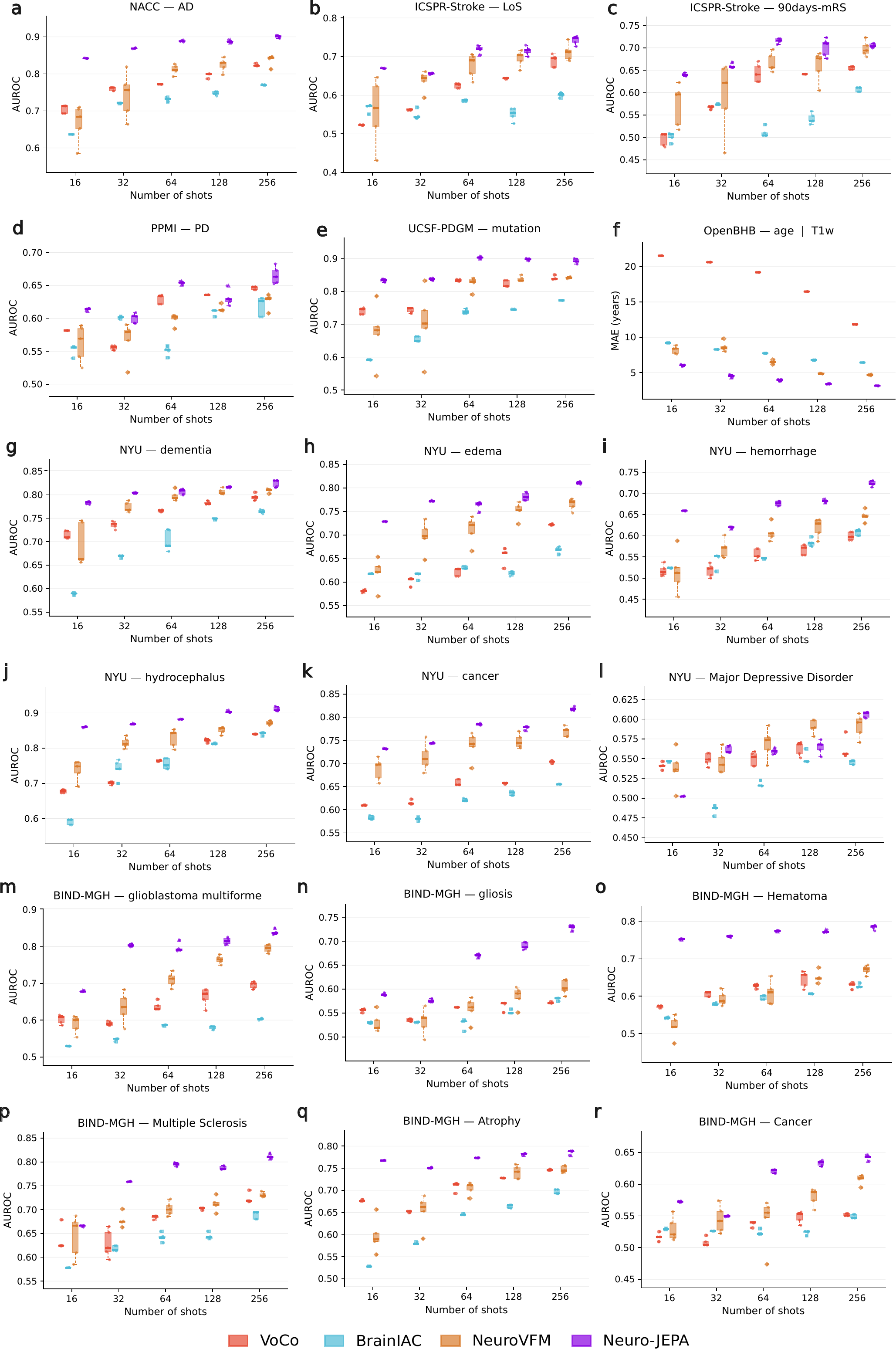}
    \caption{\textbf{Few-shot Analysis} - we examine the evaluated models label efficiency when only $k=\{16,32,64,128,256\}$ positive samples are provided on more diverse selected tasks. The performance in reported in AUROC for classification and MAE for regression. \textbf{a-f,} Few-shot performance on selected tasks from public datasets. All result is reported as averaged performance across all available modalities for each task \textbf{g-l,} Few-shot performance on selected tasks from NYU-Langone dataset. \textbf{m-r,} Few-shot performance on selected tasks from BIND-MGH dataset.}
    \label{fig:sup-few-shot-avg-auroc}
\end{figure}

\begin{figure}[htbp]
    \centering
    \includegraphics[width=0.8\textwidth]{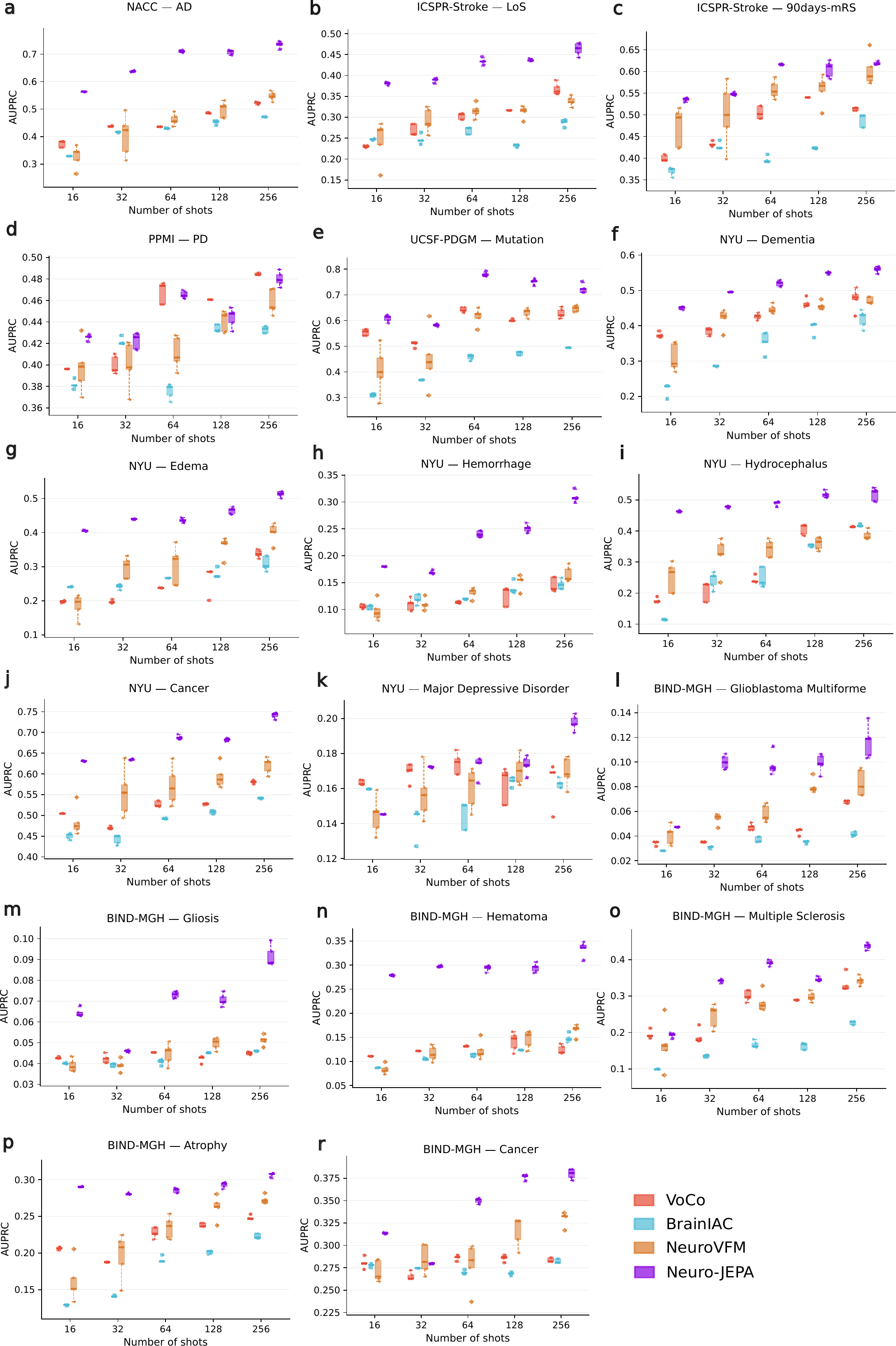}
    \caption{\textbf{Few-shot Analysis} - we examine the evaluated models label efficiency when only $k=\{16,32,64,128,256\}$ positive samples are provided on more diverse selected tasks. The performance in reported in AUPRC for classification. \textbf{a-e,} Few-shot performance on selected tasks from public datasets. All result is reported as averaged performance across all available modalities for each task \textbf{f-k,} Few-shot performance on selected tasks from NYU-Langone dataset. \textbf{l-r,} Few-shot performance on selected tasks from BIND-MGH dataset.}
    \label{fig:sup-few-shot-avg-auprc}
\end{figure}
\begin{figure}[htbp]
    \centering
    \includegraphics[width=0.75\textwidth]{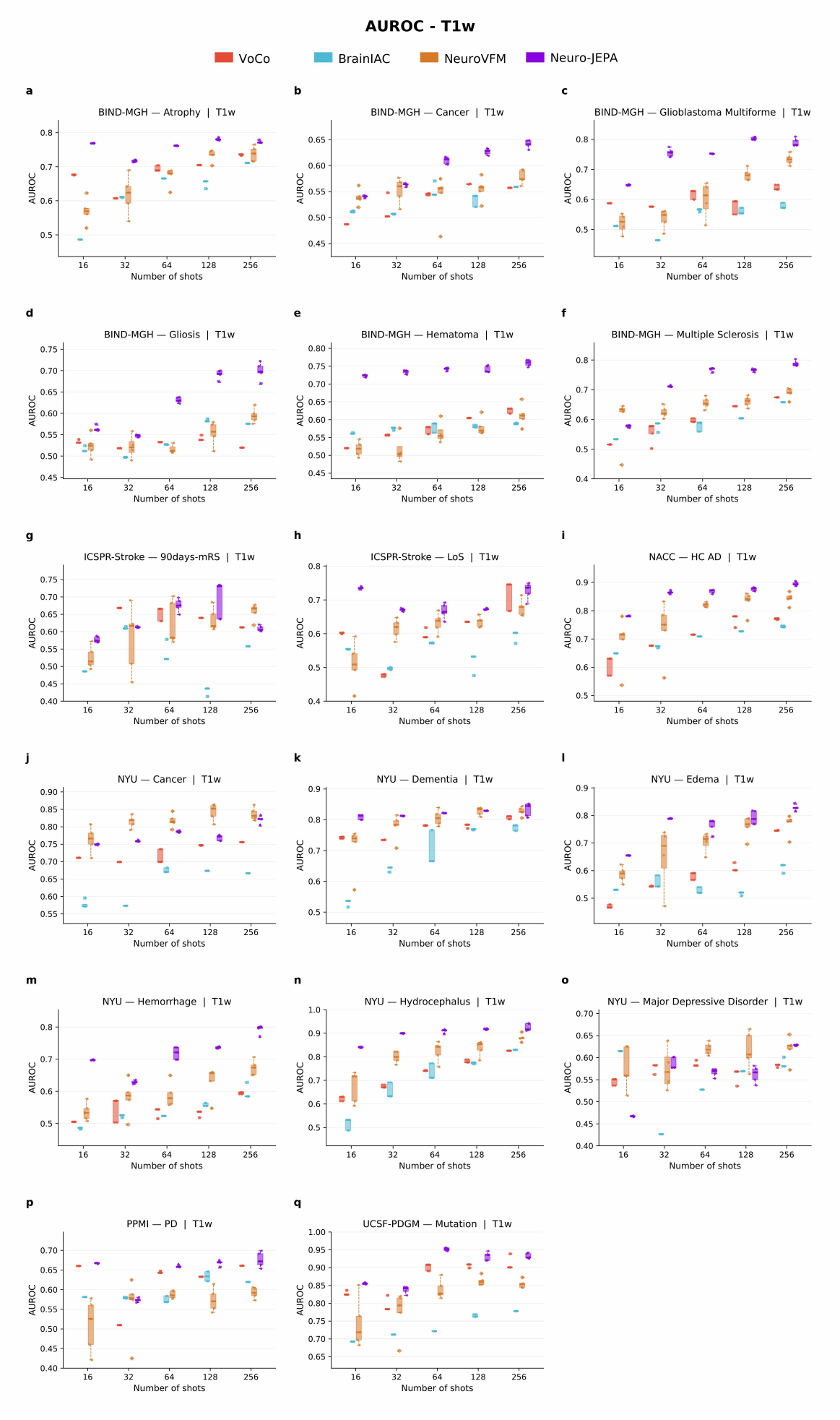}
    \caption{\textbf{AUROC Few-shot Analysis on T1w} - few-shot performance across tasks for T1w.}
    \label{fig:sup-few-shot-permod-t1w-auroc}
\end{figure}

\begin{figure}[htbp]
    \centering
    \includegraphics[width=0.75\textwidth]{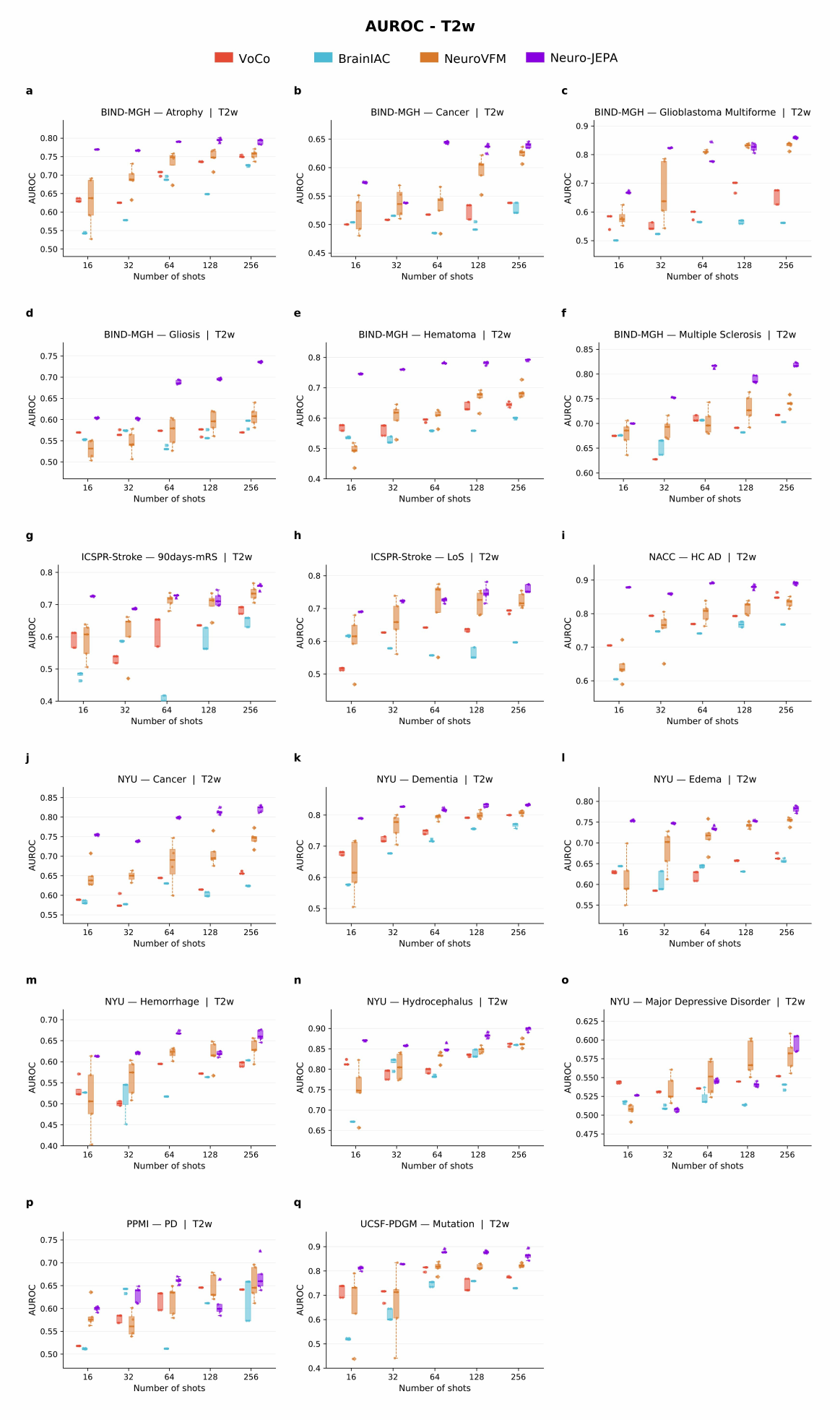}
    \caption{\textbf{AUROC Few-shot Analysis on T2w} - few-shot performance across tasks for T2w.}
    \label{fig:sup-few-shot-permod-t2w-auroc}
\end{figure}

\begin{figure}[htbp]
    \centering
    \includegraphics[width=0.75\textwidth]{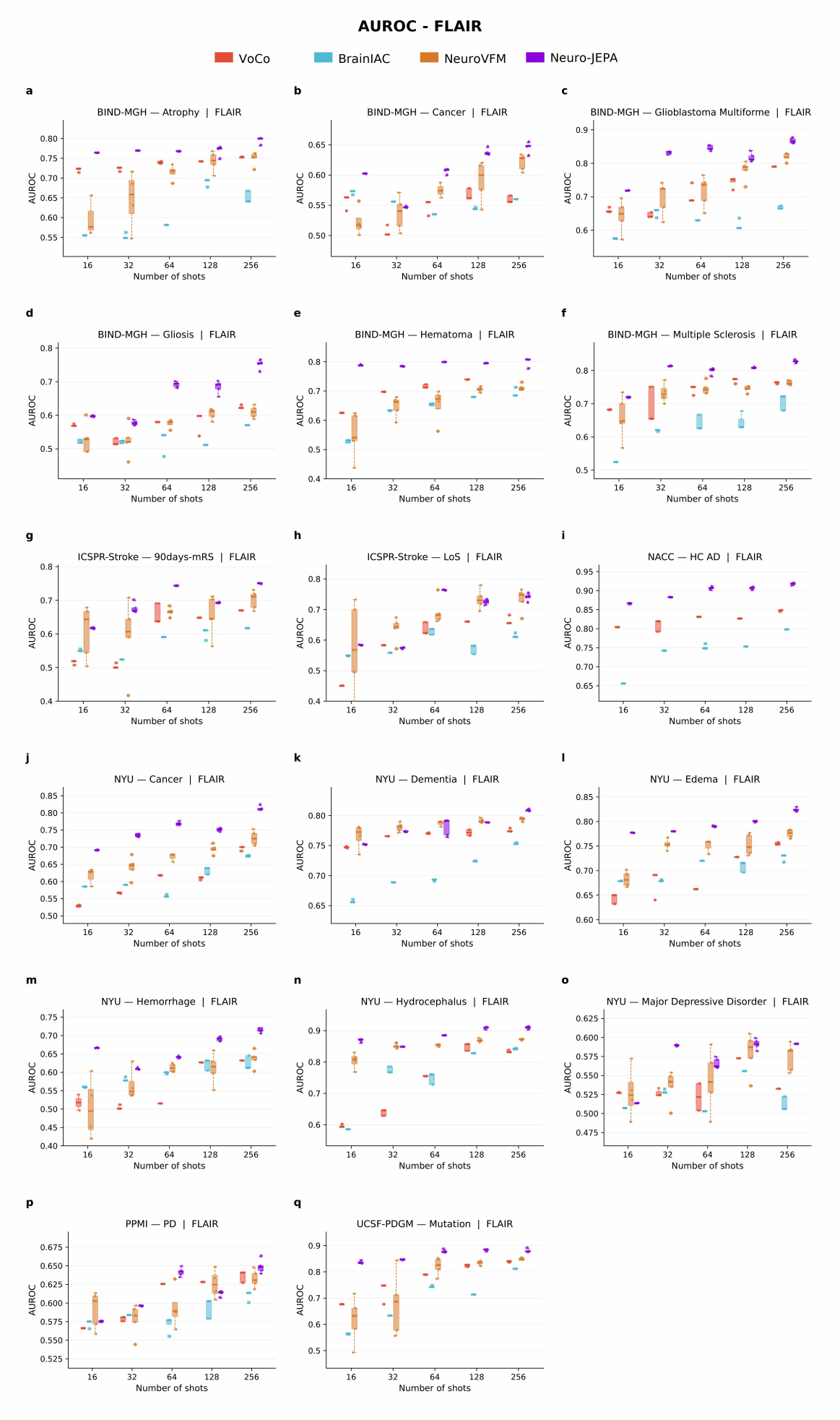}
    \caption{\textbf{AUROC Few-shot Analysis on FLAIR} - few-shot performance across tasks for FLAIR.}
    \label{fig:sup-few-shot-permod-flair-auroc}
\end{figure}

\begin{figure}[htbp]
    \centering
    \includegraphics[width=0.75\textwidth]{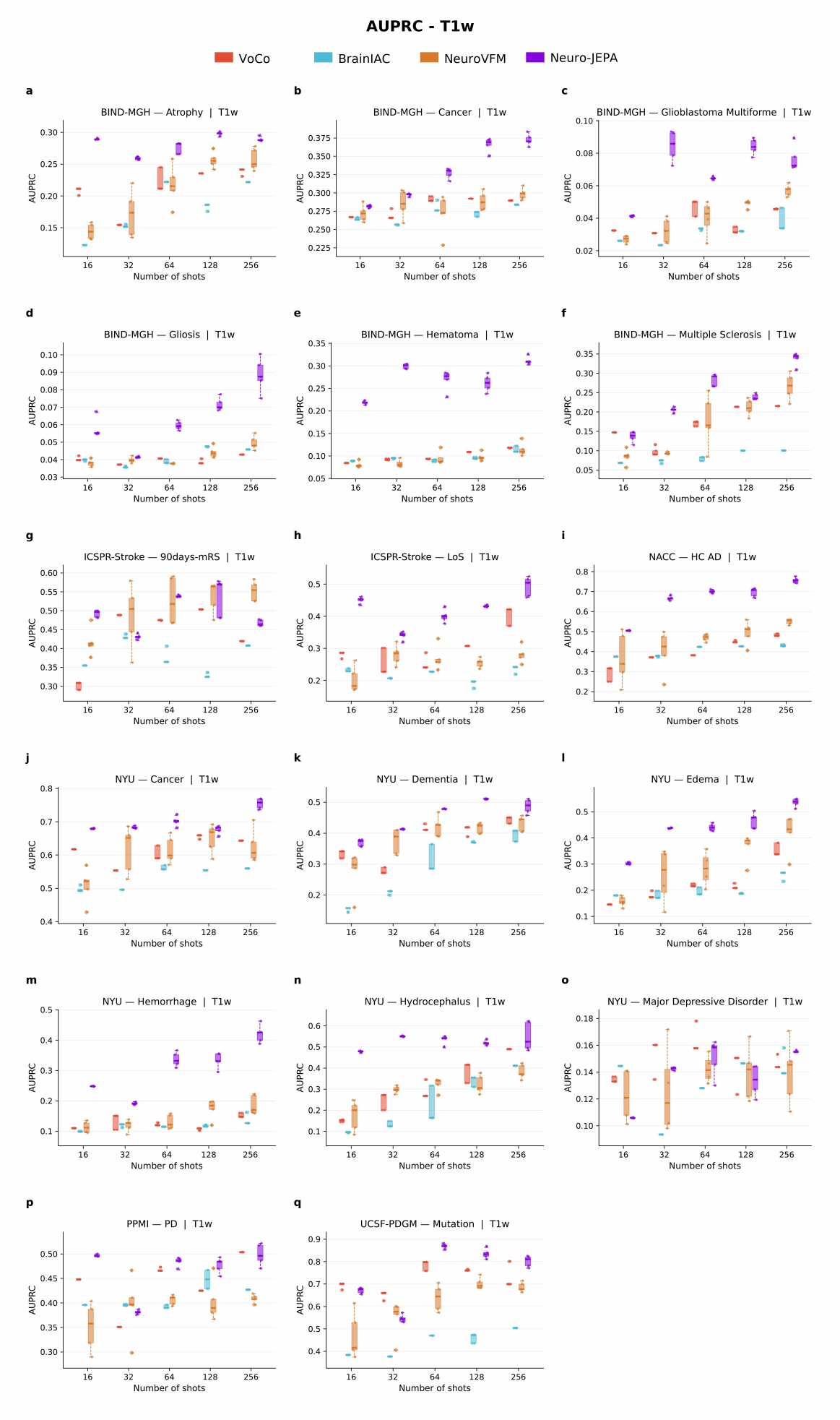}
    \caption{\textbf{AUPRC Few-shot Analysis on T1w} - few-shot performance across tasks for T1w.}
    \label{fig:sup-few-shot-permod-t1w-auprc}
\end{figure}

\begin{figure}[htbp]
    \centering
    \includegraphics[width=0.75\textwidth]{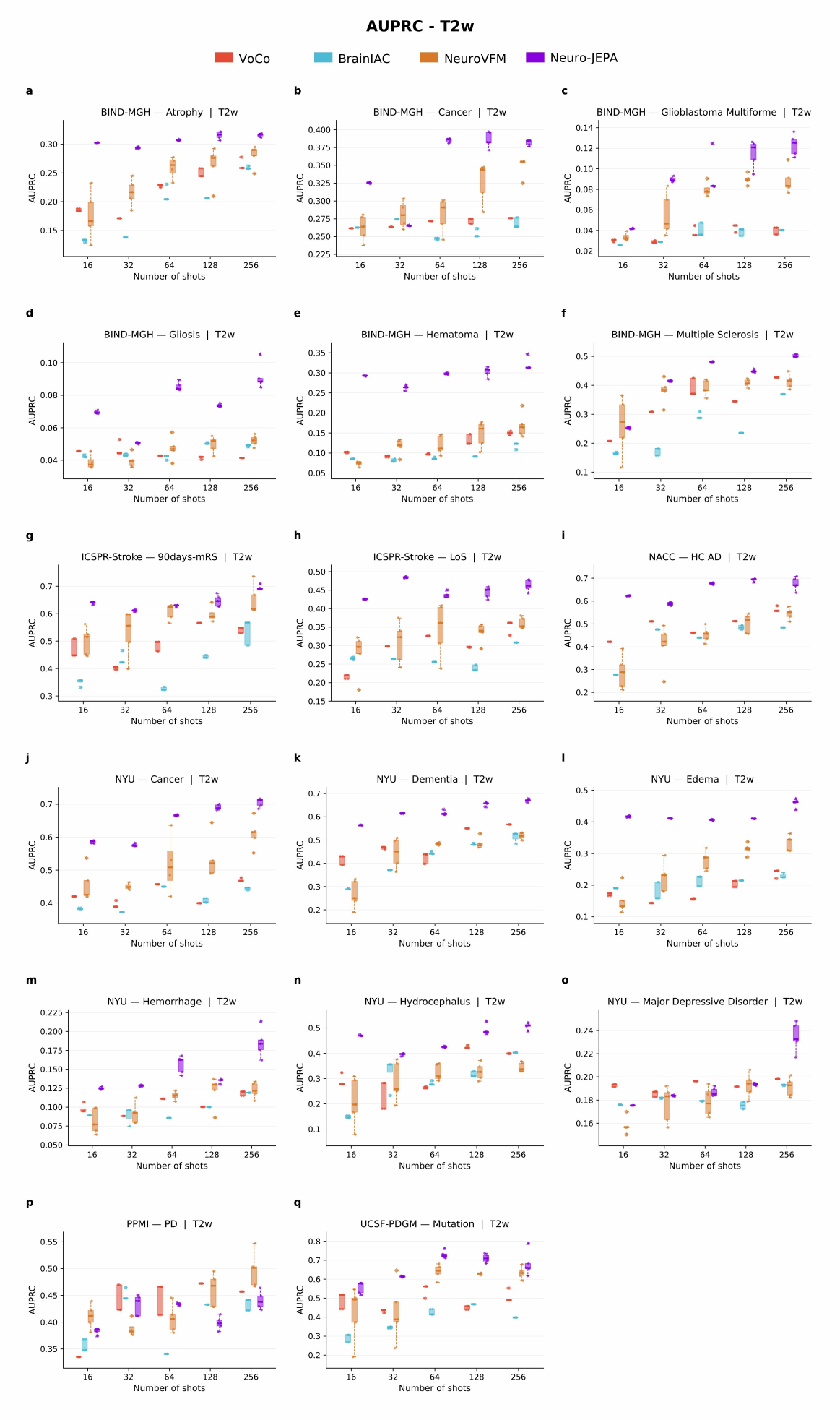}
    \caption{\textbf{AUPRC Few-shot Analysis on T2w} - few-shot performance across tasks for T2w.}
    \label{fig:sup-few-shot-permod-t2w-auprc}
\end{figure}

\begin{figure}[htbp]
    \centering
    \includegraphics[width=0.75\textwidth]{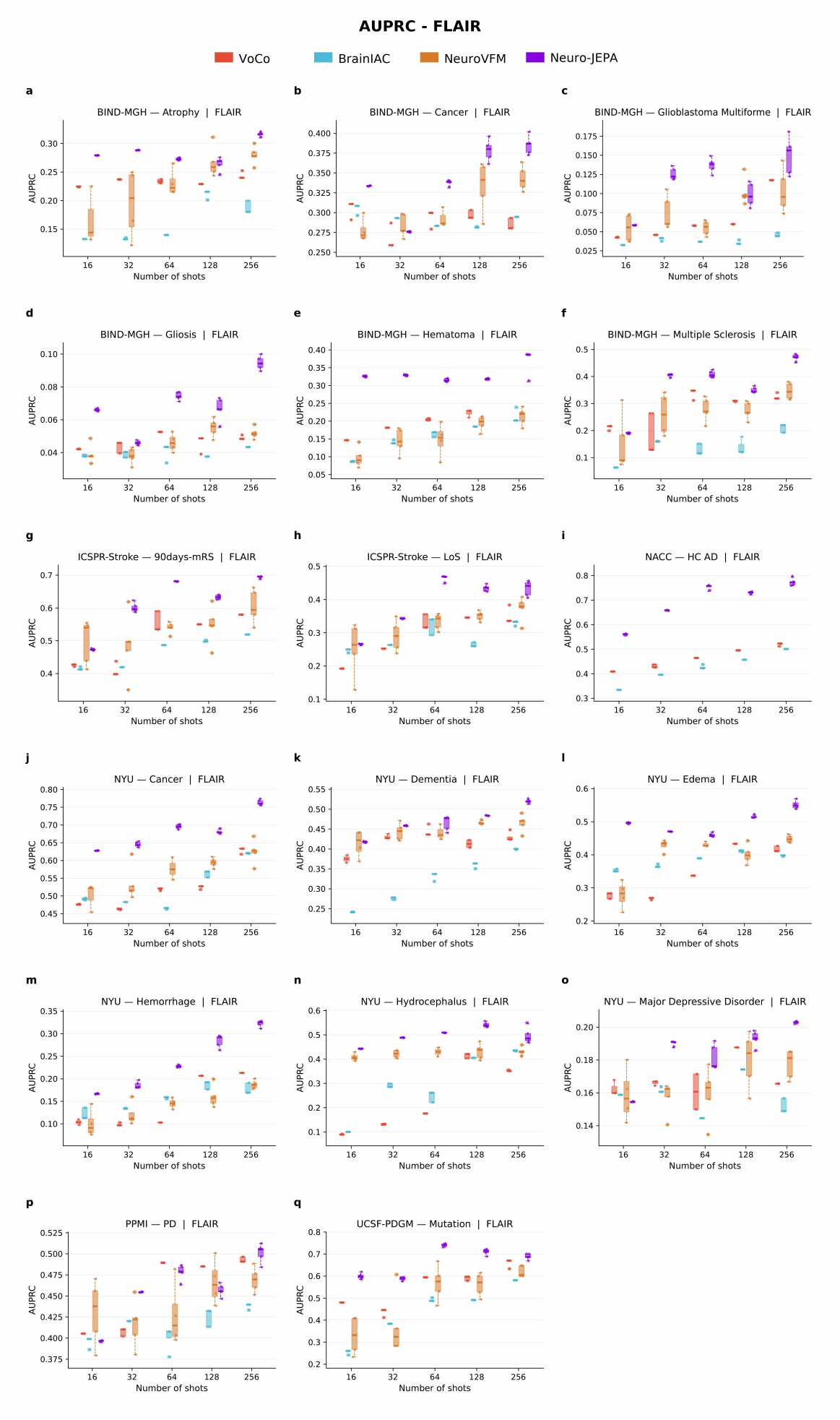}
    \caption{\textbf{AUPRC Few-shot Analysis on FLAIR} - few-shot performance across tasks for FLAIR.}
    \label{fig:sup-few-shot-permod-flair-auprc}
\end{figure}

\newpage

\subsection{Fairness Analysis}
\label{apd:fairness}
\begin{figure}[htbp]
    \centering
    \includegraphics[width=0.9\textwidth]{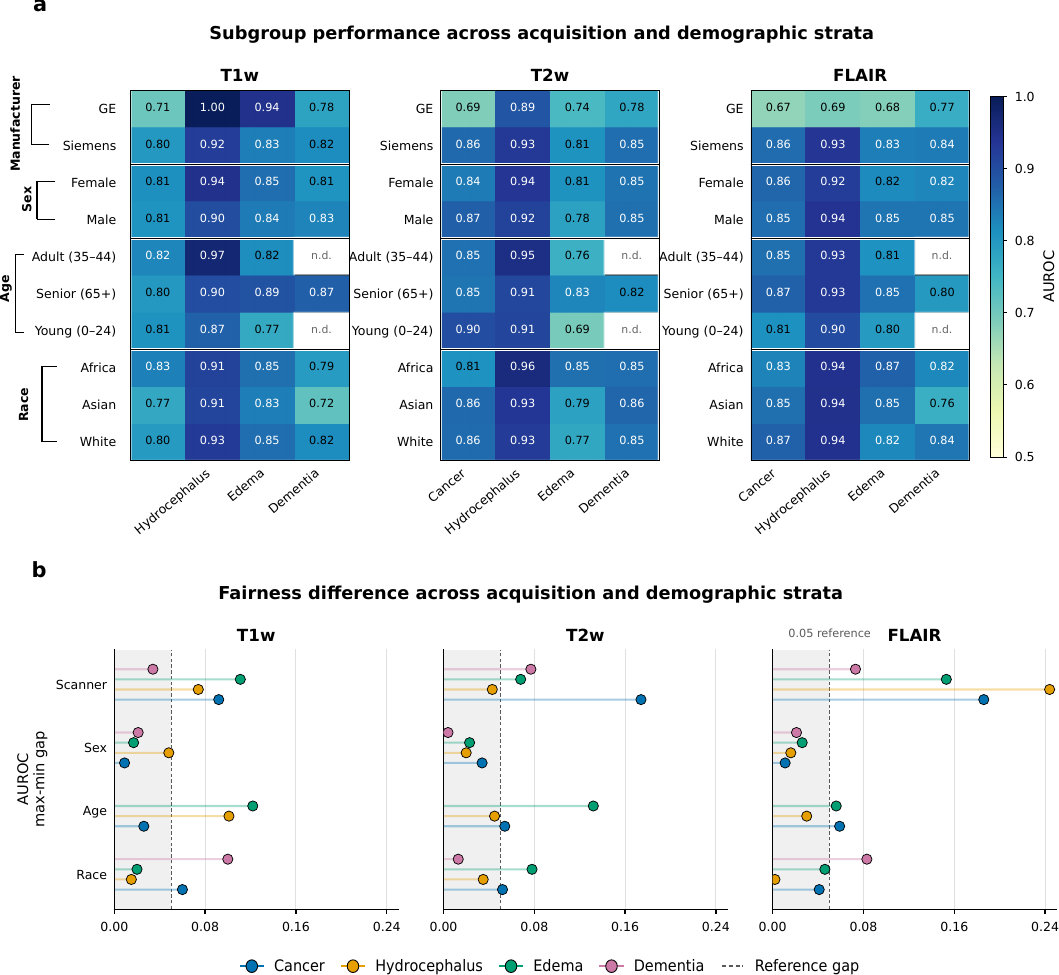}
    \caption{\textbf{Fairness Analysis across sub-cohorts.} Fairness comparison on representative diseases (Cancer, Hydrocephalus, Edema, Dementia) from NYU-Langone dataset across different sub-cohorts by attentive probing. \textbf{a,} AUROC on difference sub-cohorts and diseases. \textbf{b,} Fairness gap on maximum AUROC minus minimum AUROC for each sub-cohort and disease}
    \label{fig:sup-fairness}
\end{figure}

We performed a fairness analysis on the curated evaluation dataset across subgroups defined by age, sex, race and scanner manufacturer (Supplementary \Cref{fig:sup-fairness}). For each subgroup, we report AUROC and quantify the fairness gap as the difference between the maximum and minimum AUROC across subgroup categories. Overall, performance was broadly consistent across subgroups, with limited evidence of large disparities in most settings. The main exceptions were scanner-associated differences for T2w and FLAIR scans. The pattern is consistent with the underlying data distribution: Siemens scanners were predominant in the pretraining cohort. Thus, the observed subgroup gaps likely reflect realistic sources of distributional heterogeneity rather than systematic model failure.

\subsection{TSNE Visualization on Age Groups}
\label{apd:age_group_tsne}
We evaluated the separability of age-related representations across pretrained models using t-SNE projections of pooled embeddings, obtained by averaging token representations without further fine-tuning. Patients were grouped into three age cohorts: Young (0–24 years), Adult (35–44 years), and Senior (65+ years). The analysis was performed on a subset of NYU Langone patients with available age records who were excluded from the pretraining dataset. For each modality, 600 samples were analyzed, with 200 samples per age cohort. Results are shown in Supplementary Fig. \Cref{fig:sup-age-tsne}. Both visualization and quantitative silhouette scores indicate that Neuro-JEPA achieves the best age-group separability among the evaluated models.

\begin{figure}[htbp]
    \centering
    \includegraphics[width=1.0\textwidth]{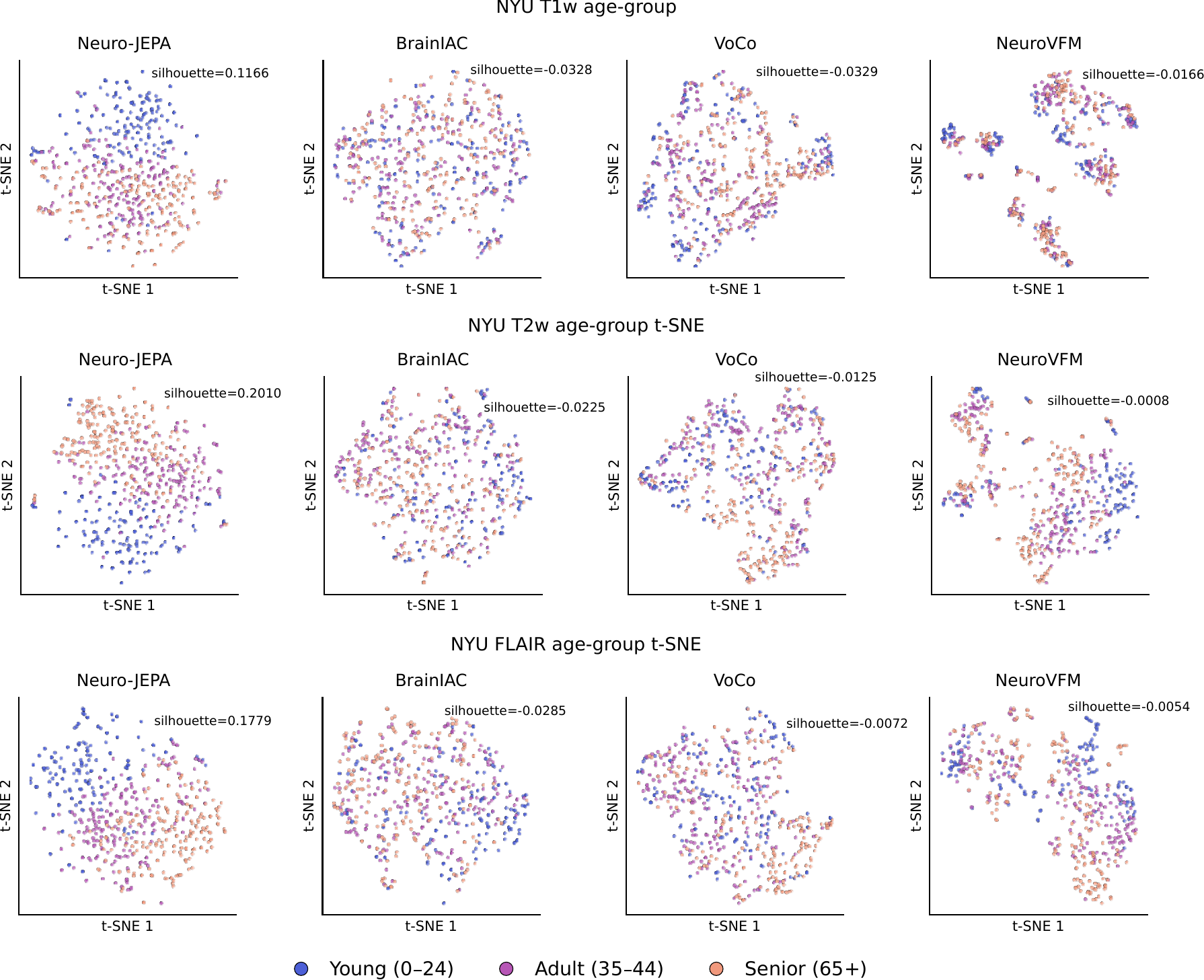}
    \caption{\textbf{Age TSNE Across Different Pretrained Models.} The plot present TSNE visualization on different age subgroups (Young (0-24), Adult (35-44) and Senior (65+)) and modalities (T1w, T2w, T2-FLAIR) for NYU Langone Dataset. The silhouette scores and visualization present that Neuro-JEPA shows best separation on age subgroups.}
    \label{fig:sup-age-tsne}
\end{figure}

\newpage
\section{Additional Result for MoE Routing Analysis and Visualization}
\label{apd:moe_routing_analysis}
\subsection{MoE Foreground vs. Background Routing}
We present result for different model size foreground vs. background routing on NYU-Langone dataset with ViT-Base-MoE and ViT-Large-MoE on all layers in Supplementary \Cref{sup-moe-routing-fg-bg-1,sup-moe-routing-fg-bg-2}. The routing distribution shows that experts are able to separate between foreground and background especially on early layers where all foreground/background tokens routed to different experts.

\subsection{MoE Different Modalities Routing}
We present result for different model size MoE routing on different modalities in Supplementary \Cref{sup-moe-modalities-nyu,sup-moe-modalities-nyu-l}. While a majority of experts have balanced routing across modalities, few experts present different routing behaviors across modalities such as Expert 16 on Layer 5 and Expert 7 on Layer 7. Given that different neuroimaging modalities can have very similar anatomies with shared information, this indicate MoE learns when to deviate from shared computation, assigning only a subset of experts to modality-sensitive contrast information.

\subsection{MoE Heatmap}
We present heatmap result for different model size MoE routing on examining if different experts assign different distribution on different tokens in Supplementary \Cref{sup-moe-routing-heatmap-1,sup-moe-routing-heatmap-2,sup-moe-routing-heatmap-3,sup-moe-routing-heatmap-4,sup-moe-routing-heatmap-5,sup-moe-routing-heatmap-6}. The result shows different experts can focus on different tokens across layers, indicating MoE learns on routing based on different anatomy structures.

\subsection{MoE Visualization}
\begin{figure}[htbp]
    \centering
    \includegraphics[width=1.0\textwidth]{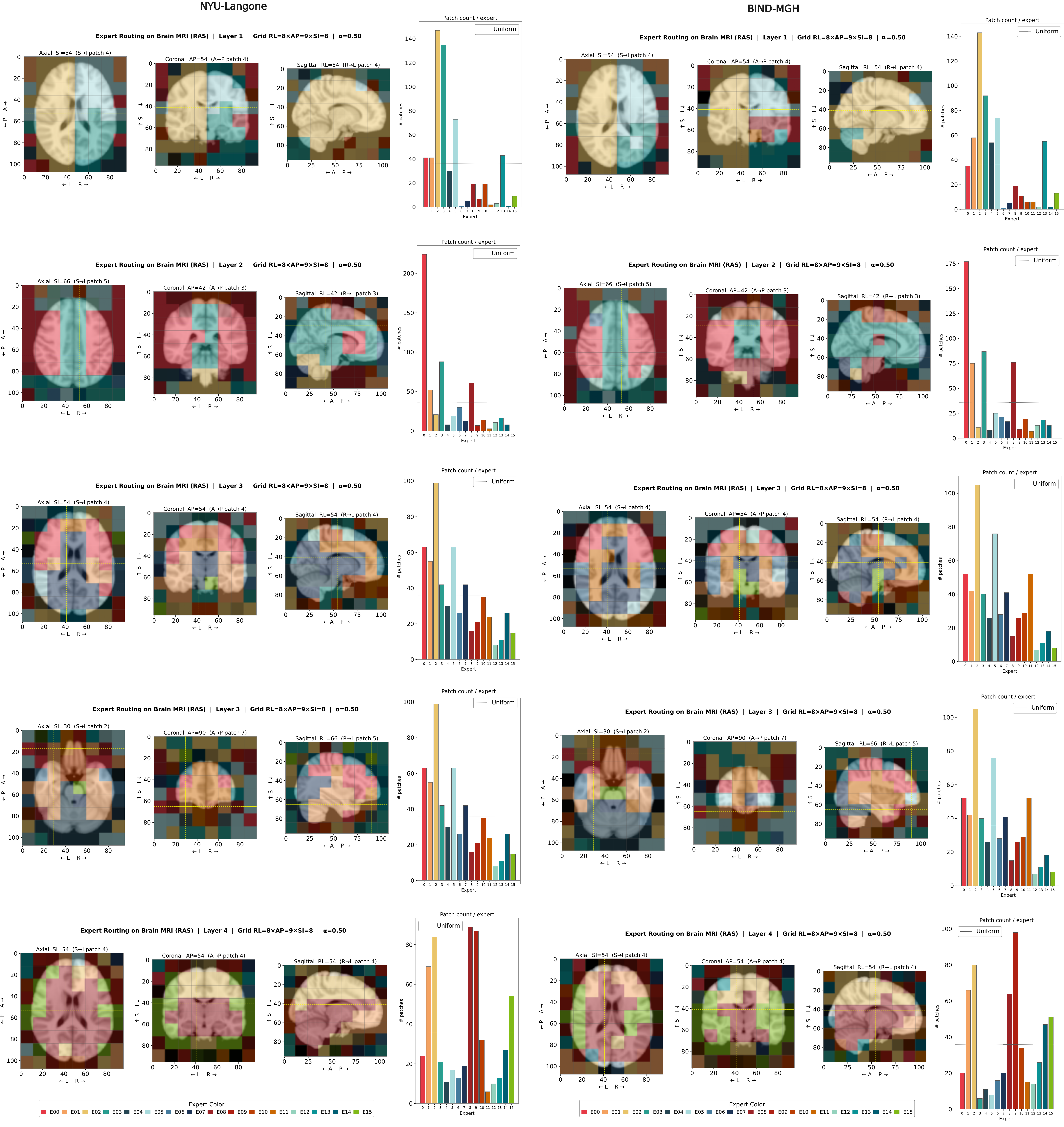}
    \caption{\textbf{MoE Visualization on T1w template - Patch Size 12.}}
    \label{fig:sup-moe-vis-t1w}
\end{figure}

\begin{figure}[htbp]
    \centering
    \includegraphics[width=1.0\textwidth]{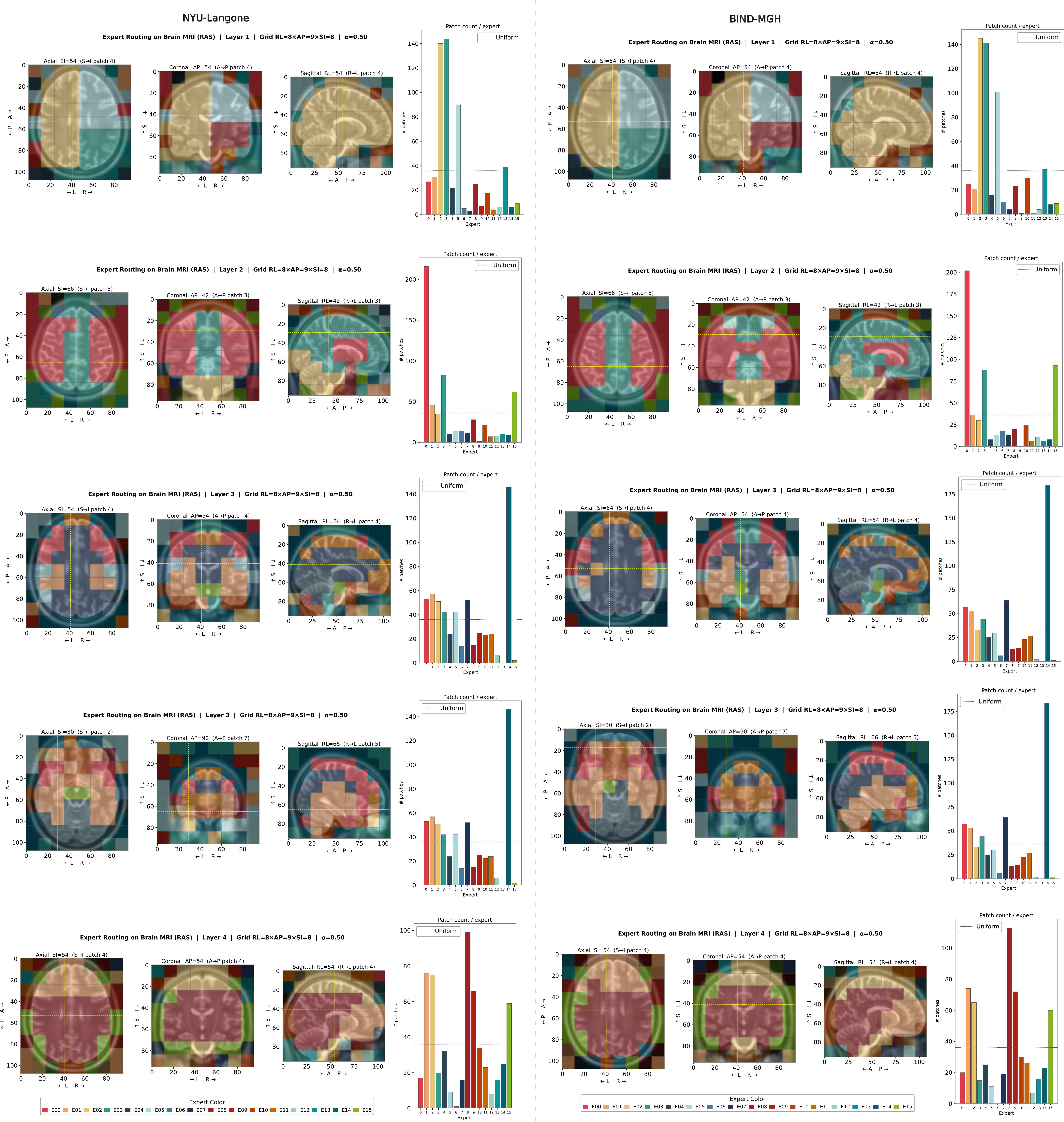}
    \caption{\textbf{MoE Visualization on T2w template - Patch Size 12.}}
    \label{fig:sup-moe-vis-t2w}
\end{figure}

Supplementary \Cref{fig:sup-moe-vis-t1w,fig:sup-moe-vis-t2w,fig:sup-moe-vis-t1w-ps8,fig:sup-moe-vis-t2w-ps8} present different patch size MoE routing visualization on selected slices on NYU-Langone and BIND-MGH datasets with T1w and T2w template. The result shows that different experts focus on different anatomical structure for MoE. Additionally, both NYU-Langone and BIND-MGH present very similar routing behaviors, indicating the revealed anatomical structure focus is not coincident. 

\subsection{Attention Map Visualization}
Supplementary \Cref{fig:sup-att_vis-stroke,fig:sup-att_vis-ms,fig:sup-att_vis-glio} present attention map visualization for stroke, multiple sclerosis and gliobastoma lesions. The visualization presents that the pretrained model is able to attend to different pathology without further adapting.
\noindent
\centering
\label{sup-moe-routing-fg-bg-1}
\includegraphics[width=0.75\linewidth, page=1]{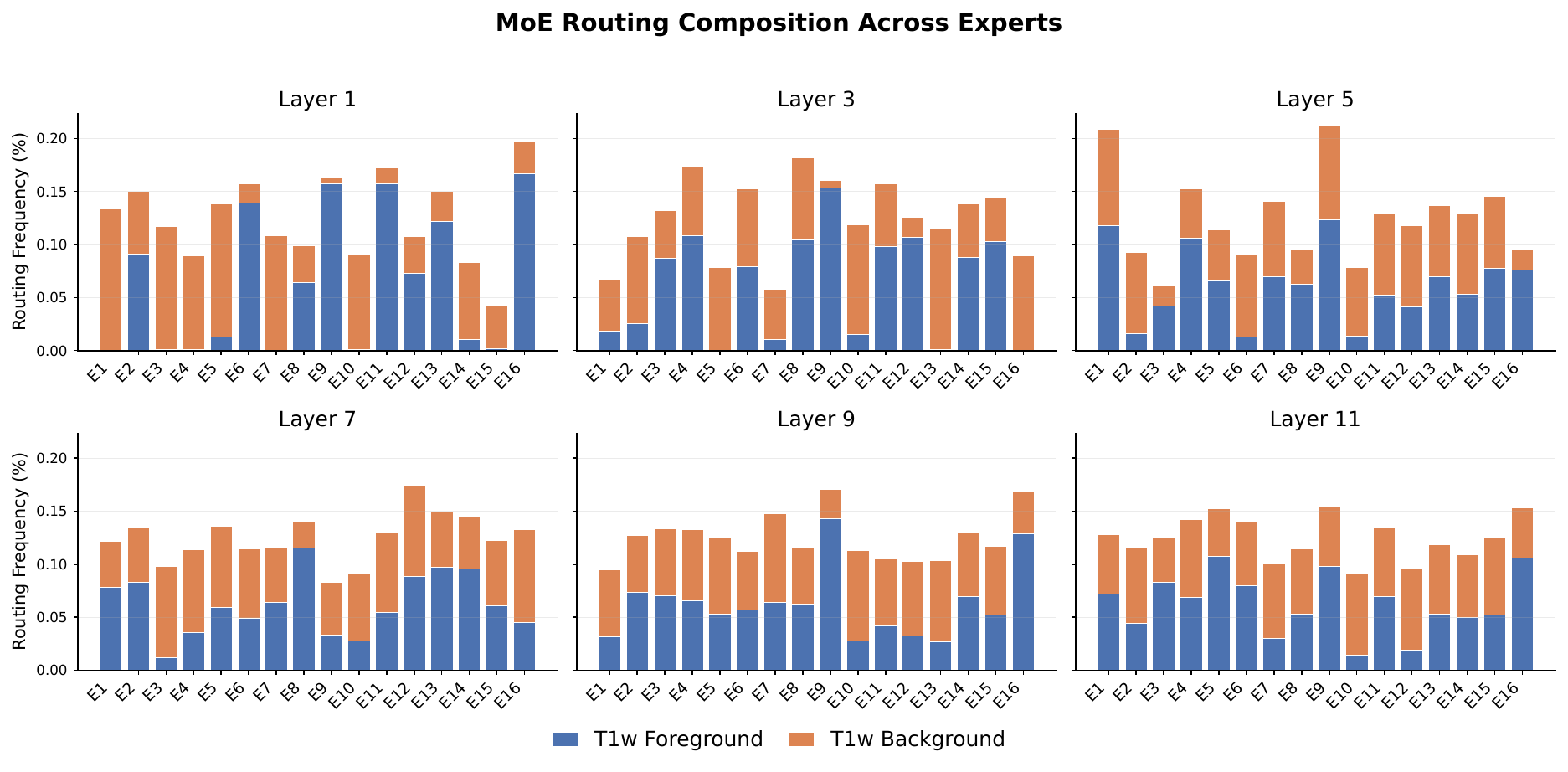}
\captionof{figure}{Supplementary MoE Routing FG vs BG -- T1w NYU - ViT-B}
\vspace{0.5cm}

\noindent
\centering
\label{sup-moe-routing-fg-bg-2}
\includegraphics[width=0.75\linewidth, page=1]{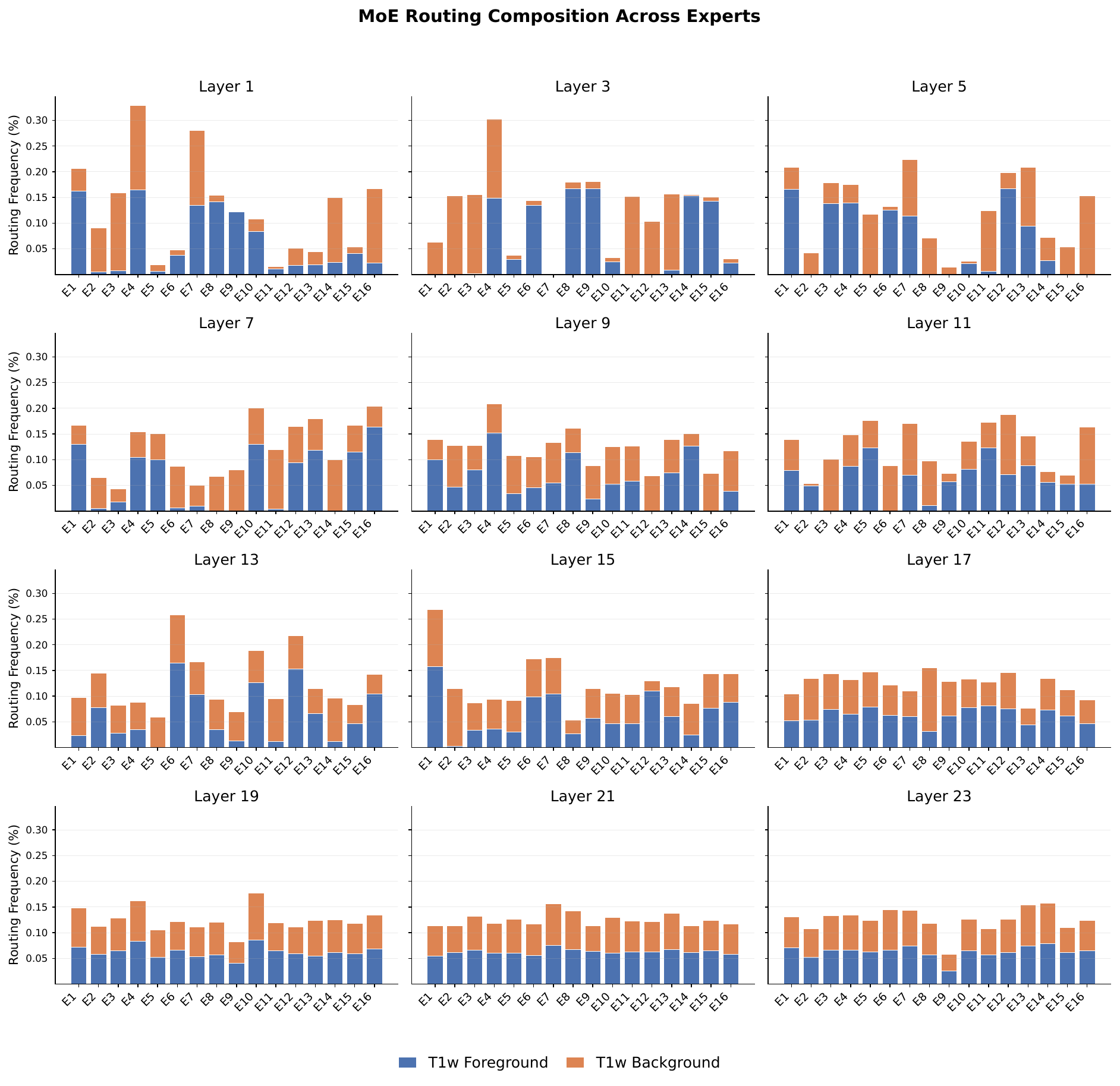}
\captionof{figure}{Supplementary MoE Routing FG vs BG -- T1w MGH - ViT-L}
\vspace{0.5cm}

\noindent
\centering
\label{sup-moe-modalities-nyu}
\includegraphics[width=0.85\linewidth, page=1]{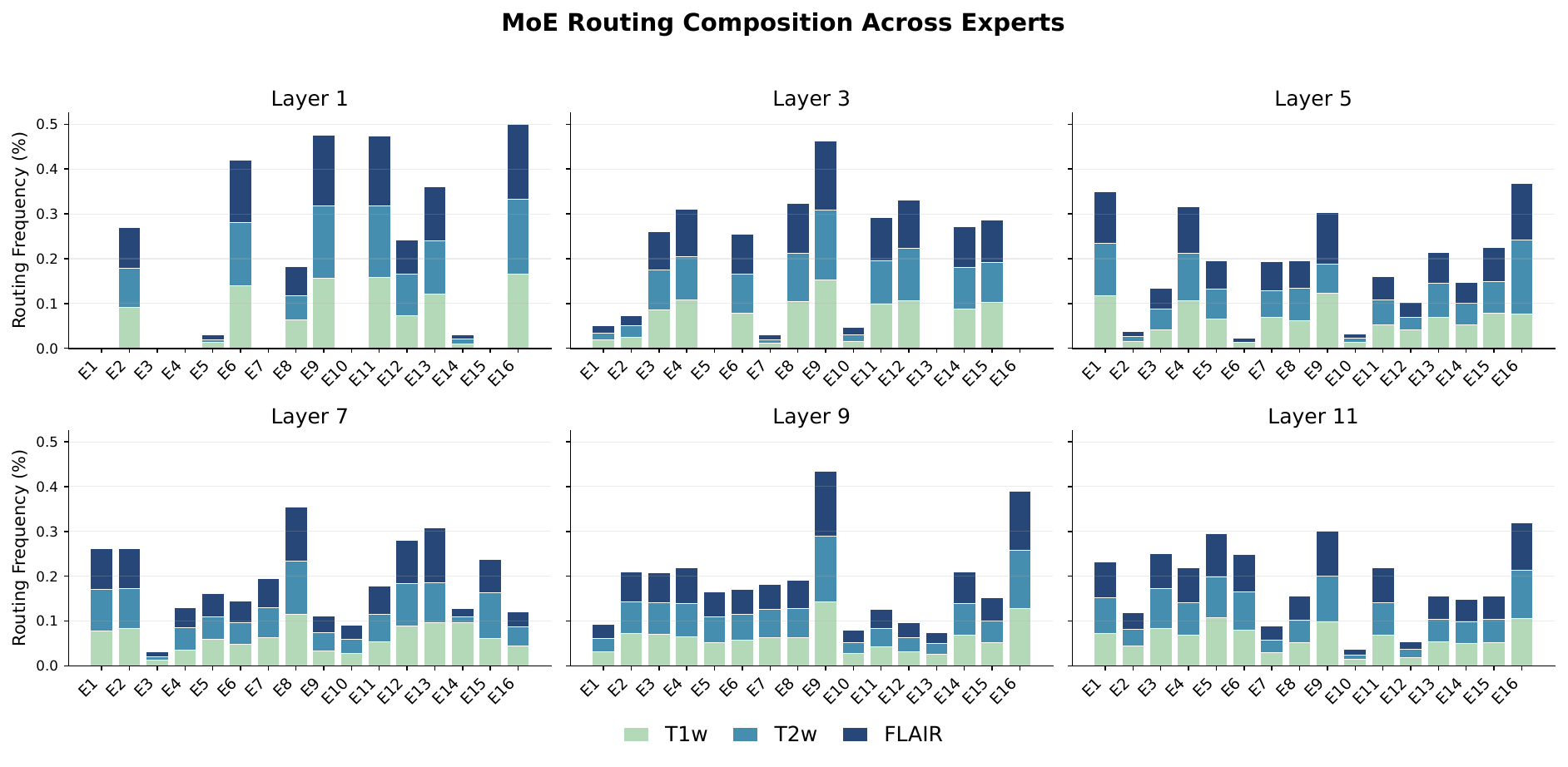}
\captionof{figure}{Supplementary MoE Routing on Different Modalities -- NYU - ViT-B}
\vspace{0.5cm}
\noindent
\centering
\label{sup-moe-modalities-nyu-l}
\includegraphics[width=0.85\linewidth, page=1]{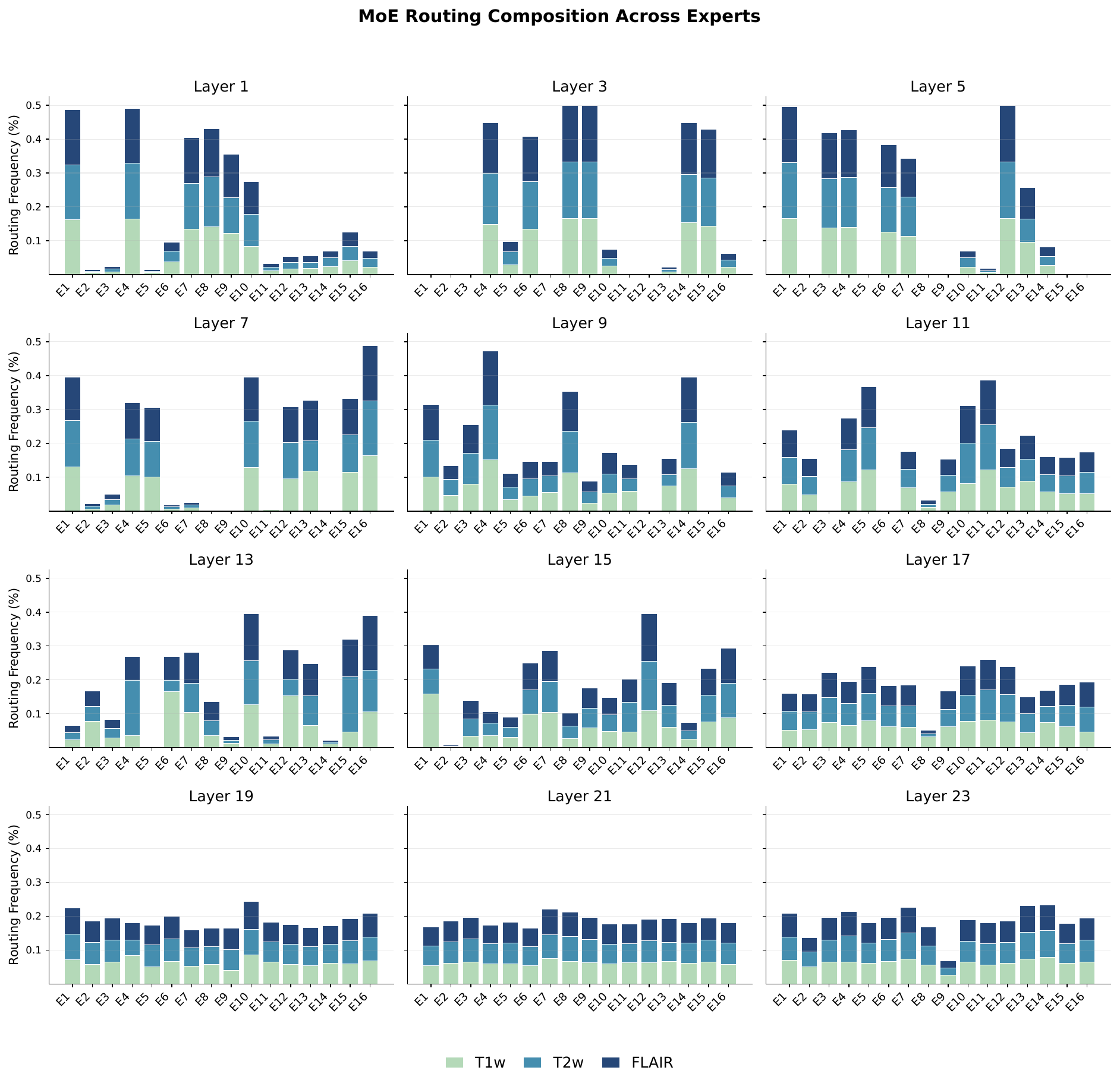}
\captionof{figure}{Supplementary MoE Routing on Different Modalities -- NYU - ViT-L}
\noindent
\centering
\label{sup-moe-routing-heatmap-1}
\includegraphics[width=0.85\linewidth, page=1]{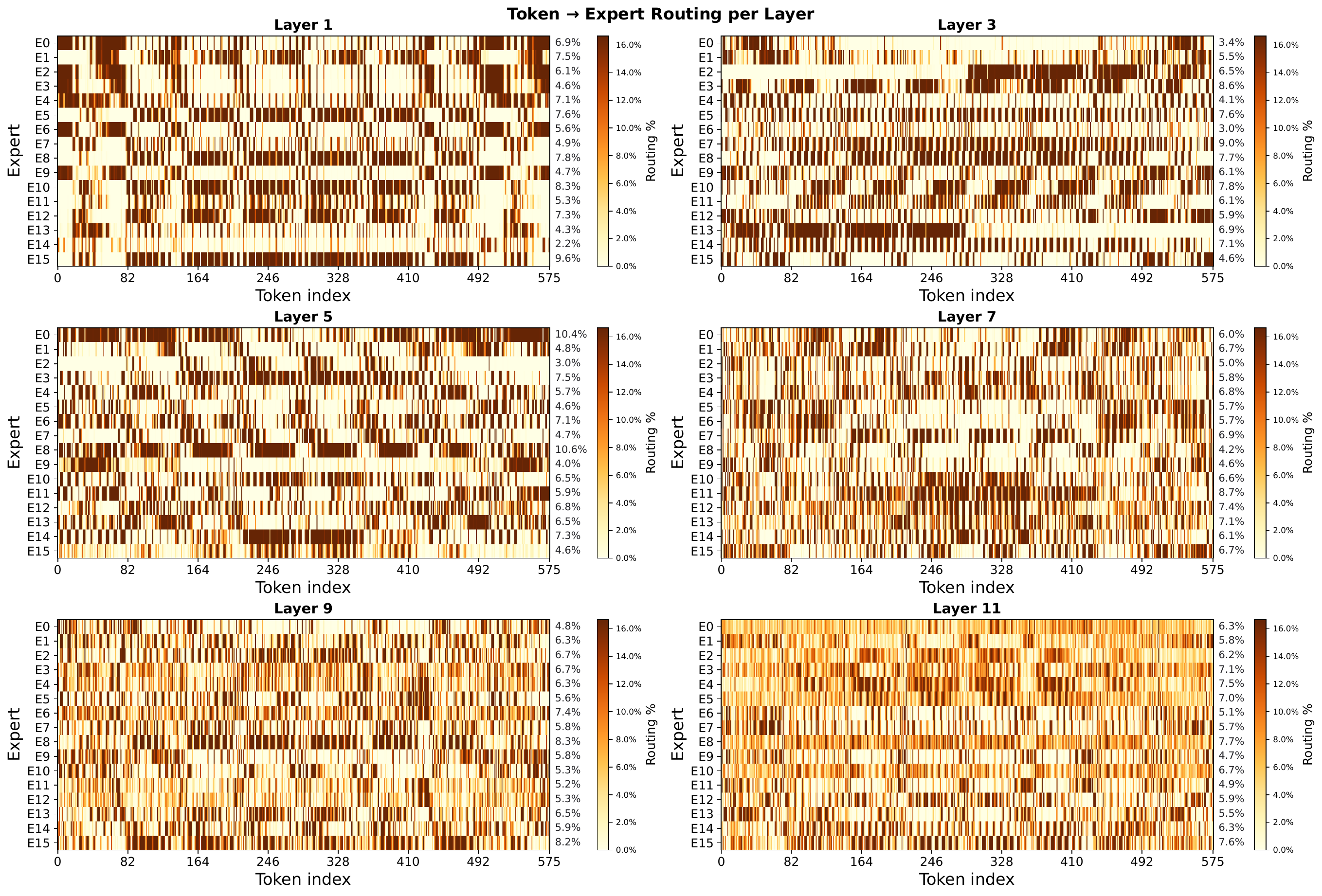}
\captionof{figure}{Supplementary MoE Routing Heatmaps -- NYU T1w - ViT-B}
\vspace{0.5cm}
\noindent
\centering
\label{sup-moe-routing-heatmap-2}
\includegraphics[width=0.85\linewidth, page=1]{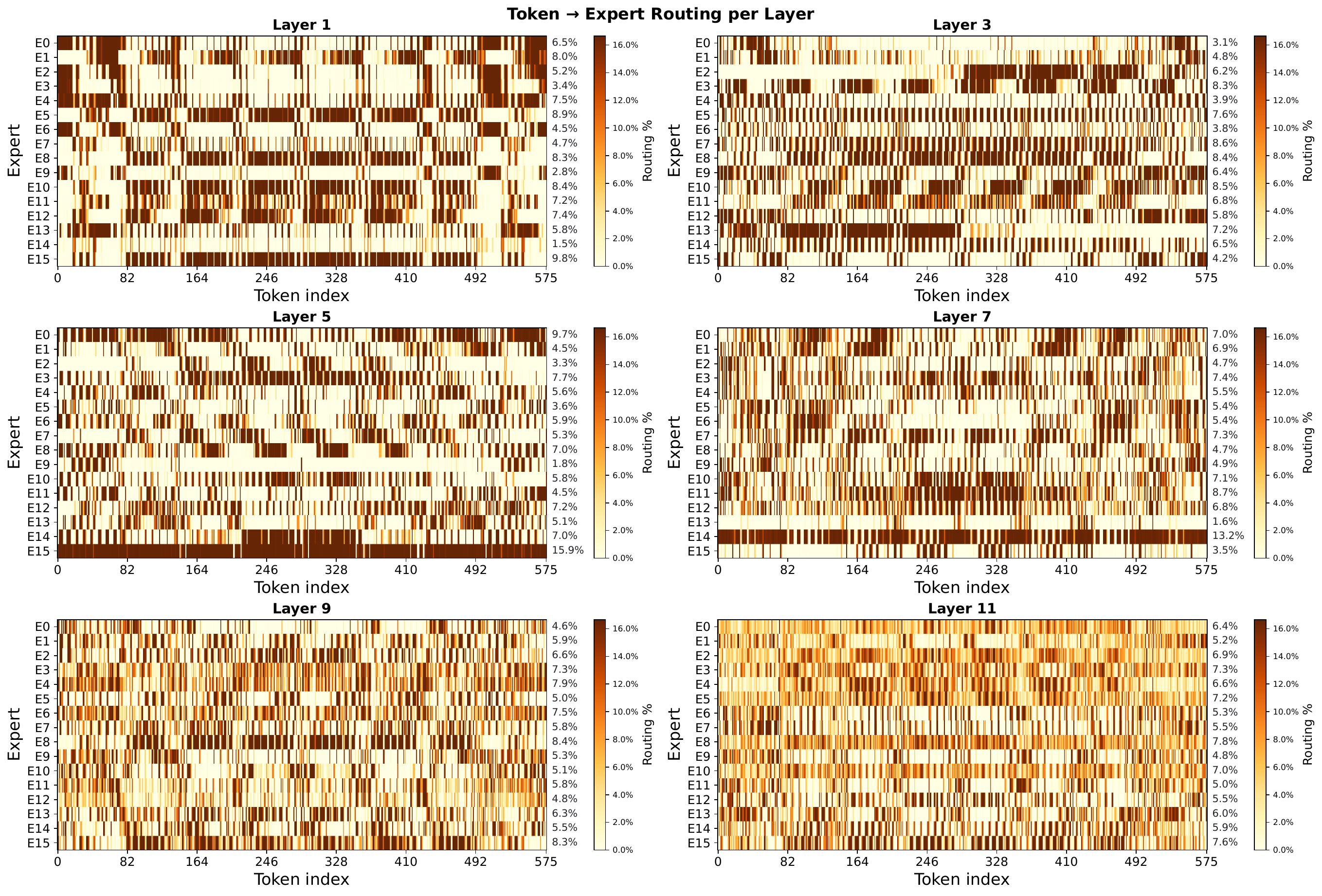}
\captionof{figure}{Supplementary MoE Routing Heatmaps -- NYU T2w - ViT-B}
\vspace{0.5cm}
\noindent
\centering
\label{sup-moe-routing-heatmap-3}
\includegraphics[width=0.85\linewidth, page=1]{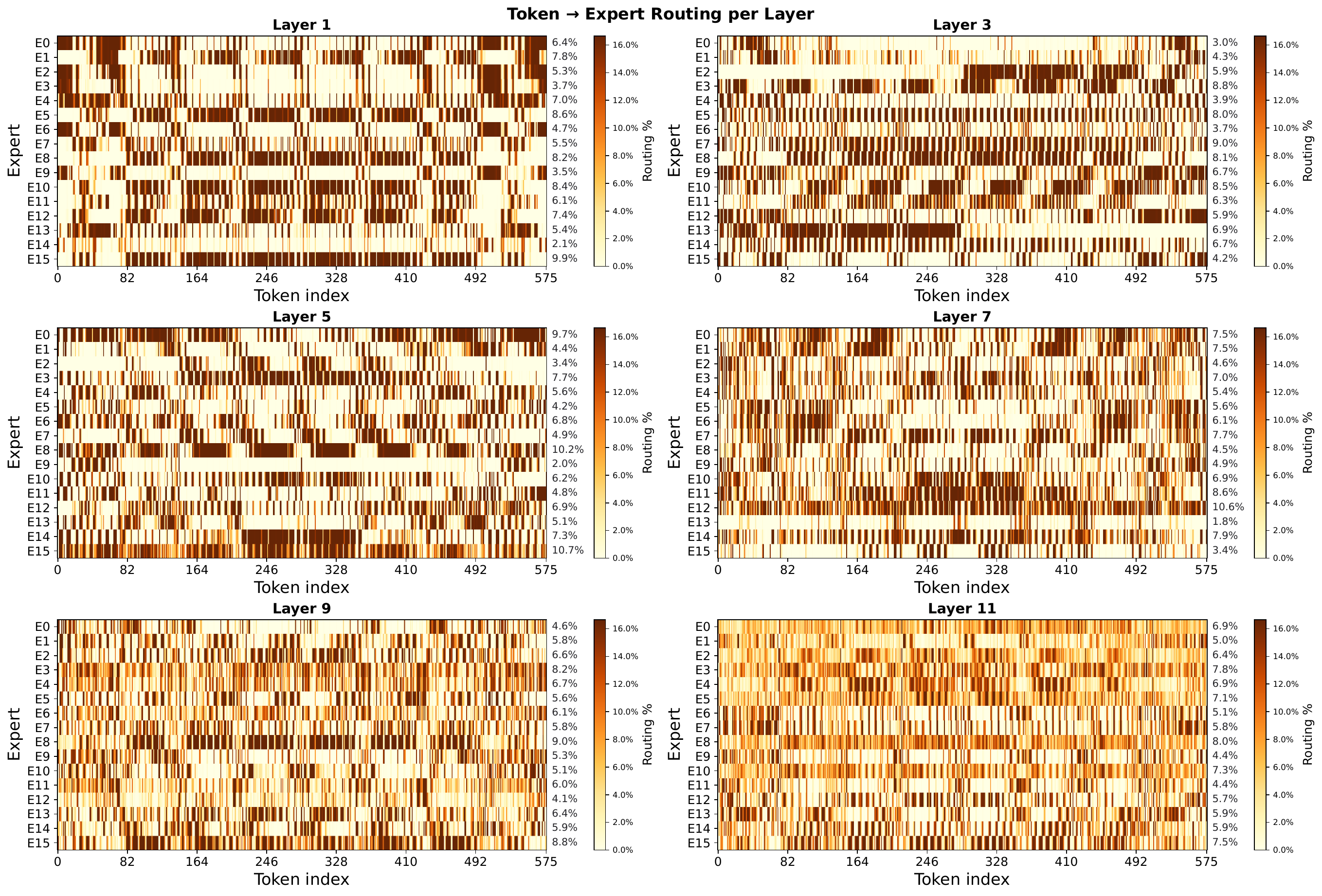}
\captionof{figure}{Supplementary MoE Routing Heatmaps -- NYU FLAIR - ViT-B}
\vspace{0.5cm}
\noindent
\centering
\label{sup-moe-routing-heatmap-4}
\includegraphics[width=0.85\linewidth, page=1]{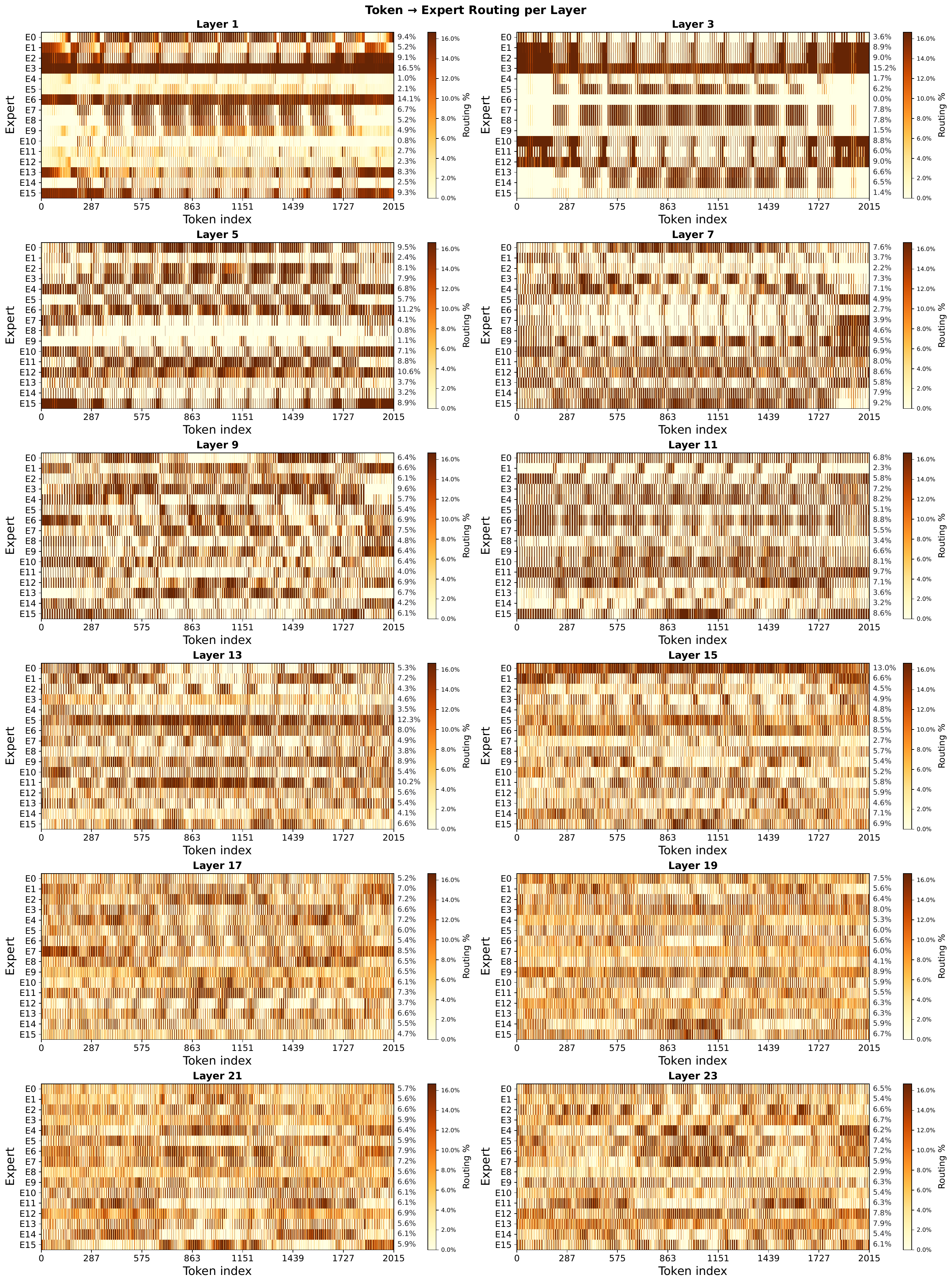}
\captionof{figure}{Supplementary MoE Routing Heatmaps -- NYU T1w - ViT-L}
\vspace{0.5cm}
\noindent
\centering
\label{sup-moe-routing-heatmap-5}
\includegraphics[width=0.85\linewidth, page=1]{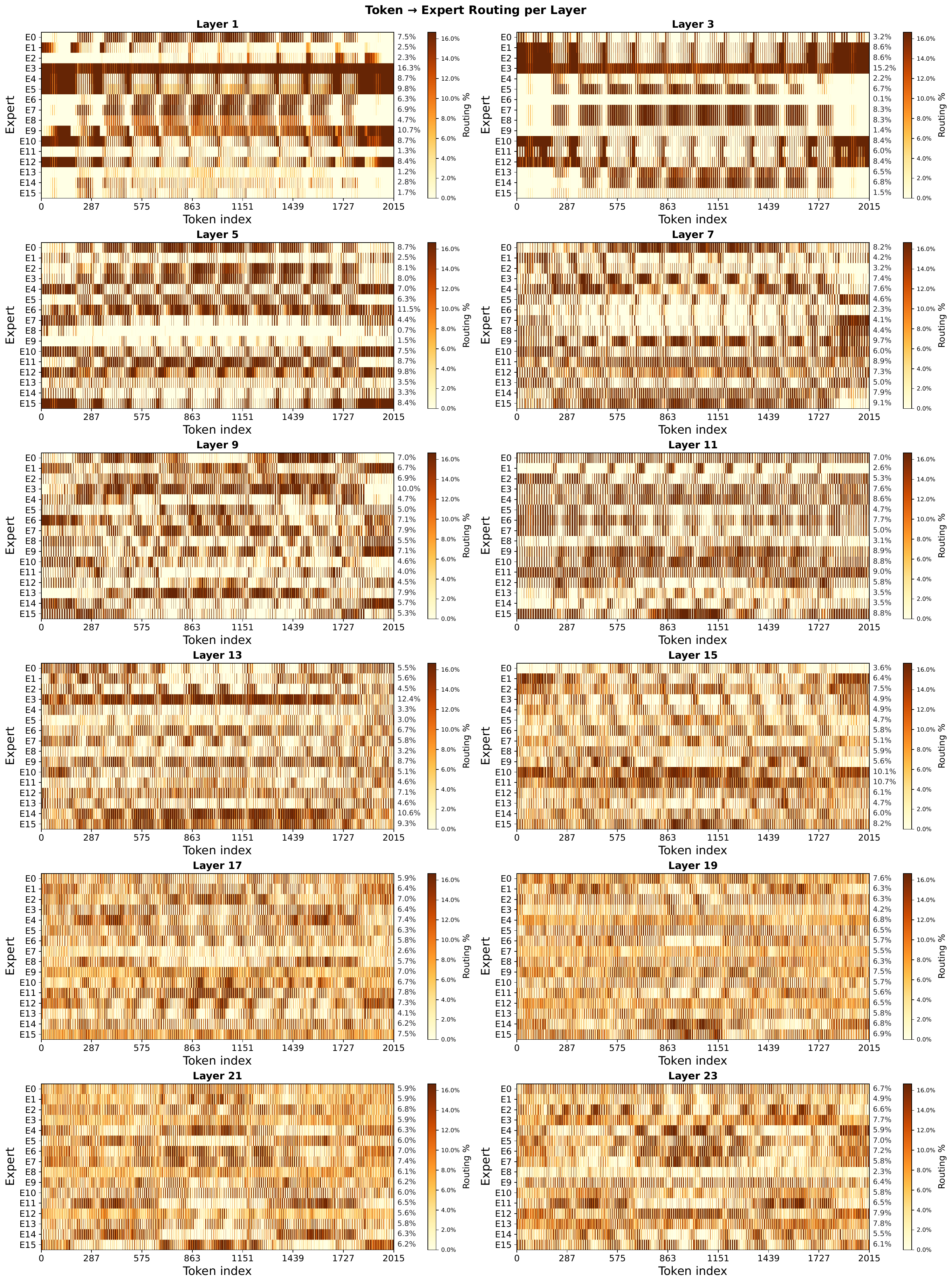}
\captionof{figure}{Supplementary MoE Routing Heatmaps -- NYU T2w - ViT-L}
\vspace{0.5cm}
\noindent
\centering
\label{sup-moe-routing-heatmap-6}
\includegraphics[width=0.85\linewidth, page=1]{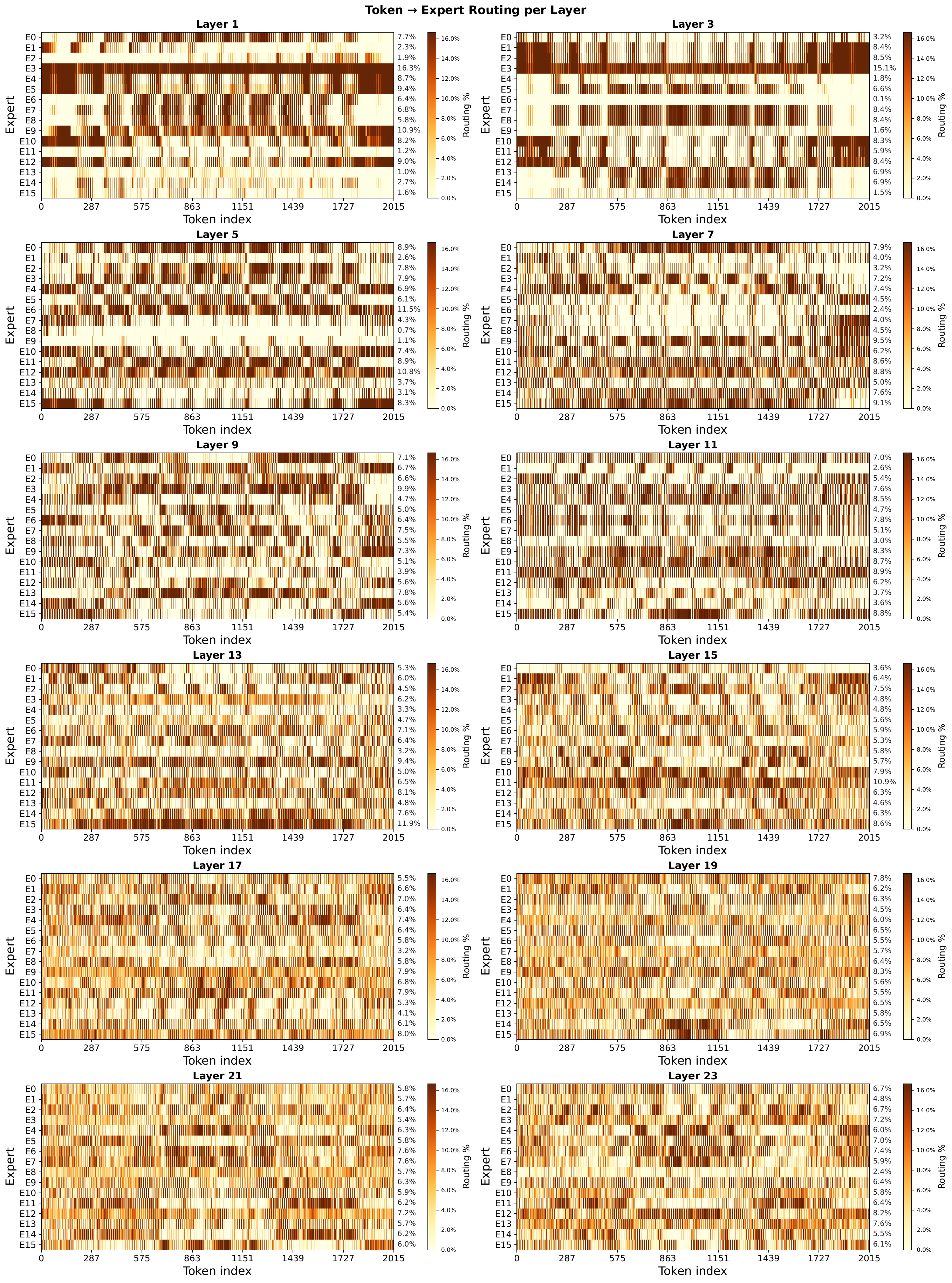}
\captionof{figure}{Supplementary MoE Routing Heatmaps -- NYU FLAIR - ViT-L}

\begin{figure}[htbp]
    \centering
    \includegraphics[width=1.0\textwidth]{figures/Supplementary-MoE-Routing-Vis/sup-moe-vis-t1w-ps8.pdf}
    \caption{\textbf{MoE Visualization on T1w template - Patch Size 8.}}
    \label{fig:sup-moe-vis-t1w-ps8}
\end{figure}

\begin{figure}[htbp]
    \centering
    \includegraphics[width=1.0\textwidth]{figures/Supplementary-MoE-Routing-Vis/sup-moe-vis-t2w-ps8.pdf}
    \caption{\textbf{MoE Visualization on T2w template - Patch Size 8.}}
    \label{fig:sup-moe-vis-t2w-ps8}
\end{figure}
\begin{figure}[htbp]
    \centering
    \includegraphics[width=0.7\textwidth]{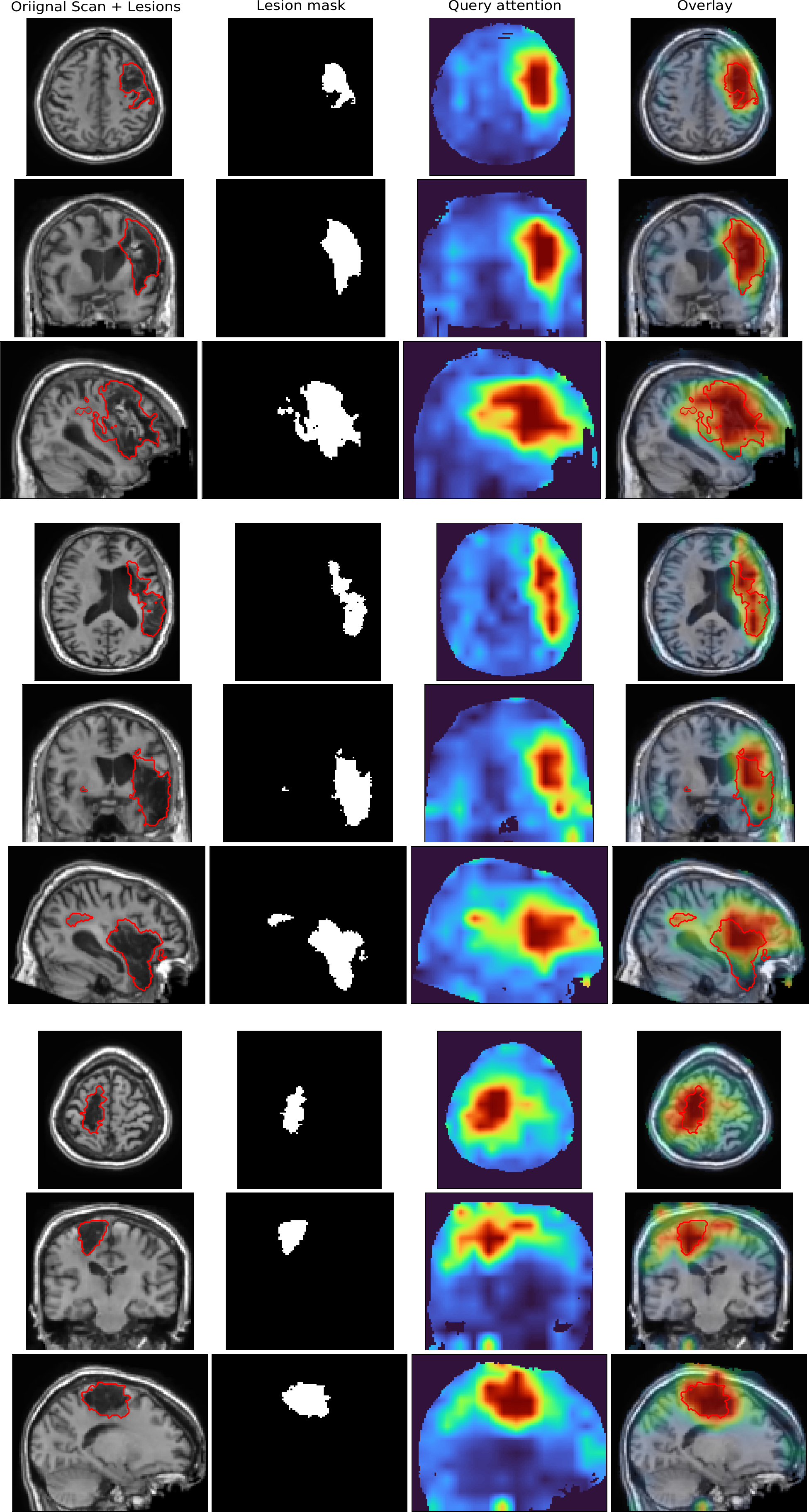}
    \caption{\textbf{Attention Visualization - Stroke Lesions.}}
    \label{fig:sup-att_vis-stroke}
\end{figure}

\begin{figure}[htbp]
    \centering
    \includegraphics[width=0.7\textwidth]{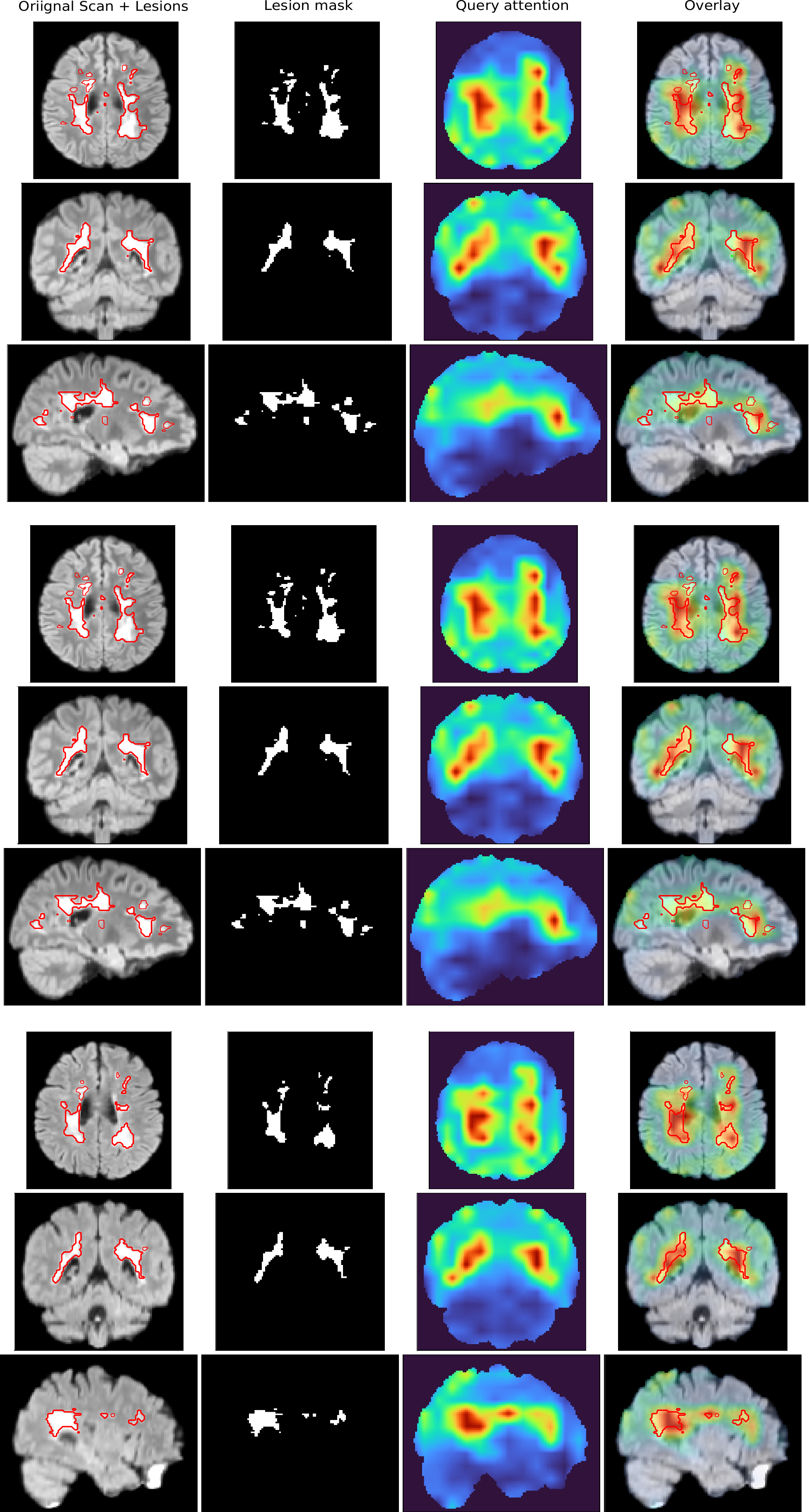}
    \caption{\textbf{Attention Visualization - Multiple Sclerosis Lesions.}}
    \label{fig:sup-att_vis-ms}
\end{figure}

\begin{figure}[htbp]
    \centering
    \includegraphics[width=0.7\textwidth]{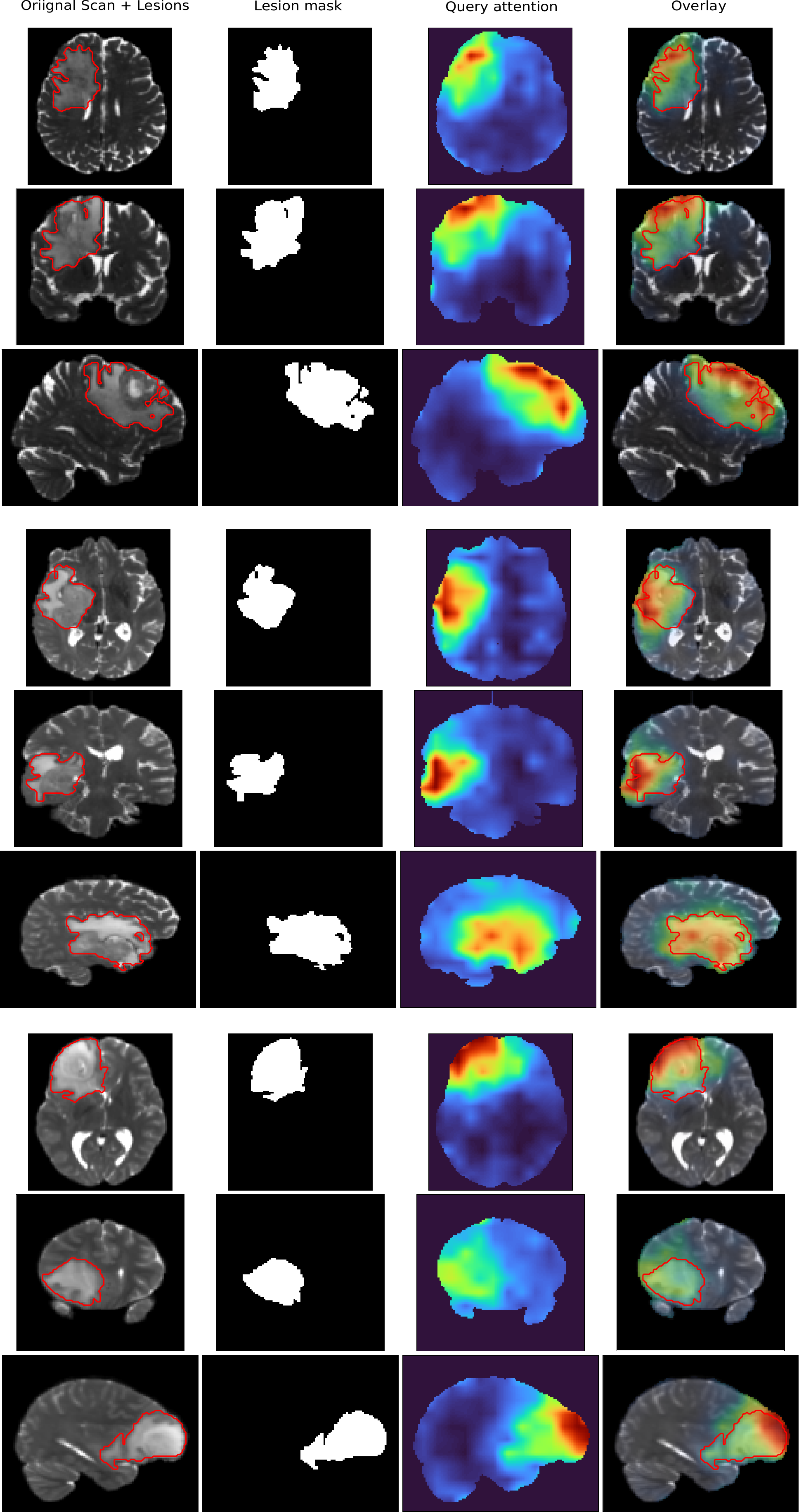}
    \caption{\textbf{Attention Visualization - Gliobastoma Lesions.} }
    \label{fig:sup-att_vis-glio}
\end{figure}

\clearpage

\end{document}